\documentclass[a4paper,12pt]{article}

\usepackage[a4paper,hmargin=25mm,textheight=660pt,centering]{geometry}
\usepackage{setspace}

\usepackage{amsmath,amssymb,amsthm,mathtools}
\numberwithin{equation}{section}
\usepackage{iftex}
\usepackage[T1]{fontenc}
\usepackage{lmodern}
\usepackage{microtype}
\UseMicrotypeSet[protrusion]{basicmath}
\ifLuaTeX
  \usepackage{selnolig}
\fi
\usepackage{graphicx,longtable,booktabs,array,caption,float,needspace}
\usepackage{enumitem}
\usepackage[svgnames]{xcolor}
\usepackage{tikz}
\usetikzlibrary{arrows.meta,calc}
\usepackage{natbib}
\usepackage{bibunits}
\defaultbibliographystyle{abbrvnat}
\defaultbibliography{ref}
\setcitestyle{aysep={,},citesep={;}}
\PassOptionsToPackage{hyphens}{url}
\usepackage[unicode]{hyperref}
\usepackage{bookmark,xurl}
\DeclareFontFamily{U}{mathx}{\hyphenchar\font45}
\DeclareFontShape{U}{mathx}{m}{n}{<-> mathx10}{}
\DeclareSymbolFont{mathx}{U}{mathx}{m}{n}
\DeclareFontSubstitution{U}{mathx}{m}{n}
\DeclareMathAccent{\widebar}{0}{mathx}{"73}
\let\hat\widehat
\let\tilde\widetilde
\let\bar\widebar

\hypersetup{
  colorlinks=true,
  linkcolor={Blue},
  filecolor={Maroon},
  citecolor={Blue},
  urlcolor={Blue}}

\newtheorem{theorem}{Theorem}
\newtheorem{proposition}{Proposition}

\newtheorem{lemma}{Lemma}
\theoremstyle{definition}
\newtheorem{assumption}{Assumption}

\newtheorem{definition}{Definition}
\theoremstyle{plain}
\newtheorem{remark}{Remark}
\newcommand{\R}{\mathbb{R}}
\newcommand{\E}{\mathbb{E}}
\newcommand{\D}{\mathbb{D}}
\newcommand{\N}{\mathbb{N}}
\renewcommand{\P}{\mathbb{P}}
\newcommand{\G}{\mathbb{G}}

\newcommand{\cB}{\mathcal{B}}
\newcommand{\cC}{\mathcal{C}}
\newcommand{\cD}{\mathcal{D}}
\newcommand{\cE}{\mathcal{E}}
\newcommand{\cF}{\mathcal{F}}
\newcommand{\cG}{\mathcal{G}}
\newcommand{\cH}{\mathcal{H}}

\newcommand{\cJ}{\mathcal{J}}
\newcommand{\cK}{\mathcal{K}}
\newcommand{\cL}{\mathcal{L}}
\newcommand{\cM}{\mathcal{M}}
\newcommand{\cN}{\mathcal{N}}

\newcommand{\cP}{\mathcal{P}}
\newcommand{\cQ}{\mathcal{Q}}
\newcommand{\cR}{\mathcal{R}}

\newcommand{\cT}{\mathcal{T}}
\newcommand{\cU}{\mathcal{U}}
\newcommand{\cV}{\mathcal{V}}

\newcommand{\cX}{\mathcal{X}}
\newcommand{\cY}{\mathcal{Y}}
\newcommand{\cZ}{\mathcal{Z}}

\DeclareMathOperator*{\argmin}{arg\,min}
\DeclareMathOperator{\Var}{Var}
\DeclareMathOperator{\Cov}{Cov}
\DeclareMathOperator{\tr}{tr}
\DeclareMathOperator{\diag}{diag}

\DeclareMathOperator{\clip}{clip}
\newcommand{\softmax}{\boldsymbol{\sigma}}

\DeclareMathOperator{\GELU}{\mathsf{GELU}}
\DeclareMathOperator{\spn}{span}
\newcommand{\op}{\mathrm{op}}
\newcommand{\T}{\mathsf T}
\newcommand{\bbar}{\bar b}

\newcommand{\1}{\mathbf 1}

\title{Can pretrained Transformers deliver semiparametrically efficient inference?}
\author{Han Yan$^1$,  Jin Zhu$^2$, Chengchun Shi$^1$ and Tengyao Wang$^1$\thanks{Address for correspondence: Tengyao Wang, Department of Statistics, London School of Economics and Political Science, 69 Aldwych, London WC2B 4RR, United Kingdom. \textbf{Email}: t.wang59@lse.ac.uk} \\ \\ 
$^1${Department of Statistics, London School of Economics, London, UK }\\ $^2$School of Mathematics, University of Birmingham, Birmingham, UK}
\date{}

\begin{document}
\begin{bibunit}
\maketitle\vspace{-1.5em}

\def\spacingset#1{\renewcommand{\baselinestretch}%
{#1}\small\normalsize} \singlespacing

\begin{abstract}
\normalsize
Semiparametric models provide a broad framework for statistical inference, yet efficient estimation typically relies on influence function correction tailored to each estimand and on nuisance estimation repeated for every sample. We show that a single pretrained Transformer can deliver efficient estimation and valid inference for a family of estimands in one forward pass, without deriving efficient influence functions or estimating nuisance functions. Pretrained on simulated data with the true parameter as the only label, the Transformer learns the posterior mean and covariance of the queried estimand and hence returns an estimate and a Wald-type confidence region. Theoretically, we prove that, for locally flat pretraining priors of slowly growing complexity, the posterior mean is efficient for every queried estimand. The Transformer inherits this efficiency with asymptotically exact coverage, in probability over evaluation populations and queries, even under distribution shift. We also reveal that the guarantee places much higher demands on pretraining when the data-generating distribution is fixed, as in classical frequentist inference. Experiments on causal inference, partial linear regression, and survival analysis show that the pretrained model generally achieves the lowest estimation error among competing methods, with nearly nominal coverage, at a computational cost of only about 0.02 seconds per dataset.

\end{abstract}

\noindent{\it Keywords:} 
amortised inference,  efficiency, pretrained model, semiparametric inference,  Transformer 
\medskip
\singlespacing

\section{Introduction}
%\iffalse
%\textcolor{teal}{[TW: The current intro is roughly: 1. transformers and tabular foundation models are successful; 2. they enable amortised learning; 3. foundation model is also applied to semiparametric inference but they require estimand-specific EIF corrections; 4. we use a single transformer for multiple estimands. My slight concern is that if we are underselling our contribution and make the contribution more incremental than it is. Specifically, I have a feeling of `this is just one more paper on using foundation model on a different problem' when I read the intro. But we actually have some very interesting results that are specific to semiparametric inference that are buried in the contributions bullet points. I think we should begin with semiparametric inference and talk about the challenge of learning different EIF in each problem and ask if this can be automated. This leads on to PFN and pretraining and amortised inference as a possible answer (basically what we have in the opening now, but slightly compacted). Then we want to talk about people have tried this but there are specific challenges of pretraining for semiparametric inference -- that estimand-specific EIF correction is needed, limiting their applicability. Then we ask the central question of the paper. When we do this, I think we want to put emphasis on the fact that semiparametric efficiency can be achieved without explicitly constructing the EIF. This I think is one of the main surprises of the result and what brings the paper beyond `another PFN work'.]}
%\fi

Semiparametric inference \citep{bickel1993} is one of the most important statistical frameworks and central to a wide range of problems, such as partial linear models, causal inference and survival analysis. A long-standing challenge is that, to ensure the efficiency of the estimator, it is often necessary to employ the efficient influence function (EIF) of the parameter of interest, which can be difficult to derive and to estimate in the presence of complex nuisance functions \citep{tsiatis2006,chernozhukov2018}. Moreover, the EIF is specific to each estimand, and it must be estimated together with its nuisance functions for each new sample, imposing computational burdens and hindering fast inference.  Hence, a natural question is whether it is possible to use a single model for automated efficient inference across samples and estimands, without deriving the EIFs or nuisance functions?  

Recent advances in tabular foundation models based on prior-data fitted networks (PFN, \citealp{muller2022, hollmann2025}) shed light on this objective due to their amortised learning \citep{sainsburydale2024} ability, in the sense that a Transformer pretrained once on large-scale simulated data can be applied to new data, returning an estimate in one forward pass without updating the Transformer parameters. This zero-shot paradigm enables fast estimation and prediction, since the learning cost is paid in pretraining instead of for new samples. Tabular foundation models have been developed for causal effect estimation \citep{balazadeh2025}, nonparametric regression \citep{ching2026,ma2026}, and uncertainty quantification \citep{ye2026can,fortini2026}, as reviewed in Section \ref{app:related} of our supplementary material (SM). 

However, when applied for semiparametric inference, the PFN may lead to inefficient estimates and invalid confidence intervals.  For the average treatment effect, \cite{melnychuk2026frequentist} proposes to correct the posterior of the PFN and remove a confounding bias induced by the prior. \cite{zhang2026} and \cite{melnychuk2026debiased} considered one-step EIF corrections in pretraining by debiasing the labels or the prior. However, these methods commonly target one estimand at a time, mostly the average treatment effect, and their estimand-specific EIF correction method can limit their applicability to general parameter families,  such as quantile treatment effects at different levels. Moreover, since pretraining is computationally demanding, it is more desirable to obtain a pretrained model that serves a whole family of estimands.

In this study, we aim to propose a single pretrained Transformer for efficient semiparametric inference across samples and estimands, without deriving any EIF or estimating any nuisance function. 
\begin{itemize}[leftmargin=*]
    \item Methodologically, we introduce a \textit{query-indexed} semiparametric framework, where the query specifies which estimand within a prespecified family to estimate and infer. For instance, in causal inference, the query class can include the average treatment effect and quantile treatment effects at different levels, so that one model serves the whole family. The designed Transformer takes a sample and a query as inputs and returns both a queried parameter estimate and its estimated covariance. Pretrained on data simulated from a pretraining prior with a Gaussian score loss, it uses only the true parameter as the label, \textit{without} any nuisance estimation or influence function derivation. Once pretrained, the Transformer can output an estimate and a confidence region for any queried parameter without updating the model on an evaluation sample. 
    \item  Theoretically, we characterise how the pretraining prior, the complexity of the Transformer and the evaluation distribution govern efficiency of amortised inference.
    (i) We establish that, under a locally flat pretraining prior of slowly growing complexity, the posterior mean of every queried estimand is semiparametrically efficient. Crucially, the posterior attains this efficiency without any estimand-specific EIF correction. Hence, a single pretraining objective, with the true parameter as the only label, suffices for the whole family of estimands. 
    %prove a semiparametric efficiency theorem for the posterior moments that reveals, under a locally flat pretraining prior of slowly growing complexity, the posterior mean of every queried estimand is efficient.% Therefore, pretraining once with the true parameter as the only label yields an efficient estimator of every estimand in the family, without any EIF correction.
    (ii) It is shown that, under such a prior, a pretrained Transformer can inherit the efficiency of the posterior mean, provided that its parameter count is quadratic in the prior complexity and the number of pretraining samples is at least quadratic in the prior complexity, up to logarithmic factors whose powers increase with the data dimension. 
    (iii) We provide the efficiency and inference validity guarantees of the Transformer output that holds in probability over evaluation populations and queries, even when their distribution differs from the pretraining prior. It is also revealed that for a fixed evaluation population, as in the classical frequentist inference framework, the efficiency is harder to attain, because guarantees that cover every approximate minimiser of the pretraining loss then need a number of pretraining samples exponential in the prior complexity.
    \item Empirically, we test the proposal on four concrete tasks: estimating the average and quantile treatment effects, the coefficient in a semiparametric model, and the mean survival time. These four tasks cover important problems in causal inference, semiparametric regression, and survival analysis. Comparisons of our method with various existing baselines show that (i) our method achieves superior estimation accuracy and (ii) it delivers valid statistical inference, empirically covering the true parameters at close to the nominal level. Computationally, it is also much faster than existing machine learning methods, with a runtime below $10^{-1}$ seconds. 
\end{itemize}

The remainder of the paper is organised as follows. Section \ref{sec:framework} introduces the framework and the proposed method, Section \ref{sec:theory} establishes  theoretical guarantees, Section \ref{sec:numerical-experiments} reports numerical experiments, before closing the paper with a discussion in Section \ref{sec:discussion}. %Appendix \ref{app:related} reviews related work in detail, and the other appendices contain proofs and implementation details.
Detailed reviews of related works, proofs and implementation details are presented in the supplement.

\section{Framework and methodology}\label{sec:framework}

\subsection{Problem setup}
\label{sec:setup}

%We first consider a standard setup for semiparametric inference problems.
Let $\cP$ be a statistical model for the distribution $\P$ of an observation $Z \in \cZ \subset \R^{d_z}$. Semiparametric inference concerns a finite-dimensional parameter $\beta_0$ of $\P$ in the presence of a potentially infinite-dimensional nuisance component $\eta_\P$ \citep{bickel1993,tsiatis2006}. %In many problems, $\beta_0$ is characterised by an estimating equation $\E_\P\{g(Z;\beta_0,\eta_\P)\} = 0$, where $\eta_\P$ may include conditional means, propensity scores or conditional distribution functions. % \citep{chernozhukov2018}. %For example, partially linear regression coefficients are defined through estimating equations involving conditional means, and treatment effects and quantiles are functionals of conditional means or distribution functions. %Such parameters can be studied under nonparametric models without imposing finite-dimensional restrictions on the underlying distribution.
Classical semiparametric inference methods typically involve an influence function correction and variance estimation, tailored to different parameters and recomputed for every dataset. %To amortise such computational costs, we aim to obtain a pretrained model on synthetic data that can perform efficient inference on different evaluation samples and a family of parameters without retraining. 
Motivated by the zero-shot learning ability of foundation models, we ask whether this whole procedure can be learned via pretraining, so that a single pretrained model delivers fast and efficient inference for new evaluation samples and for each parameter in a prespecified family, even without knowledge of its EIF.

To this end, we consider a family of parameters $$\{\beta_q(\P):~q \in \cQ, ~\P \in \cP\},$$ with the target parameter $\beta_q(\P)\in \R^{d_k}$ indexed by some query $q = (k,u)$ in an estimand query set $\cQ \coloneqq \{(k,u) : k \in \cK,\ u \in \cU_k\}$, where $k$ selects an estimand class from a prespecified set $\cK$ of semiparametric targets and $u \in \cU_k$ is an optional argument for the parameter class. For instance, $\cK$ may contain the average treatment effect (ATE), the quantile treatment effect (QTE) and the partially linear projection (PLR), with $\cU_k \subset (0,1)$ as the quantile level set for the QTE, while $u$ can be suppressed for estimands without a level argument, such as the ATE. %The query set $\cQ$ is specified before pretraining and determines the parameters for which the pretrained model can perform inference.
We aim to obtain a single model that can be used for inference of all parameters in $\cQ$, where computation burdens are amortised in pretraining. 

\subsection{Pretraining and inference procedure}\label{sec:method}

Having specified the estimands and their queries, we now describe how a  Transformer can be pretrained and conduct inference of all queried parameters in $\cQ$.  %We pretrain a Transformer to map a dataset and a query to a point estimate and an associated covariance estimate, using  a Gaussian score loss function and  true parameters as labels. Once pretrained, the Transformer can be reused to construct point estimates and confidence regions for new datasets and queries without further fitting. We describe
Our procedure consists of three steps, including  synthetic data generation, Transformer pretraining, and inference on evaluation samples. Figure \ref{fig:pretraining-inference} summarises the procedure.

\paragraph{Step 1: Generate pretraining data.}
We specify a \emph{pretraining prior} $\pi \coloneqq \pi_P\otimes\pi_q$ on $\cP\times\cQ$ and a distribution $\pi_n$ on the sample sizes $\cN \coloneqq \{n_{\min},\cdots,n_{\max}\}$, where $\pi_P$ and $\pi_q$ are the priors on populations and queries, respectively. %Section \ref{sec:target-efficiency} specifies the population priors covered by the theoretical analysis.
Under the pretraining prior $\pi$, the pretraining data $\cT \coloneqq \{(\cD^{(t)}_{n_t},q_t,\beta_{0,t})\}_{t = 1}^T$ is generated as follows.
\begin{enumerate}[leftmargin=*]
  \item Draw random distribution triples $\{(\P_t,q_t,n_t)\}_{t = 1}^T \overset{\rm i.i.d.}{\sim} \pi\otimes\pi_n$, with $q_t\coloneqq (k_t,u_t)$.
  \item For each $t \in [T]$,  independently generate $\cD^{(t)}_{n_t} = \{Z_i^{(t)}\}_{i = 1}^{n_t} \overset{\rm i.i.d.}{\sim} \P_t$, and compute the true parameter $\beta_{0,t} \coloneqq \beta_{q_t}(\P_t)$.
\end{enumerate}

\paragraph{Step 2: Pretrain the Transformer.}
The Transformer $f = (f^\beta,f^V)$ takes a dataset $\cD_n = \{Z_i\}_{i = 1}^n$ and a query $q = (k,u)$ as inputs and returns $(\beta,V) \coloneqq (f^\beta(\cD_n,q),f^V(\cD_n,q))$, where  $\beta \in \R^{d_k}$ estimates $\beta_q(\P)$, and $V \in \R^{d_k\times d_k}$ is positive definite and $V/n$ estimates the sampling covariance. The detailed architecture of the Transformer class $\cF_{\mathrm{TF}}$ is presented in Section \ref{app:architecture} of the SM.
%Its readout projects the estimate onto a fixed ball and clips the eigenvalues of $V$ to a fixed interval $[v_-,v_+]$, which our theory assumes to contain those of the efficient covariance.
%With the simulated data $\cT = \{(\cD^{(t)}_{n_t},q_t,\beta_{0,t})\}_{t = 1}^T$, 
We pretrain a Transformer $f$ using the Gaussian score loss 
\begin{equation}\label{eq:score}
\ell_n(\beta_0;\beta,V) \coloneqq \frac{1}{2}\log\det V+\frac{n}{2}(\beta_0-\beta)^\T V^{-1}(\beta_0-\beta),
\end{equation}
where $\beta_0$ is the true parameter and $(\beta,V)$ is a candidate output. The empirical risk is
\[
\hat R_T(f) \coloneqq \frac{1}{T}\sum_{t = 1}^T\ell_{n_t}\{\beta_{0,t};f^\beta(\cD^{(t)}_{n_t},q_t),f^V(\cD^{(t)}_{n_t},q_t)\}.
\]
We obtain $\hat f \in \cF_{\mathrm{TF}}$ by approximate empirical risk minimisation so that
\begin{equation}\label{eq:erm}
\hat R_T(\hat f) \leq \inf_{f \in \cF_{\mathrm{TF}}}\hat R_T(f)+\epsilon_{\mathrm{opt}},
\end{equation}
where $\epsilon_{\mathrm{opt}} \geq 0$ is some optimisation tolerance.

\paragraph{Step 3: Inference on an evaluation sample.}
With the pretrained  $\hat f$, we can conduct inference on an evaluation sample for any semiparametric target $q \in \cQ$, without updating $\hat f$.
Let $N \in \cN$ be the size of the evaluation sample and let $\nu$ be an evaluation distribution on $\cP\times\cQ$. Independently of $\cT$, draw $(\P,q) \sim \nu$ and then $\cD_N=\{Z_i\}_{i=1}^N$ with $Z_i\overset{\rm i.i.d.}{\sim}\P$. %The choice $\nu=\delta_{(\P,q)}$ corresponds to inference for $\beta_q(\P)$ from the sample $\cD_N$, with $\P$ fixed but unknown and $q$ specified.
The fitted Transformer returns $\hat\beta \coloneqq \hat f^\beta(\cD_N,q)$ and $\hat V \coloneqq \hat f^V(\cD_N,q)$. %, where $\hat V/N$ can be used to estimate the covariance of $\hat\beta$. Then, for every $\alpha \in (0,1)$,   
A Wald-type confidence region with nominal $(1-\alpha)$ confidence level can be constructed as  
\begin{equation}\label{eq:confidence-region}
\hat\cC_N(\cD_N,q) \coloneqq \{\beta \in \R^{d_k}:N(\hat\beta-\beta)^\T\hat V^{-1}(\hat\beta-\beta) \leq \chi^2_{d_k,1-\alpha}\},
\end{equation}
where $\chi^2_{d_k,1-\alpha}$ is the $(1-\alpha)$ quantile of the $\chi^2_{d_k}$ distribution.

\begingroup
\setlength{\intextsep}{4pt}
\begin{figure}[!t]
\centering
\captionsetup{skip=5pt}
\resizebox{\linewidth}{!}{\input{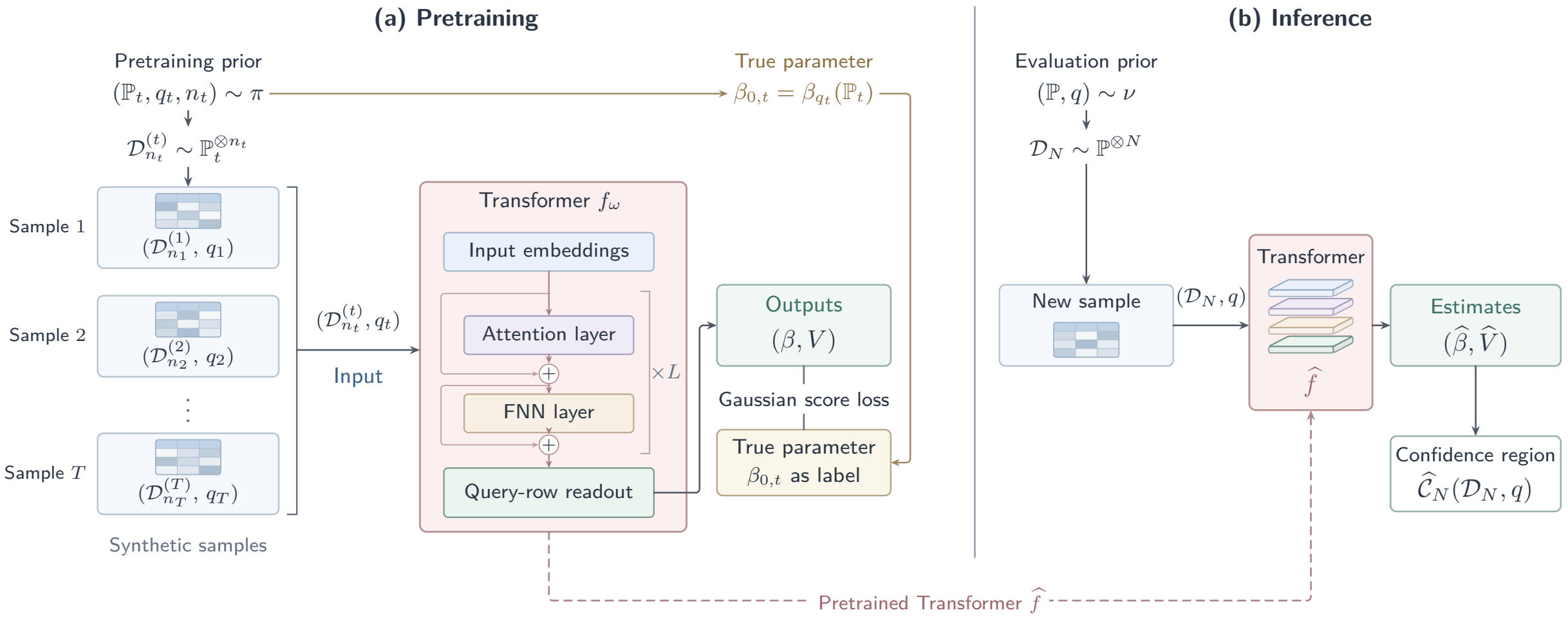}}
\caption{Graphical illustration of the pretraining and amortised inference pipeline.}
\label{fig:pretraining-inference}
\end{figure}
\endgroup

Our proposal has three main methodological advantages.
(i) The labels in \eqref{eq:score} are simply the true parameters, available for every population and query drawn in Step 1, whereas labels involving the EIF, such as in \cite{zhang2026},  must be computed for each estimand from its own nuisance functions, such as the outcome density at each quantile level for the QTE. Therefore,  our approach is better suited than EIF corrections to amortised inference across a family of estimands. 
(ii) By accommodating the parameter query as one of its inputs, a single Transformer serves the whole prespecified family of target parameters, such as the QTE at different levels, so that the cost of pretraining is shared across datasets and targets. (iii) One application of the pretrained Transformer to an evaluation sample returns $\hat\beta$ and $\hat V$ jointly, and hence the Wald region \eqref{eq:confidence-region}, avoiding the nuisance estimation, bias correction and variance estimation that classical methods repeat for each dataset.

\section{Theoretical results}\label{sec:theory}
%Our pretraining fits the Transformer to simulated datasets using only the true parameter $\beta_q(\P)$ as the label, and requires neither the efficient influence function (EIF) of $\beta_q(\P)$ nor any nuisance estimate.
In this section, we investigate whether $\hat\beta = \hat f^\beta(\cD_N, q)$ attains the semiparametric efficiency bound and $\hat\cC_N(\cD_N,q)$ has asymptotically exact coverage, for $(\P,q)$ drawn from the evaluation distribution $\nu$. %, which may reweight the pretraining prior $\pi$. 

To formalise the efficiency benchmark, let $\psi^\star_{\P,q}$ denote the EIF of $\beta_q(\P)$, with efficiency bound $V_q^\star(\P) \coloneqq \E_\P\{(\psi^\star_{\P,q})^{\otimes2}\}$, and let ${\P_Nh \coloneqq N^{-1}\sum_{i=1}^Nh(Z_i)}$. We say the fitted Transformer $\hat f$ is \emph{efficient under $\nu$} if
\begin{equation}\label{eq:goal}
\hat\beta = \beta_q(\P)+\P_N\psi^\star_{\P,q}+o_\nu(N^{-\frac{1}{2}})\quad\text{and}\quad \hat V = V_q^\star(\P)+o_\nu(1),
\end{equation}
namely that $\hat\beta$ attains the semiparametric efficiency bound and $\hat V$ estimates it consistently. Because the evaluation population and query $(\P,q)$ are drawn from $\nu$, this is efficiency \textit{in probability over populations and queries}, so that \eqref{eq:goal} holds for all but a vanishing $\nu$ proportion of $(\P,q)$.
Throughout, we assume that the support $\cN$ of $\pi_n$ consists of finitely many sample sizes, with  $n_{\min} \asymp n_{\max}$ and $\pi_n(N)$ bounded away from zero, %as when $\pi_n$ spreads over finitely many sample sizes, 
so that $n \asymp N$ for every $n \in \cN$ and all rates are stated in terms of $N$. We consider both $\pi_P$ and $\nu$ to vary with $N$. The index $\nu$ in $o_\nu$, $O_\nu$, $\overset{\nu}{\to}$ and $\overset{\nu}{\rightsquigarrow}$ refers to $\Pr_\nu$, the joint distribution of $(\cT,\P,q,\cD_N)$ under which $(\P,q) \sim \nu$ is independent of $\cT$ and $\cD_N \mid (\P,q,\cT) \sim \P^{\otimes N}$.

\subsection{The pretraining target and its efficiency}\label{sec:target-efficiency}

Because the loss \eqref{eq:score} is averaged over populations drawn from $\pi$, pretraining essentially estimates the Bayes rule under $\pi$, which minimises the posterior expected loss for each dataset and query, as the following proposition shows.

\begin{proposition}\label{prop:population-loss}
Define the posterior mean and scaled covariance by
\begin{equation}\label{eq:posterior-moment}
m_\pi(\cD_n,q) \coloneqq \E_{\pi_P}\{\beta_q(\P)\mid\cD_n\}\quad\text{and}\quad S_\pi(\cD_n,q) \coloneqq n\Cov_{\pi_P}\{\beta_q(\P)\mid\cD_n\},
\end{equation}
respectively, and let $f_\pi \coloneqq (m_\pi,S_\pi)$. For every $n \in \cN$, query $q \in \cQ$ and dataset $\cD_n \in \cZ^n$ such that $\beta_q(\P)$ has a finite posterior second moment conditional on $(\cD_n,q)$ and $S_\pi(\cD_n,q) \succ 0$,
\begin{equation}\label{eq:posterior-loss-target}
\argmin_{\beta \in \R^{d_k}, V \succ 0} \E_\pi\{\ell_n(\beta_q(\P);\beta,V)\mid\cD_n,q\} = f_\pi(\cD_n,q).
\end{equation}
\end{proposition}

The above proposition reveals that the oracle targets of pretraining are the posterior moments of $\beta_q(\P)$ given the evaluation sample $\cD_n$, %s% that the true parameter alone as the label suffices to train the network to report both an estimate and its uncertainty. This 
as a property of the Dawid--Sebastiani score \citep{dawid1999}. %, and makes the pretrained Transformer a neural Bayes estimator \citep{sainsburydale2024} for a prior on populations, with the query as an additional input. 
Whether the $m_\pi(\cD_n,q)$ is efficient depends on the prior $\pi$. Since priors of tabular foundation models are often defined implicitly by stochastic mechanisms without analytic densities, we approximate $\pi_P$ by a class of sieve priors.

Specifically,  following \cite{rivoirard2012}, let $(b_k)_{k=1}^\infty$ be a bounded orthonormal basis of the mean zero functions in $L_2(\P_0)$, where $\P_0$ is a reference distribution on $\cZ$ specified in Section \ref{app:prior} of the SM. For every $K \geq 1$, every $\P \in \cP$ %, whose log density relative to $\P_0$ is bounded as in \cite{barron1991}, 
can be written uniquely as
\[
\frac{d\P}{d\P_0} \propto \exp\left(\sum_{k=1}^K\theta_{\P,k}b_k+g_{\P, K}\right)
\]
with coefficients $\theta_\P \in \R^K$ and a mean zero remainder $g_{\P, K}$ orthogonal to $(b_{k})_{k \in [K]}$ in $L_2(\P_0)$. We define the truncated Gaussian series prior $\pi_K$ as the distribution on the sieve $\cP_K \coloneqq \{\P \in \cP : g_{\P, K} = 0\}$ by drawing $\theta_\P \sim \cN(0,\sigma^2\diag(k^{-2a})_{k \in [K]})$ truncated to $\|\theta_\P\|_1 \leq B$, for some constants $a > 1$ and $\sigma,B > 0$. %Its support, the sieve, is the exponential family with sufficient statistics $b_1,\cdots,b_K$ and natural parameter in the $\ell_1$ ball of radius $B$, and
We call $\{\pi_K\}_{K \geq 1}$ the \textit{sieve prior class}.  Just as $\pi_K$ is supported on the sieve $\cP_K$, we let $\cP_\pi$ denote the support of the simulator prior $\pi_P$, which the conditions below compare with the sieve. Our results hold under the following conditions, formalised as Assumptions \ref{cond:general-queries}--\ref{cond:local-evaluation} in Section \ref{app:assumptions} of the SM.
\paragraph*{Assumptions (informal)}
\begin{enumerate}[label=(A\arabic*),ref=(A\arabic*)]
\item\label{A1} Each $\beta_q(\P)$ is pathwise differentiable, with $\psi^\star_{\P,q}\in L_2(\P)$ continuous in $(\P,q)$, has a Lipschitz Jacobian in $\theta_\P$, and is computable as in Definition \ref{def:integral-queries} of the SM.
\item\label{A2} Eigenvalues of $V_q^\star(\P)$ are bounded away from zero and infinity, uniformly in $(\P,q)$. %, and so are those of $S_\pi$ with probability tending to one.
\item\label{A3} There exists $K = K_N$ such that $\|\theta_\P\|_1 \leq B$ for every $\P \in \cP_\pi$, $\sup_{\P \in \cP_\pi}\|g_{\P, K}\|_\infty = o(1)$ and $\delta_\pi \coloneqq \sup_{\P \in \cP_\pi}\|g_{\P, K}\|_{L_2(\P_0)} = o(N^{-\frac{1}{2}})$.
\item\label{A4} %Given $g_{\P, K}$, the coefficients $\theta_\P$ have a density $w_\pi$ relative to their distribution under $\pi_K$, with $\|\log w_\pi\|_\infty = O(1)$ and $\|\log w_\pi\|_{\mathrm{Lip}} \leq L_N = o(\sqrt N)$, uniformly in $g_{\P, K}$.
The density of $\theta_\P$ conditional on $g_{\P, K}$ relative to $\pi_K$ is uniformly bounded away from zero and infinity, and its logarithm is Lipschitz with constant $L_N = o(\sqrt N)$.
\item\label{A5} The evaluation distribution satisfies $\sup_N\chi^2(\nu\,\|\,\pi) < \infty$.
\end{enumerate}

Beyond pathwise differentiability, \ref{A1} controls the second order term of $\theta \mapsto \beta_q(\P_\theta)$ and asks for a computable representation of the queries, used by Theorem \ref{thm:TF-error} to construct Transformers and satisfied by all our examples, while \ref{A2} is the usual nondegeneracy of the efficiency bound. %, and Theorem \ref{thm:TF-error} also needs the mild computability condition of Assumption \ref{cond:general-queries}, which all our examples satisfy. 
Under \ref{A3}, the sieve approximates every population in $\cP_\pi$ with an error negligible relative to the sampling error $N^{-\frac{1}{2}}$, and we call $K$ the \textit{pretraining prior complexity} of $\pi_P$. %, since the sampling error decreases with $N$ and the approximation error with $K$. 
Condition \ref{A4} requires $\pi_P$ to reweight $\pi_K$ by a density nearly flat at the scale $N^{-\frac{1}{2}}$ of posterior fluctuations, as assumed in the Bernstein--von Mises theorem for exponential families of growing dimension \citep{ghosal2000}. Together, \ref{A3} and \ref{A4} make the posteriors under $\pi_P$ and $\pi_K$ equivalent to first order, which justifies the \textit{sieve approximation} of $\pi_P$ by $\pi_K$ and reduces the analysis of every such simulator prior to that of a Gaussian series prior whose complexity grows with $N$. Section \ref{app:assumptions} of the SM verifies \ref{A3} and \ref{A4} for the Gaussian series prior, and \ref{A3} for remainders in a Sobolev ball. Finally, as in \cite{ma2026}, \ref{A5} allows distribution shift within a $\chi^2$ ball around $\pi$, including evaluation at a single query that $\pi_q$ selects with positive probability.
Under these conditions, the efficiency of the posterior moments hinges on how fast the prior complexity $K$ grows with $N$, as the next theorem establishes. 
\begin{theorem}[Efficiency of the target]\label{thm:target}
Suppose \ref{A1} and \ref{A3}--\ref{A5} hold, and the pretraining prior complexity satisfies $K \to \infty$ and $K^{3\vee2a} = o(N)$. Then $f_\pi$ is efficient under $\nu$, that is, the posterior moments at $(\cD_N,q)$ satisfy
\begin{equation}\label{eq:target-efficiency}
m_\pi = \beta_q(\P)+\P_N\psi^\star_{\P,q}+o_\nu(N^{-\frac{1}{2}})\quad\text{and}\quad S_\pi = V_q^\star(\P)+o_\nu(1).
\end{equation}
\end{theorem}

Theorem \ref{thm:target} is a semiparametric efficiency theorem for the posterior moments, in which a common prior $\pi_P$, specified without any EIF, serves every estimand $q \in \cQ$. The posterior attains such efficiency by correcting every estimand along the same sieve scores. For the sieve prior $\pi_K$, our proof shows that 
\begin{equation}\label{eq:sieve-expansion}
m_\pi = \beta_q(\P)+\P_N\psi^K_{\P,q}+o_\nu(N^{-\frac{1}{2}})\quad\text{and}\quad S_\pi = \E_\P\{(\psi^K_{\P,q})^{\otimes2}\}+o_\nu(1),
\end{equation}
whenever $K^{3\vee2a} = o(N)$,
where $\psi^K_{\P,q}$ is the projection of $\psi^\star_{\P,q}$ onto the span of the sieve scores $\{b_k-\E_\P b_k\}_{k \in [K]}$. Hence, to the first order,  $m_\pi$ matches a one-step estimator that corrects with the projected EIF, and becomes efficient in $\cP$ as $K \to \infty$ so that sieve scores approximate every EIF. Such an implicit correction requires a prior that adds no bias of order $N^{-\frac{1}{2}}$, which the Gaussian series prior ensures in every sieve direction, since the prior standard deviations $\sigma k^{-a}$ exceed $N^{-\frac{1}{2}}$ when $K^{2a} = o(N)$. As a result, in this smooth regime, the EIF corrections of the prior or the labels \citep{ray2020,zhang2026} are not needed, and the posterior mean agrees to first order with the efficient sieve plug-in estimator \citep{shen1997}, which pretraining amortises across datasets and queries. %, whereas \cite{ray2020} correct a prior along the least favourable direction of a single functional.
%or by more when it ignores the propensity score \citep{robins1997,ritov2014},
%The simulator prior $\pi_P$ enters \eqref{eq:target-efficiency} only through the remainders.

%whereas one step estimators \citep{chernozhukov2018} attain it through a correction built from the EIF of each parameter.

\begin{remark}%[Role of the pretraining prior complexity]
\label{rem:prior-complexity}
The above theorem reveals that the prior complexity $K$ must increase with the sample size for efficiency across a general query family, since for bounded $K$ the posterior moments in \eqref{eq:sieve-expansion} are efficient only in the parametric sieve $\cP_K$. Slow growth with $K^{3\vee2a} = o(N)$ suffices to keep the posterior of $\theta_\P$ nearly normal and free of prior bias, but lets the sieve error in \ref{A3} fall below the sampling error only for joint log densities of Sobolev smoothness $s > d(3/2\vee a)$ in $d$ continuous coordinates. This smoothness requirement, while restrictive for large $d$, is the price of controlling the approximation bias of every queried estimand through the sieve error of the density alone.  By contrast, for a single estimand, a smooth EIF can offset a rough density, as only the product of their sieve errors must be $o(N^{-\frac{1}{2}})$ \citep{castillo2015}.
%(iii) More pretraining cannot compensate for a prior that is too simple. With $K$ bounded, the target is efficient for the parametric model $\cP_K$ rather than for $\cP$, so that $S_\pi$ lies below $V_q^\star(\P)$ and Wald regions can undercover when $\P$ departs from $\cP_K$.
\end{remark}

\subsection{Efficiency of the pretrained Transformer}\label{sec:efficiency}

We now study whether the pretrained Transformer $\hat f$ can accurately estimate $f_\pi$ and perform efficient inference on  evaluation samples. %Pretraining controls only an average risk over simulated datasets, whereas efficiency requires $\hat\beta$ to match $m_\pi$ more precisely than the sampling error. Our next theorem shows that a vanishing pretraining error $r_N$ suffices and is attainable at polynomial cost.
The following theorem establishes the convergence rate of the estimation error of $(\hat\beta,\hat V)$ for the posterior moment $(m_\pi, S_\pi)$.  

\begin{theorem}[Transformer estimation error]\label{thm:TF-error}
Suppose \ref{A1}--\ref{A5} hold, $K \to \infty$ with $K^{3\vee2a} = o(N)$, $\epsilon_{\mathrm{opt}} \to 0$ and $K^2\log^{2d+7}N = o(T)$. Then,  for the Transformer class $\cF_{\mathrm{TF}}$ with a single attention head and $O(K^2\log^{2d+2}N)$ parameters, 
we have 
\begin{equation}\label{eq:tracking}
\sqrt N\|\hat\beta-m_\pi\|+\|\hat V-S_\pi\|_F = O_\nu(\sqrt{r_N}),
\end{equation}
where 
\begin{equation}\label{eq:learning-rate}
r_N \coloneqq \frac{K^3}{N}+K\sqrt{\frac{\log N}{N}}+\frac{K^2}{T}\log^{2d+7}(NT)+\frac{L_N^2}{N}+N\delta_\pi^2+\epsilon_{\mathrm{opt}}.
\end{equation}
% with $c = 2d+7$ and an implied constant that does not depend on $N$, $K$ or $T$.
\end{theorem}

%Theorem \ref{thm:TF-error} shows that the pretrained Transformer inherits the efficiency of $f_\pi$ provided that $r_N$ tends to zero, because the Gaussian score penalises the error $\hat\beta-m_\pi$ at the scale of the sampling error. 
The rate $r_N$ consists of the approximation, stochastic and optimisation errors of the Transformer for learning $f_\pi$. The approximation error includes three parts, where $L_N^2/N+N\delta_\pi^2$ is the price of $\pi_P$ that departs from the sieve prior $\pi_K$, $K^3/N$ arises from a normal approximation to the posterior in dimension $K$ after one attention head pools the data into the sieve basis sample means $\{\P_Nb_k\}_{k \in [K]}$, which are sufficient for $\theta_\P$ under $\pi_K$, and $K\sqrt{\log N/N}$ from populations whose coefficients lie near the boundary of the $\ell_1$ ball truncating the prior. The term $(K^2/T)\log^{2d+7}(NT)$ is the stochastic error of fitting $O(K^2)$ Transformer parameters with $T$ pretraining samples. Therefore, once the optimisation error vanishes, order $K^2$ Transformer parameters and pretraining samples suffice up to some logarithmic factor, so that the statistical cost of pretraining grows quadratically with the prior complexity. %, which must grow for Theorem \ref{thm:target} and be larger for rougher populations.
\begin{theorem}[Efficient amortised inference]\label{thm:main}
Suppose \ref{A1}--\ref{A5} hold, $K \to \infty$ with $K^{3\vee2a} = o(N)$, and $\hat f$ satisfies \eqref{eq:erm} over $\cF_{\mathrm{TF}}$ with $r_N \to 0$. Then
\begin{equation}\label{eq:efficient-normality}
\sqrt N\{\hat\beta-\beta_q(\P)\} \mid (\P,q,\cT) \overset{\nu}{\rightsquigarrow} \cN\{0,V_q^\star(\P)\}\quad\text{and}\quad \hat V \overset{\nu}{\to} V_q^\star(\P),
\end{equation}
and, for every $\alpha \in (0,1)$, the Wald region \eqref{eq:confidence-region} satisfies
\begin{equation}\label{eq:goal-coverage}
\P\{\beta_q(\P) \in \hat\cC_N(\cD_N,q)\mid q,\cT\} \overset{\nu}{\to} 1-\alpha.
\end{equation}
\end{theorem}

Theorem \ref{thm:main} shows that the pretrained $\hat f$ returns an efficient estimate and an asymptotically exact Wald region, in probability over populations and queries drawn from $\nu$. Because the probability in \eqref{eq:goal-coverage} is taken under $\P$, the coverage holds population by population, except for a vanishing proportion in $\nu$, and hence is stronger than coverage averaged over the prior. However, classical frequentist inference usually concerns the efficiency and inference validity at a fixed population $\P_\star$, which Theorem \ref{thm:main} does not cover because a point mass at $\P_\star$ has infinite $\chi^2$ divergence from $\pi$, violating Condition \ref{A5}. The pretraining cost and efficiency guarantee in this case is established in the following theorem.  

\begin{theorem}[Inference at a fixed population]\label{thm:fixed-population}
Suppose that \ref{A1}--\ref{A4} hold with $\delta_\pi = 0$, and let $\nu = \delta_{\P_\star}\otimes\pi_q$ for a fixed population $\P_\star$ satisfying Assumption \ref{cond:fixed-population} in the SM. If $\hat f$ satisfies \eqref{eq:erm} over a Transformer class with $\exp\{O(K^2\log N)\}$ parameters, $T \geq \exp(CK^2\log N)$ for a large enough constant $C$ and $\epsilon_{\mathrm{opt}} = O(N^{-\frac{K}{2}})$, then
\begin{equation}\label{eq:single-efficiency}
\sqrt N\{\hat\beta-\beta_q(\P_\star)\} \mid (q,\cT) \overset{\nu}{\rightsquigarrow} \cN\{0,V_q^\star(\P_\star)\}\quad\text{and}\quad \hat V \overset{\nu}{\to} V_q^\star(\P_\star),
\end{equation}
and, for every $\alpha \in (0,1)$, the Wald region \eqref{eq:confidence-region} satisfies
\begin{equation}\label{eq:single-coverage}
\P_\star\{\beta_q(\P_\star) \in \hat\cC_N(\cD_N,q)\mid q,\cT\} \overset{\nu}{\to} 1-\alpha.
\end{equation}
\end{theorem}

Theorem \ref{thm:fixed-population} extends the efficiency guarantee of Theorem \ref{thm:main} to a fixed evaluation population, while the number of pretraining samples grows from order $K^2$ to $\exp(CK^2\log N)$. %The growth reflects the passage from a guarantee in probability over populations drawn from $\nu$ to a pointwise guarantee at $\P_\star$.
Hence,  the posterior moments remain efficient at $\P_\star$,  but are costly to learn, because a pointwise guarantee requires the Transformer to approximate them accurately on samples drawn from $\P_\star$ itself. Intuitively, since the pretraining loss averages over pretraining samples, it controls the Transformer error mainly on samples that are simulated often. However, samples from a fixed $\P_\star$ are rarely simulated in the pretraining data. As a result, pretraining needs exponentially many pretraining samples to ensure the efficiency at a fixed $\P_\star$.

\begin{table}[!htbp]
\centering
\caption{Sufficient sizes and optimisation error for efficient inference.}\label{tab:guarantees}
\small
\begin{tabular}{@{}lcccc@{}}
\toprule
Population & $\chi^2(\nu\|\pi)$ & Parameters & $T$ & $\epsilon_{\mathrm{opt}}$ \\
\midrule
Random, $\P \sim \nu$ & bounded & $O(K^2\log^{2d+2}N)$ & $\gg K^2\log^{2d+7}N$ & $o(1)$ \\
Fixed, $\P = \P_\star$ & infinite & $\exp\{O(K^2\log N)\}$ & $\exp(CK^2\log N)$ & $O(N^{-\frac{K}{2}})$ \\
\bottomrule
\end{tabular}
\end{table}

Table \ref{tab:guarantees} summarises the pretraining requirements under which Theorems \ref{thm:main} and \ref{thm:fixed-population} guarantee efficiency at random and fixed evaluation populations, respectively. 
For a fixed population, the lower bound in Proposition \ref{prop:fixed-lower-bound} of our SM further reveals that exponentially many pretraining samples are also necessary for guarantees covering every approximate empirical risk minimiser. 

\iffalse
\begin{remark}
In the regime of Theorem \ref{thm:main}, the posterior moments agree to first order with the explicit estimator of Definition \ref{def:reference-estimator}, which plugs the posterior mode in the exponential family sieve of \cite{barron1991} into $\beta_q$ and its Delta method covariance. Since the Transformer answers every query in one forward pass without likelihoods or derivatives, amortisation adds automation and speed at no cost in efficiency on average over populations.
\end{remark}
\fi 

\section{Numerical experiments}
\label{sec:numerical-experiments}

We study  inference tasks in causal inference, semiparametric regression, and survival analysis. Each experiment uses a task-specific instance of the pretraining and inference procedure in Section \ref{sec:method}; the corresponding estimator is labelled ``Ours'' in every figure. The task generators, preprocessing and optimisation details are deferred to Section \ref{app:experiments} of the SM. Throughout these experiments, we use two criteria to evaluate performance: the root mean squared error (RMSE), $\{R^{-1}\sum_{r=1}^R\|\hat\beta^{(r)}-\beta_q(\P_r)\|^2\}^{1/2}$, and the empirical coverage rate $R^{-1}\sum_{r=1}^R\1\{\beta_q(\P_r)\in\hat\cC_N(\cD_N^{(r)},q)\}$ of the 95\% Wald region \eqref{eq:confidence-region}, where $r$ indexes $R = 100$ independent replications, each with its own simulated population $\P_r$, evaluation sample $\cD_N^{(r)}$ and estimate $\hat\beta^{(r)}$. To save space, the results on survival analysis are deferred to Section \ref{sec:exp-mst} of the SM.

\subsection{Average treatment effect}
\label{sec:exp-ate}

The average treatment effect (ATE) is the mean change in the outcome that would result if every unit in the target population were switched from control to treatment.  Formally, let $Z=(Y,X,D)\sim\P$, where $D\in\{0,1\}$ is the treatment, $X$ contains pretreatment covariates and $Y$ is the observed outcome, and define $m_d(x) \coloneqq \E_\P(Y\mid D=d,X=x)$.  Under consistency, conditional exchangeability and overlap \citep{rosenbaum1983}, the target is identified by$\beta_{\mathrm{ATE}}(\P) =\E_\P\{m_1(X)-m_0(X)\}$. The evaluation data have the form $\cD_N=\{(Y_i,X_i,D_i)\}_{i=1}^N\overset{\mathrm{i.i.d.}}{\sim}\P$.

We compare our estimator with CausalPFN \citep{balazadeh2025}, Do-PFN \citep{robertson2025}, inverse-probability weighting \citep{hirano2003}, the S-, T- and X-learners \citep{kunzel2019}, augmented inverse-probability weighting (AIPW) \citep{bang2005}, and the difference in means \citep{neyman1990}.  The left panel of Figure~\ref{fig:ate-results}(A) shows that our estimator has the smallest RMSE at every sample size. In particular, it outperforms both AIPW, a doubly robust estimator, and CausalPFN, an amortised plug-in estimator based on outcome regression. AIPW has a larger RMSE than CausalPFN suggests that its random-forest nuisance estimates are less accurate. CausalPFN, in turn, has a larger RMSE than our estimator, which is consistent with Theorem~\ref{thm:main}, under which our estimator attains the semiparametric efficiency bound. The right panel shows that the coverage of our Wald interval remains close to the nominal 95\% level, as Theorem~\ref{thm:main} guarantees asymptotically, whereas several comparison intervals substantially undercover.

\subsection{Partially linear projection}
\label{sec:exp-plm}

\begin{figure}[t]
\centering
\begin{tikzpicture}
\node[anchor=north west,inner sep=0] (a) {\includegraphics[width=\linewidth]{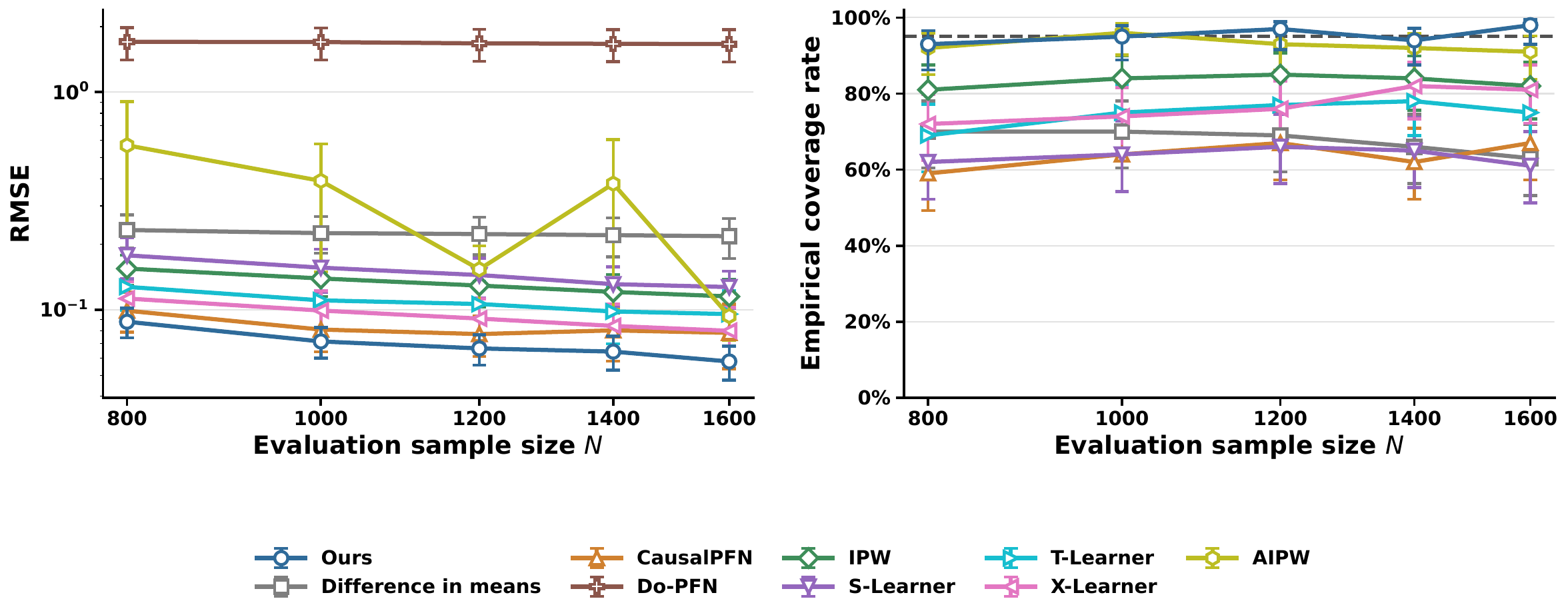}};
\node[anchor=north west,inner sep=0,font=\bfseries] at (a.north west) {(A)};
\end{tikzpicture}
\begin{tikzpicture}
\node[anchor=north west,inner sep=0] (b) {\includegraphics[width=\linewidth]{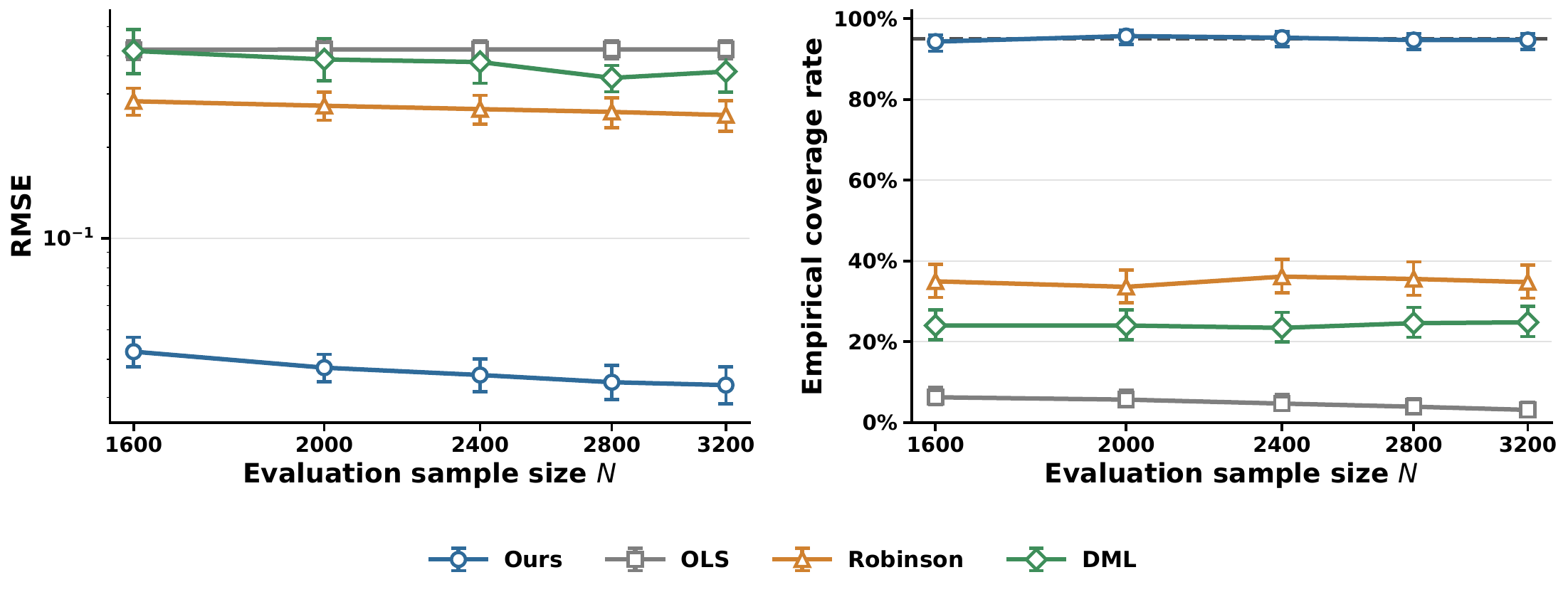}};
\node[anchor=north west,inner sep=0,font=\bfseries] at (b.north west) {(B)};
\end{tikzpicture}
\vspace*{-12pt}
\caption{Performance of ATE estimation (Subfigure A) and coefficient estimation in the partially linear projection (Subfigure B). Left: RMSE with pointwise 95\% bootstrap intervals. Right: empirical coverage with 95\% Wilson intervals.}
\label{fig:ate-results}
\end{figure}

The partially linear projection measures the linear association between a scalar regressor $W$ and an outcome $Y$ after allowing the effect of the covariates $X$ on both variables to be nonlinear.  Let $Z=(Y,W,X)\sim\P$, $\widetilde W \coloneqq W- \E_\P(W\mid X)$ and $\widetilde Y \coloneqq Y- \E_\P(Y\mid X)$. The target is $\beta_{\mathrm{PLR}}(\P) \coloneqq [\E_\P(\widetilde W^2)]^{-1}\E_\P(\widetilde W\widetilde Y)$, the one-dimensional instance of the projection in Section \ref{app:pl-example} of the SM.  If $\E_\P(Y\mid W,X)=W\beta+g(X)$, then $\beta_{\mathrm{PLR}}(\P)=\beta$.  The evaluation sample is $\cD_N=\{(Y_i,W_i,X_i)\}_{i=1}^N\overset{\mathrm{i.i.d.}}{\sim}\P$. We compare our estimator with ordinary least squares, Robinson residualisation \citep{robinson1988} and cross-fitted double machine learning \citep{chernozhukov2018}.  Figure \ref{fig:ate-results}(B) shows that our estimator again has the smallest coefficient RMSE. The coverage of its Wald interval remains very close to the nominal 95\% level.

\subsection{Quantile treatment effect}
\label{sec:exp-qte}

The quantile treatment effect (QTE) at level $q\in(0,1)$ is the difference between the $q$-quantiles of the outcome distributions that would result if every unit in the target population were switched to treatment and to control.  The observation, the identification conditions and the evaluation sample $\cD_N$ are as in Section \ref{sec:exp-ate}; only the target differs.  Let $F_{d,\P}(y) \coloneqq \E_X \P(Y\leq y\mid D=d,X=x)$, and let $\xi_{d,\P}\coloneqq F_{d,\P}^{-1}$. The target is $\beta_{\mathrm{QTE},q}(\P)\coloneqq\xi_{1,\P}(q)-\xi_{0,\P}(q)$.  Unlike the ATE, this target also involves a query: the quantile level $q$ is supplied with $\cD_N$ as $q=(\mathrm{QTE},u)$.

\begin{figure}[t]
\centering
\vspace*{-12pt}
\includegraphics[width=\linewidth]{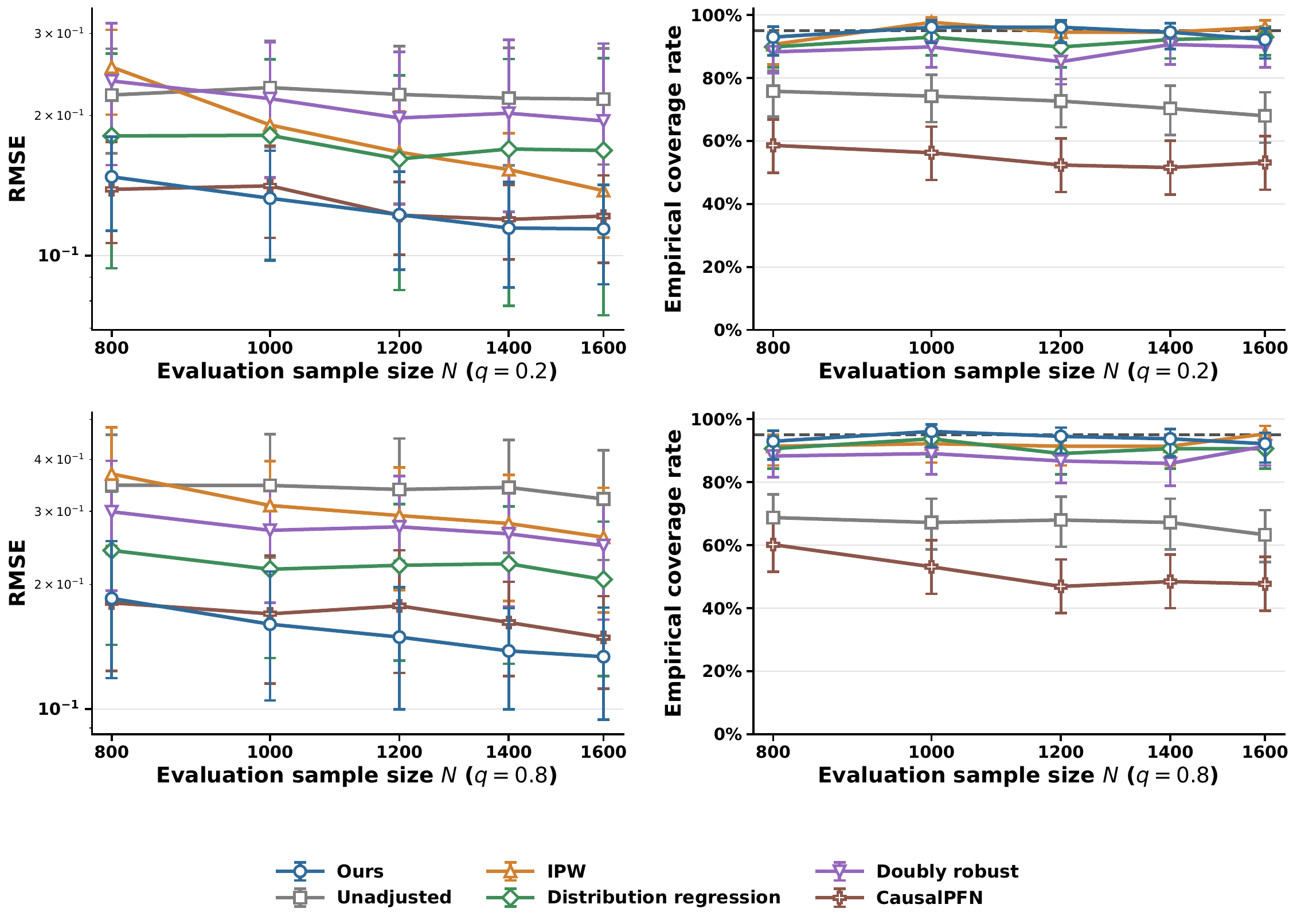}
\vspace*{-12pt}
\caption{Point and interval estimation of the quantile treatment effect on held-out datasets at quantile levels $q=0.2$ (top row) and $q=0.8$ (bottom row). }
\label{fig:qte-results}
\end{figure}

We compare our estimator with several classical baselines, including distribution regression \citep{chernozhukov2013}, inverse-probability weighting (IPW) for marginal quantile treatment effects \citep{firpo2007}, and a doubly robust (DR) estimator of marginal quantiles \citep{diaz2017}, as well as with CausalPFN \citep{balazadeh2025}. We evaluate these methods in terms of both point and interval estimation at two quantile levels, $q=0.2$ and $q=0.8$. Figure~\ref{fig:qte-results} shows that our estimator attains the smallest RMSE at most combinations of quantile level and context length, with CausalPFN its closest competitor. Moreover, its coverage rate is close to the nominal 95\% level. In contrast, the distribution regression, CausalPFN and IPW intervals sometimes fall short of the nominal level.

\iffalse
\subsection{Mean survival time}
\label{sec:exp-mst}

The mean survival time is the expected time to an event, such as death, relapse or system failure; its estimation must account for incomplete follow-up caused by right censoring.  Let $T$ and $C$ denote the latent failure and censoring times, and let $X$ contain baseline covariates.  Their joint law induces an observed-data distribution $\P$ for $Z=(Y,\Delta,X)$, where
 $Y \coloneqq \min(T,C)$ and $\Delta \coloneqq \1\{T\leq C\}$.
At the population level, the mean failure time $\beta_{\mathrm{MST}}(\P)=\E(T)$ is identified from $\P$ under conditionally independent censoring and suitable positivity and support conditions.  The evaluation data are $\cD_N=\{(Y_i,\Delta_i,X_i)\}_{i=1}^N$. 

We compare our estimator with the complete-case mean $\sum_{i=1}^N\Delta_iY_i/\sum_{i=1}^N\Delta_i$, the Kaplan--Meier estimator \citep{kaplan1958}, an inverse-probability-of-censoring-weighted estimator \citep{satten2001}, and plug-in estimators based on the Cox model \citep{cox1972} and a random survival forest \citep{ishwaran2008}.  In Figure \ref{fig:mst-results}, our estimator has the lowest RMSE throughout, and its decrease broadly follows the $N^{-1/2}$ reference trend.  Its Wald coverage remains close to the nominal 95\% level.  The rate and coverage patterns are qualitatively consistent with Theorem \ref{thm:main}.

\begin{figure}[t]
\centering
\includegraphics[width=\linewidth]{figure/mst_results.pdf}
\caption{Performance on mean-survival-time estimation.}
\label{fig:mst-results}
\end{figure}

\fi

\subsection{Computational analysis}\label{sec:computation-analysis}
\begin{figure}[t]
\vspace*{-12pt}
\includegraphics[width=\linewidth]{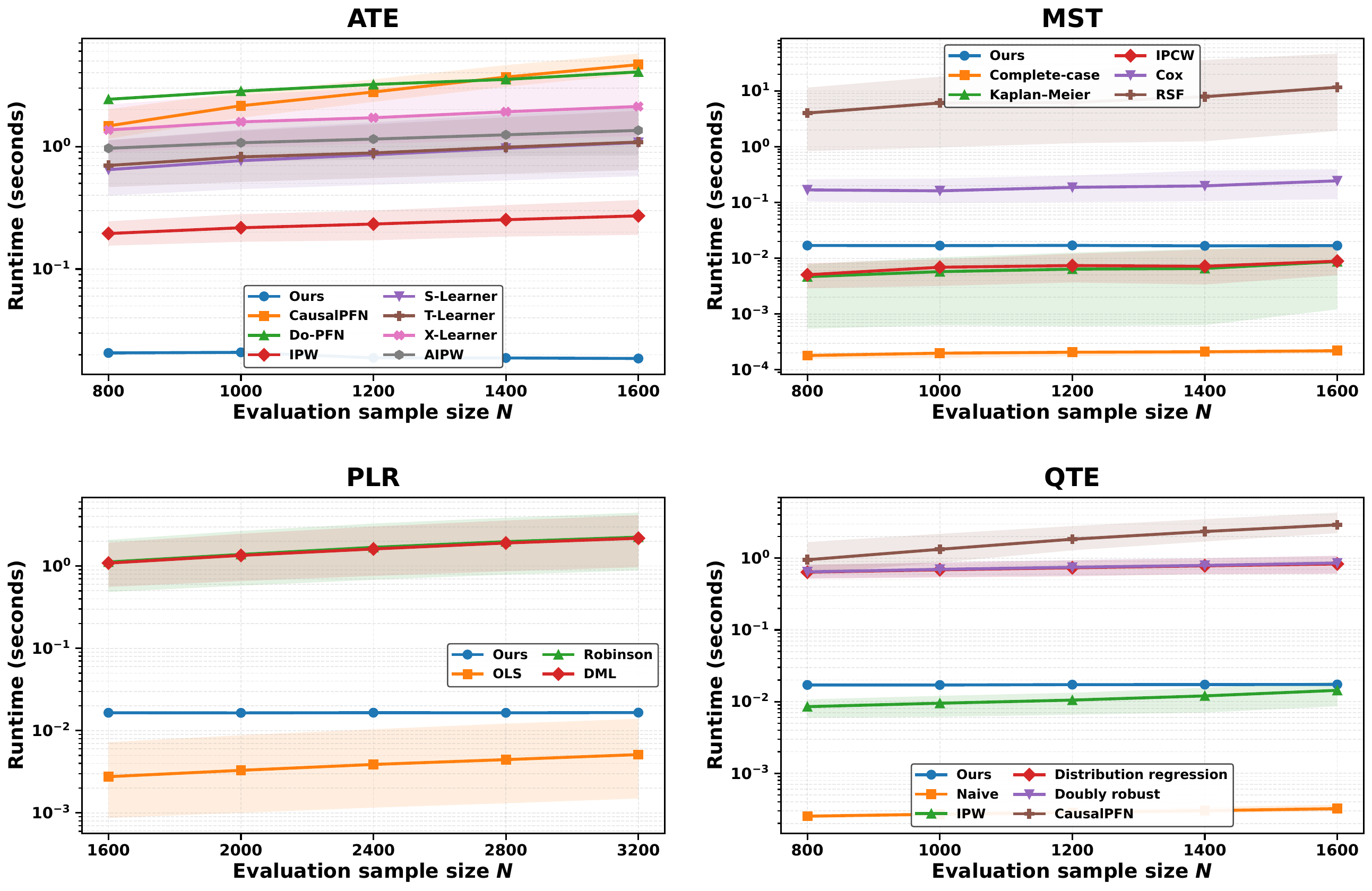}
\caption{Runtime comparison for the four tasks considered in numerical experiments. The $x$-axis is the context length and the $y$-axis is the runtime in seconds ($\log_{10}$-scaled).}\label{fig:runtime}
\end{figure}

We now study the computational cost of our procedure, together with that of the baselines in Sections \ref{sec:exp-ate}--\ref{sec:exp-qte}, and Section \ref{sec:exp-mst} of the SM. Figure \ref{fig:runtime} shows that our method takes about $2\times10^{-2}$ seconds per dataset at every context length, and that it is faster than the other machine-learning-based methods. 

\subsection{Analysis of the Infant Health and Development Program}\label{sec:exp-ihdp}
The Infant Health and Development Program (IHDP) was an eight-site randomised trial that evaluated whether comprehensive early intervention, such as home visits and parent support groups, improved the cognitive and health outcomes of premature infants \citep{ihdp1990}.  \cite{hill2011} turned the trial into a benchmark for observational studies by simulating the outcomes from known nonlinear response surfaces of the real covariates. Each realisation of the benchmark contains the same $N=747$ children, of whom 139 are treated, together with 25 pretreatment covariates describing the child and the mother. Because the potential outcomes are simulated from functions of $X$ alone, the treatment is unconfounded conditional on $X$ by construction, and the conditional means $m_{0}$ and $m_{1}$ are known. Following \cite{shalit2017}, we compute the true effect from the noiseless potential outcomes, $\tau \coloneqq \frac{1}{N}\sum_{i=1}^N\{m_{1}(X_i)-m_{0}(X_i)\}$. We measure the accuracy of our estimator and of the baselines of Section \ref{sec:exp-ate} by their absolute estimation error. Our estimator uses the checkpoint pretrained for Section \ref{sec:exp-ate} without any modification, so this is a distribution-shift setting: neither the covariates nor the response surfaces are generated by the simulator prior.

Table \ref{tab:ihdp} shows that our estimator attains the smallest mean absolute error, below that of CausalPFN, another amortised method. CausalPFN is the best baseline, which may reflect more accurate estimates of the conditional means $m_0$ and $m_1$ than those of the other baselines.
\begin{table}[t]
\centering
\caption{Mean absolute error of ATE estimation together with standard error on the IHDP data.}
\label{tab:ihdp}
\small
\setlength{\tabcolsep}{3pt}
\begin{tabular}{@{}ccccccccc@{}}
\toprule
AIPW & CausalPFN &
\begin{tabular}[c]{@{}c@{}}Difference\\in means\end{tabular} &
Do-PFN & IPW & S-Learner & T-Learner & X-Learner & Ours \\
\midrule
0.094 & 0.069 & 0.171 & 1.256 & 0.095 & 0.186 & 0.080 & 0.122 &
\textbf{0.055} \\
$\pm$ 0.009 & $\pm$ 0.007 & $\pm$ 0.024 & $\pm$ 0.098 &
$\pm$ 0.010 & $\pm$ 0.045 & $\pm$ 0.008 & $\pm$ 0.022 &
$\pm$ \textbf{0.006} \\
\bottomrule
\end{tabular}
\end{table}

\section{Discussion}\label{sec:discussion}
In this study, we show that a Transformer pretrained once on simulated data, with true parameters as the only label, can deliver efficient estimates and valid inference for a family of semiparametric estimands, in probability over evaluation populations and queries, even under distribution shift. The posterior under a locally flat prior of slowly growing complexity is efficient for every estimand, and the Transformer recovers the posterior moments as an amortised sieve estimator at a statistical cost quadratic in the prior complexity, up to logarithmic factors whose powers grow with the data dimension. %, since rescaling the Gaussian series prior by any fixed factor other than one already falls outside them once its complexity is large. 

Theoretically, our guarantees need a pretraining prior that grows with the sample size, whereas foundation models are usually pretrained once, and whether they can recover the efficient posteriors of suitable priors that do not grow seems to be an open question.  The one-step correction method based on the EIF restores inference validity \citep{melnychuk2026frequentist}, but for a family of estimands they need the influence function and nuisance estimates of every estimand on every dataset. Consequently, they forgo amortisation and need not cohere across estimands. It is of our future interest to explore methods that attain valid inference for a whole family at lower cost than the EIF correction approach, while under weaker prior requirements than this study. 

It is also worth remarking that the semiparametric efficiency is an asymptotic property, yet it is found that tabular foundation models outperform tuned predictors on small datasets \citep{hollmann2025}. In our experiments, the pretrained Transformer also achieved smaller RMSEs than asymptotically efficient doubly robust and double machine learning estimators. We conjecture it is because their nuisance estimates are inaccurate in small samples, whereas the simulator priors in our experiments, unlike the locally flat priors of our theory, carry information about the nuisance functions. Characterising whether and when pretrained models outperform asymptotically optimal statistical estimators in finite samples is an interesting theoretical question, which may call for new optimality criteria to explain such a phenomenon. 

\singlespacing

\begingroup
\normalsize

\putbib
%\bibliography{ref}
\endgroup
\end{bibunit}

\clearpage
\appendix
\begin{bibunit}
\renewcommand{\thesection}{\Alph{section}}
\renewcommand{\thesubsection}{\thesection.\arabic{subsection}}
\renewcommand{\theHsection}{appendix.\arabic{section}}
\numberwithin{theorem}{section}
\numberwithin{proposition}{section}
\numberwithin{corollary}{section}
\numberwithin{lemma}{section}
\numberwithin{definition}{section}
\numberwithin{remark}{section}

\singlespacing

\begin{center}
{\Large Supplementary material for\\
``Can pretrained Transformers deliver semiparametrically \\ efficient inference?''\par}
\end{center}

This supplementary material is organised as follows. Section \ref{app:related} reviews related work. Section \ref{app:implementation} specifies the sieve prior of Section \ref{sec:target-efficiency} and the Transformer architecture of Section \ref{sec:method}, Section \ref{app:queries} presents examples of the queried estimands of Section \ref{sec:setup}, and Section \ref{app:assumptions} formalises the assumptions of Section \ref{sec:target-efficiency}. Sections \ref{app:proofs}, \ref{app:proofs-efficiency} and \ref{app:proofs-examples} prove the results of Sections \ref{sec:target-efficiency}, \ref{sec:efficiency} and \ref{app:queries}, respectively. Section \ref{app:experiments} details the experiments of Section \ref{sec:numerical-experiments}.

\section{Related work}\label{app:related}

This section reviews the three lines of research on which our work draws, compares our results with the closest works and describes the techniques on which our proofs build.

\subsection{Tabular foundation models and  statistical applications}\label{app:related-amortised}

\paragraph{Development of tabular foundation models.}
Amortised inference maps a dataset to an estimate or posterior through one forward pass of a network trained on simulations \citep{zammitmangion2025,sainsburydale2024}. Prior-data fitted networks (PFNs) apply this idea to the posterior predictive distribution \citep{muller2022,muller2025position}, and TabPFN and its successors made them tabular foundation models \citep{hollmann2023,hollmann2025,grinsztajn2025}. Later models scale to larger tables through more efficient architectures \citep{qu2025,qu2026,zeng2025tabflex}, or change the pretraining data, from a mixture of synthetic priors to real tables \citep{zhang2025mitra,ma2025tabdpt,garg2025realtabpfn}. Others exploit column names and text \citep{kim2024carte,gardner2024tabula} or model the joint distribution of all variables \citep{zhang2025limix}. On the TabArena benchmark \citep{erickson2025tabarena},  several tabular foundation models outperform many machine learning methods, such as tree-based and
boosting algorithms, on regression and classification problems for tabular data.

%Foundation models for time series follow the same route. TimesFM, Chronos, Moirai and Lag-Llama pretrain Transformers on large collections of time series and forecast new series without further training \citep{das2024timesfm,ansari2024chronos,woo2024moirai,rasul2023lagllama}, and MOMENT extends such pretraining to general-purpose time series analysis \citep{goswami2024moment}. Closer to PFNs, ForecastPFN is trained only on synthetic series \citep{dooley2023}, and TabPFN-TS forecasts by applying TabPFN v2 to temporal features, without time series pretraining \citep{hoo2025}. Chronos-2 extends in-context learning to multivariate series and covariates \citep{ansari2025chronos2}. On the statistical side, the Transformer architecture makes the variance of PFN predictions vanish as the context grows, but not their bias \citep{nagler2023}, and pretrained PFNs mostly recover their Bayesian target \citep{hulskamp2026}. These works concern prediction, whereas we ask whether the pretraining target yields efficient inference about functionals. %Pretraining for the posterior mean and covariance of a queried functional makes efficiency a property of one target, which Theorem \ref{thm:target} characterises.

\paragraph{Statistical applications of PFNs.}
Many PFN models can be naturally used for statistical analysis, since they can return posteriors of model parameters in context \citep{reuter2025,mittal2025amortized} and conditional densities for simulation-based inference \citep{vetter2025}. 
Pretrained Transformers have since been adapted to many statistical tasks, including Bayesian optimisation \citep{muller2023}, clustering \citep{zhao2026}, variable selection \citep{zhu2026amortized} and contextual bandits \citep{tan2026}. 
%Other PFNs cluster data with an unknown number of clusters \citep{zhao2026}, select variables through posterior inclusion probabilities \citep{zhu2026amortized}, forecast time series \citep{dooley2023} and guide Bayesian optimisation \citep{muller2023} and Thompson sampling in contextual bandits \citep{tan2026}. Each network addresses one task, whereas ours takes the query as an input to serve a whole family of estimands. 
Among various statistical applications of the PFNs, those on causal effect estimation are closest to our semiparametric framework and are reviewed in
Section \ref{app:related-closest}.

\paragraph{Uncertainty quantification for PFNs.}
Since amortised posteriors can be overconfident \citep{hermans2022}, validity for inference/uncertainty quantification is usually restored by conformal calibration \citep{patel2024} or Neyman construction \citep{dalmasso2024}, which WALDO \citep{masserano2023} applies to the Wald statistic similar to \eqref{eq:confidence-region}. For tabular foundation models, martingale posteriors \citep{fong2023} quantify the uncertainty of PFN estimates by imputing future observations \citep{nagler2026,ng2026}, and a predictive central limit theorem yields Gaussian credible bands \citep{fortini2026}. TabCon outputs intervals in one forward pass and calibrates them by reinforcement learning \citep{ye2026can}, and TabCUQ perturbs the outcomes in the context and applies a wild bootstrap \citep{shi2026tabcuq}. By Theorem \ref{thm:main}, our covariance output estimates the efficiency bound itself, so that the Wald-type confidence region, without calibration, is centred at an efficient estimator and attains asymptotically exact coverage in probability over populations, that is, conditional on each population for all but a vanishing proportion of them. This is stronger than the expected coverage over the prior used to assess amortised posteriors \citep{hermans2022}, much as conditional coverage is stronger than marginal coverage in conformal prediction \citep{lei2014,barber2021}. %The two notions can differ sharply in nonparametric regression, where the frequentist coverage of credible regions falls infinitely often below any positive level for almost every truth under the prior \citep{cox1993}.

\subsection{Semiparametric estimation and inference}\label{app:related-semiparametric}

\paragraph{One step correction with the EIF.}
For an estimand $\beta_q(\P)$ with EIF $\psi^\star_{\P,q}$, the one step estimator $\beta_q(\hat\P)+\P_N\psi^\star_{\hat\P,q}$ removes the first order bias $-\P\psi^\star_{\hat\P,q}$ of a plug-in estimate by the empirical mean of the estimated EIF \citep{bickel1993,tsiatis2006,hines2022}. Estimating equation and targeted maximum likelihood estimators attain the same efficiency by solving the estimated EIF equation \citep{robins1994,bang2005,vanderlaan2006}, and with sample splitting, flexible nuisance estimates suffice when the product of their errors is $o(N^{-\frac{1}{2}})$ \citep{chernozhukov2018}. Recent work automates the EIF by numerical or automatic differentiation \citep{carone2019,jordan2022,luedtke2026}, Monte Carlo approximation \citep{agrawal2024} or learned Riesz representers \citep{chernozhukov2022auto,chernozhukov2022riesz}, but still fits nuisance functions and a correction for every dataset and every estimand. For a family of estimands, such as quantile treatment effects at all levels, each member needs its own EIF and nuisance functions. Separate corrections also need not cohere, so that estimated quantile curves can cross \citep{chernozhukov2010} and survival curves need not be monotone \citep{cai2020}, unless one distribution is updated for all estimands at once \citep{vanderlaan2016}.  %Our network uses no influence function, since its only label is the parameter and one forward pass serves any query, at the price of the condition on the whole joint density in Remark \ref{rem:prior-complexity}.

\paragraph{Semiparametric BvM theorems.}
These theorems show that the posterior of a smooth functional is asymptotically normal and centred at an efficient estimator \citep{bickel2012,castillo2015}, but a nonparametric prior can bias it at the order $N^{-\frac{1}{2}}$ \citep{rivoirard2012}. The remedies tailor the prior to the functional \citep{ray2020,breunig2025}, correct the posterior with the EIF \citep{yiu2025}, or use priors on parametric models of suitably growing dimension \citep{ghosal2000,rivoirard2012}, and \cite{rosa2026stein} locates the bias through nonasymptotic bounds. Sieve plug-in estimators, such as sieve maximum likelihood estimators, are also efficient for smooth functionals under undersmoothing, without any EIF correction \citep{newey1994,shen1997,chen2007}, and the posterior moments under the sieve prior agree with the plug-in of Definition \ref{def:reference-estimator} to first order, since both are efficient at every smooth population by Proposition \ref{prop:spectral-truth}. Our Theorem \ref{thm:target} belongs to the third class, whose priors need no EIF because they are flat along every sieve direction at the scale $N^{-\frac{1}{2}}$, and adapts it to amortisation. %It concerns the posterior mean and covariance rather than the whole posterior. %covers every query under one prior and simulator priors that are only close to a sieve prior, and holds in probability over populations rather than at each fixed population.

\subsection{Statistical theory of in-context learning}\label{app:related-icl}

Theory for in-context learning views a pretrained Transformer either as an approximation to a Bayes rule or as an algorithm run on the context, and our results draw on both views, the first for the target of pretraining and the second for learning it

\paragraph{The Bayesian view.}
Since pretraining tasks are drawn from a prior, the pretraining loss is minimised by the Bayes rule, and a well-trained Transformer performs approximate Bayesian inference in context \citep{xie2022,muller2022}. Theory in this view relates attention to Bayesian model averaging \citep{zhang2025bma}, bounds the gap to the Bayes rule \citep{wakayama2026} and constructs Transformers for Gaussian process posterior predictive distributions \citep{kang2026}, while experiments indicate that task diversity decides which prior the network follows \citep{raventos2023}.  For nonparametric regression, \cite{ma2026} takes the nonparametric Bayesian view and shows that pretraining on a prior mixing over smoothness levels attains adaptive optimal rates over test distributions within a $\chi^2$ ball around the prior, a distribution shift similar to our \ref{A5}.  Closest to our setting, \cite{li2026experimenters} imitate Bayesian updating under Gaussian series priors by attention to estimate the average treatment effect in adaptive experiments. %Our target, the posterior mean and covariance of the queried functional, is also a Bayes rule by Proposition \ref{prop:population-loss}, but whereas these works study prediction risk or a single estimand, Theorem \ref{thm:target} asks when it is efficient for a whole family of functionals.

\paragraph{The algorithmic view.}
This view emerged from the finding that Transformers learn linear functions in context \citep{garg2022} and from attention layers constructed to perform gradient descent and ridge regression \citep{akyurek2023,vonoswald2023}. \cite{bai2023} develop it into a systematic statistical theory, showing that Transformers can implement least squares, ridge regression, the Lasso and generalised linear models in context with near-optimal prediction risk. Their constructions rely on in-context gradient descent and can be learned from polynomially many pretraining sequences. They further show that a single Transformer can perform in-context algorithm selection, adapting its procedure to each input, much as a statistician chooses a method for the data at hand. Subsequent work bounds the number of pretraining tasks needed for linear regression \citep{wu2024} and shows that pretrained Transformers attain minimax rates in nonparametric regression \citep{kim2024,ching2026}.%Our  Theorem \ref{thm:TF-error} follows this route, but because efficiency requires the first order expansion \eqref{eq:goal}, the theorem bounds the distance from the Transformer's output to the posterior moments at a scale finer than the sampling error, rather than prediction error at the optimal rate.

\subsection{Pretrained networks for semiparametric inference}\label{app:related-closest}

\paragraph{Pretrained causal estimators.}
CausalPFN \citep{balazadeh2025} learns posteriors of conditional potential outcome means, and hence of the ATE and conditional average treatment effects (CATEs), from simulated causal data. Its posterior mean is consistent for almost every data generating process under the prior exactly when the prior identifies these means, and its credible intervals are calibrated. Do-PFN \citep{robertson2025}, CausalFM \citep{ma2026causalfm} and CInA \citep{zhang2023cina} pretrain related networks for interventional distributions, for front door and instrumental variable designs and for covariate balancing weights, respectively. \cite{bynum2025} label simulated datasets by their true effects, as we do, and decompose the mean squared error of the trained network over the prior. Pretrained models also serve inside influence function estimators, as nuisance learners for doubly robust estimators \citep{weckstein2026} and for the natural indirect effect (NIE) in MediPFN \citep{bo2026medipfn}, or as predictors in prediction-powered inference \citep{angelopoulos2023,ji2025surrogates}. The guarantees for networks trained without the EIF concern consistency or accuracy, and the intervals of causal PFNs can undercover \citep{mourao2026}, whereas influence function estimators keep their guarantees but must be refitted for every dataset and estimand.

\paragraph{EIF corrections of pretrained networks.}
The three works closest to ours use the EIF to make pretrained networks efficient. \cite{melnychuk2026frequentist} show that PFN estimates of the ATE can retain a confounding bias induced by the prior. They remove it after pretraining by a one step correction of the posterior on each evaluation dataset, which combines the EIF with nuisance posteriors recovered from the network. \cite{melnychuk2026debiased} instead correct, during pretraining, either the labels through a Bayesian bootstrap of the EIF or the prior along a least favourable direction, and analyse an idealised network that returns the exact posterior. \cite{zhang2026} labels each simulated dataset by the EIF corrected estimate $\beta_q(\P)+\P_n\psi^\star_{\P,q}$ of the ATE and learns its variance through a separate output. Its error bound for the trained network holds on average over the prior and transfers, for instance, to a population $\P_\star$ with positive prior mass at a cost proportional to $1/\pi_P(\{\P_\star\})$. Each of these corrections is constructed for one estimand, so a family of estimands needs the EIF and nuisance values of every member at every simulated observation, with the difficulties described in Section \ref{app:related-semiparametric}.

\paragraph{Our position.}
Different from the aforementioned studies, our approach aims to attain efficiency for a whole family of estimands with the true parameter as the only label, without any EIF or nuisance function. The reason is that the posterior targeted by pretraining already corrects every estimand along the same sieve scores. Under a locally flat prior of slowly growing complexity, this implicit correction makes the posterior moments efficient in probability over populations and queries, reflected in Theorem \ref{thm:target}. The pretrained Transformer inherits their efficiency once the number of pretraining samples is at least quadratic in the prior complexity, up to logarithmic factors shown in our Theorem \ref{thm:main}.

The price of this versatility is smoothness of the whole joint density, whereas, for a single estimand, a smooth EIF can offset a rough density, as discussed in Remark \ref{rem:prior-complexity}. At a fixed population, the posterior moments remain efficient, but learning them becomes costly. Exponentially many pretraining samples suffice by Theorem \ref{thm:fixed-population}, and Proposition \ref{prop:fixed-lower-bound} shows that they are also necessary for a guarantee that covers every approximate empirical risk minimiser.

\begin{table}[!htbp]
\centering
\caption{Pretrained Transformer for causal and semiparametric inference. The columns indicate whether a method needs no efficient influence function, returns its output in one forward pass, supports inference, establishes efficiency, bounds its pretraining sample complexity  and holds at a fixed population, with $\checkmark$, $\times$ and $\circ$ indicating yes, no and partly.}\label{tab:related}
\small
\setlength{\tabcolsep}{4.5pt}
\begin{tabular}{@{}l|l|c|c|c|c|c|c|@{}}
\toprule
 & Estimands & \begin{tabular}[b]{@{}c@{}}No\\EIF\end{tabular} & \begin{tabular}[b]{@{}c@{}}One\\pass\end{tabular} & Inference & Efficiency & \begin{tabular}[b]{@{}c@{}}Pretraining\\complexity\end{tabular} & \begin{tabular}[b]{@{}c@{}}Fixed\\$\P$\end{tabular} \\
\midrule
\cite{balazadeh2025} & ATE, CATE & $\checkmark$ & $\checkmark$ & $\checkmark$ & $\times$ & $\times$ & $\circ$ \\
\hline
\cite{bynum2025} & ATE, CATE & $\checkmark$ & $\checkmark$ & $\times$ & $\times$ & $\times$ & $\times$ \\
\hline
\cite{bo2026medipfn} & NIE & $\times$ & $\times$ & $\checkmark$ & $\times$ & $\times$ & $\times$ \\
\hline
\cite{melnychuk2026frequentist} & ATE & $\times$ & $\times$ & $\checkmark$ & $\checkmark$ & $\times$ & $\checkmark$ \\
\hline
\cite{melnychuk2026debiased} & one estimand & $\times$ & $\checkmark$ & $\checkmark$ & $\checkmark$ & $\times$ & $\checkmark$ \\
\hline
\cite{zhang2026} & ATE & $\times$ & $\checkmark$ & $\checkmark$ & $\checkmark$ & $\checkmark$ & $\circ$ \\
\hline
This paper & query family & $\checkmark$ & $\checkmark$ & $\checkmark$ & $\checkmark$ & $\checkmark$ & $\circ$ \\
\bottomrule
\end{tabular}
\end{table}

\subsection{Techniques}\label{app:related-techniques}

Our analysis employs several different existing theoretical techniques from the statistics and learning theory literature to establish our efficiency, approximation and generalisation guarantees. The analysis of the posterior target uses the exponential series of \cite{rivoirard2012} with the Gaussian coefficients of \cite{castillo2015} and a logarithmic Sobolev inequality for the coefficient posterior \citep{ledoux1999}. We show that Transformer network averages the data by its attention layer that is similar in \cite{zhang2026}, unrolls proximal gradient and projected Richardson steps as motivated by \cite{bai2023} and \cite{kang2026}, and approximates smooth maps by GELU networks and Chebyshev series \citep{yakovlev2025,trefethen2013}. The generalisation bound rests on an exponential inequality for fast rates \citep{vanerven2015,grunwald2020}, and the transfer to evaluation distributions on the $\chi^2$ argument of \cite{ma2026}.

\section{Prior and Transformer architecture}\label{app:implementation}

This section specifies the Transformer architecture and the sieve prior.

\subsection{Transformer architecture}\label{app:architecture}

One parameter vector $\omega \in \Omega$, with $\|\omega\|_\infty \leq W$, serves all queries $q \in \cQ$ and sample sizes $N \in \cN$.
For $q = (k,u) \in \cQ$ and $j \in \cK$, let $u_j(q) \coloneqq u$ if $j = k$ and $u_j(q) \coloneqq \bar u_j$ if $j \ne k$, where $\bar u_j \in \cU_j$ are given points. We encode the query family, its argument and the scaled context size by
\[
\xi(q,N) \coloneqq \left(1,(1\{j = k\})_{j \in \cK},(u_j(q)^\T)_{j \in \cK},\frac{N}{n_{\max}}\right)^\T \in \R^{d_\xi},
\]
where $d_\xi \coloneqq 2+|\cK|+\sum_{j \in \cK}\ell_j$ and $n_{\max} = \max\cN$.
%Let $d_e \geq 3$ and $W \geq 1$.

\begin{definition}[Input embedding]
For $N \in \cN$, each map $\mathsf{Emb}_\omega \in \cF_{\mathrm{Emb}}(d_e,W)$ sends $(\cD_N,q) \in \cZ^N\times\cQ$, with $\cD_N = \{Z_i\}_{i = 1}^N$, to
\[
\mathsf{Emb}_\omega(\cD_N,q) \coloneqq \begin{pmatrix}(1,Z_1^\T)E_o&1&0\\\vdots&\vdots&\vdots\\(1,Z_N^\T)E_o&1&0\\\xi(q,N)^\T E_q&0&1\end{pmatrix} \in \R^{(N+1)\times d_e},
\]
where $E_o \in [-W,W]^{(d_z+1)\times(d_e-2)}$ and $E_q \in [-W,W]^{d_\xi\times(d_e-2)}$ are components of $\omega$ learned during pretraining.
\end{definition}

For $S \in [-\infty,\infty)^{(N+1)\times(N+1)}$, let $\softmax(S)_{ij} \coloneqq e^{S_{ij}}/\sum_{r = 1}^{N+1}e^{S_{ir}}$, for $i,j \in [N+1]$, denote the row-wise softmax, with $e^{-\infty} = 0$.

\begin{definition}[Masked attention]
For $J \geq 1$ heads and masks $\mathsf M = (\mathsf M_1,\cdots,\mathsf M_J)$, each map $\mathsf{Attn}_\omega \in \cF_{\mathrm{Attn}}(J,d_e,W;\mathsf M)$ acts on $H \in \R^{(N+1)\times d_e}$ by
\[
\mathsf{Attn}_\omega(H) \coloneqq H+\sum_{j = 1}^J\softmax\left\{\frac{H\mathsf Q_j(H\mathsf K_j)^\T}{\sqrt{d_e}}+\mathsf M_j\right\}H\mathsf V_j,
\]
where $\mathsf Q_j,\mathsf K_j,\mathsf V_j \in [-W,W]^{d_e\times d_e}$, for $j \in [J]$, are components of $\omega$.
\end{definition}

The masks $\mathsf M_j = \mathsf M_j^{(N)} \in \{0,-\infty\}^{(N+1)\times(N+1)}$, given by one rule for all $N \in \cN$, have an unmasked entry in every row and are invariant under relabelling of observations, that is,
\[
\max_{r \in [N+1]}\mathsf M_j(i,r) = 0\quad\text{and}\quad \mathsf M_j(\rho(i),\rho(r)) = \mathsf M_j(i,r),
\]
for all $j \in [J]$, $i,r \in [N+1]$ and permutations $\rho$ of $[N+1]$ with $\rho(N+1) = N+1$.

\begin{definition}[Feedforward layers and blocks]
For width $d_f \geq 1$, each map $\mathsf{FFN}_\omega \in \cF_{\mathrm{FFN}}(d_e,d_f,W)$ acts on $H \in \R^{(N+1)\times d_e}$ by
\[
\mathsf{FFN}_\omega(H) \coloneqq H+\GELU(H\mathsf U_1+\mathbf1_{N+1}\mathsf c_1^\T)\mathsf U_2+\mathbf1_{N+1}\mathsf c_2^\T,
\]
where $\mathsf U_1 \in [-W,W]^{d_e\times d_f}$, $\mathsf U_2 \in [-W,W]^{d_f\times d_e}$, $\mathsf c_1 \in [-W,W]^{d_f}$ and $\mathsf c_2 \in [-W,W]^{d_e}$ are components of $\omega$, $\mathbf1_{N+1} \coloneqq (1,\cdots,1)^\T \in \R^{N+1}$, and the activation $\GELU(x) \coloneqq x\Phi(x)$ acts entrywise, with $\Phi$ the standard normal distribution function. The blocks form the class
\[
\cF_{\mathrm{Block}}(J,d_e,d_f,W;\mathsf M) \coloneqq \left\{F\circ A:A \in \cF_{\mathrm{Attn}}(J,d_e,W;\mathsf M),\,F \in \cF_{\mathrm{FFN}}(d_e,d_f,W)\right\}.
\]
\end{definition}

Let $d_0 \coloneqq \max_{k \in \cK}d_k$, where $\beta_{(k,u)}(\P) \in \R^{d_k}$, let $P_k \in \{0,1\}^{d_k\times d_0}$ with $P_kP_k^\T = I_{d_k}$, for $k \in \cK$, be coordinate projections, and let $P_q \coloneqq P_k$ for $q = (k,u) \in \cQ$.
For constants $0 < v_- < v_+ < \infty$, we define the spectral clipping of $A = A^\T = U\diag(\lambda_1,\cdots,\lambda_r)U^\T \in \R^{r\times r}$ with $U^\T U = I_r$ by
\[
\clip_{[v_-,v_+]}(A) \coloneqq U\diag\{(\lambda_j\vee v_-)\wedge v_+:j \in [r]\}U^\T.
\]
For the point output, let $C_\beta \in (0,\infty)$ be a constant with $\|\beta_q(\P)\| \leq C_\beta$ for all $\P \in \cP$ and $q \in \cQ$, which exists by \eqref{eq:query-lipschitz} and Assumption \ref{cond:general-queries}(i), and let $\clip_{C_\beta}(\beta) \coloneqq C_\beta\beta/\max\{C_\beta,\|\beta\|\}$ denote the Euclidean projection of $\beta \in \R^{d_k}$ onto $\left\{\beta' \in \R^{d_k}:\|\beta'\| \leq C_\beta\right\}$.

\begin{definition}[Parameter and covariance readout]\label{def:transformer-output}
Each map $\mathsf{Read}_\omega \in \cF_{\mathrm{Read}}(d_e,W)$ sends $(H,q) \in \R^{(N+1)\times d_e}\times\cQ$ to
\[
\mathsf{Read}_\omega(H,q) \coloneqq \left(\clip_{C_\beta}\{P_qa_\beta(z)\},\,\clip_{[v_-,v_+]}\left\{\frac{C_\omega+C_\omega^\T}{2}\right\}\right),
\]
where $z \coloneqq H_{N+1,:}^\T \in \R^{d_e}$, $C_\omega \coloneqq P_qa_C(z)P_q^\T$, and $a_\beta:\R^{d_e} \to \R^{d_0}$ and $a_C:\R^{d_e} \to \R^{d_0\times d_0}$ are affine with coefficients in $[-W,W]$.
\end{definition}

The point projection and the spectral clipping are nonexpansive, that is, $\|\clip_{C_\beta}(\beta)-\clip_{C_\beta}(\beta')\| \leq \|\beta-\beta'\|$ and $\|\clip_{[v_-,v_+]}(A)-\clip_{[v_-,v_+]}(A')\|_F \leq \|A-A'\|_F$ for all $\beta,\beta' \in \R^{d_k}$ and symmetric $A,A' \in \R^{r\times r}$.

\begin{definition}[Transformer class]\label{def:transformer-readout}
For $L \geq 1$ blocks, let $\cF_{\mathrm{TF}}(L,J,d_e,d_f,W) \coloneqq \left\{f_\omega = (f_\omega^\beta,f_\omega^V):\omega \in \Omega\right\}$, where, for all $N \in \cN$ and $(\cD_N,q) \in \cZ^N\times\cQ$,
\[
f_\omega(\cD_N,q) \coloneqq \mathsf{Read}_\omega\left(\mathsf{Block}_{L,\omega}\circ\cdots\circ\mathsf{Block}_{1,\omega}\{\mathsf{Emb}_\omega(\cD_N,q)\},q\right),
\]
with $\mathsf{Emb}_\omega \in \cF_{\mathrm{Emb}}(d_e,W)$, $\mathsf{Block}_{l,\omega} \in \cF_{\mathrm{Block}}(J,d_e,d_f,W;\mathsf M_l)$ with masks $\mathsf M_l$ for $l \in [L]$, and $\mathsf{Read}_\omega \in \cF_{\mathrm{Read}}(d_e,W)$.
\end{definition}

\iffalse
For the constructive results, $\Omega = [-W,W]^p$ leaves all embedding, attention, feedforward and readout parameters free, including every matrix or bias entry set to zero or to a quadrature constant in the comparator. In Propositions \ref{thm:finite-transformer} and \ref{thm:learning}, $W = (C_WNK)^{C_W}$ for a sufficiently large $C_W \geq 1$ depending only on the fixed prior, query specification, parameter dimensions and covariance bounds.

Because the observation embeddings and feedforward maps are shared across rows and the mask is symmetric, we have, for every permutation $\sigma$ of $[N]$,
\[
f_\omega((Z_{\sigma(i)})_{i = 1}^N,q) = f_\omega(\cD_N,q).
\]
\fi

\subsection{Pretraining prior}\label{app:prior}

%This appendix fixes the base distribution and basis of the exponential tilt family \eqref{eq:exp-tilt}, and specifies the pretraining prior used in Section \ref{sec:theory}.
Let $Z = (V,D)$ take values in $\cZ = \cV\times\cZ_D \subset \R^{d_z}$, where $V = (V_1,\cdots,V_d)$ collects the $d \geq 1$ continuous coordinates, with $\cV = \prod_{\ell = 1}^d\cV_\ell$ and $\cV_\ell \in \left\{\R,[0,1]\right\}$, and $D$ takes values in a finite set $\cZ_D \subset [0,1]^{d_D}$. Under the reference distribution $\P_0$, the variables $V_1,\cdots,V_d,D$ are independent, $V_\ell \sim \cN(0,1)$ if $\cV_\ell = \R$, $V_\ell \sim \mathrm{Unif}[0,1]$ if $\cV_\ell = [0,1]$, and $D \sim \mathrm{Unif}(\cZ_D)$.
Let $\Phi_\ell(v) \coloneqq \P_0(V_\ell \leq v)$ for $v \in \cV_\ell$, let $(\chi_0,\cdots,\chi_{|\cZ_D|-1})$ be an orthonormal basis of $L_2\{\mathrm{Unif}(\cZ_D)\}$ with $\chi_0 \equiv 1$, and define, for $u \in [0,1]$ and $h \in \N$,
\[
\varphi_0(u) \coloneqq 1,\quad \varphi_{2h-1}(u) \coloneqq \sqrt{2}\cos(2\pi hu) \quad\text{and}\quad \varphi_{2h}(u) \coloneqq \sqrt{2}\sin(2\pi hu).
\]

\begin{definition}[Orthonormal trigonometric basis]
Let $(b_k)_{k \geq 1}$ enumerate, in nondecreasing order of $\|r\|_\infty$, with ties broken lexicographically in $(r,\eta)$, the functions
\[
(v,D) \longmapsto \chi_\eta(D)\prod_{\ell = 1}^{d}\varphi_{r_\ell}\{\Phi_\ell(v_\ell)\}
\]
over $r = (r_1,\cdots,r_d) \in \N_0^d$ and $\eta \in \{0,\cdots,|\cZ_D|-1\}$ with $(r,\eta) \ne (0,0)$.
\end{definition}

Then $(1,b_1,b_2,\cdots)$ is an orthonormal basis of $L_2(\P_0)$ with $\sup_{k \geq 1,\,z \in \cZ}|b_k(z)| \leq C_\infty$, where $C_\infty \coloneqq 2^{\frac{d}{2}}\max_{0 \leq \eta < |\cZ_D|}\|\chi_\eta\|_\infty$.
For $K \geq 1$, let $b_{(K)} \coloneqq (b_1,\cdots,b_K)^\T$ and $A_K(\theta) \coloneqq \log\E_{\P_0}e^{\theta^\T b_{(K)}(Z)}$ for $\theta \in \R^K$. For $a > 1$ and $\sigma,B > 0$, let $\lambda_k \coloneqq \sigma^2k^{-2a}$ for $k \in \N$, $\Lambda_K \coloneqq \diag(\lambda_1,\cdots,\lambda_K)$ and $\Theta_K \coloneqq \left\{\theta \in \R^K:\|\theta\|_1 \leq B\right\}$.

\begin{definition}[Exponential sieve prior]\label{def:sieve-prior}
For $K \geq 1$, each $\theta \in \R^K$ defines the exponential tilt $\P_\theta$ by
\begin{equation}\label{eq:exp-tilt}
\frac{d\P_\theta}{d\P_0}(z) \coloneqq \exp\{\theta^\T b_{(K)}(z)-A_K(\theta)\},\quad z \in \cZ,
\end{equation}
and the coefficient $\theta$ has the prior
\begin{equation}\label{eq:prior}
\theta \sim \pi_K \coloneqq \cN(0,\Lambda_K)\bigm|\left\{\theta \in \Theta_K\right\}.
\end{equation}
\end{definition}

Let $c_0 \in (0,e^{-2(B+2)C_\infty}]$ and $C_0 \in [e^{(B+2)C_\infty},\infty)$ be constants, and let $\cP$ denote the set of distributions $\P \ll \P_0$ on $\cZ$ whose density satisfies $c_0 < \operatorname{ess\,inf}(d\P/d\P_0) \leq \operatorname{ess\,sup}(d\P/d\P_0) < C_0$ and, for quantile queries, has a version continuous in the outcome value $y$.

Since $0 \leq A_K(\theta) \leq C_\infty\|\theta\|_1$, the model $\cP$ contains $\P_\theta$ for all $K \geq 1$ and $\theta \in \R^K$ with $\|\theta\|_1 < B+2$, and every $\P$ with $\|d\P/d\P_0-d\P_{\theta_j}/d\P_0\|_\infty \to 0$ for some sequence of $\theta_j \in \R^{K_j}$ with $\|\theta_j\|_1 \leq B+1$. For $\P \in \cP$ and bounded $s \in L_2^0(\P) \coloneqq \left\{h \in L_2(\P):\E_\P h = 0\right\}$, continuous in $y$ for quantile queries, the path $\P_\tau$ with $d\P_\tau/d\P \coloneqq 1+\tau s$ lies in $\cP$ for all sufficiently small $|\tau|$, and these scores $s$ are dense in $L_2^0(\P)$. Hence the tangent space of $\cP$ at $\P$ is $L_2^0(\P)$, so that each pathwise differentiable query has a unique gradient, which is the corresponding EIF \citep{bickel1993,tsiatis2006}.

For a finite set $\cK$ and $k \in \cK$, let $\cU_k \subset \R^{\ell_k}$ be nonempty and compact, and let $\pi_{q,k}$ be a distribution on $\cU_k$, uniform in our experiments. %, and let the family probabilities $\varpi_k > 0$ sum to one.
The prior $\pi_q$ on the query set $\cQ$ is specified as follows. 

\begin{definition}[Query prior]\label{def:query-prior}
The query prior is the distribution $\pi_q$ on $\cQ$ with $\pi_q(\{k\}\times A) \coloneqq \varpi_k\pi_{q,k}(A)$ for all $k \in \cK$ and Borel sets $A \subseteq \cU_k$, where $\varpi_k \geq 0$ and $\sum_{k \in \cK}\varpi_k = 1$.
\end{definition}

\section{Semiparametric examples}\label{app:queries}

This section presents motivating examples of the semiparametric queried estimands $\beta_q(\P)$ of Section \ref{sec:framework}, in the notation of Section \ref{app:prior}, all of which, together with their finite stacks, satisfy Assumption \ref{cond:general-queries} by Proposition \ref{prop:query-verification}, which Section \ref{app:proofs-examples} proves.

\paragraph*{Average and group treatment effects.}\phantomsection\label{app:ate-example}
\begin{itemize}[leftmargin=*,nosep]
\item \emph{Data.} $Z = (Y,X,D)$, with outcome $Y \in \R$, covariates $X \in [0,1]^{d_X}$ and treatment $D \in \{0,1\}$, so that $V = (Y,X)$, $d = 1+d_X$, $\cZ_D = \{0,1\}$ and $\chi_1(D) = 2D-1$.
\item \emph{Estimand.} For a group $G \subseteq [0,1]^{d_X}$ with indicator $w_G(x) \coloneqq 1\{x \in G\}$ and probability $p_G \coloneqq \E_\P w_G(X) > 0$, the group effect is
\[
\beta_G(\P) \coloneqq \frac{\E_\P[w_G(X)\{m_1(X)-m_0(X)\}]}{p_G},
\]
where $m_a(x) \coloneqq \E_\P(Y\mid D = a,X = x)$ for $a \in \{0,1\}$. Under consistency, conditional exchangeability and overlap \citep{rosenbaum1983}, $\beta_G(\P)$ is the average treatment effect within $G$, and $\beta_{[0,1]^{d_X}}(\P)$ is the ATE of Section \ref{sec:exp-ate}.
\item \emph{EIF.} With the propensity score $e_a(x) \coloneqq \P(D = a\mid X = x)$, the EIF, which reduces to that of \cite{hahn1998} for the ATE, is
\[
\psi^\star_{\P,G}(Z) = \frac{w_G(X)}{p_G}\left[\frac{(2D-1)\{Y-m_D(X)\}}{e_D(X)}+m_1(X)-m_0(X)-\beta_G(\P)\right].
\]
\item \emph{Queries.} The argument is a rectangle $G = \prod_{j = 1}^{d_X}[u_j^-,u_j^+] \subseteq [0,1]^{d_X}$ with $\P_0(X \in G) \geq p_0$ for a given $p_0 > 0$, specified by its endpoints $u = (u^-,u^+)$ or selected from a finite collection.
\end{itemize}

\paragraph*{Quantile treatment effect.}
\begin{itemize}[leftmargin=*,nosep]
\item \emph{Data.} As for average treatment effects.
\item \emph{Estimand.} Let $F_a(y\mid x) \coloneqq \P(Y \leq y\mid D = a,X = x)$ and $F_{a,\P}(y) \coloneqq \E_\P F_a(y\mid X)$, which, under the same conditions, is the distribution function of the potential outcome under treatment $a$. At a quantile level $u \in (0,1)$, the QTE is
\[
\beta_u(\P) \coloneqq \xi_{1,\P}(u)-\xi_{0,\P}(u),\quad\text{where } \xi_{a,\P} \coloneqq F_{a,\P}^{-1}.
\]
\item \emph{EIF.} By the continuity of $d\P/d\P_0$ in $y$ on $\cP$, the density $f_{a,\P}(y) \coloneqq dF_{a,\P}(y)/dy$ is continuous and positive, and, differentiating the identity $F_{a,\P}\{\xi_{a,\P}(u)\} = u$ along paths through $\P$, we obtain the EIF \citep{firpo2007}
\[
\psi^\star_{\P,u}(Z) = -\frac{\psi_{1,\P}\{Z;\xi_{1,\P}(u)\}}{f_{1,\P}\{\xi_{1,\P}(u)\}}+\frac{\psi_{0,\P}\{Z;\xi_{0,\P}(u)\}}{f_{0,\P}\{\xi_{0,\P}(u)\}},
\]
where $\psi_{a,\P}(Z;y) \coloneqq 1\{D = a\}\{1\{Y \leq y\}-F_a(y\mid X)\}/e_a(X)+F_a(y\mid X)-F_{a,\P}(y)$ is the EIF of $F_{a,\P}(y)$.
\item \emph{Queries.} The argument is the level $u \in [\underline u,1-\underline u]$ for a given $\underline u \in (0,1/2)$.
\end{itemize}

\paragraph*{Mean survival time.}
\begin{itemize}[leftmargin=*,nosep]
\item \emph{Data.} $Z = (Y,\Delta,X)$, with observed time $Y \coloneqq \min(T,C)$, event indicator $\Delta \coloneqq 1\{T \leq C\}$ and covariates $X \in [0,1]^{d_X}$, where the failure and censoring times $T,C \in [J]$ are recorded at visits $1,\cdots,J$. The pair $(Y,\Delta)$ takes values in $\left\{(l,1):l \in [J]\right\}\cup\left\{(l,0):l \in [J-1]\right\}$, and the discrete factor is $D \coloneqq (Y/J,\Delta) \in [0,1]^2$, so that $V = X$ and $d = d_X$.
\item \emph{Estimand.} For $l \in [J]$ and $0 \leq j \leq J$, let
\[
\lambda_l(x) \coloneqq \frac{\P(Y = l,\Delta = 1\mid X = x)}{\P(Y \geq l\mid X = x)}\quad\text{and}\quad S(j\mid x) \coloneqq \prod_{l = 1}^{j}\{1-\lambda_l(x)\}
\]
denote the discrete hazard and survival functions, respectively. At a horizon $u \in [J]$, the restricted mean survival time is
\[
\beta_u(\P) \coloneqq \E_\P\mu_u(X),\quad\text{where } \mu_u(x) \coloneqq \sum_{j = 0}^{u-1}S(j\mid x).
\]
If censoring is conditionally independent, that is, $T$ and $C$ are independent given $X$, then $\lambda_l(x) = \P(T = l\mid T \geq l,X = x)$, $S(j\mid X) = \P(T > j\mid X)$ by the Kaplan--Meier product formula \citep{kaplan1958}, and $\beta_u(\P) = \E\{\min(T,u)\}$, so that $\beta_J(\P) = \E(T)$ is the mean survival time of Section \ref{sec:exp-mst}. There is a constant $c > 0$ such that $\P\{(Y,\Delta) = a\mid X\} \geq c$ for all $\P \in \cP$ and values $a$ of $(Y,\Delta)$, so that $\P(C \geq l\mid X) \geq \P(Y \geq l\mid X) \geq c$, which is the positivity condition, and $1-\lambda_l(X) \geq c$ for $l \in [J-1]$.
\item \emph{EIF.} The EIF weights the hazard residuals by the sensitivity of $\mu_u$ to each hazard,
\[
\psi^\star_{\P,u}(Z) = \mu_u(X)-\beta_u(\P)-\sum_{l = 1}^{u-1}\frac{\sum_{j = l}^{u-1}S(j\mid X)}{1-\lambda_l(X)}\cdot\frac{1\{Y = l,\Delta = 1\}-\lambda_l(X)1\{Y \geq l\}}{\P(Y \geq l\mid X)}.
\]
\item \emph{Queries.} The argument is the horizon $u \in \{2,\cdots,J\}$.
\end{itemize}

\paragraph*{Partially linear projection.}\phantomsection\label{app:pl-example}
\begin{itemize}[leftmargin=*,nosep]
\item \emph{Data.} $Z = (Y,W,X)$, with outcome $Y \in \R$, regressors $W \in [0,1]^{d_W}$ and covariates $X \in [0,1]^{d_X}$, so that $V = (Y,W,X)$, $d = 1+d_W+d_X$ and $\cZ_D$ is a singleton.
\item \emph{Estimand.} Let $m_W(x) \coloneqq \E_\P(W\mid X = x)$ and $m_Y(x) \coloneqq \E_\P(Y\mid X = x)$, and let $\tilde W \coloneqq W-m_W(X)$ and $\tilde Y \coloneqq Y-m_Y(X)$. The partially linear projection is the least squares coefficient
\[
\beta(\P) \coloneqq \argmin_{b \in \R^{d_W}}\E_\P\{(\tilde Y-b^\T\tilde W)^2\} = \Sigma_W^{-1}\E_\P(\tilde W\tilde Y),\quad\text{where } \Sigma_W \coloneqq \E_\P(\tilde W^{\otimes2}),
\]
and $\lambda_{\min}(\Sigma_W) \geq c > 0$ by the lower density bound $c_0$ of $\cP$. If $\E_\P(Y\mid W,X) = W^\T\beta+g(X)$, then $\beta(\P) = \beta$, and, for $d_W = 1$, $\beta(\P)$ is the PLR coefficient of Section \ref{sec:exp-plm}.
\item \emph{EIF.} The EIF is
\[
\psi^\star_\P(Z) = \Sigma_W^{-1}\tilde W\{\tilde Y-\tilde W^\T\beta(\P)\}.
\]
\item \emph{Queries.} The argument is a matrix $M$ with $d_W$ columns in a given compact set, and the query returns the contrast $M\beta(\P)$, which equals $\beta(\P)$ for $M = I_{d_W}$.
\end{itemize}

%Proposition \ref{prop:query-verification} shows that these families and their fixed finite stacks, such as quantile effects at several levels, satisfy Assumptions \ref{cond:general-queries}, \ref{cond:query-evaluation} and \ref{cond:query-lipschitz}(i).

For $m \in \N$ and query families $k_1,\cdots,k_m$, the finite stack $k = (k_1,\cdots,k_m)$ is the query family with arguments $u = (u_1,\cdots,u_m) \in \cU_k \coloneqq \prod_{j = 1}^m\cU_{k_j}$ and values $\beta_{(k,u)}(\P) \coloneqq \{\beta_{(k_1,u_1)}(\P)^\T,\cdots,\beta_{(k_m,u_m)}(\P)^\T\}^\T$, such as quantile effects at several levels.

\begin{proposition}[Query conditions for the examples]\label{prop:query-verification}
Let $\cK$ be a finite set of finite stacks of the four query families above, with the arguments specified there. Then, on the model $\cP$ of Section \ref{app:prior}, the queries $q \in \cQ$ satisfy Assumption \ref{cond:general-queries}(i)--(iv).
\end{proposition}

\section{Assumptions}\label{app:assumptions}

Assumptions \ref{cond:general-queries}--\ref{cond:local-evaluation} formalise Conditions \ref{A1}--\ref{A5} of Section \ref{sec:target-efficiency}, and Assumption \ref{cond:fixed-population} in Section \ref{app:single-population} states the conditions of Theorem \ref{thm:fixed-population} on the fixed population. In these assumptions, $p$ and $p'$ denote the densities $d\P/d\P_0$ and $d\P'/d\P_0$ of $\P,\P' \in \cP$, and $\Theta_K$, $\P_\theta$ and $\pi_K$ are as in Section \ref{app:prior}. Moreover, $\cP_\pi$ denotes the support of $\pi_P$, and $\theta_\P \in \R^K$ and $g_{\P,K}$ denote the sieve coefficients and remainder of $\P$ in Section \ref{sec:target-efficiency}, given by $d\P/d\P_0 \propto \exp\{\theta_\P^\T b_{(K)}+g_{\P,K}\}$ with $\E_{\P_0}g_{\P,K} = 0$ and $\E_{\P_0}\{g_{\P,K}b_{(K)}\} = 0$. %In the following assumptions,  we use $p$ and $p'$ to  denote the densities of $\P,\P' \in \cP$ relative to $\P_0$, respectively.

\begin{assumption}[Regular parameters]\label{cond:general-queries}
There exists $C < \infty$ such that, for all $K \geq 1$, $\P \in \cP$ and $q = (k,u) \in \cQ$, the following hold.
\begin{enumerate}[label=(\roman*),nosep]
\item The map $\beta_q:\cP \to \R^{d_k}$ is pathwise differentiable at $\P$ with EIF $\psi^\star_{\P,q}$, and $\|\beta_q(\P_0)\|+\|\psi^\star_{\P,q}\|_{L_2(\P)} \leq C$.
\item The Jacobian $D_q(\theta) \coloneqq \partial\beta_q(\P_\theta)/\partial\theta^\T \in \R^{d_k\times K}$ satisfies $\|D_q(\theta)-D_q(\theta')\|_{\op} \leq C\sqrt K\|\theta-\theta'\|$ for all $\theta,\theta' \in \R^K$ with $\|\theta\|_1,\|\theta'\|_1 \leq B+1$.
\item The EIF satisfies $\|\psi^\star_{\P',(k,u')}-\psi^\star_{\P,q}\|_{L_2(\P_0)} \to 0$ as $\|p'-p\|_\infty+\|u'-u\| \to 0$ with $\P' \in \cP$ and $u' \in \cU_k$.
\item The query family $k$ admits the integral representation of Definition \ref{def:integral-queries} in Section \ref{app:comparator-primitives}.
\end{enumerate}
\end{assumption}

\begin{assumption}[Covariance bounds]\label{cond:covariance-bounds}
With $v_\pm$ the clipping bounds of the covariance readout $\clip_{[v_-,v_+]}$ in Definition \ref{def:transformer-output}, the efficiency bound $V_q^\star(\P)$ satisfies the following.
\begin{enumerate}[label=(\roman*),nosep]
\item For $\nu$-almost every $(\P,q)$, $v_-I \preceq V_q^\star(\P) \preceq v_+I$.
% \item[(ii)] For each $n \in \cN$, there is a measurable subset $\cG_n$ of the regular inputs of Lemma \ref{lem:laplace} such that
% \[
% v_-I \preceq S_\pi(\cD_n,q) \preceq v_+I \quad \text{for all }(\cD_n,q) \in \cG_n,
% \quad
% \inf_{n \in \cN}\cL_{\pi,n}(\cG_n) \to 1,
% \]
% where $\cL_{\pi,n}$ is the pretraining input distribution at size $n$.
% \item[(iii)] Under the evaluation input distribution,
% \begin{equation}\label{eq:posterior-spectrum}
% \cL_{\nu,N}\{v_-I \preceq S_\pi(\cD_N,q) \preceq v_+I\} \longrightarrow 1.
% \end{equation}
\item There exist constants $v_- < \underline v \leq \bar v < v_+$, independent of $N$, such that $\underline vI \preceq V_q^\star(\P) \preceq \bar vI$ for $\pi$-almost every $(\P,q)$.
\end{enumerate}
\end{assumption}
%Part~(ii) follows if the simulator's efficiency covariances lie a fixed distance inside $[v_-,v_+]$ and $K \to \infty$ with $K^{3\vee2a} = o(n_{\min})$. Indeed, \eqref{eq:irregular-context} gives regularity with probability tending to one, and Proposition \ref{prop:triangular} gives the same for the spectral bound on $S_\pi$. These conclusions apply along every sequence of pretraining sizes, which gives the uniform probability in part~(ii). Let $\Gamma_N(\nu) \coloneqq 1+\chi^2(\cL_{\nu,N}\,\|\,\cL_{\pi,N})$ denote the overlap factor of the evaluation input distribution relative to the pretraining one. Under bounded $\Gamma_N(\nu)$, Cauchy--Schwarz transfers this probability bound to evaluation and establishes part~(iii). More generally, part~(iii) follows from posterior covariance consistency when $V_q^\star(\P)$ stays a fixed distance inside the output interval. An evaluation bound is needed when $\Gamma_N(\nu)$ grows, since part~(ii) alone then need not transfer.

\begin{assumption}[Sieve prior approximation]\label{cond:simulator-prior}
There exist $K = K_N$ and sequences $c_N,h_N \to 0$ such that $\theta_\P \in \Theta_K$, $N\|g_{\P,K}\|_{L_2(\P_0)}^2 \leq c_N$ and $\|g_{\P,K}\|_\infty \leq h_N$ for all $\P \in \cP_\pi$.
\end{assumption}

\begin{assumption}[Prior regularity]\label{cond:posterior-tails}
There exist constants $0 < c_w \leq C_w < \infty$, independent of $N$, and $L_N = o(\sqrt N)$ such that, for every $g \in \left\{g_{\P,K}:\P \in \cP_\pi\right\}$, the conditional density $w_\pi(\cdot\mid g)$ defined by $\pi_P(\theta_\P \in d\theta\mid g_{\P,K} = g) = w_\pi(\theta\mid g)\,\pi_K(d\theta)$ exists and satisfies, for all $\theta,\theta' \in \Theta_K$,
\[
c_w \leq w_\pi(\theta\mid g) \leq C_w\quad\text{and}\quad|\log w_\pi(\theta\mid g)-\log w_\pi(\theta'\mid g)| \leq L_N\|\theta-\theta'\|.
\]
\end{assumption}

\begin{assumption}[Evaluation distribution]\label{cond:local-evaluation}
There exists $C < \infty$ such that $\chi^2(\nu_N\,\|\,\pi) \leq C$ for every $N$, where $\nu_N$ is the evaluation distribution $\nu$ at sample size $N$ and $\chi^2(\nu_N\,\|\,\pi) \coloneqq \int(d\nu_N/d\pi-1)^2\,d\pi$ if $\nu_N \ll \pi$ and $\chi^2(\nu_N\,\|\,\pi) \coloneqq \infty$ otherwise.
\end{assumption}

Assumptions \ref{cond:general-queries} and \ref{cond:covariance-bounds} concern the queries. Assumption \ref{cond:general-queries}(i) asks each query to be pathwise differentiable with an efficient influence function bounded in $L_2(\P)$, the standard setting of semiparametric efficiency theory \citep{bickel1993,tsiatis2006}. Of the remaining parts, (ii) bounds the second order term $\|\beta_q(\P_{\theta'})-\beta_q(\P_\theta)-D_q(\theta)(\theta'-\theta)\|$ by $C\sqrt K\|\theta'-\theta\|^2$, as shown in Lemma \ref{lem:regularity}, like the remainder conditions of \cite{castillo2015}, (iii) ensures that the sieve scores approximate the efficient influence function as $K$ grows, and (iv) is used only to construct Transformers that evaluate the queries. All four parts hold for the examples of Section \ref{app:queries} by Proposition \ref{prop:query-verification}. Assumption \ref{cond:covariance-bounds} is the usual nondegeneracy of the efficiency bound, relative to the eigenvalue range $[v_-,v_+]$ of the covariance readout, and follows from \ref{A2} when this range contains the bounds of \ref{A2} in its interior, while its part~(ii) makes the clipping of $S_\pi$ rarely active in pretraining, by \eqref{eq:pretraining-spectrum}.

Assumptions \ref{cond:simulator-prior} and \ref{cond:posterior-tails} require the simulator prior $\pi_P$ to stay close to the sieve prior $\pi_K$ at the resolution of the data, and both hold trivially for $\pi_K$, with $g_{\P,K} = 0$, $w_\pi \equiv 1$ and $L_N = 0$. Assumption \ref{cond:simulator-prior} bounds the sieve approximation error of every population in $\cP_\pi$ by $o(N^{-\frac{1}{2}})$ in $L_2(\P_0)$, below the sampling error, and by $o(1)$ in supremum norm, which lets Assumption \ref{cond:general-queries}(iii) relate $\P$ to $\P_{\theta_\P}$. This assumption plays the role of the no-bias condition in semiparametric Bernstein--von Mises (BvM) theorems \citep{castillo2015}, and holds, for example, when $\cP_\pi$ lies in a Sobolev class $\cP_s(\Xi,\eta)$ of the form \eqref{eq:sobolev-class} with $s > d/2$ and $\sqrt NK^{-\frac{s}{d}} \to 0$, because then $\|g_{\P,K}\|_{L_2(\P_0)} \leq \Xi K^{-\frac{s}{d}}$ and $\|g_{\P,K}\|_\infty \lesssim K^{\frac{1}{2}-\frac{s}{d}}$. Assumption \ref{cond:posterior-tails} lets $\pi_P$ reweight $\pi_K$ by a density that is bounded away from zero and infinity and almost constant over shifts of order $N^{-\frac{1}{2}}$, since $L_NN^{-\frac{1}{2}} \to 0$, a quantitative form of the usual BvM requirements that the prior density be continuous and positive \citep{bickel2012} and nearly invariant under such shifts \citep{ghosal2017}.
Assumption \ref{cond:local-evaluation} bounds the $\chi^2$ divergence of the evaluation distribution from the pretraining prior, as in the distribution shift analysis of \cite{ma2026}. %, and holds, for example, when $\nu_N$ upweights no pair $(\P,q)$ by more than a factor $1+C$. %Evaluation may thus emphasise particular populations or queries, or even fix a query that $\pi_q$ selects with positive probability, but it cannot concentrate on a population of prior probability zero, the case of Theorem \ref{thm:fixed-population}. It transfers pretraining bounds to evaluation through \eqref{eq:chi2-transfer}.

\section{Proofs for Section \ref{sec:target-efficiency}}\label{app:proofs}

This section proves Proposition \ref{prop:population-loss} and Theorem \ref{thm:target}, the latter through the expansion of Proposition \ref{prop:triangular} and the change of prior in Section \ref{app:simulator-prior}. We say that a statement holds under $\pi_K$ if it holds when $\pi_P$ is the distribution of $\P_\theta$ with $\theta \sim \pi_K$ of Definition \ref{def:sieve-prior}. At size $n$, we let $\E_\pi\E_\P\varphi \coloneqq \int\int\varphi(\P,q,\cD_n)\,\P^{\otimes n}(d\cD_n)\,\pi(d\P,dq)$ for measurable $\varphi \geq 0$. Let $\G_Nh \coloneqq \sqrt N(\P_Nh-\E_\P h)$ denote the empirical process. Throughout, $N \geq 3$ and $T \geq 2$, and constants $c$ and $C$ may vary between bounds.

Besides the density envelope $c_0 \leq d\P/d\P_0 \leq C_0$ for $\P \in \cP$ of Section \ref{app:prior}, two consequences of the assumptions recur in the proofs. First, for $\P,\P' \in \cP$ with densities $p,p'$, the path $\P_t \in \cP$ with density $p_t \coloneqq (1-t)p+tp'$, $t \in [0,1]$, has score $(p'-p)/p_t$ with $L_2(\P_t)$ norm at most $c_0^{-\frac{1}{2}}\|p-p'\|_{L_2(\P_0)}$, so that, integrating $\partial_t\beta_q(\P_t) = \E_{\P_t}\{\psi^\star_{\P_t,q}(p'-p)/p_t\}$ of Assumption \ref{cond:general-queries}(i) over $t \in [0,1]$, we obtain, for $q \in \cQ$,
\begin{equation}\label{eq:query-lipschitz}
\|\beta_q(\P)-\beta_q(\P')\| \leq Cc_0^{-\frac{1}{2}}\|p-p'\|_{L_2(\P_0)}.
\end{equation}
Second, by Assumption \ref{cond:local-evaluation}, $\nu_N$, the evaluation distribution at size $N$, has a density $w$ relative to $\pi_P\otimes\pi_q$ with $\int w^2\,d(\pi_P\otimes\pi_q) \leq 1+C$, so that, by the Cauchy--Schwarz inequality and $(\E_\P\varphi)^2 \leq \E_\P\varphi$, every measurable $\varphi = \varphi(\P,q,\cD_N)$ with values in $[0,1]$ satisfies
\begin{equation}\label{eq:chi2-transfer}
\E_{\nu_N}\E_\P\varphi \leq \{(1+C)\,\E_{\pi_P\otimes\pi_q}\E_\P\varphi\}^{\frac{1}{2}}.
\end{equation}

\subsection{Proof of Proposition \ref{prop:population-loss}}\label{app:population-loss}

\begin{proof}
For symmetric matrices  $C$ and $A$ in $\R^{d_k \times d_k}$ with $A \succ 0$, define $F_C(A) \coloneqq -\log\det A+\tr(AC)$. We first minimise $F_C$ under spectral constraints and then apply the minimisers to the posterior risk.

Let $(e_j)_{j \in [d_k]}$ be an orthonormal basis of $\R^{d_k}$ with $Ce_j = \lambda_je_j$, and, with $v_\pm$ as in Section \ref{app:architecture}, let $J$ be $[v_+^{-1},v_-^{-1}]$ or, if $C \succ 0$, $(0,\infty)$. For symmetric $A$ with eigenvalues in $J$, the entries $A_{jj} \coloneqq e_j^\T Ae_j$ lie in $J$, and, by Hadamard's inequality, we have
\[
F_C(A) \geq \sum_{j = 1}^{d_k}(\lambda_jA_{jj}-\log A_{jj}),\quad\text{with equality only if $e_i^\T Ae_j = 0$ for all $i \ne j$}.
\]
Since $\phi_j(a) \coloneqq \lambda_ja-\log a$ is strictly convex in $a > 0$ with $\phi_j'(a) = \lambda_j-1/a$, the unique minimiser of $\phi_j$ over $J$ is $\lambda_j^{-1}$ if $J = (0,\infty)$ and $\{(\lambda_j\vee v_-)\wedge v_+\}^{-1}$ otherwise, so that
\begin{align}
\argmin_{A \succ 0}F_C(A) &= \{C^{-1}\}\quad\text{if } C \succ 0, \label{eq:unconstrained-precision}\\
\argmin_{v_+^{-1}I \preceq A \preceq v_-^{-1}I} F_C(A) &= \{(\clip C)^{-1}\}, \label{eq:clipped-precision}
\end{align}
where $\clip \coloneqq \clip_{[v_-,v_+]}$ is the spectral clipping of Section \ref{app:architecture}.

Since the posterior of $\beta_q(\P)$ has mean $m_\pi$ and covariance $S_\pi/n$, expanding the quadratic form in \eqref{eq:score} around $m_\pi$ with $A \coloneqq V^{-1}$, we have
\begin{equation}\label{eq:posterior-risk-decomposition}
2\E_\pi\{\ell_n(\beta_q(\P);\beta,V)\mid\cD_n,q\} = F_{S_\pi}(A)+n(\beta-m_\pi)^\T A(\beta-m_\pi),
\end{equation}
whose quadratic term vanishes only at $\beta = m_\pi$. Hence, by \eqref{eq:unconstrained-precision} with $C = S_\pi$, \eqref{eq:posterior-loss-target} holds, and $f_\pi^\circ \coloneqq (m_\pi,\clip S_\pi)$ satisfies $f_\pi^\circ = f_\pi$ whenever $v_-I \preceq S_\pi \preceq v_+I$. Subtracting \eqref{eq:posterior-risk-decomposition} at $f_\pi^\circ$ from \eqref{eq:posterior-risk-decomposition} at $(u,\Sigma)$, for $u \in \R^{d_k}$ and $v_-I \preceq \Sigma \preceq v_+I$, we obtain the conditional excess score
\begin{equation}\label{eq:posterior-loss-regret}
\begin{aligned}
&\,\E_\pi\{\ell_n(\beta_0;u,\Sigma)-\ell_n(\beta_0;m_\pi,\clip S_\pi)\mid\cD_n,q\}\\
= &\,\frac{n}{2}(u-m_\pi)^\T\Sigma^{-1}(u-m_\pi)+\frac{1}{2}\{F_{S_\pi}(\Sigma^{-1})-F_{S_\pi}(A_*)\} \geq 0,
\end{aligned}
\end{equation}
where $\beta_0 \coloneqq \beta_q(\P)$, $A_* \coloneqq (\clip S_\pi)^{-1}$ and the inequality follows from \eqref{eq:clipped-precision}. Equality holds only at $(u,\Sigma) = f_\pi^\circ$, which completes the proof.
\end{proof}

\subsection{Proof of Theorem \ref{thm:target}}\label{app:main-results}

\begin{proof}
We first prove the theorem for the sieve prior, for each complexity $K$ and then as $K$ grows, and then extend the theorem to every prior satisfying Assumptions \ref{cond:simulator-prior} and \ref{cond:posterior-tails}.

Under $\pi_K$ with $K^{3\vee2a} = o(N)$, by Proposition \ref{prop:triangular} and \eqref{eq:chi2-transfer} with $\varphi = \cR_N^K \in [0,1]$, the projected remainder defined after \eqref{eq:posterior-moment-remainder}, we have
\[
\E_\nu\E_\P\cR_N^K \leq \{(1+C)\E_\pi\E_\P\cR_N^K\}^{\frac{1}{2}} = o(1),
\]
which proves \eqref{eq:sieve-expansion} for the sieve prior by Markov's inequality, without using $K \to \infty$.

We next let $K \to \infty$ under the sieve prior. Since $\psi^\star_{\P_\theta,q}$ is centred under $\P_\theta$, the projection error $e_K(\theta,q) \coloneqq \|\psi^\star_{\P_\theta,q}-\psi^K_{\P_\theta,q}\|_{L_2(\P_\theta)}^2$, with $\psi^K_{\P,q}$ as in Section \ref{app:posterior-expansion}, satisfies
\[
e_K(\theta,q)^{\frac{1}{2}} \leq \|\psi^\star_{\P_\theta,q}-\Pi_K\psi^\star_{\P_\theta,q}\|_{L_2(\P_\theta)} \leq C\|\psi^\star_{\P_\theta,q}-\Pi_K\psi^\star_{\P_\theta,q}\|_{L_2(\P_0)},
\]
where $\Pi_K$ denotes the componentwise $L_2(\P_0)$ projection onto $\spn(1,b_1,\cdots,b_K)$, the first inequality holds because $\Pi_K\psi^\star_{\P_\theta,q}-\E_{\P_\theta}\Pi_K\psi^\star_{\P_\theta,q}$ lies componentwise in $\spn\{b_k-\E_{\P_\theta}b_k:k \in [K]\}$, and the second follows from the density envelope.

With $\vartheta$, $\cB_\infty$ and $\P_\vartheta$ as defined before Lemma \ref{lem:prior-boundary}, for each $q \in \cQ$ and on the event $\cB_\infty$, since $\Pi_K$ is an orthogonal projection, we have
\[
e_K(\vartheta_{1:K},q)^{\frac{1}{2}} \leq C\|\psi^\star_{\P_{\vartheta_{1:K}},q}-\psi^\star_{\P_\vartheta,q}\|_{L_2(\P_0)}+C\|\psi^\star_{\P_\vartheta,q}-\Pi_K\psi^\star_{\P_\vartheta,q}\|_{L_2(\P_0)} = o(1),
\]
where the first term vanishes by Assumption \ref{cond:general-queries}(iii), as $\|d\P_{\vartheta_{1:K}}/d\P_0-d\P_\vartheta/d\P_0\|_\infty \leq C\sum_{k > K}|\vartheta_k|$, and the second by the completeness of the basis, while $e_K \leq C$ by Assumption \ref{cond:general-queries}(i). By conditional centring and \eqref{eq:projection-identity}, we have
\begin{equation}\label{eq:projection-error-moments}
\begin{aligned}
\E_{\P_\theta}\|\G_N(\psi^K_{\P_\theta,q}-\psi^\star_{\P_\theta,q})\|^2 &= e_K(\theta,q)\quad\text{and}\\
\|V_q^\star(\P_\theta)-\E_{\P_\theta}\{(\psi^K_{\P_\theta,q})^{\otimes2}\}\|_{\op} &\leq e_K(\theta,q).
\end{aligned}
\end{equation}
By the triangle inequality, \eqref{eq:projection-error-moments} and $1\wedge(2x+y) \leq 2(1\wedge x)+y$ for $x,y \geq 0$, we have
\[
\cR_N \leq 2\cR_N^K+2\|\G_N(\psi^K_{\P_\theta,q}-\psi^\star_{\P_\theta,q})\|^2+Ce_K(\theta,q).
\]
Averaging over $(\theta,q) \sim \pi_K\otimes\pi_q$ and $\cD_N \sim \P_\theta^{\otimes N}$, by Proposition \ref{prop:triangular}, \eqref{eq:projection-error-moments} and \eqref{eq:series-coupling}, we obtain
\begin{equation}\label{eq:simulator-target}
\E_\pi\E_\P\cR_N \leq 2\E_\pi\E_\P\cR_N^K+C\E_{\pi_K\otimes\pi_q}e_K = o(1),
\end{equation}
which, by \eqref{eq:chi2-transfer} with $\varphi = \cR_N$ and Markov's inequality, proves \eqref{eq:target-efficiency} for the sieve prior.

We finally extend the expansion to the priors of Assumptions \ref{cond:simulator-prior} and \ref{cond:posterior-tails}, with the sequences $c_N,h_N$ and the bound $C_w$ therein. Let $\bar\cR_N$ denote \eqref{eq:posterior-moment-remainder} with $f_{\pi_K}$ in place of $f_\pi$, where $f_{\pi_K}$, $\P_{\theta,g}$ and $\epsilon_{\pi,N}$ are as in Section \ref{app:simulator-prior}. By \eqref{eq:prior-stability} and Lemma \ref{lem:prior-reduction}, we have
\begin{equation}\label{eq:tilt-reduction}
\begin{aligned}
\E_{\pi_P\otimes\pi_q}\E_\P\cR_N &\leq 2\E_{\pi_P\otimes\pi_q}\E_\P\bar\cR_N+C\epsilon_{\pi,N}\\
&\leq 2C_w\sup_g\E_{\pi_K\otimes\pi_q}\E_{\P_{\theta,g}}\bar\cR_N(\P_{\theta,g},q,\cD_N)+C\epsilon_{\pi,N},
\end{aligned}
\end{equation}
where $\epsilon_{\pi,N} = o(1)$ by Assumptions \ref{cond:simulator-prior} and \ref{cond:posterior-tails}. Hence it suffices to bound the last expectation at $\P \coloneqq \P_{\theta,g_N}$ for every sequence $g_N \in \left\{g_{\P',K}:\P' \in \cP_\pi\right\}$, for which we have
\begin{equation}\label{eq:tilt-mean-decomposition}
\begin{aligned}
&\,\sqrt N\{m_{\pi_K}-\beta_q(\P)\}-\G_N\psi^\star_{\P,q}\\
= &\,\sqrt N\{m_{\pi_K}-\beta_q(\P_\theta)-\P_N\psi^\star_{\P_\theta,q}\}-\sqrt N\{\beta_q(\P)-\beta_q(\P_\theta)-\E_\P\psi^\star_{\P_\theta,q}\}\\
&\quad+\G_N(\psi^\star_{\P_\theta,q}-\psi^\star_{\P,q}),
\end{aligned}
\end{equation}
where $\G_N$ is centred under $\P$, the first term is the mean error in $\bar\cR_N(\P_\theta,q,\cD_N)$ as $\E_{\P_\theta}\psi^\star_{\P_\theta,q} = 0$, and the last term has second moment at most $\Delta_\psi^2$ for $\Delta_\psi \coloneqq \|\psi^\star_{\P,q}-\psi^\star_{\P_\theta,q}\|_{L_2(\P)}$. Moreover,
\begin{align}
\sqrt N\|\beta_q(\P)-\beta_q(\P_\theta)-\E_\P\psi^\star_{\P_\theta,q}\| &\overset{\text{(i)}}{\leq} C\{N\chi^2(\P\,\|\,\P_\theta)\}^{\frac{1}{2}} \overset{\text{(ii)}}{\leq} Cc_N^{\frac{1}{2}}, \quad \text{and} \label{eq:tilt-mean-bound}\\
\|V_q^\star(\P)-V_q^\star(\P_\theta)\|_F &\overset{\text{(iii)}}{\leq} \|\E_\P\{(\psi^\star_{\P,q})^{\otimes2}-(\psi^\star_{\P_\theta,q})^{\otimes2}\}\|_F+(e^{2h_N}-1)\E_{\P_\theta}\|\psi^\star_{\P_\theta,q}\|^2 \notag\\
&\overset{\text{(iv)}}{\leq} C(\Delta_\psi+h_N), \label{eq:tilt-covariance-bound}
\end{align}
where (i) follows from \eqref{eq:query-lipschitz}, $\|p-d\P_\theta/d\P_0\|_{L_2(\P_0)}^2 \leq C\chi^2(\P\,\|\,\P_\theta)$, $\E_\P\psi^\star_{\P_\theta,q} = \E_{\P_\theta}\{(d\P/d\P_\theta-1)\psi^\star_{\P_\theta,q}\}$, Cauchy--Schwarz and Assumption \ref{cond:general-queries}(i), (ii) and (iii) from \eqref{eq:tilt-chi-square}, and (iv) from Cauchy--Schwarz and Assumption \ref{cond:general-queries}(i). Consequently, we obtain
\begin{align}
\E_{\pi_K\otimes\pi_q}\E_\P\bar\cR_N(\P,q,\cD_N) &\overset{\text{(i)}}{\leq} 3\E_{\pi_K\otimes\pi_q}\E_\P\bar\cR_N(\P_\theta,q,\cD_N)+C(c_N+h_N^2+\E_{\pi_K\otimes\pi_q}\Delta_\psi^2) \notag\\
&\overset{\text{(ii)}}{\leq} 3\E_{\pi_K\otimes\pi_q}\E_{\P_\theta}\bar\cR_N(\P_\theta,q,\cD_N)+C\E_{\pi_K\otimes\pi_q}\Delta_\psi^2+o(1) \notag\\
&\overset{\text{(iii)}}{=} C\E_{\pi_K\otimes\pi_q}\Delta_\psi^2+o(1), \label{eq:tilt-remainder-bound}
\end{align}
where (i) follows from \eqref{eq:tilt-mean-decomposition}--\eqref{eq:tilt-covariance-bound}, $\|a+b+c\|^2 \leq 3(\|a\|^2+\|b\|^2+\|c\|^2)$ and $1\wedge(3x+y) \leq 3(1\wedge x)+y$ for $x,y \geq 0$, (ii) from Lemma \ref{lem:shift} and \eqref{eq:tilt-chi-square}, and (iii) from \eqref{eq:simulator-target}.

It remains to show that $\E_{\pi_K\otimes\pi_q}\Delta_\psi^2 = o(1)$. For $q \in \cQ$, let $\theta \coloneqq \vartheta_{1:K}$ on $\cB_\infty$. Since $\|d\P_\theta/d\P_0-d\P_\vartheta/d\P_0\|_\infty \to 0$, by \eqref{eq:tilt-chi-square} and the density envelope we have
\[
\|p-d\P_\vartheta/d\P_0\|_\infty+\|d\P_\theta/d\P_0-d\P_\vartheta/d\P_0\|_\infty \leq C(e^{2h_N}-1)+2\|d\P_\theta/d\P_0-d\P_\vartheta/d\P_0\|_\infty = o(1),
\]
so that $\Delta_\psi = o(1)$ by Assumption \ref{cond:general-queries}(iii) at $\P_\vartheta$ and the density envelope. Since $\Delta_\psi \leq C$ by Assumption \ref{cond:general-queries}(i), applying \eqref{eq:series-coupling} to $\Delta_\psi^2$, we obtain $\E_{\pi_K\otimes\pi_q}\Delta_\psi^2 = o(1)$. Consequently, by \eqref{eq:chi2-transfer}, \eqref{eq:tilt-reduction} and \eqref{eq:tilt-remainder-bound},
\[
\E_\nu\E_\P\cR_N \leq \{(1+C)\E_{\pi_P\otimes\pi_q}\E_\P\cR_N\}^{\frac{1}{2}} = o(1),
\]
which proves \eqref{eq:target-efficiency} for these priors by Markov's inequality and completes the proof.
\end{proof}

\subsection{Technical lemmas for Theorem \ref{thm:target}}\label{app:theory-technical}

%The lemmas below describe the coefficient posterior under $\pi_K$ and lead to Proposition \ref{prop:triangular}, and Section \ref{app:simulator-prior} compares it with the posterior under the priors of Assumptions \ref{cond:simulator-prior} and \ref{cond:posterior-tails}.

We first introduce some notations that will be used in this part. 
At a context $(\cD_N,q)$, with $b_{(K)}$, $A_K$, $\Lambda_K$ and $\Theta_K$ as in Section \ref{app:prior}. For $\theta \in \R^K$, we define, $\bbar \coloneqq \P_Nb_{(K)}$, $s_\theta \coloneqq b_{(K)}-\E_{\P_\theta}b_{(K)}$, $I_K(\theta) \coloneqq \Cov_{\P_\theta}\{b_{(K)}\}$ and
\[
\cV_N(\theta) \coloneqq N\{A_K(\theta)-\theta^\T\bbar\}+\theta^\T\Lambda_K^{-1}\theta/2. 
\]

Under $\theta \sim \pi_K$ and $\cD_N \mid \theta \sim \P_\theta^{\otimes N}$, the coefficient posterior of $\theta$ given $\cD_N$ has density $p_N \coloneqq e^{-\cV_N}/\int_{\Theta_K}e^{-\cV_N(\theta)}\,d\theta$ on $\Theta_K$, and we let $\E_N$, $\Var_N$ and $\Cov_N$ denote the expectation, variance and covariance of $\theta \sim p_N$, respectively. The posterior mode $\tilde\theta \coloneqq \argmin_{\theta \in \Theta_K}\cV_N(\theta)$ is unique by the strong convexity in Lemma \ref{lem:geometry}.

For a sufficiently large constant $C_r$, let $\rho_N \coloneqq C_r\sqrt{K\log N/N}$, and call a context $(\cD_N,q)$ regular if $(\cD_N,q) \in \cE_N$, where
\[
\cE_N \coloneqq \left\{(\cD_N,q) \in \cZ^N\times\cQ:\left\{\theta' \in \R^K:\|\theta'-\tilde\theta\| \leq \rho_N\right\} \subset \Theta_K\right\}.
\]

\begin{definition}[Mode approximation]\label{def:reference-estimator}
The mode pair $(g,V)$ at a context $(\cD_N,q) \in \cZ^N\times\cQ$ is given by
\begin{equation}\label{eq:plugin}
g(\cD_N,q) \coloneqq \beta_q(\P_{\tilde\theta})\quad\text{and}\quad V(\cD_N,q) \coloneqq D_q(\tilde\theta)\left\{I_K(\tilde\theta)+\frac{\Lambda_K^{-1}}{N}\right\}^{-1}D_q(\tilde\theta)^\T,
\end{equation}
where $D_q(\theta)$ is the Jacobian of $\theta \mapsto \beta_q(\P_\theta)$ in Assumption \ref{cond:general-queries}(ii).
\end{definition}

\subsubsection{Sieve curvature and query derivatives}\label{app:sieve-geometry}

\begin{lemma}\label{lem:geometry}
There exist constants $0 < c \leq C < \infty$, depending only on $B$ and the basis bound $C_\infty$ of Section \ref{app:prior}, such that, for all $K \geq 1$ and $\theta,\theta' \in \R^K$ with $\|\theta\|_1,\|\theta'\|_1 \leq B+1$,
\[
c \leq \frac{d\P_\theta}{d\P_0} \leq C,\quad cI \preceq I_K(\theta) \preceq CI \quad\text{and}\quad \|I_K(\theta)-I_K(\theta')\|_{\op} \leq C\sqrt K\|\theta-\theta'\|.
\]
\end{lemma}

\begin{proof}
We derive the density bounds from $\sup_{k,z}|b_k(z)| \leq C_\infty$, compare $I_K(\theta)$ with $I_K(0) = I$ and then integrate third moments of $s_\theta$ along coefficient segments.

For $\|\theta\|_1 \leq B+1$, both $|\theta^\T b_{(K)}|$ and $|A_K(\theta)|$ are at most $C_\infty(B+1)$, so that
\[
e^{-2C_\infty(B+1)} \leq \frac{d\P_\theta}{d\P_0} \leq e^{2C_\infty(B+1)},
\]
which is the first claim. Since $\E_{\P_0}b_{(K)} = 0$ and $\Cov_{\P_0}\{b_{(K)}\} = I$, by these density bounds, we obtain, for every $v \in \R^K$,
\begin{equation}\label{eq:information-bounds}
\begin{aligned}
v^\T I_K(\theta)v &= \inf_{t \in \R}\E_{\P_\theta}\{(v^\T b_{(K)}-t)^2\} \geq c\inf_{t \in \R}\E_{\P_0}\{(v^\T b_{(K)}-t)^2\} = c\|v\|^2,\\
v^\T I_K(\theta)v &\leq \E_{\P_\theta}\{(v^\T b_{(K)})^2\} \leq C\,\E_{\P_0}\{(v^\T b_{(K)})^2\} = C\|v\|^2,
\end{aligned}
\end{equation}
which is the second claim. Since $I_K = \nabla^2A_K$, by $\|u^\T s_\theta\|_\infty \leq 2C_\infty\sqrt K\|u\|$ and \eqref{eq:information-bounds}, the directional derivative $\partial_u\{v^\T I_K(\theta)v\} \coloneqq \partial_t\{v^\T I_K(\theta+tu)v\}|_{t = 0}$ along $u \in \R^K$ satisfies
\[
|\partial_u\{v^\T I_K(\theta)v\}| = |\E_{\P_\theta}\{(v^\T s_\theta)^2u^\T s_\theta\}| \leq 2C_\infty\sqrt K\|u\|\,v^\T I_K(\theta)v \leq C\sqrt K\|u\|\|v\|^2.
\]
Integrating this bound at $\theta+t(\theta'-\theta)$ with $u \coloneqq \theta'-\theta$ over $t \in [0,1]$, as $\|\theta+t(\theta'-\theta)\|_1 \leq B+1$, and using the symmetry of $I_K$, we obtain
\[
\|I_K(\theta')-I_K(\theta)\|_{\op} = \sup_{\|v\| = 1}|v^\T\{I_K(\theta')-I_K(\theta)\}v| \leq C\sqrt K\|\theta'-\theta\|,
\]
which is the third claim and completes the proof.
\end{proof}

\begin{lemma}\label{lem:regularity}
Under Assumption \ref{cond:general-queries}(i), there exists a constant $C$, depending only on $B$, $C_\infty$ and the constant in Assumption \ref{cond:general-queries}, such that, for all $K \geq 1$, $q \in \cQ$ and $\theta \in \R^K$ with $\|\theta\|_1 \leq B+1$, the Jacobian $D_q(\theta)$ of $\theta \mapsto \beta_q(\P_\theta)$ satisfies
\begin{equation}\label{eq:score-identity}
D_q(\theta) = \E_{\P_\theta}(\psi^\star_{\P_\theta,q}s_\theta^\T)\quad\text{and}\quad \|D_q(\theta)\|_{\op} \leq C,
\end{equation}
and, if Assumption \ref{cond:general-queries}(ii) also holds, then, for all $\theta' \in \R^K$ with $\|\theta'\|_1 \leq B+1$ and $\Delta \coloneqq \theta'-\theta$,
\begin{equation}\label{eq:query-taylor}
\|\beta_q(\P_{\theta'})-\beta_q(\P_\theta)-D_q(\theta)\Delta\| \leq C\sqrt K\|\Delta\|^2.
\end{equation}
\end{lemma}

\begin{proof}
We identify $D_q$ through the scores of the paths $t \mapsto \P_{\theta+tv}$ and then integrate the Lipschitz bound of Assumption \ref{cond:general-queries}(ii) along coefficient segments.

For $v \in \R^K$, by the density envelope, the path $t \mapsto \P_{\theta+tv}$ lies in $\cP$ for $|t|\|v\|_1 < 1$ and has score $v^\T s_\theta$ at $t = 0$. Hence, by Assumption \ref{cond:general-queries}(i), Cauchy--Schwarz and \eqref{eq:information-bounds}, we have
\[
\|D_q(\theta)v\| = \|\E_{\P_\theta}(\psi^\star_{\P_\theta,q}v^\T s_\theta)\| \leq \left\{\E_{\P_\theta}\|\psi^\star_{\P_\theta,q}\|^2\right\}^{\frac{1}{2}}\{v^\T I_K(\theta)v\}^{\frac{1}{2}} \leq C\|v\|.
\]
Since $\|\theta+t\Delta\|_1 \leq B+1$ for $t \in [0,1]$, integrating $\partial_t\beta_q(\P_{\theta+t\Delta}) = D_q(\theta+t\Delta)\Delta$ over $t \in [0,1]$ and applying Assumption \ref{cond:general-queries}(ii), we obtain
\[
\|\beta_q(\P_{\theta'})-\beta_q(\P_\theta)-D_q(\theta)\Delta\| \leq \int_0^1\|D_q(\theta+t\Delta)-D_q(\theta)\|_{\op}\|\Delta\|\,dt \leq C\sqrt K\|\Delta\|^2,
\]
which is \eqref{eq:query-taylor} and completes the proof.
\end{proof}

\subsubsection{Posterior concentration}\label{app:posterior-concentration}

\begin{lemma}[Posterior concentration]\label{lem:subgauss}
Under $\pi_K$ and Assumption \ref{cond:general-queries}(i), there exists a constant $s_0 < \infty$, depending only on $B$, $C_\infty$ and the constant in Assumption \ref{cond:general-queries}, such that, for all $N$, $K$, contexts $(\cD_N,q)$, $u \in \R^{d_k}$ with $\|u\| = 1$ and $t \in \R$,
\[
\log\E_N\exp\left[t\sqrt N\,u^\T\{\beta_q(\P_\theta)-m_\pi\}\right] \leq \frac{s_0^2t^2}{2},
\]
where $\E_N$ is the expectation over $\theta \sim p_N$.
\end{lemma}

\begin{proof}
We derive logarithmic Sobolev and Poincar\'e inequalities from the strong log-concavity of the posterior and then apply Herbst's argument to the Lipschitz function $\theta \mapsto \sqrt N\,u^\T\beta_q(\P_\theta)$.

By Lemma \ref{lem:geometry}, $\nabla^2\cV_N \succeq cNI$ on $\Theta_K$, so that $W(\theta) \coloneqq \cV_N(\theta)-cN\|\theta\|^2/2$ is convex on $\Theta_K$. We extend $W$ by its supporting planes and penalise the distance to $\Theta_K$, defining, for $x \in \R^K$ and $j \geq 1$,
\[
\widetilde W(x) \coloneqq \sup_{y \in \Theta_K}\{W(y)+\nabla W(y)^\T(x-y)\}\quad\text{and}\quad V_j(x) \coloneqq \frac{cN}{2}\|x\|^2+\widetilde W(x)+j\operatorname{dist}(x,\Theta_K)^2,
\]
where $\widetilde W = W$ on $\Theta_K$ by convexity and the slopes are bounded by compactness. Mollifying the last two terms of $V_j$ with radii at most one preserves the curvature $cN$ and, by a supporting plane of $\widetilde W$, the envelope, with $C$ depending on $N$, $K$ and $\cD_N$,
\[
e^{-V_j(x)} \leq \exp\{-cN\|x\|^2/2+C\|x\|+C\},\quad\text{for } x \in \R^K \text{ and } j \geq 1.
\]
Applying the logarithmic Sobolev inequality~(2.17) of \cite{ledoux1999} for these potentials to the $L$-Lipschitz extension $\inf_{y \in \Theta_K}\{f(y)+L\|\cdot-y\|\}$ of an $L$-Lipschitz $f$ on $\Theta_K$, and then removing the smoothing and letting $j \to \infty$, we obtain
\begin{equation}\label{eq:posterior-lsi}
\operatorname{Ent}_N(f^2) \leq \frac{2}{cN}\E_N\|\nabla f\|^2\quad\text{and}\quad\Var_N(f) \leq \frac{1}{cN}\E_N\|\nabla f\|^2,
\end{equation}
where $\operatorname{Ent}_N(f^2) \coloneqq \E_N(f^2\log f^2)-\E_N(f^2)\log\E_N(f^2)$, the limits follow by dominated convergence \citep{durrett2019}, and the variance bound by linearisation, as in Proposition~2.1 of \cite{ledoux1999}.

For an $L$-Lipschitz $F:\Theta_K \to \R$, applying Herbst's argument in Proposition~2.3 of \cite{ledoux1999} to \eqref{eq:posterior-lsi}, we have
\begin{equation}\label{eq:posterior-herbst}
\log\E_N e^{t\{F-\E_N(F)\}} \leq \frac{t^2L^2}{2cN},\quad \text{for } t \in \R.
\end{equation}
By Lemma \ref{lem:regularity}, $F \coloneqq \sqrt N\,u^\T\beta_q(\P_\theta)$ is $C\sqrt N$-Lipschitz on $\Theta_K$ for $\|u\| = 1$, and, applying \eqref{eq:posterior-herbst} with $L = C\sqrt N$ and $\E_NF = \sqrt N\,u^\T m_\pi$, we obtain the asserted bound with $s_0^2 \coloneqq C^2/c$.
\end{proof}

\subsubsection{The posterior mode}\label{app:posterior-mode}

The coefficient posterior density $p_N$, which is continuous on $\Theta_K$, the posterior expectation $\E_N$ and mode $\tilde\theta$, the average $\bbar = \P_Nb_{(K)}$, the radius $\rho_N$ and the set $\cE_N$ of regular contexts are as in Section \ref{app:theory-technical}. Let $\eta(\theta)$ denote the outward unit normal at $\theta \in \partial\Theta_K$, and define the integrated boundary density by
\[
B_{\partial,N} \coloneqq \int_{\partial\Theta_K}p_N(\theta)\,d\mathcal H^{K-1}(\theta),
\]
where $\mathcal H^{K-1}$ is the $(K-1)$-dimensional Hausdorff measure.

\begin{lemma}[Posterior mode comparison]\label{lem:laplace}
Under $\pi_K$ and Assumption \ref{cond:general-queries}(i)--(ii), if $K^3 \leq N$ and $\|\Lambda_K^{-1}\|_{\op} \leq C_\Lambda N$ for a constant $C_\Lambda > 0$, then there exists a constant $C$, depending only on $B$, $C_\infty$, $C_\Lambda$, $C_r$, $d_0 = \max_{k \in \cK}d_k$ and the constant in Assumption \ref{cond:general-queries}, such that, for every context $(\cD_N,q)$, the posterior moments $(m_\pi,S_\pi)$ and the mode pair $(g,V)$ of Definition \ref{def:reference-estimator} satisfy $\|S_\pi\|_F+\|V\|_F \leq C$ and
\[
N\|m_\pi-g\|^2 \leq CK\quad\text{and}\quad N\|m_\pi-g\| \leq C(B_{\partial,N}+K^{\frac{3}{2}}),
\]
and, if $(\cD_N,q) \in \cE_N$, then
\[
\sqrt N\|m_\pi-g\| \leq CK^{\frac{3}{2}}/\sqrt N\quad\text{and}\quad \|S_\pi-V\|_F \leq CK/\sqrt N.
\]
\end{lemma}

\begin{proof}
At every context, we bound the posterior moments about $\tilde\theta$ and compare $m_\pi$ with $g$ by the Gauss--Green formula, and on $\cE_N$ we show $B_{\partial,N} \leq N^{-8}$ and bound $\|S_\pi-V\|_F$.

By the variational inequality $(\theta-\tilde\theta)^\T\nabla\cV_N(\tilde\theta) \geq 0$ at the mode, strong convexity and the convexity of $\Theta_K$, we have, for $\theta \in \Theta_K$ and for $\theta$ on each boundary facet, respectively,
\[
(\theta-\tilde\theta)^\T\nabla\cV_N(\theta) \geq cN\|\theta-\tilde\theta\|^2\quad\text{and}\quad (\theta-\tilde\theta)^\T\eta(\theta) \geq 0.
\]
Applying the Gauss--Green formula \citep{evans2015} to the field $(\theta-\tilde\theta)e^{-\cV_N(\theta)}$, whose divergence is $\{K-(\theta-\tilde\theta)^\T\nabla\cV_N(\theta)\}e^{-\cV_N(\theta)}$, we obtain
\begin{equation}\label{eq:mode-moments}
cN\E_N\|\theta-\tilde\theta\|^2 \leq \E_N\{(\theta-\tilde\theta)^\T\nabla\cV_N(\theta)\} = K-\int_{\partial\Theta_K}(\theta-\tilde\theta)^\T\eta\,p_N\,d\mathcal H^{K-1} \leq K,
\end{equation}
so that $N\|m_\pi-g\|^2 \leq CN\E_N\|\theta-\tilde\theta\|^2 \leq CK$ by $m_\pi = \E_N\beta_q(\P_\theta)$, Jensen's inequality and $\|D_q\|_{\op} \leq C$ in \eqref{eq:score-identity}, and $\|S_\pi\|_F+\|V\|_F \leq C$ by Lemmas \ref{lem:subgauss}, \ref{lem:geometry} and \ref{lem:regularity}, as $d_k \leq d_0$.

To compare $\E_N\theta$ with $\tilde\theta$, let $H \coloneqq I_K(\tilde\theta)+\Lambda_K^{-1}/N$ and define the gradient remainder $r(\theta)$ by
\begin{equation}\label{eq:gradient-remainder}
\nabla\cV_N(\theta) = \nabla\cV_N(\tilde\theta)+NH(\theta-\tilde\theta)+r(\theta),
\end{equation}
where, by Lemma \ref{lem:geometry}, $\|r(\theta)\| \leq CN\sqrt K\|\theta-\tilde\theta\|^2$ and $\|\D_\theta r(\theta)\|_{\op} \leq CN\sqrt K\|\theta-\tilde\theta\|$. By \eqref{eq:mode-moments} and the Gauss--Green formula, we obtain
\begin{equation}\label{eq:mean-gradient-bounds}
\|\E_Nr\| \leq CK^{\frac{3}{2}}\quad\text{and}\quad \|\E_N\nabla\cV_N(\theta)\| = \left\|\int_{\partial\Theta_K}\eta p_N\,d\mathcal H^{K-1}\right\| \leq B_{\partial,N}.
\end{equation}
Since $\E_N\theta \in \Theta_K$ and $H \succeq cI$, we have
\begin{equation}\label{eq:mean-mode-gap}
\begin{aligned}
cN\|\E_N\theta-\tilde\theta\|^2 &\leq (\E_N\theta-\tilde\theta)^\T NH(\E_N\theta-\tilde\theta)\\
&\overset{\text{(i)}}{=} (\E_N\theta-\tilde\theta)^\T\{\E_N\nabla\cV_N(\theta)-\nabla\cV_N(\tilde\theta)-\E_Nr\}\\
&\overset{\text{(ii)}}{\leq} \|\E_N\theta-\tilde\theta\|(B_{\partial,N}+CK^{\frac{3}{2}}),
\end{aligned}
\end{equation}
where (i) averages \eqref{eq:gradient-remainder} under $\E_N$, and (ii) uses \eqref{eq:mean-gradient-bounds} and the variational inequality $(\E_N\theta-\tilde\theta)^\T\nabla\cV_N(\tilde\theta) \geq 0$. Combining \eqref{eq:mean-mode-gap} with \eqref{eq:query-taylor} and \eqref{eq:mode-moments}, we obtain
\begin{equation}\label{eq:boundary-mode-mean}
\|m_\pi-g\| \leq C\|\E_N\theta-\tilde\theta\|+C\sqrt K\,\E_N\|\theta-\tilde\theta\|^2 \leq \frac{C}{N}(B_{\partial,N}+K^{\frac{3}{2}}).
\end{equation}

We now let $(\cD_N,q) \in \cE_N$, so that $\nabla\cV_N(\tilde\theta) = 0$. As $cNI \preceq \nabla^2\cV_N \preceq CNI$ on $\Theta_K$ by Lemma \ref{lem:geometry} and $\|\Lambda_K^{-1}\|_{\op} \leq C_\Lambda N$, the normalising constant satisfies
\begin{equation}\label{eq:normalising-lower}
\int_{\Theta_K}e^{-\cV_N(\theta)+\cV_N(\tilde\theta)}\,d\theta \geq \int_{\|\theta-\tilde\theta\| \leq \rho_N}e^{-\frac{CN\|\theta-\tilde\theta\|^2}{2}}\,d\theta \geq \left(\frac{c}{N}\right)^{\frac{K}{2}},
\end{equation}
as $\Pr(\|G\| \leq \rho_N) \geq 1/2$ for $G \sim \cN(0,(CN)^{-1}I)$ by Markov's inequality and large $C_r$. As $\|\theta-\tilde\theta\| \geq \rho_N$ on $\partial\Theta_K$, combining \eqref{eq:normalising-lower} with $\mathcal H^{K-1}(\partial\Theta_K) = 2^K\sqrt KB^{K-1}/(K-1)!$, we obtain
\begin{align*}
B_{\partial,N} &\leq \left(\frac{N}{c}\right)^{\frac{K}{2}}\frac{2^K\sqrt K B^{K-1}}{(K-1)!} e^{-\frac{cN\rho_N^2}{2}}\\
&\leq \exp\left\{CK+\frac{K}{2}\log N+\frac{1}{2}\log K -cC_r^2K\log N\right\} \leq N^{-8},
\end{align*}
for sufficiently large $C_r$. Substituting $B_{\partial,N} \leq N^{-8}$ into \eqref{eq:boundary-mode-mean}, we obtain $\sqrt N\|m_\pi-g\| \leq CK^{\frac{3}{2}}/\sqrt N$.

It remains to bound the covariance error. For $v,w \in \R^K$ with $\|v\| = \|w\| = 1$, by \eqref{eq:posterior-lsi} and the bound on $\D_\theta r$, we have $\Var_N(v^\T\theta) \leq C/N$ and
\[
\Var_N(w^\T r) \leq \frac{C}{N}\E_N\|(\D_\theta r)^\T w\|^2 \leq CNK\,\E_N\|\theta-\tilde\theta\|^2 \leq CK^2,
\]
so that, by Cauchy--Schwarz, $\|\Cov_N(\theta,r)\|_{\op} \leq CK/\sqrt N$. Since $\nabla p_N = -p_N\nabla\cV_N$, integrating $\{\theta-\E_N(\theta)\}\nabla p_N(\theta)^\T$ by parts over $\Theta_K$, we have $\Cov_N\{\theta,\nabla\cV_N(\theta)\} = I-B_N$, where
\[
B_N \coloneqq \int_{\partial\Theta_K}\{\theta-\E_N(\theta)\}\eta(\theta)^\T p_N(\theta)\,d\mathcal H^{K-1}(\theta)\quad\text{satisfies}\quad \|B_N\|_{\op} \leq 2BN^{-8}.
\]
Substituting \eqref{eq:gradient-remainder} with $\nabla\cV_N(\tilde\theta) = 0$ and multiplying on the right by $H^{-1}$, we obtain
\begin{equation}\label{eq:coefficient-covariance}
\|N\Cov_N(\theta)-H^{-1}\|_{\op} = \|\{B_N+\Cov_N(\theta,r)\}H^{-1}\|_{\op} \leq C\left(N^{-8}+\frac{K}{\sqrt N}\right) \leq C\frac{K}{\sqrt N}.
\end{equation}
For the query, let $A \coloneqq D_q(\tilde\theta)$ and define the remainder $R_q(\theta)$ by
\begin{equation}\label{eq:query-remainder}
\beta_q(\P_\theta) = g+A(\theta-\tilde\theta)+R_q(\theta),
\end{equation}
where $\|R_q(\theta)\| \leq C\sqrt K\|\theta-\tilde\theta\|^2$ and $\|\D_\theta R_q(\theta)\|_{\op} \leq C\sqrt K\|\theta-\tilde\theta\|$ by \eqref{eq:query-taylor} and Assumption \ref{cond:general-queries}(ii). Applying \eqref{eq:posterior-lsi} to the $d_k \leq d_0$ coordinates and using \eqref{eq:mode-moments}, we have
\begin{equation}\label{eq:query-remainder-variance}
\E_N\|R_q-\E_N R_q\|^2 \leq \frac{C}{N}\E_N\|\D_\theta R_q\|_F^2 \leq \frac{CK}{N}\E_N\|\theta-\tilde\theta\|^2 \leq \frac{CK^2}{N^2},
\end{equation}
and, likewise, $\E_N\|A\{\theta-\E_N(\theta)\}\|^2 \leq C/N$. Expanding $N\Cov_N\{\beta_q(\P_\theta)\}$ by \eqref{eq:query-remainder} and applying Cauchy--Schwarz to the cross terms, we obtain
\begin{align*}
\|S_\pi-V\|_F &\leq \|A\{N\Cov_N(\theta)-H^{-1}\}A^\T\|_F+N\E_N\|R_q-\E_N R_q\|^2\\
&\quad+2N[\E_N\|A\{\theta-\E_N(\theta)\}\|^2\,\E_N\|R_q-\E_N R_q\|^2]^{\frac{1}{2}}\\
&\overset{\text{(i)}}{\leq} C\left\{\frac{K}{\sqrt N}+\frac{K^2}{N}\right\} \overset{\text{(ii)}}{\leq} C\frac{K}{\sqrt N},
\end{align*}
where (i) follows from \eqref{eq:coefficient-covariance}, \eqref{eq:query-remainder-variance} and the bound on $\E_N\|A\{\theta-\E_N(\theta)\}\|^2$, and (ii) from $K^3 \leq N$. This proves the bound on $\|S_\pi-V\|_F$ on $\cE_N$ and completes the proof.
\end{proof}

Let $\vartheta_k \sim \cN(0,\lambda_k)$, $k \in \N$, be independent, with $\lambda_k = \sigma^2k^{-2a}$ as in Section \ref{app:prior}, let $\vartheta_{1:K} \coloneqq (\vartheta_1,\cdots,\vartheta_K)^\T$ and, for $K \in \N\cup\{\infty\}$, define the truncation events $\cB_K \coloneqq \left\{\sum_{k = 1}^K|\vartheta_k| \leq B\right\}$, so that $\pi_K$ is the distribution of $\vartheta_{1:K}$ given $\cB_K$, with normalising constant $Z_K \coloneqq \Pr(\cB_K)$. On $\cB_\infty$, the series $\sum_{k \geq 1}\vartheta_kb_k$ converges uniformly on $\cZ$, and the population $\P_\vartheta$ with $d\P_\vartheta/d\P_0 \propto \exp(\sum_{k \geq 1}\vartheta_kb_k)$ lies in $\cP$ and satisfies $\|d\P_{\vartheta_{1:K}}/d\P_0-d\P_\vartheta/d\P_0\|_\infty \to 0$ as $K \to \infty$. Let $p_K$ denote the Lebesgue density of $\pi_K$ on $\R^K$.

\begin{lemma}\label{lem:prior-boundary}
Under $\pi_K$, there exists a constant $C$, depending only on $\sigma$, $a$ and $B$, such that, for all $N,K \geq 1$ and $t > 0$, we have $Z_K \geq C^{-1}$, $\pi_K\left\{\theta \in \R^K:B-t < \|\theta\|_1 \leq B\right\} \leq Ct$ and $\E_\pi\E_\P B_{\partial,N} \leq C\sqrt K$. Moreover, if $K = K_N \to \infty$ and measurable functions $\varphi_N:\R^K\times\cQ \to [0,\infty)$ satisfy $\sup_N\|\varphi_N\|_\infty < \infty$ and $\lim_{N \to \infty}\varphi_N(\vartheta_{1:K},q) = 0$ on $\cB_\infty$ for every $q \in \cQ$, then
\begin{equation}\label{eq:series-coupling}
\E_{\pi_K\otimes\pi_q}\varphi_N = o(1).
\end{equation}
\end{lemma}

\begin{proof}
We first bound $Z_K$ below by $\Pr(\cB_\infty)$, then bound the boundary layer of $\pi_K$ and the average boundary density, and finally deduce \eqref{eq:series-coupling}.

Since $a > 1$, we may choose $J$ with $\sum_{k > J}\E|\vartheta_k| < B/4$. By independence and Markov's inequality, we have
\[
\Pr(\cB_\infty) \geq \Pr\left(\sum_{k = 1}^J|\vartheta_k| \leq \frac{B}{2}\right)\Pr\left(\sum_{k > J}|\vartheta_k| \leq \frac{B}{2}\right) \geq \frac{1}{2}\Pr\left(\sum_{k = 1}^J|\vartheta_k| \leq \frac{B}{2}\right) > 0,
\]
and $Z_K \geq \Pr(\cB_\infty) > 0$ as $\cB_\infty \subseteq \cB_K$. By convolution, the density $\bar p_K$ of $\|\vartheta_{1:K}\|_1$ obeys the bound $\|\bar p_K\|_\infty \leq \sqrt{2/(\pi\lambda_1)}$ of the density of $|\vartheta_1|$, and hence, by \eqref{eq:prior},
\[
\pi_K\{B-t < \|\theta\|_1 \leq B\} = Z_K^{-1}\int_{(B-t)_+}^B\bar p_K(s)\,ds \leq Ct.
\]
For the boundary density, the likelihood ratio $\cL_N(\theta) \coloneqq e^{N\{\theta^\T\bbar-A_K(\theta)\}}$ satisfies $\E_{\P_0^{\otimes N}}\cL_N(\theta) = 1$. Averaging Bayes' formula over $\cD_N$ under $\E_\pi\E_\P$, in which the predictive density cancels, we have, for $\theta \in \Theta_K$,
\[
\E_\pi\E_\P\{p_N(\theta)\} = \int p_K(\theta)\cL_N(\theta)\,d\P_0^{\otimes N}(\cD_N) = p_K(\theta).
\]
Consequently, by Tonelli's theorem \citep{durrett2019} and the coarea formula \citep{evans2015} with factor $\sqrt K$ on each facet, we obtain
\begin{equation}\label{eq:boundary-density-average}
\E_\pi\E_\P B_{\partial,N} = \int_{\partial\Theta_K}p_K(\theta)\,d\mathcal H^{K-1}(\theta) = \frac{\sqrt K\,\bar p_K(B)}{Z_K} \leq C\sqrt K.
\end{equation}
Finally, let $q \sim \pi_q$ be independent of $\vartheta$. Since $\cB_\infty \subseteq \cB_K$ and $Z_K \geq \Pr(\cB_\infty)$, we have
\[
\E_{\pi_K\otimes\pi_q}\varphi_N = \frac{\E\{1_{\cB_K}\varphi_N(\vartheta_{1:K},q)\}}{Z_K} \leq \frac{\E\{1_{\cB_\infty}\varphi_N(\vartheta_{1:K},q)\}+C\Pr(\cB_K\setminus\cB_\infty)}{\Pr(\cB_\infty)},
\]
where $\E\{1_{\cB_\infty}\varphi_N(\vartheta_{1:K},q)\} = o(1)$ by dominated convergence \citep{durrett2019} and $\Pr(\cB_K\setminus\cB_\infty) = o(1)$ as $\cB_K \to \cB_\infty$, which proves \eqref{eq:series-coupling} and completes the proof.
\end{proof}

\begin{lemma}\label{lem:mode-localisation}
For all $N$, $K$, $\cD_N \in \cZ^N$ and $\theta \in \Theta_K$,
\begin{equation}\label{eq:mode-truth-displacement}
\|\tilde\theta-\theta\| \leq C\left(\|\bbar-\E_{\P_\theta}b_{(K)}\|+\frac{\|\Lambda_K^{-1}\theta\|}{N}\right),
\end{equation}
where $C$ depends only on $B$ and $C_\infty$. If, moreover, $K^3 \leq N$ and $\|\Lambda_K^{-1}\|_{\op} \leq C_\Lambda N$ for a constant $C_\Lambda > 0$, then
\[
\E_{\pi_K}\{\P_\theta(\cE_N^c)\} \leq C\left(K\sqrt{\frac{\log N}{N}}+N^{-4}\right),
\]
where $\P_\theta(\cE_N^c) \coloneqq \P_\theta^{\otimes N}\left\{\cD_N:(\cD_N,q) \notin \cE_N\right\}$ does not depend on $q$ and $C$ depends only on $\sigma$, $a$, $B$, $C_\infty$, $C_\Lambda$ and $C_r$.
\end{lemma}

\begin{proof}
We derive \eqref{eq:mode-truth-displacement} from the strong monotonicity of $\nabla\cV_N$ and then bound $\E_{\pi_K}\{\P_\theta(\cE_N^c)\}$ through the tails of $\bbar$ and of $\Lambda_K^{-1}\theta$.

By the strong monotonicity of $\nabla\cV_N$ on $\Theta_K$ in Lemma \ref{lem:geometry} and the variational inequality at $\tilde\theta$, we have
\[
cN\|\tilde\theta-\theta\|^2 \leq (\tilde\theta-\theta)^\T\{\nabla\cV_N(\tilde\theta)-\nabla\cV_N(\theta)\} \leq -(\tilde\theta-\theta)^\T\nabla\cV_N(\theta),
\]
which proves \eqref{eq:mode-truth-displacement}, as $\nabla\cV_N(\theta) = N\{\E_{\P_\theta}b_{(K)}-\bbar\}+\Lambda_K^{-1}\theta$. As $|b_k| \leq C_\infty$, applying Hoeffding's inequality \citep{hoeffding1963} under $\P_\theta^{\otimes N}$ to each $k \in [K]$ and a union bound, we obtain
\begin{equation}\label{eq:sufficient-statistic-tail}
\P_\theta\left\{\|\bbar-\E_{\P_\theta}b_{(K)}\| > c_b\sqrt{\frac{K\log N}{N}}\right\} \leq 2K\exp\left(-\frac{c_b^2\log N}{2C_\infty^2}\right) \leq 2N^{-4},
\end{equation}
for a large constant $c_b$, where the last inequality uses $K \leq N$. Let $Q_K \coloneqq \sum_{k = 1}^K\vartheta_k^2/\lambda_k \sim \chi_K^2$, so that $\|\Lambda_K^{-1}\vartheta_{1:K}\|^2 \leq C_\Lambda NQ_K$. By Markov's inequality and $\E e^{Q_K/4} = 2^{\frac{K}{2}}$, we have
\begin{equation}\label{eq:prior-score-tail}
\Pr\{Q_K > 2K\log 2+16\log N\} \leq N^{-4}.
\end{equation}
Since conditioning on $\cB_K$ multiplies probabilities by at most $Z_K^{-1} \lesssim 1$ by Lemma \ref{lem:prior-boundary}, combining \eqref{eq:sufficient-statistic-tail} and \eqref{eq:prior-score-tail} with \eqref{eq:mode-truth-displacement}, we obtain, for a constant $C_3$,
\[
\E_{\pi_K}\left[\P_\theta\left\{\|\tilde\theta-\theta\| > C_3\sqrt{\frac{K\log N}{N}}\right\}\right] \leq CN^{-4}.
\]
On the event $\left\{\|\tilde\theta-\theta\| \leq C_3\sqrt{K\log N/N}\right\}$, by the triangle inequality and $\|x\|_1 \leq \sqrt K\|x\|$ for $x \in \R^K$, we have
\[
\sup_{\|\theta'-\tilde\theta\| \leq \rho_N}\|\theta'\|_1 \leq \|\theta\|_1+\sqrt K(\|\tilde\theta-\theta\|+\rho_N) \leq \|\theta\|_1+(C_3+C_r)K\sqrt{\frac{\log N}{N}},
\]
so that $(\cD_N,q) \in \cE_N$ unless $\|\theta\|_1 > B-(C_3+C_r)K\sqrt{\log N/N}$. Consequently, by Lemma \ref{lem:prior-boundary}, we obtain
\begin{equation}\label{eq:irregular-context}
\begin{aligned}
\E_{\pi_K}\{\P_\theta(\cE_N^c)\} &\leq CN^{-4}+\pi_K\left\{B-(C_3+C_r)K\sqrt{\frac{\log N}{N}} < \|\theta\|_1 \leq B\right\}\\
&\leq C\left(K\sqrt{\frac{\log N}{N}}+N^{-4}\right),
\end{aligned}
\end{equation}
which is the asserted bound on $\E_{\pi_K}\{\P_\theta(\cE_N^c)\}$ and completes the proof.
\end{proof}

\subsubsection{Posterior expansion under the sieve prior}\label{app:posterior-expansion}

For a population $\P$ with $c_0 \leq d\P/d\P_0 \leq C_0$, where $c_0$ and $C_0$ are the density bounds of Section \ref{app:prior}, let $s_{\P,K} \coloneqq b_{(K)}-\E_\P b_{(K)}$ denote the centred features, with covariance $J_{\P,K} \coloneqq \E_\P(s_{\P,K}s_{\P,K}^\T)$, so that $cI \preceq J_{\P,K} \preceq CI$ as in \eqref{eq:information-bounds}. We define the projection coefficients and the projected influence function by
\[
C_{K,q}(\P) \coloneqq \E_\P(\psi^\star_{\P,q}s_{\P,K}^\T)J_{\P,K}^{-1}\quad\text{and}\quad \psi^K_{\P,q} \coloneqq C_{K,q}(\P)s_{\P,K},
\]
respectively, and, by the normal equations, $\psi^K_{\P,q}$ is the componentwise $L_2(\P)$ projection of $\psi^\star_{\P,q}$ onto $\spn\{b_k-\E_\P b_k:k \in [K]\}$. Since $\E_\P\{\psi^K_{\P,q}(\psi^\star_{\P,q}-\psi^K_{\P,q})^\T\} = 0$, we have
\begin{equation}\label{eq:projection-identity}
V_q^\star(\P)-\E_\P\{(\psi^K_{\P,q})^{\otimes2}\} = \E_\P\{(\psi^\star_{\P,q}-\psi^K_{\P,q})^{\otimes2}\}.
\end{equation}
Finally, with the empirical process $\G_N$ of Section \ref{app:proofs}, we define the bounded error of the efficient expansion of the posterior moments by
\begin{equation}\label{eq:posterior-moment-remainder}
\cR_N(\P,q,\cD_N) \coloneqq 1\wedge\{\|\sqrt N(m_\pi-\beta_q(\P))-\G_N\psi^\star_{\P,q}\|^2+\|S_\pi-V_q^\star(\P)\|_F^2\},
\end{equation}
and its projected counterpart $\cR_N^K$ by the same formula with $\psi^K_{\P,q}$ and $\E_\P\{(\psi^K_{\P,q})^{\otimes2}\}$ in place of $\psi^\star_{\P,q}$ and $V_q^\star(\P)$. We call the joint distribution of $(\theta,q,\cD_N)$ with $(\theta,q) \sim \pi_K\otimes\pi_q$ and $\cD_N \mid (\theta,q) \sim \P_\theta^{\otimes N}$ the simulator experiment.

\begin{proposition}[Posterior moments under the sieve prior]\label{prop:triangular}
Under $\pi_K$ and Assumption \ref{cond:general-queries}(i)--(ii), if $K^{3\vee2a} = o(N)$, then $\E_\pi\E_\P\cR_N^K = o(1)$, that is, under the simulator experiment,
\[
\sqrt N\{m_\pi-\beta_q(\P_\theta)\} = \G_N\psi^K_{\P_\theta,q}+o_\P(1)\quad\text{and}\quad S_\pi = \E_{\P_\theta}\{(\psi^K_{\P_\theta,q})^{\otimes2}\}+o_\P(1).
\]
\end{proposition}

\begin{proof}
We localise $\tilde\theta$ around the generating coefficient $\theta$, expand $\tilde\theta-\theta$ in the sieve scores and carry the expansion over to $(m_\pi,S_\pi)$, without using $K \to \infty$.

Since $\inf_KZ_K > 0$ by Lemma \ref{lem:prior-boundary}, conditioning on $\cB_K$ preserves moment bounds up to a constant, and hence
\begin{equation}\label{eq:prior-score-moment}
\E_{\pi_K}\|\Lambda_K^{-1}\theta\|^2 \leq C\E\|\Lambda_K^{-1}\vartheta_{1:K}\|^2 = C\sum_{k = 1}^K\lambda_k^{-1} \leq CK^{2a+1}.
\end{equation}
For $h \coloneqq \tilde\theta-\theta$, squaring \eqref{eq:mode-truth-displacement} and using $\E_{\P_\theta}\|\bbar-\E_{\P_\theta}b_{(K)}\|^2 = \tr\{I_K(\theta)\}/N \leq CK/N$, we obtain
\begin{equation}\label{eq:simulator-displacement}
\E_{\pi_K}\E_{\P_\theta}\|h\|^2 \leq C\left(\frac{K}{N}+\frac{K^{2a+1}}{N^2}\right) \leq C\frac{K}{N}.
\end{equation}
For large $N$, $K^3 \leq N$ and $\|\Lambda_K^{-1}\|_{\op} = \sigma^{-2}K^{2a} \leq N$, so that, by Lemma \ref{lem:mode-localisation} with $C_\Lambda = 1$, $\E_{\pi_K}\{\P_\theta(\cE_N^c)\} \leq C(K\sqrt{\log N/N}+N^{-4}) = o(1)$.

On $\cE_N$, let $I \coloneqq I_K(\theta)$, $H \coloneqq I+\Lambda_K^{-1}/N$, $A \coloneqq D_q(\theta)$ and $U_N \coloneqq N^{-\frac{1}{2}}\sum_{i = 1}^N\{b_{(K)}(Z_i)-\E_{\P_\theta}b_{(K)}\}$. Since $\nabla\cV_N(\tilde\theta) = 0$, with $\mathrm{rem}(h) \coloneqq \nabla A_K(\tilde\theta)-\nabla A_K(\theta)-Ih$, we obtain
\begin{equation}\label{eq:mode-expansion}
\sqrt N\,h = H^{-1}U_N-\frac{H^{-1}\Lambda_K^{-1}\theta}{\sqrt N}-H^{-1}\sqrt N\,\mathrm{rem}(h),
\end{equation}
where $\|\mathrm{rem}(h)\| \leq C\sqrt K\|h\|^2$ by Lemma \ref{lem:geometry}. To bound $H^{-1}\Lambda_K^{-1}\theta/\sqrt N$ in \eqref{eq:mode-expansion}, for $u \in \R^{d_k}$ with $\|u\| = 1$ and $z \coloneqq \Lambda_K^{-\frac{1}{2}}\vartheta_{1:K} \sim \cN(0,I)$, we define
\[
w(\theta) \coloneqq H^{-1}A^\T u\quad\text{and}\quad F(z) \coloneqq \Lambda_K^{-\frac{1}{2}}w(\Lambda_K^{\frac{1}{2}}z),
\]
where $w$ is extended from $\Theta_K$ by Euclidean projection, so that $\|w\| \leq C$ and $w$ is $C\sqrt K$-Lipschitz by Lemma \ref{lem:geometry}, \eqref{eq:score-identity} and Assumption \ref{cond:general-queries}(ii). For smooth $F$ and $\delta F \coloneqq z^\T F-\operatorname{div}F$, integrating by parts in $z \sim \cN(0,I)$, we have
\[
\E\{(\delta F)f\} = \E(F^\T\nabla f)\quad\text{and}\quad \partial_k\delta F = F_k+\delta(\partial_kF),
\]
and hence
\[
\E(\delta F)^2 = \sum_{k = 1}^K\E\{F_k[F_k+\delta(\partial_kF)]\} = \E\{\|F\|^2+\tr[(DF)^2]\}.
\]
By mollification and dominated convergence, this identity extends to the present $F$, and, substituting $z = \Lambda_K^{-\frac{1}{2}}\vartheta_{1:K}$, we obtain
\[
\E\{(\vartheta_{1:K}^\T\Lambda_K^{-1}w-\operatorname{div}w)^2\} = \E\{w^\T\Lambda_K^{-1}w+\tr[(Dw)^2]\} \leq C(K^{2a}+K^2),
\]
where $|\tr[(Dw)^2]| \leq K\|Dw\|_{\op}^2 \leq CK^2$ and $w^\T\Lambda_K^{-1}w \leq CK^{2a}$. Since also $(\operatorname{div}w)^2 \leq K^2\|Dw\|_{\op}^2 \leq CK^3$, conditioning on $\cB_K$ and summing over the $d_k \leq d_0$ query coordinates, we have
\begin{equation}\label{eq:prior-score-term}
\E_{\pi_K}\left\|\frac{AH^{-1}\Lambda_K^{-1}\theta}{\sqrt N}\right\|^2 \leq C\frac{K^{3\vee2a}}{N} = o(1).
\end{equation}
By \eqref{eq:simulator-displacement}, we have
\begin{equation}\label{eq:mode-remainder-mean}
\E_{\pi_K}\E_{\P_\theta}\{\sqrt N\|\mathrm{rem}(h)\|\} \leq C\sqrt{NK}\,\E_{\pi_K}\E_{\P_\theta}\|h\|^2 \leq C\frac{K^{\frac{3}{2}}}{\sqrt N} = o(1).
\end{equation}
Moreover, $M \coloneqq H^{-1}-I^{-1} = -H^{-1}\Lambda_K^{-1}I^{-1}/N$ and, conditionally on $(\theta,q)$, $U_N$ is centred with covariance $I$, so that
\begin{equation}\label{eq:resolvent-error}
\E_{\P_\theta}\|AMU_N\|^2 = \tr(AMIMA^\T) \leq Cd_0\|A\|_{\op}^2\|M\|_{\op}^2 \leq C\frac{K^{4a}}{N^2} = o(1).
\end{equation}
By \eqref{eq:score-identity}, $AI^{-1} = C_{K,q}(\P_\theta)$ and hence $AI^{-1}U_N = \G_N\psi^K_{\P_\theta,q}$. Therefore, on $\cE_N$, we obtain
\begin{equation}\label{eq:simulator-mean-expansion}
\begin{aligned}
\sqrt N\{m_\pi-\beta_q(\P_\theta)\} &= \sqrt N(m_\pi-g)+\sqrt N\{\beta_q(\P_{\tilde\theta})-\beta_q(\P_\theta)-Ah\}+A\sqrt N\,h\\
&\overset{\text{(i)}}{=} O\left(\frac{K^{\frac{3}{2}}}{\sqrt N}+\sqrt{NK}\|h\|^2\right)+AI^{-1}U_N+o_\P(1) \\ &\overset{\text{(ii)}}{=} \G_N\psi^K_{\P_\theta,q}+o_\P(1),
\end{aligned}
\end{equation}
where (i) follows from Lemma \ref{lem:laplace}, \eqref{eq:query-taylor}, \eqref{eq:mode-expansion}, \eqref{eq:prior-score-term}, \eqref{eq:mode-remainder-mean} and \eqref{eq:resolvent-error}, and (ii) from \eqref{eq:simulator-displacement}. By Lemma \ref{lem:laplace}, Assumption \ref{cond:general-queries}(ii) and the resolvent identity, we have
\begin{equation}\label{eq:simulator-covariance}
\begin{aligned}
\|S_\pi-\E_{\P_\theta}\{(\psi^K_{\P_\theta,q})^{\otimes2}\}\|_F &\leq \|S_\pi-V\|_F+\|V-AI^{-1}A^\T\|_F\\
&\leq C\left(\frac{K}{\sqrt N}+\sqrt K\|h\|+\frac{K^{2a}}{N}\right) = o_\P(1),
\end{aligned}
\end{equation}
where $\E_{\P_\theta}\{(\psi^K_{\P_\theta,q})^{\otimes2}\} = AI^{-1}A^\T$ by \eqref{eq:score-identity}. As $\E_{\pi_K}\{\P_\theta(\cE_N^c)\} = o(1)$ and $0 \leq \cR_N^K \leq 1$, by \eqref{eq:simulator-mean-expansion} and \eqref{eq:simulator-covariance}, we obtain $\E_\pi\E_\P\cR_N^K = o(1)$, which proves the proposition.
\end{proof}

\subsubsection{Change of prior}\label{app:simulator-prior}

The lemmas below compare the posterior under a prior $\pi_P$ with that under $\pi_K$ and reduce expectations under $\pi_P$ to the simulator experiment. In this subsection, $\theta_\P$ and $g_{\P,K}$ are the sieve coefficients and remainder of $\P \in \cP$ in Section \ref{sec:target-efficiency}, $\cP_\pi$ is the support of $\pi_P$, and $w_\pi$, $C_w$ and $L_N$ are as in Assumption \ref{cond:posterior-tails}.

At every context $(\cD_n,q)$, let $f_{\pi_K} \coloneqq (m_{\pi_K},S_{\pi_K})$ denote the posterior moments \eqref{eq:posterior-moment} under $\pi_K$ and $\E_n$ the expectation over the coefficient posterior of Section \ref{app:theory-technical} at size $n$, with density proportional to $e^{-\cV_n}$ on $\Theta_K$. We define the sieve approximation error in \eqref{eq:learning-rate} by $\epsilon_{\pi,N} \coloneqq L_N^2/N+N\delta_\pi^2$, with $\delta_\pi = \sup_{\P \in \cP_\pi}\|g_{\P,K}\|_{L_2(\P_0)}$ of \ref{A3}, and, except in the mode pair $(g,V)$, $g$ denotes a remainder $g_{\P',K}$ with $\P' \in \cP_\pi$.

For $\theta \in \Theta_K$ and bounded $g:\cZ \to \R$, let $\P_{\theta,g}$ denote the tilt with $d\P_{\theta,g}/d\P_\theta \coloneqq e^g/\E_{\P_\theta}e^g$. If $g = g_{\P',K}$ for some $\P' \in \cP_\pi$, then $|\theta^\T b_{(K)}+g| \leq C_\infty(B+1)$ once $h_N \leq C_\infty$, for the bound $\|g\|_\infty \leq h_N$ of Assumption \ref{cond:simulator-prior}, so that $\P_{\theta,g} \in \cP$ by the envelope argument in Section \ref{app:prior}.

\begin{lemma}\label{lem:shift}
Let $P_1$ and $P_2$ be probability measures on a common measurable space with $P_2 \ll P_1$ and $\Gamma \coloneqq 1+\chi^2(P_2\,\|\,P_1) < \infty$. Then every measurable $\varphi$ with values in $[0,1]$ satisfies $\E_{P_2}\varphi \leq \E_{P_1}\varphi+\{(\Gamma-1)\E_{P_1}\varphi\}^{\frac{1}{2}}$.
\end{lemma}

\begin{proof}
We compare the two expectations through $w \coloneqq dP_2/dP_1$, which satisfies $\E_{P_1}\{(w-1)^2\} = \Gamma-1$. Since $\varphi^2 \leq \varphi$, by Cauchy--Schwarz, we have
\[
\E_{P_2}\varphi-\E_{P_1}\varphi = \E_{P_1}\{(w-1)\varphi\} \leq [\E_{P_1}\{(w-1)^2\}\,\E_{P_1}(\varphi^2)]^{\frac{1}{2}} \leq \{(\Gamma-1)\E_{P_1}\varphi\}^{\frac{1}{2}},
\]
which proves the lemma.
\end{proof}

\begin{lemma}[Reduction to the sieve prior]\label{lem:prior-reduction}
Under Assumptions \ref{cond:simulator-prior} and \ref{cond:posterior-tails}, every measurable $\varphi:\cP\times\cQ\times\cZ^N \to [0,\infty]$ satisfies
\begin{equation}\label{eq:prior-reduction}
\E_{\pi_P\otimes\pi_q}\E_\P\varphi \leq C_w\sup_g\E_{\pi_K\otimes\pi_q}\E_{\P_{\theta,g}}\varphi(\P_{\theta,g},q,\cD_N),
\end{equation}
where $g$ ranges over $\left\{g_{\P',K}:\P' \in \cP_\pi\right\}$. Moreover, there exists a constant $C$ such that, for every sequence $g_N \in \left\{g_{\P',K}:\P' \in \cP_\pi\right\}$, the tilts $\P \coloneqq \P_{\theta,g_N}$ satisfy $\|d\P/d\P_\theta-1\|_\infty \leq e^{2h_N}-1$, $\chi^2(\P\,\|\,\P_\theta) \leq Cc_N/N$ and $\chi^2(\P^{\otimes N}\,\|\,\P_\theta^{\otimes N}) = o(1)$ uniformly in $\theta \in \Theta_K$, where $c_N$ and $h_N$ are the sequences of Assumption \ref{cond:simulator-prior}.
\end{lemma}

\begin{proof}
We derive \eqref{eq:prior-reduction} from the conditional density of the coefficient and then bound the divergences of the tilts from the sieve populations.

By Assumption \ref{cond:posterior-tails}, the coefficient $\theta_\P$ has density $w_\pi \leq C_w$ relative to $\pi_K$ given $g_{\P,K}$, and $\P = \P_{\theta_\P,g_{\P,K}}$, which proves \eqref{eq:prior-reduction}. For bounded $\varphi$, the supremum is $o(1)$ if the expectation is $o(1)$ along every sequence $g_N$. For the tilts, by Assumption \ref{cond:simulator-prior}, we have
\begin{equation}\label{eq:tilt-chi-square}
\begin{aligned}
\left|\frac{d\P}{d\P_\theta}-1\right| &= \left|\frac{e^{g_N}}{\E_{\P_\theta}e^{g_N}}-1\right| \leq e^{2h_N}-1,\\
\chi^2(\P\,\|\,\P_\theta) &= \frac{\Var_{\P_\theta}(e^{g_N})}{(\E_{\P_\theta}e^{g_N})^2} \overset{\text{(i)}}{\leq} e^{4h_N}\E_{\P_\theta}(g_N^2) \overset{\text{(ii)}}{\leq} \frac{Cc_N}{N},\\
\chi^2(\P^{\otimes N}\,\|\,\P_\theta^{\otimes N}) &= \{1+\chi^2(\P\,\|\,\P_\theta)\}^N-1 \overset{\text{(iii)}}{\leq} e^{Cc_N}-1 = o(1),
\end{aligned}
\end{equation}
where (i) follows from $\Var(e^{g_N}) \leq \E(e^{g_N}-1)^2$, $|e^x-1| \leq |x|e^{|x|}$ and $\E_{\P_\theta}e^{g_N} \geq e^{-h_N}$, (ii) from the density envelope and Assumption \ref{cond:simulator-prior}, and (iii) from $1+x \leq e^x$, uniformly in $\theta \in \Theta_K$, which completes the proof.
\end{proof}

\begin{lemma}[Stability of the posterior]\label{lem:prior-stability}
Under Assumptions \ref{cond:general-queries}(i), \ref{cond:simulator-prior} and \ref{cond:posterior-tails}, there exist constants $C$ and $s_0$, independent of $N$, $K$, $n$, $q$ and $\cD_n$, such that, for all sufficiently large $N$, all $n \in \cN$ and all contexts $(\cD_n,q)$,
\begin{equation}\label{eq:prior-stability}
n\|m_\pi-m_{\pi_K}\|^2+\|S_\pi-S_{\pi_K}\|_F^2 \leq C\epsilon_{\pi,N},
\end{equation}
and, for all $u \in \R^{d_k}$ with $\|u\| = 1$ and $t \in \R$,
\[
\log\E_\pi\left(\exp\left[t\sqrt n\,u^\T\{\beta_q(\P)-m_\pi\}\right]\,\middle|\,\cD_n,q\right) \leq \frac{s_0^2t^2}{2}.
\]
\end{lemma}

\begin{proof}
We bound the density $\rho$ of the posterior of $\theta_\P$ under $\pi_P$ relative to that under $\pi_K$, deduce a bound on $\E_n\{(\rho-1)^2\}$, and then compare moments and tails.

For each $n$, $\cD_n$ and $q$, let $\mu$ denote the distribution of $g_{\P, K}$ under $\pi_P$. Since $\P_{\theta,g}$ has density $e^g/\E_{\P_\theta}e^g$ relative to $\P_\theta$, the posterior of $\theta_\P$ under $\pi_P$ has a density $\rho$ relative to that under $\pi_K$, which satisfies
\begin{equation}\label{eq:posterior-density-ratio}
\rho(\theta) \propto \int w_\pi(\theta\mid g)\exp\left\{\sum_{i=1}^ng(Z_i)-n\log\E_{\P_\theta}e^g\right\}\mu(dg).
\end{equation}
Since the functions $b_k$ are orthonormal in $L_2(\P_0)$, by Bessel's inequality we have
\begin{equation}\label{eq:tilt-moment-shift}
\|\nabla_\theta\log\E_{\P_\theta}e^g\| = \|\E_{\P_{\theta,g}}b_{(K)}-\E_{\P_\theta}b_{(K)}\| \leq \|d\P_{\theta,g}/d\P_0-d\P_\theta/d\P_0\|_{L_2(\P_0)} \leq C\|g\|_{L_2(\P_0)},
\end{equation}
where the last inequality uses the density envelope and $\|g\|_\infty \leq h_N \leq 1$. By Assumption \ref{cond:posterior-tails} and \eqref{eq:tilt-moment-shift}, the logarithm of each integrand in \eqref{eq:posterior-density-ratio} is $L_N'$-Lipschitz in $\theta$ for $L_N' \coloneqq L_N+Cn\delta_\pi$, and hence so is $\log\rho$ on $\Theta_K$.

We next bound the $\chi^2$ divergence. Normalising $\rho$ so that $\E_n(\rho) = 1$ and applying \eqref{eq:posterior-herbst} to $\log\rho$ at $t = 2$, we have
\[
\E_n(\rho^2) \overset{\text{(i)}}{\leq} \frac{\E_n(\rho^2)}{e^{2\E_n(\log\rho)}} \leq \exp\left(\frac{2L_N'^2}{cn}\right),
\]
where (i) holds because $\E_n(\log\rho) \leq \log\E_n(\rho) = 0$ by Jensen's inequality. Since $n \asymp N$ and $L_N'^2/n \lesssim L_N^2/N+N\delta_\pi^2$ is bounded, it follows that
\begin{equation}\label{eq:posterior-chi-square}
\chi_n^2 \coloneqq \E_n\{(\rho-1)^2\} \leq \frac{CL_N'^2}{n} \leq C\epsilon_{\pi,N}.
\end{equation}

For $u \in \R^{d_k}$ with $\|u\| = 1$, let $G \coloneqq \sqrt n\,u^\T\{\beta_q(\P_\theta)-m_{\pi_K}\}$, so that $\E_n(G) = 0$, $\E_n(G^2) \leq s_0^2$ and $\E_n(G^4) \leq 16s_0^4$ by Lemma \ref{lem:subgauss}. By Cauchy--Schwarz and \eqref{eq:posterior-chi-square}, we have
\begin{equation}\label{eq:tilted-G-moments}
|\E_n(\rho G)| = |\E_n\{(\rho-1)G\}| \leq s_0\chi_n\quad\text{and}\quad|\E_n(\rho G^2)-\E_n(G^2)| \leq 4s_0^2\chi_n.
\end{equation}
By \eqref{eq:query-lipschitz}, $\|\beta_q(\P_{\theta,g})-\beta_q(\P_\theta)\| \leq C\|g\|_{L_2(\P_0)} \leq C\delta_\pi$, and, since $\E_n(\rho G^2) \leq C$ by \eqref{eq:tilted-G-moments}, we obtain
\begin{align*}
|\sqrt n\,u^\T(m_\pi-m_{\pi_K})-\E_n(\rho G)| &\leq C\sqrt n\,\delta_\pi,\\
|u^\T(S_\pi-S_{\pi_K})u-\E_n(\rho G^2)+\E_n(G^2)+\{\E_n(\rho G)\}^2| &\leq C(\sqrt n\,\delta_\pi+n\delta_\pi^2).
\end{align*}
Taking the supremum over $u \in \R^{d_k}$ with $\|u\| = 1$ and using $d_k \leq d_0$, we obtain \eqref{eq:prior-stability}.

For the sub-Gaussian bound, let $F \coloneqq \sqrt n\,u^\T\beta_q(\P_\theta)$ for $\|u\| = 1$, which is $C\sqrt n$-Lipschitz in $\theta$ by Lemma \ref{lem:regularity}. By Cauchy--Schwarz, for $t \in \R$, we have
\[
\E_n\{\rho e^{t(F-\E_nF)}\} \leq \{\E_n(\rho^2)\}^{\frac{1}{2}}\{\E_ne^{2t(F-\E_nF)}\}^{\frac{1}{2}} \leq Ce^{Ct^2},
\]
where the last inequality uses \eqref{eq:posterior-chi-square} and \eqref{eq:posterior-herbst} for $F$. Under the posterior for $\pi_P$, by the bound on $\|\beta_q(\P_{\theta,g})-\beta_q(\P_\theta)\|$ and Cauchy--Schwarz, $X \coloneqq \sqrt n\,u^\T\{\beta_q(\P)-m_\pi\}$ satisfies
\begin{align*}
|X-(F-\E_nF)| &\leq |X-\{F-\E_n(\rho F)\}|+|\E_n(\rho F)-\E_nF|\\
&\leq C\sqrt n\,\delta_\pi+\chi_n[\E_n\{(F-\E_nF)^2\}]^{\frac{1}{2}} \leq C,
\end{align*}
so that the centred variable $X$ has moment generating function at most $\exp\{C(1+|t|+t^2)\}$. Since $e^x \leq 1+x+x^2e^{|x|}/2$, this function is at most $1+Ct^2$ for $|t| \leq 1$, and the asserted bound follows with $s_0$ depending only on $C$.
\end{proof}

\section{Proofs for Section \ref{sec:efficiency}}\label{app:proofs-efficiency}

This section proves Theorems \ref{thm:TF-error}, \ref{thm:main} and \ref{thm:fixed-population} for every prior satisfying Assumptions \ref{cond:simulator-prior} and \ref{cond:posterior-tails}.

As in Section \ref{app:proofs}, $N \geq 3$, $T \geq 2$, constants $c$ and $C$ may vary between bounds and, at size $n$, $\E_\pi\E_\P$ averages over $(\P,q) \sim \pi$ and $\cD_n \sim \P^{\otimes n}$. For outputs $f = (f^\beta,f^V)$ and $f' = (f'^\beta,f'^V)$ at the same context $(\cD_N,q)$, we define the moment error
\begin{equation}\label{eq:moment-error}
e(f,f') \coloneqq N\|f^\beta-f'^\beta\|^2+\|f^V-f'^V\|_F^2.
\end{equation}
For the empirical risk $\hat R_T$ of Section \ref{sec:method}, let $R(f) \coloneqq \E\{\hat R_T(f)\}$ denote the population pretraining risk. For the class $\cF_{\mathrm{TF}} = \cF_{\mathrm{TF}}(L,J,d_e,d_f,W)$ of Definition \ref{def:transformer-readout}, with $\Omega \subseteq [-W,W]^p$ and $W \geq 1$, we define the approximation error and the pretraining error by
\begin{align}
&\epsilon_{\mathrm{app}} \coloneqq \inf_{f \in \cF_{\mathrm{TF}}}\{R(f)-R(f_\pi^\circ)\}\quad\text{and}\label{eq:risk-benchmark}\\
&\tilde r_N \coloneqq \epsilon_{\mathrm{app}}+\epsilon_{\mathrm{opt}}+\frac{pL^2\log(2pTWn_{\max})}{T},\label{eq:learning-budget}
\end{align}
respectively.

\subsection{Proof of Theorem \ref{thm:TF-error}}\label{app:proof-tf-error}

Let $X \coloneqq (\cD_n,q)$ denote a context independent of $\cT$, with $(\P,q,n) \sim \pi\otimes\pi_n$ and $\cD_n \sim \P^{\otimes n}$, at which $e$ is evaluated with $n$ in place of $N$. For $\beta_0 \in \R^{d_k}$ and an output $f$ with $v_-I \preceq f^V \preceq v_+I$, where $[v_-,v_+]$ is the covariance range of Section \ref{app:architecture}, we define the score difference
\begin{equation}\label{eq:score-difference}
g_f(X,\beta_0) \coloneqq \ell_n\{\beta_0;f^\beta(X),f^V(X)\}-\ell_n\{\beta_0;m_\pi(X),\clip S_\pi(X)\}.
\end{equation}
We let $h_f(X) \coloneqq \E_\pi\{g_f(X,\beta_q(\P))\mid X\}$ and $\hat\E_T(\varphi) \coloneqq T^{-1}\sum_{t = 1}^T\varphi\{(\cD^{(t)}_{n_t},q_t),\beta_{0,t}\}$ denote the conditional excess score in \eqref{eq:posterior-loss-regret} and the empirical mean over $\cT$ of a function $\varphi$ of $(X,\beta_0)$, respectively, so that $h_f \geq 0$, $\E h_f = R(f)-R(f_\pi^\circ)$ and $\hat\E_T(g_f) = \hat R_T(f)-\hat R_T(f_\pi^\circ)$.

\begin{proof}
We decompose the truncated excess risk of $\hat f$ into approximation, optimisation and stochastic errors, bound the first by approximating the mode pair $(g,V)$ of Definition \ref{def:reference-estimator} and the last by metric entropy and localisation, and then transfer the bound to $\nu$.

\paragraph{Step 1: Risk decomposition \eqref{eq:risk-decomposition}.}
Let $\cJ(f) \coloneqq \E(1\wedge h_f)$ denote the truncated excess risk, so that $\E\{1\wedge e(\hat f,f_\pi^\circ)\mid\cT\} \lesssim \cJ(\hat f)$ by Lemma \ref{lem:central}. For a deterministic $f^\circ \in \cF_{\mathrm{TF}}$ and $c' > 0$, we have
\begin{equation}\label{eq:risk-decomposition}
\begin{aligned}
\cJ(\hat f) &= c'\hat\E_T(g_{\hat f})+\{\cJ(\hat f)-c'\hat\E_T(g_{\hat f})\}\\
&\leq c'\hat\E_T(g_{f^\circ})+c'\epsilon_{\mathrm{opt}}+\sup_{f \in \cF_{\mathrm{TF}}}\left\{\cJ(f)-c'\hat\E_T(g_f)\right\},
\end{aligned}
\end{equation}
where the inequality follows from the basic inequality $\hat\E_T(g_{\hat f}) \leq \hat\E_T(g_{f^\circ})+\epsilon_{\mathrm{opt}}$ implied by \eqref{eq:erm}. The three terms in the last line are the approximation, optimisation and stochastic errors, respectively.

\paragraph{Step 2: Approximation error \eqref{eq:approximation-error}.}
For large $N$, $K^3\vee\sigma^{-2}K^{2a} \leq n_{\min}$, so that, by Lemma \ref{lem:network}, a network $\bar f = (\bar f^\beta,\clip C^\circ)$ satisfies \eqref{eq:architecture-rate} and $\sup_{n \in \cN}a_n(\bar f,C^\circ) \leq (4n_{\min})^{-1}$. As $\bar f \in \cF_N$ of Section \ref{app:proof-finite-transformer} for $\Omega \coloneqq [-W,W]^p$, with $W$ its coefficient bound, we obtain
\begin{equation}\label{eq:approximation-error}
\begin{aligned}
\epsilon_{\mathrm{app}} \leq R(\bar f)-R(f_\pi^\circ) &= \sum_{n \in \cN}\pi_n(n)\E_\pi\E_\P h_{\bar f}\\
&\overset{\text{(i)}}{\lesssim} \sum_{n \in \cN}\pi_n(n)\E_\pi\E_\P\{n_{\min}^{-2}+n\|m_\pi-g\|^2+\|S_\pi-V\|_F^2\}\\
&\overset{\text{(ii)}}{\lesssim} \frac{K^3}{N}+K\sqrt{\frac{\log N}{N}}+\epsilon_{\pi,N},
\end{aligned}
\end{equation}
where (i) follows from \eqref{eq:raw-covariance-comparison} at size $n$ and the triangle inequality, as in \eqref{eq:comparator-score-split}, and (ii) from the proof of Lemma \ref{lem:posterior-comparison} under $\pi_K$, where $\epsilon_{\pi,N} = 0$, from Lemma \ref{lem:prior-mode} otherwise, and from $n_{\min}^{-2} \leq K^3/n_{\min}$ and $n_{\min} \asymp N$.

\paragraph{Step 3: Metric entropy of the loss class.}
Let $\delta \in (0,1/2)$, and let $\bar z$, $H_\delta$ and $u_\delta$ be as in \eqref{eq:pretraining-support}. By Lemma \ref{lem:entropy} with $u = T^{-1}$, $J+d_e+d_f \leq pW$, $W \geq 1$ and $\log(1+\bar z) \lesssim \log(2n_{\max}T/\delta)$, we have
\begin{equation}\label{eq:entropy-bound}
H_\delta+u_\delta \lesssim pL^2\log(2pTWn_{\max}\delta^{-1}).
\end{equation}
For the last term of \eqref{eq:learning-budget}, $\cF_N$ has the architecture in the statement, with $J = 1$ and constant input and output dimensions, so that $p \asymp Ld_e(d_e+d_f)$, and, as $K \leq n_{\max} \asymp N$, we have
\begin{equation}\label{eq:entropy-rate}
pL^2\log(2pTWn_{\max}) \lesssim K^2\log^{2d+6}n_{\max}\log(2NT) \lesssim K^2\log^{2d+7}(2NT),
\end{equation}
where the first inequality follows from \eqref{eq:architecture-rate} and $p,W \lesssim n_{\max}^C$, and the second from $n_{\max} \asymp N$. Substituting \eqref{eq:approximation-error} and \eqref{eq:entropy-rate} into \eqref{eq:learning-budget}, we obtain
\begin{equation}\label{eq:explicit-learning-budget}
\tilde r_N \lesssim \frac{K^3}{N}+K\sqrt{\frac{\log N}{N}}+\frac{K^2\log^{2d+7}(2NT)}{T}+\epsilon_{\pi,N}+\epsilon_{\mathrm{opt}} \lesssim r_N,
\end{equation}
where $r_N$ is as in \eqref{eq:learning-rate} and the last inequality uses $\log(2NT) \leq 2\log(NT)$ and the definition $\epsilon_{\pi,N} = L_N^2/N+N\delta_\pi^2$ in Section \ref{app:simulator-prior}, with $L_N$ of Assumption \ref{cond:posterior-tails} and $\delta_\pi$ of \ref{A3}. Moreover, we have
\[
r_N \leq \frac{K^3}{N}+K\sqrt{\frac{\log N}{N}}+4^{2d+7}K^2\log^{2d+7}N\left(\frac{1}{T}+\frac{1}{N^3}\right)+\epsilon_{\pi,N}+\epsilon_{\mathrm{opt}} = o(1),
\]
where the inequality uses $\log(NT) \leq 4\log N$ if $T \leq N^3$ and the decrease of $t \mapsto t^{-1}\log^{2d+7}(Nt)$ for $\log(Nt) > 2d+7$ if $T > N^3$, and the equality the conditions of Theorem \ref{thm:TF-error}.

\paragraph{Step 4: Localised stochastic error \eqref{eq:learning-delta}.}
Let $f^\circ \in \cF_{\mathrm{TF}}$ be deterministic with $\E h_{f^\circ} \leq \epsilon_0 = \epsilon_{\mathrm{app}}+T^{-1}$. Bounding the terms of \eqref{eq:risk-decomposition} with $c' = \tau_0/c_2$ as in \eqref{eq:net-comparison} by Lemma \ref{lem:general-comparison} and \eqref{eq:entropy-bound}, we obtain, on an event $\cE_T(\delta)$ with $\Pr\{\cE_T(\delta)\} \geq 1-\delta$,
\[
\begin{aligned}
\pi_n(N)\E_\pi\E_\P\{1\wedge e(\hat f,f_\pi^\circ)\} &\overset{\text{(i)}}{\leq} \E\{1\wedge e(\hat f,f_\pi^\circ)\mid\cT\}\\
&\overset{\text{(ii)}}{\lesssim} \epsilon_{\mathrm{opt}}+\frac{pL^2\log(2pTWn_{\max}\delta^{-1})}{T}+\frac{\epsilon_{\mathrm{app}}+T^{-1}}{\delta},
\end{aligned}
\]
where (i) keeps the term $n = N$ and (ii) follows from Lemma \ref{lem:general-comparison} and \eqref{eq:entropy-bound}. As $\pi_n(N) \geq c$, $T^{-1} \lesssim \tilde r_N$ and $\log(1/\delta) \leq \delta^{-1}$, we have, on $\cE_T(\delta)$,
\begin{equation}\label{eq:learning-delta}
\E_\pi\E_\P\{1\wedge e(\hat f,f_\pi^\circ)\} \leq C\delta^{-1}\tilde r_N.
\end{equation}
Besides bounds depending only on $\cP$, this step uses the prior only through Lemma \ref{lem:subgauss}, which extends to the priors of Assumptions \ref{cond:simulator-prior} and \ref{cond:posterior-tails} by Lemma \ref{lem:prior-stability}, and hence so does \eqref{eq:learning-delta}.

\paragraph{Step 5: Transfer to $\nu$ in \eqref{eq:learning}.}
We let $\cL_{\pi,n}$ and $\cL_{\nu,N}$ denote the distributions of $(\cD_n,q)$ for $(\P,q) \sim \pi$ and of $(\cD_N,q)$ for $(\P,q) \sim \nu$, respectively, where $\cD_n \sim \P^{\otimes n}$ given $(\P,q)$, and define the overlap factor
\begin{equation}\label{eq:overlap-factor}
\Gamma_N(\nu) \coloneqq 1+\chi^2(\cL_{\nu,N}\,\|\,\cL_{\pi,N}).
\end{equation}
Let $\cT \in \cE_T(\delta)$ and $0 < u \leq 1$. Since $1_{\{e > u^2\}} \leq (1\wedge e)/u^2$, by \eqref{eq:learning-delta}, we have
\[
\gamma_u \coloneqq \cL_{\pi,N}\{e(\hat f,f_\pi^\circ) > u^2\} \leq \frac{C\tilde r_N}{\delta u^2}.
\]
As $(\P,q,\cD_N)$ is independent of $\cT$, applying Lemma \ref{lem:shift} to $\varphi = 1_{\{e(\hat f,f_\pi^\circ) > u^2\}}$, we obtain
\begin{equation}\label{eq:tracking-transfer}
\Pr\nolimits_\nu\{e(\hat f,f_\pi^\circ)^{\frac{1}{2}} > u\mid\cT\} \leq \gamma_u+\{(\Gamma_N(\nu)-1)\gamma_u\}^{\frac{1}{2}} \leq \frac{C\tilde r_N}{\delta u^2}+\frac{\{C(\Gamma_N(\nu)-1)\tilde r_N\}^{\frac{1}{2}}}{\sqrt\delta\,u}.
\end{equation}
For a constant $M > 0$ and $u \coloneqq M\{\Gamma_N(\nu)\tilde r_N\}^{\frac{1}{2}}$, which is at most one for large $N$ if $\Gamma_N(\nu)\tilde r_N = o(1)$, adding $\Pr\{\cE_T(\delta)^c\} \leq \delta$ to \eqref{eq:tracking-transfer} and using $\Gamma_N(\nu) \geq 1$, we obtain
\[
\Pr\nolimits_\nu\left\{\sqrt N\|\hat\beta-m_\pi\|+\|\hat V-\clip S_\pi\|_F > \sqrt2\,u\right\} \leq \delta+\frac{C}{\delta M^2}+\frac{C}{\sqrt\delta\,M},
\]
where we use $\{\sqrt N\|\hat\beta-m_\pi\|+\|\hat V-\clip S_\pi\|_F\}^2 \leq 2e(\hat f,f_\pi^\circ)$. Taking $\delta$ small and then $M$ large, we obtain, for every $\cF_{\mathrm{TF}}$ and $\nu$ with $\Gamma_N(\nu)\tilde r_N = o(1)$,
\begin{equation}\label{eq:learning}
\sqrt N\|\hat\beta-m_\pi\|+\|\hat V-\clip S_\pi\|_F = O_\nu\left(\{\Gamma_N(\nu)\tilde r_N\}^{\frac{1}{2}}\right).
\end{equation}
Under Assumption \ref{cond:local-evaluation}, $\cL_{\nu,N}$ has density $\E_\pi\{w(\P,q)\mid\cD_N,q\}$ relative to $\cL_{\pi,N}$, where $w \coloneqq d\nu_N/d(\pi_P\otimes\pi_q)$, and hence, by conditional Jensen's inequality,
\begin{equation}\label{eq:overlap-local-evaluation}
\Gamma_N(\nu) \leq \int w^2\,d(\pi_P\otimes\pi_q) \leq 1+C.
\end{equation}

\paragraph{Step 6: From $f_\pi^\circ$ to $f_\pi$ in \eqref{eq:tracking-general}.}
Let $\cE_n$ denote the event $\cE_N$ of regular contexts in Section \ref{app:theory-technical} at size $n$ and define
\[
\cG_n \coloneqq \left\{(\cD_n,q) \in \cE_n:v_-I \preceq S_\pi(\cD_n,q) \preceq v_+I\right\}.
\]
Under $\pi_K$ by \eqref{eq:irregular-context} and \eqref{eq:simulator-target}, and otherwise by the proofs of Lemma \ref{lem:prior-mode} and Theorem \ref{thm:target}, $\cL_{\pi,n}(\cE_n) = 1-o(1)$ and $S_\pi = V_q^\star(\P)+o(1)$ in pretraining probability, and hence, by Assumption \ref{cond:covariance-bounds}(ii),
\begin{equation}\label{eq:pretraining-spectrum}
\inf_{n \in \cN}\cL_{\pi,n}(\cG_n) = 1-o(1).
\end{equation}
By Cauchy--Schwarz and \eqref{eq:overlap-local-evaluation}, we obtain
\begin{equation}\label{eq:posterior-spectrum}
\cL_{\nu,N}\{f_\pi \ne f_\pi^\circ\} \leq \cL_{\nu,N}(\cG_N^c) \leq \{\Gamma_N(\nu)\cL_{\pi,N}(\cG_N^c)\}^{\frac{1}{2}} = o(1).
\end{equation}
For $\cF_{\mathrm{TF}}$ with $\tilde r_N = o(1)$ and a constant $M > 0$, since $e(\hat f,f_\pi) = e(\hat f,f_\pi^\circ)$ on $\{f_\pi = f_\pi^\circ\}$, we have
\[
\Pr\nolimits_\nu\{e(\hat f,f_\pi) > M^2\tilde r_N\} \leq \Pr\nolimits_\nu\{e(\hat f,f_\pi^\circ) > M^2\tilde r_N\}+\cL_{\nu,N}\{f_\pi \ne f_\pi^\circ\},
\]
where the second term is $o(1)$ by \eqref{eq:posterior-spectrum} and the first is arbitrarily small for large $M$ by \eqref{eq:learning} and \eqref{eq:overlap-local-evaluation}. Hence, for every $\cF_{\mathrm{TF}}$ with $\tilde r_N = o(1)$ and $\hat f$ satisfying \eqref{eq:erm},
\begin{equation}\label{eq:tracking-general}
\sqrt N\|\hat\beta-m_\pi\|+\|\hat V-S_\pi\|_F \leq \{2e(\hat f,f_\pi)\}^{\frac{1}{2}} = O_\nu(\tilde r_N^{\frac{1}{2}}).
\end{equation}
For $\cF_N$, we have $\tilde r_N \lesssim r_N = o(1)$ by Step~3, and hence, by \eqref{eq:tracking-general}, we obtain \eqref{eq:tracking}, which completes the proof.
\end{proof}

\subsubsection{Approximation error}\label{app:proof-finite-transformer}

For the approximation error in Step~2 of the proof of Theorem \ref{thm:TF-error}, we construct a Transformer that approximates the mode pair $(g,V)$ of Definition \ref{def:reference-estimator} uniformly over contexts, while recording its depth, width and coefficient bounds, and then bound the prior-averaged excess score of this Transformer. As in Sections \ref{app:prior} and \ref{app:theory-technical}, $b_{(K)}$, $A_K$, $\Lambda_K$, $\Theta_K$, $\bbar$, $I_K$, $\cV_N$, $\E_N$, $\tilde\theta$, $\cE_N$ and $B_{\partial,N}$ denote the sieve basis, log partition function, prior covariance and support, sample mean, information, potential, expectation and mode of the coefficient posterior under $\pi_K$, event of regular contexts and integrated boundary density at size $N$, with $n$ in place of $N$ at size $n$.

Let $\cF_N$ denote a class $\cF_{\mathrm{TF}}(L,1,d_e,d_f,W)$ of Definition \ref{def:transformer-readout} with $L \lesssim \log^2 N$, $d_e\vee d_f \lesssim K\log^d N$ and $W \lesssim N^C$. For an output $f = (f^\beta,\clip C)$ with raw covariance $C = C^\T$, we measure the uniform error relative to $(g,V)$ at size $N$ by
\[
a_N(f,C) \coloneqq \sup_{(\cD_N,q) \in \cZ^N\times\cQ}\{\sqrt N\|f^\beta-g\|+\|C-V\|_F\}.
\]

\begin{lemma}[Approximation of the mode pair]\label{lem:network}
Suppose that Assumption \ref{cond:general-queries}(i), (ii) and (iv) hold, $K^3\vee\sigma^{-2}K^{2a} \leq n_{\min}$ and $n_{\max} \lesssim n_{\min}$. Then there exist $L$, $d_e$, $d_f$ and $W$ with
\begin{equation}\label{eq:architecture-rate}
L \lesssim \log^2 n_{\max},\quad d_e\vee d_f \lesssim K\log^d n_{\max}\quad\text{and}\quad W \lesssim n_{\max}^C,
\end{equation}
and a deterministic $f^\circ \in \cF_{\mathrm{TF}}(L,1,d_e,d_f,W)$, with raw covariance $C^\circ$, such that $\sup_{N \in \cN}a_N(f^\circ,C^\circ) \leq (4n_{\min})^{-1}$, with constants independent of $K$, $n_{\min}$ and $n_{\max}$.
\end{lemma}

\begin{proof}
The network $f^\circ$ averages the features in one attention head, approximates $\tilde\theta$ by projected gradient steps and evaluates $\beta_q$ and the covariance $V$ at this approximation, with degree and accuracy
\[
r \coloneqq \lceil C_1\log(C_1n_{\max}K)\rceil \leq 2C_1\log(C_1n_{\max}K)\quad\text{and}\quad\tau \coloneqq e^{-r} \leq (C_1n_{\max}K)^{-C_1}
\]
with a constant $C_1 \geq 2$. As the constants $c$ and $C$ below do not depend on $C_1$, we choose $C_1$ so large that the finitely many conditions used below hold, all of the forms
\[
n_{\max}^{\frac{3}{2}}(Kr)^C\tau \leq c\quad\text{and}\quad r \geq C\log(2K).
\]
Moreover, $\log(2K) \leq r \lesssim \log n_{\max}$, because $K \leq n_{\max}$.

\paragraph{Step 1: Features $\widehat b$ and penalty $\widehat a_N$.}
With the features $\widehat b_{K,\tau}$ of Lemma \ref{lem:raw-basis-efficient} and zero query and key maps, the masked head returns $\widehat b \coloneqq N^{-1}\sum_{i = 1}^N\widehat b_{K,\tau}(Z_i)$, as in Lemma~G.2 of \cite{zhang2026}, so that
\begin{equation}\label{eq:network-feature-average}
\|\widehat b-\bbar\| \leq \sqrt K\,\tau.
\end{equation}
The query row encodes $N/n_{\max} \in [c,1]$ through $\xi(q,N)$ of Section \ref{app:architecture}, and, by Lemmas \ref{lem:gelu-arithmetic} and \ref{lem:chebyshev}, we compute $\widehat s_N$ with $|\widehat s_N-n_{\max}/N| \leq C\tau$ and $\widehat a_N(x)$ with $\|\widehat a_N(x)-\widehat s_N\Lambda_K^{-1}x/n_{\max}\|_\infty \leq \tau$. For $\|x\|_\infty \leq C$, since $\|\Lambda_K^{-1}\|_{\op} = \sigma^{-2}K^{2a} \leq n_{\min} \leq n_{\max}$, we have
\begin{equation}\label{eq:network-penalty-error}
\left\|\widehat a_N(x)-\frac{\Lambda_K^{-1}x}{N}\right\| \leq \sqrt K\,\tau+\frac{\|\Lambda_K^{-1}x\|}{n_{\max}}\left|\widehat s_N-\frac{n_{\max}}{N}\right| \leq C\sqrt K\,\tau.
\end{equation}
All later value maps are zero, which preserves $\widehat b$.

\paragraph{Step 2: Quadrature error \eqref{eq:network-quadrature-error}.}
For $\|\theta\|_1 \leq B+1$, the function $g_\theta \coloneqq \theta^\T b_{(K)}$ satisfies $\|g_\theta\|_\infty \leq C_\infty(B+1)$ for the basis bound $C_\infty$ of Section \ref{app:prior}, so that, by Taylor's theorem with $m_0 \coloneqq \lceil Cr\rceil$, we have
\begin{equation}\label{eq:network-taylor}
\left\|e^{g_\theta}-\sum_{j = 0}^{m_0}\frac{g_\theta^j}{j!}\right\|_\infty \leq \frac{C^{m_0+1}}{(m_0+1)!} \leq \tau.
\end{equation}
Let $\P_{0,Q}$ denote the uniform distribution on the grid
\[
\cZ_Q \coloneqq \left\{(\Phi_1^{-1}(x_1),\cdots,\Phi_d^{-1}(x_d),D):x_1,\cdots,x_d \in \left\{\frac{2j-1}{2m_Q}:j \in [m_Q]\right\},\ D \in \cZ_D\right\},
\]
where $m_Q \coloneqq (m_0+2)m_K+1$, $m_K$ is the largest coordinate frequency in \eqref{eq:frequency-bound} and $\Phi_\ell$ is the distribution function of $V_\ell$ under $\P_0$ in Section \ref{app:prior}. As $\P_{0,Q}$ integrates the Taylor polynomial in \eqref{eq:network-taylor} times each $f \in \{1,b_k,b_kb_l\}$ exactly, we have
\[
\left|(\E_{\P_{0,Q}}-\E_{\P_0})(fe^{g_\theta})\right| \leq 2\|f\|_\infty\left\|e^{g_\theta}-\sum_{j = 0}^{m_0}\frac{g_\theta^j}{j!}\right\|_\infty \leq C\tau.
\]
Let $M_f$ and $M_{f,Q}$ denote $\E_{\P_0}(fe^{g_\theta})$ and $\E_{\P_{0,Q}}(fe^{g_\theta})$, respectively. Since $M_1,M_{1,Q} \in [c,C]$, the normalised moments satisfy
\[
\left|\frac{M_{f,Q}}{M_{1,Q}}-\frac{M_f}{M_1}\right| \leq \frac{|M_{f,Q}-M_f|}{M_{1,Q}}+\frac{|M_f|\,|M_{1,Q}-M_1|}{M_{1,Q}M_1} \leq C\tau.
\]
The gradients and Hessians of $A_Q(\theta) \coloneqq \log M_{1,Q}$ and $A_K(\theta) = \log M_1$ are the means and covariances of $b_{(K)}$ under $e^{g_\theta}\,d\P_{0,Q}/M_{1,Q}$ and $e^{g_\theta}\,d\P_0/M_1 = d\P_\theta$, respectively, and, applying the ratio bound coordinatewise, we obtain
\begin{equation}\label{eq:network-quadrature-error}
\sup_{\|\theta\|_1 \leq B+1}\bigl\{K^{-\frac{1}{2}}\|\nabla A_Q(\theta)-\nabla A_K(\theta)\|+K^{-1}\|\nabla^2 A_Q(\theta)-I_K(\theta)\|_F\bigr\} \leq C\tau.
\end{equation}
Let $z_1,\cdots,z_{S_Q}$ denote the nodes of $\cZ_Q$, where, by \eqref{eq:frequency-bound}, $S_Q = |\cZ_D|m_Q^d \lesssim K(m_0+2)^d \lesssim Kr^d$, and define the tilted expectation by
\begin{equation}\label{eq:network-softmax-average}
\E_{Q,\theta}(f) \coloneqq \frac{S_Q^{-1}\sum_{i = 1}^{S_Q}f(z_i)e^{g_\theta(z_i)}}{S_Q^{-1}\sum_{i = 1}^{S_Q}e^{g_\theta(z_i)}}.
\end{equation}
As $\E_{\P_{0,Q}}b_{(K)} = \E_{\P_0}b_{(K)} = 0$, $\E_{\P_{0,Q}}b_{(K)}b_{(K)}^\T = \E_{\P_0}b_{(K)}b_{(K)}^\T = I$ and $e^{g_\theta}/M_{1,Q} \in [c,C]$, by \eqref{eq:information-bounds}, $cI \preceq \nabla^2A_Q(\theta) \preceq CI$ for $\|\theta\|_1 \leq B+1$.

\paragraph{Step 3: Mode error \eqref{eq:network-mode-error}.}
As in the proof of Theorem~D.2 of \cite{bai2023}, where successive Transformer layers implement proximal gradient steps, we compute $\theta_{Q,N} \coloneqq \argmin_{\theta \in \Theta_K}F_{Q,N}(\theta)$, where
\[
F_{Q,N}(\theta) \coloneqq A_Q(\theta)-\theta^\T\widehat b+\frac{1}{2N}\theta^\T\Lambda_K^{-1}\theta.
\]
As $\|\Lambda_K^{-1}\|_{\op} \leq N$, $cI \preceq \nabla^2F_{Q,N} \preceq CI$ on $\left\{\theta \in \R^K:\|\theta\|_1 \leq B+1\right\}$. Since $\tilde\theta$ minimises $\cV_N/N$ over $\Theta_K$ and $\nabla F_{Q,N}-\nabla\cV_N/N = \nabla A_Q-\nabla A_K-\widehat b+\bbar$, by Lemma \ref{lem:perturbed-iterations}(ii), \eqref{eq:network-feature-average} and \eqref{eq:network-quadrature-error}, we have
\begin{equation}\label{eq:network-minimiser-error}
c\|\theta_{Q,N}-\tilde\theta\| \leq \|\widehat b-\bbar\|+\|\nabla A_Q(\tilde\theta)-\nabla A_K(\tilde\theta)\| \leq C\sqrt K\,\tau.
\end{equation}
Computing $e^{g_\theta(z_i)}$ and the reciprocal of their average in \eqref{eq:network-softmax-average} within $C\tau$ per coordinate by Lemmas \ref{lem:gelu-arithmetic} and \ref{lem:chebyshev}, and adding \eqref{eq:network-penalty-error}, we obtain
\begin{equation}\label{eq:network-gradient-error}
\left\|\widehat{\nabla A_Q}(\theta)-\widehat b+\widehat a_N(\theta)-\nabla F_{Q,N}(\theta)\right\| \leq C\sqrt K\,\tau\quad\text{for } \|\theta\|_1 \leq B+1.
\end{equation}
We use a step size $0 < \gamma \leq 1/C$, start at $\widehat\theta_0 = 0$ and let $\widehat\theta_{j+1} \coloneqq \widehat P(z_j)$, where $\widehat P$ is the map of Lemma \ref{lem:gelu-projection} with $\lambda \coloneqq C\sqrt K\log(2K)/\tau$, so that $\|\widehat P(z)-\operatorname{proj}_{\Theta_K}(z)\| \leq \tau$ for $\|z\|_\infty \leq C$, and
\[
z_j \coloneqq \widehat\theta_j-\gamma\{\widehat{\nabla A_Q}(\widehat\theta_j)-\widehat b+\widehat a_N(\widehat\theta_j)\}.
\]
If $\|\widehat\theta_j\|_1 \leq B+1/2$, then $\|z_j\|_\infty \leq C$, and, since $\sqrt K\tau \leq 1/2$,
\[
\|\widehat\theta_{j+1}\|_1 \leq \|\operatorname{proj}_{\Theta_K}(z_j)\|_1+\sqrt K\|\widehat\theta_{j+1}-\operatorname{proj}_{\Theta_K}(z_j)\| \leq B+\frac{1}{2},
\]
so that, by induction, $\|\widehat\theta_j\|_1 \leq B+1/2$ for all $j$. The map $T_{Q,N}(\theta) \coloneqq \operatorname{proj}_{\Theta_K}\{\theta-\gamma\nabla F_{Q,N}(\theta)\}$ satisfies $T_{Q,N}(\theta_{Q,N}) = \theta_{Q,N}$ by the variational inequality at $\theta_{Q,N}$, and, for $\|\theta\|_1 \leq B+1$, as $\operatorname{proj}_{\Theta_K}$ is nonexpansive \citep{bubeck2015},
\begin{align*}
\|T_{Q,N}(\theta)-\theta_{Q,N}\| &\leq \left\|\left[I-\gamma\int_0^1\nabla^2F_{Q,N}\{\theta_{Q,N}+s(\theta-\theta_{Q,N})\}\,ds\right](\theta-\theta_{Q,N})\right\|\\
&\leq \rho\|\theta-\theta_{Q,N}\|,
\end{align*}
where $\rho \coloneqq 1-\gamma c < 1$. Since $\|\widehat\theta_{j+1}-T_{Q,N}(\widehat\theta_j)\| \leq \tau+C\gamma\sqrt K\tau$ by \eqref{eq:network-gradient-error}, by Lemma \ref{lem:perturbed-iterations}(i) with $\|\widehat\theta_0-\theta_{Q,N}\| \leq B$ and $J \coloneqq \lceil Cr\rceil$, and by \eqref{eq:network-minimiser-error}, we obtain
\begin{equation}\label{eq:network-mode-error}
\|\widehat\theta_J-\tilde\theta\| \leq B\rho^J+\frac{C\sqrt K\,\tau}{1-\rho}+C\sqrt K\,\tau \leq C\sqrt K\,\tau.
\end{equation}

\paragraph{Step 4: Outputs $\widehat\beta$, $\widehat D$ and $\widehat X$.}
At $\widehat\theta \coloneqq \widehat\theta_J$, by Lemma \ref{lem:query-computation} with degree $r$ and accuracy $\tau = e^{-r}$ for each product and positive part, including the errors of selecting the query family, we have, for the Jacobian $D_q$ of $\theta \mapsto \beta_q(\P_\theta)$ in Assumption \ref{cond:general-queries}(ii),
\begin{equation}\label{eq:network-functional-error}
\|\widehat\beta-\beta_q(\P_{\widehat\theta})\|+\|\widehat D-D_q(\widehat\theta)\|_F \leq C(Kr)^C\tau.
\end{equation}
By Lemma \ref{lem:regularity} and $d_k \leq d_0 = \max_{j \in \cK}d_j$, $\|\widehat D\|_F \leq C$, and the matrix $H_Q \coloneqq \nabla^2A_Q(\widehat\theta)+\Lambda_K^{-1}/N$, with $cI \preceq H_Q \preceq CI$, acts as
\[
H_Qx = \E_{Q,\widehat\theta}\left[b_{(K)}\{b_{(K)}-\E_{Q,\widehat\theta}b_{(K)}\}^\T x\right]+\frac{\Lambda_K^{-1}x}{N}.
\]
As $|b_{(K)}(z)^\T x| \leq C_\infty K\|x\|_\infty \leq CK$ for $z \in \cZ$ and $\|x\|_\infty \leq C$, computing the tilted mean and grid weights, with $S_Q^{-1}$ factored out, within $C\tau$ per coordinate as in Step~3 and using \eqref{eq:network-penalty-error}, we obtain
\[
\|\widehat{H_Qx}-H_Qx\| \leq C\sqrt K(K\tau+\tau)+\left\|\widehat a_N(x)-\frac{\Lambda_K^{-1}x}{N}\right\| \leq CK^{\frac{3}{2}}\tau.
\]
Let $X_* \coloneqq H_Q^{-1}\widehat D^\T$ and let $\operatorname{proj}$ denote the entrywise projection onto $[-C,C]^{K\times d_k}$, which contains $X_*$. As $\operatorname{proj}$ is nonexpansive and $X_* = \operatorname{proj}\{X_*-\gamma(H_QX_*-\widehat D^\T)\}$, the projected Richardson step \citep{saad2003} of \cite{kang2026} satisfies
\[
\|\operatorname{proj}\{X-\gamma(H_QX-\widehat D^\T)\}-X_*\|_F \leq \|(I-\gamma H_Q)(X-X_*)\|_F \leq \rho\|X-X_*\|_F.
\]
Starting at $\widehat X_0 = 0$, we let $\widehat X_{j+1}$ denote the computed value of this step at $\widehat X_j$, so that $\|\widehat X_{j+1}-\operatorname{proj}\{\widehat X_j-\gamma(H_Q\widehat X_j-\widehat D^\T)\}\|_F \leq CK^{\frac{3}{2}}\tau$, and hence, by Lemma \ref{lem:perturbed-iterations}(i), for $\widehat X \coloneqq \widehat X_J$,
\begin{equation}\label{eq:network-solve-error}
\|\widehat X-H_Q^{-1}\widehat D^\T\|_F \leq \rho^J\|X_*\|_F+\frac{CK^{\frac{3}{2}}\tau}{1-\rho} \leq CK^{\frac{3}{2}}\tau.
\end{equation}

\paragraph{Step 5: Bound on $a_N(f^\circ,C^\circ)$ and \eqref{eq:architecture-rate}.}
For $H_K \coloneqq I_K(\tilde\theta)+\Lambda_K^{-1}/N$, by the resolvent identity, \eqref{eq:network-quadrature-error}, \eqref{eq:network-mode-error} and Lemma \ref{lem:geometry}, we have
\[
\|H_Q^{-1}-H_K^{-1}\|_{\op} \leq \|H_Q^{-1}\|_{\op}\|H_Q-H_K\|_{\op}\|H_K^{-1}\|_{\op} \leq C(K\tau+\sqrt K\|\widehat\theta-\tilde\theta\|) \leq CK\tau.
\]
Since the point readout $\clip_{C_\beta}$ of Definition \ref{def:transformer-output} is nonexpansive and $\clip_{C_\beta}(g) = g$, by \eqref{eq:network-functional-error}, \eqref{eq:network-mode-error}, Lemma \ref{lem:regularity} and Assumption \ref{cond:general-queries}(ii), for $D \coloneqq D_q(\tilde\theta)$ we have
\[
\|f^{\circ,\beta}-g\| \leq \|\widehat\beta-g\| \leq C(Kr)^C\tau\quad\text{and}\quad\|\widehat D-D\|_F \leq C(Kr)^C\tau.
\]
Inserting $\widehat DH_Q^{-1}\widehat D^\T$ in $C^\circ-V$, where $C^\circ = \{\widehat{DX}+(\widehat{DX})^\T\}/2$ for the computed product $\widehat{DX}$ of $\widehat D$ and $\widehat X$, we obtain
\[
\|C^\circ-V\|_F \leq C\{\|\widehat X-H_Q^{-1}\widehat D^\T\|_F+\|\widehat D-D\|_F+\|H_Q^{-1}-H_K^{-1}\|_{\op}+(Kr)^C\tau\} \leq C(Kr)^C\tau,
\]
where the second inequality follows from \eqref{eq:network-solve-error} and the two preceding bounds. Since $N \leq n_{\max}$, by the choice of $C_1$ and $n_{\min} \leq n_{\max}$, we have, for every $N \in \cN$,
\[
a_N(f^\circ,C^\circ) \leq C\sqrt{n_{\max}}(Kr)^C\tau \leq (4n_{\min})^{-1}.
\]
For the size, the features, the reciprocal and each of the $J$ gradient and $J$ Richardson updates use $O(r)$ layers, as $\log(2K) \leq r$, the query evaluation uses $O(r^2)$ layers and each layer has $O(S_Q+K)$ coordinates, so that, since $J \lesssim r \lesssim \log n_{\max}$,
\[
L \lesssim r^2 \lesssim \log^2 n_{\max}\quad\text{and}\quad d_e\vee d_f \lesssim S_Q+K \lesssim Kr^d \lesssim K\log^d n_{\max}.
\]
The coefficients are polynomial in $n_{\max}K/\tau \leq e(C_1n_{\max}K)^{C_1+1}$, which proves \eqref{eq:architecture-rate}. As the coefficients and the mask rule do not depend on $N$, one $f^\circ$ serves all $N \in \cN$, which completes the proof.
\end{proof}

\begin{lemma}\label{lem:prior-mean-comparison}
Suppose that $\pi_P$ is the distribution of $\P_\theta$ under $\theta \sim \pi_K$, Assumption \ref{cond:general-queries}(i)--(ii) holds and $K^3\vee\|\Lambda_K^{-1}\|_{\op} \leq N$. Then, with a constant independent of $N$ and $K$, the mode mean $g$ satisfies
\begin{equation}\label{eq:prior-mean-comparison}
\E_\pi\E_\P\{N\|m_\pi-g\|^2\} \lesssim \frac{K^3}{N}+\frac{K}{\sqrt N}.
\end{equation}
\end{lemma}

\begin{proof}
We interpolate between the mean bounds $N\|m_\pi-g\|^2 \leq CK$ and $N\|m_\pi-g\| \leq C(B_{\partial,N}+K^{\frac{3}{2}})$ of Lemma \ref{lem:laplace}, valid at every context, and obtain
\[
N\|m_\pi-g\|^2 \leq C\left\{\frac{K^3}{N}+\min\left(K,\frac{B_{\partial,N}^2}{N}\right)\right\} \leq C\left(\frac{K^3}{N}+\sqrt{\frac{K}{N}}\,B_{\partial,N}\right),
\]
where the last inequality uses $\min(x,y) \leq \sqrt{xy}$. Applying $\E_\pi\E_\P$ to this bound, which is uniform in $q$, and using $\E_\pi\E_\P B_{\partial,N} \leq C\sqrt K$ in \eqref{eq:boundary-density-average}, we obtain \eqref{eq:prior-mean-comparison}, which completes the proof.
\end{proof}

\begin{lemma}\label{lem:posterior-comparison}
Under the conditions of Lemma \ref{lem:prior-mean-comparison}, for every $\varepsilon \geq 0$ and every output $f^\circ = (f^{\circ,\beta},\clip C^\circ)$ with symmetric $C^\circ$ and $a_N(f^\circ,C^\circ) \leq \varepsilon$, the conditional excess score $h_{f^\circ}$ of Section \ref{app:proof-tf-error} at size $N$ satisfies
\[
\E_\pi\E_\P h_{f^\circ} \lesssim \varepsilon^2+\frac{K^3}{N}+K\sqrt{\frac{\log N}{N}},
\]
with a constant independent of $N$, $K$, $\varepsilon$ and $f^\circ$.
\end{lemma}

\begin{proof}
We bound $h_{f^\circ}$ by $\varepsilon^2$ and the errors $N\|m_\pi-g\|^2$ and $\|S_\pi-V\|_F^2$ of the mode pair, and then average the latter over contexts.

For symmetric $C$, let $A_C$ and $A_S$ denote $(\clip C)^{-1}$ and $(\clip S_\pi)^{-1}$, respectively, with the spectral clipping $\clip = \clip_{[v_-,v_+]}$ of Section \ref{app:architecture}. Since $F_C(A_C) \leq F_C(A_S)$ for $F_C(A) = -\log\det A+\tr(AC)$ by \eqref{eq:clipped-precision}, it holds that
\[
F_{S_\pi}(A_C)-F_{S_\pi}(A_S) = F_C(A_C)-F_C(A_S)+\tr\{(S_\pi-C)(A_C-A_S)\} \leq v_-^{-2}\|C-S_\pi\|_F^2,
\]
where the inequality also uses $A_C-A_S = A_C(\clip S_\pi-\clip C)A_S$ and the nonexpansiveness of spectral clipping. Substituting this bound into \eqref{eq:posterior-loss-regret} at size $N$, we obtain
\begin{equation}\label{eq:raw-covariance-comparison}
h_{(\beta,\clip C)} \leq \frac{N\|\beta-m_\pi\|^2}{2v_-}+\frac{\|C-S_\pi\|_F^2}{2v_-^2},
\end{equation}
and, by the triangle inequality through $g$ and $V$, it follows that
\begin{equation}\label{eq:comparator-score-split}
0 \leq h_{f^\circ} \leq C\{\varepsilon^2+N\|m_\pi-g\|^2+\|S_\pi-V\|_F^2\}.
\end{equation}
By Lemma \ref{lem:laplace}, $\|S_\pi-V\|_F^2 \leq CK^2/N$ on $\cE_N$ and $\|S_\pi-V\|_F^2 \leq C$ on $\cE_N^c$, and hence, by Lemma \ref{lem:mode-localisation} and $N^{-4} \leq K\sqrt{\log N/N}$,
\[
\E_\pi\E_\P\|S_\pi-V\|_F^2 \leq \frac{CK^2}{N}+C\E_{\pi_K}\{\P_\theta(\cE_N^c)\} \leq C\left(\frac{K^2}{N}+K\sqrt{\frac{\log N}{N}}\right).
\]
Substituting \eqref{eq:prior-mean-comparison} and this estimate into \eqref{eq:comparator-score-split}, and using $K^2 \leq K^3$ and $K/\sqrt N \leq K\sqrt{\log N/N}$, we obtain the asserted bound.
\end{proof}

The next lemma extends Lemma \ref{lem:prior-mean-comparison} and the covariance bound in the proof of Lemma \ref{lem:posterior-comparison} to the priors of Assumptions \ref{cond:simulator-prior} and \ref{cond:posterior-tails}.

\begin{lemma}\label{lem:prior-mode}
Suppose that Assumptions \ref{cond:general-queries}(i)--(ii), \ref{cond:simulator-prior} and \ref{cond:posterior-tails} hold and $K^3\vee\|\Lambda_K^{-1}\|_{\op} \leq n_{\min}$. Then, for all sufficiently large $N$ and every $n \in \cN$, the mode pair $(g,V)$ at size $n$ satisfies
\[
\E_\pi\E_\P\{n\|m_\pi-g\|^2+\|S_\pi-V\|_F^2\} \lesssim \frac{K^3}{N}+K\sqrt{\frac{\log N}{N}}+\epsilon_{\pi,N},
\]
where $\epsilon_{\pi,N} = L_N^2/N+N\delta_\pi^2$ is the sieve approximation error of Section \ref{app:simulator-prior}, with a constant independent of $N$, $K$ and $n$.
\end{lemma}

\begin{proof}
We first compare $m_\pi$ with $g$, as in Lemma \ref{lem:prior-mean-comparison}, and then bound $\E_\pi\{\P(\cE_n^c)\}$ under $\pi_P$.

With the density ratio $\rho$ in \eqref{eq:posterior-density-ratio} and $L_N' = L_N+Cn\delta_\pi$ as in the proof of Lemma \ref{lem:prior-stability}, where $\delta_\pi$ is as in \ref{A3}, we have $\|\nabla\log\rho\| \leq L_N'$ almost everywhere, and we denote the density and expectation of the posterior of $\theta_\P$ under $\pi_P$ by
\[
p_n^\pi(\theta) \propto \rho(\theta)e^{-\cV_n(\theta)}\quad\text{on }\Theta_K\quad\text{and}\quad\E_\rho \coloneqq \E_n(\rho\,\cdot),
\]
respectively. Since $L_N'^2/n$ is bounded, applying the arguments for \eqref{eq:mode-moments} and \eqref{eq:boundary-mode-mean} to $p_n^\pi$, with $\nabla\cV_n-\nabla\log\rho$ in place of $\nabla\cV_n$ in the Gauss--Green formula, we obtain
\[
\E_\rho\|\theta-\tilde\theta\|^2 \leq C\left(\frac{K}{n}+\frac{L_N'^2}{n^2}\right)\quad\text{and}\quad n\|\E_\rho\{\beta_q(\P_\theta)\}-g\|^2 \leq C\left(\frac{K^3}{n}+\frac{L_N'^2}{n}+\sqrt{\frac{K}{n}}\,B_{\partial,n}^\pi\right),
\]
where $B_{\partial,n}^\pi \coloneqq \int_{\partial\Theta_K}p_n^\pi\,d\mathcal H^{K-1}$. Averaging Bayes' formula over $\P \sim \pi_P$ and $\cD_n \sim \P^{\otimes n}$, we find that $\E_\pi\E_\P p_n^\pi$ is the density of $\theta_\P$ under $\pi_P$, which is at most $C_wp_K$ for the density $p_K$ of $\pi_K$ by Assumption \ref{cond:posterior-tails}, so that, by \eqref{eq:boundary-density-average},
\[
\E_\pi\E_\P B_{\partial,n}^\pi \leq C_w\int_{\partial\Theta_K}p_K\,d\mathcal H^{K-1} \leq C\sqrt K.
\]
Since $\|m_\pi-\E_\rho\{\beta_q(\P_\theta)\}\| \leq C\delta_\pi$ by the bound on $\|\beta_q(\P_{\theta,g})-\beta_q(\P_\theta)\|$ in the proof of Lemma \ref{lem:prior-stability}, the bound on the mean term follows.

For the covariance term, by \eqref{eq:prior-stability} and Lemma \ref{lem:laplace} applied to $S_{\pi_K}$, we have
\[
\|S_\pi-V\|_F^2 \leq \begin{cases}C(\epsilon_{\pi,N}+K^2/n), & (\cD_n,q) \in \cE_n,\\ C, & (\cD_n,q) \notin \cE_n.\end{cases}
\]
It remains to bound $\E_\pi\{\P(\cE_n^c)\}$. At every dataset, \eqref{eq:mode-truth-displacement} holds with $\theta = \theta_\P$, and \eqref{eq:sufficient-statistic-tail} holds under every $\P$ with $\E_\P b_{(K)}$ in place of $\E_{\P_\theta}b_{(K)}$, while, by \eqref{eq:tilt-moment-shift},
\[
\|\E_\P b_{(K)}-\E_{\P_{\theta_\P}}b_{(K)}\| \leq C\delta_\pi = o\left(\sqrt{K\log N/N}\right).
\]
As the density of $\theta_\P$ under $\pi_P$ is at most $C_wp_K$, repeating the argument for \eqref{eq:irregular-context}, we obtain $\E_\pi\{\P(\cE_n^c)\} \leq C(K\sqrt{\log N/N}+N^{-4})$, which proves the covariance bound and completes the proof.
\end{proof}

\subsubsection{Network approximation of elementary functions, features and queries}\label{app:comparator-primitives}

For $k \geq 1$, let $(r^{(k)},\eta^{(k)})$ denote the pair $(r,\eta)$ that labels $b_k$ in the enumeration of the trigonometric basis in Section \ref{app:prior}, so that $b_k(v,D) = \chi_{\eta^{(k)}}(D)\prod_{\ell = 1}^d\varphi_{r^{(k)}_\ell}\{\Phi_\ell(v_\ell)\}$, where $\Phi_\ell$ is the distribution function of $V_\ell$ under the reference distribution $\P_0$. Since $\varphi_{2h-1}$ and $\varphi_{2h}$ have frequency $h$, we define the largest coordinate frequency of $b_1,\cdots,b_K$, truncated below at $1$, by
\begin{equation}\label{eq:max-frequency}
m_K \coloneqq \max\left\{1,\max_{k \in [K]}\left\lceil\frac{\|r^{(k)}\|_\infty}{2}\right\rceil\right\},
\end{equation}
and, since the enumeration is nondecreasing in $\|r\|_\infty$, we have
\begin{equation}\label{eq:frequency-bound}
m_K^d \leq CK.
\end{equation}
For $t,s \in \N$, let $\cM_{t,s}$ denote the class of maps $\Psi = \Psi_t\circ\cdots\circ\Psi_1$ of depth $t$ and width $s$ such that, for all $l \in [t]$ and $i \in [s]$, the $i$th coordinate $\Psi_{l,i}$ of $\Psi_l:\R^s \to \R^s$ is one of the maps
\[
x \mapsto a^\T x+b,\quad x \mapsto x_jx_{j'}\quad\text{and}\quad x \mapsto (x_j)_+,
\]
with $a \in \R^s$, $b \in \R$ and $j,j' \in [s]$, and we call the entries of these $a$ and $b$ the coefficients of $\Psi$. For $\tau \geq 0$ and $x \in \R^s$, a $\tau$-perturbation of $\Psi$ at $x$ is a sequence $x^{(0)} = x,x^{(1)},\cdots,x^{(t)} \in \R^s$ of intermediate values with output $x^{(t)}$ such that, for all $l \in [t]$ and $i \in [s]$,
\[
x^{(l)}_i = \Psi_{l,i}(x^{(l-1)})\ \ \text{if $\Psi_{l,i}$ is affine}\quad\text{and}\quad|x^{(l)}_i-\Psi_{l,i}(x^{(l-1)})| \leq \tau\ \ \text{otherwise}.
\]
For $L,w \in \N$ and $W \geq 1$, a GELU network of depth $L$, width $w$ and coefficient bound $W$ is a composition of $L$ maps on $\R^{1\times w}$ of the form
\[
h \mapsto h+\GELU(h\mathsf U_1+\mathsf c_1^\T)\mathsf U_2+\mathsf c_2^\T,\quad\text{with}\quad\mathsf U_1,\mathsf U_2 \in [-W,W]^{w\times w}\ \text{and}\ \mathsf c_1,\mathsf c_2 \in [-W,W]^w,
\]
which are the row maps of the feedforward layers in $\cF_{\mathrm{FFN}}(w,w,W)$ of Section \ref{app:architecture}. In both kinds of maps, an input in $\R^{s_0}$ occupies the first $s_0$ coordinates, the others being zero, and an output in $\R^{s_1}$ consists of the first $s_1$ coordinates.

\begin{lemma}\label{lem:perturbed-iterations}
Let $\cX \subset \R^m$.
\begin{enumerate}[label=(\roman*),nosep]
\item Let $T:\cX \to \R^m$, $x_* \in \cX$ and $\rho \in (0,1)$ satisfy $T(x_*) = x_*$ and $\|T(x)-x_*\| \leq \rho\|x-x_*\|$ for all $x \in \cX$. If $J \in \N$, $e \geq 0$, $\hat x_0,\cdots,\hat x_{J-1} \in \cX$ and $\hat x_J \in \R^m$ satisfy $\|\hat x_{j+1}-T(\hat x_j)\| \leq e$ for $0 \leq j < J$, then $\|\hat x_J-x_*\| \leq \rho^J\|\hat x_0-x_*\|+e/(1-\rho)$.
\item Let $c > 0$ and $G,G':\cX \to \R^m$ satisfy $(x-y)^\T\{G(x)-G(y)\} \geq c\|x-y\|^2$ for all $x,y \in \cX$. If $x_*,x' \in \cX$ satisfy $G(x_*)^\T(x'-x_*) \geq 0$ and $G'(x')^\T(x_*-x') \geq 0$, then $c\|x_*-x'\| \leq \|G(x')-G'(x')\|$.
\end{enumerate}
\end{lemma}

\begin{proof}
We prove (i) by iterating the perturbed contraction and (ii) by combining strong monotonicity with the two variational inequalities.

For (i), by the triangle inequality, $\|\hat x_{j+1}-x_*\| \leq e+\|T(\hat x_j)-x_*\| \leq e+\rho\|\hat x_j-x_*\|$, and iterating over $j < J$ with $\sum_{i \geq 0}\rho^i = (1-\rho)^{-1}$ proves (i).

For (ii), since $(x_*-x')^\T G(x_*) \leq 0 \leq (x_*-x')^\T G'(x')$, by strong monotonicity of $G$ at $(x_*,x')$ we obtain
\begin{align*}
c\|x_*-x'\|^2 &\leq (x_*-x')^\T\{G(x_*)-G(x')\}\\
&\leq (x_*-x')^\T\{G'(x')-G(x')\} \\ &\leq \|x_*-x'\|\,\|G(x')-G'(x')\|,
\end{align*}
which proves (ii) after division by $\|x_*-x'\|$, as (ii) is trivial when $x_* = x'$.
\end{proof}

\begin{lemma}[GELU approximation of products and positive parts]\label{lem:gelu-arithmetic}
Let $0 < \tau < 1/2$, $M \geq 1$ and $\cX \subset \R^s$, and let $\Psi \in \cM_{t,s}$ have coefficients in $[-M,M]$ such that, for every $x \in \cX$, every $\tau$-perturbation $(x^{(0)},\cdots,x^{(t)})$ of $\Psi$ at $x$ satisfies $x^{(l-1)}_j,x^{(l-1)}_{j'} \in [-M,M]$ whenever $\Psi_{l,i}(y) = y_jy_{j'}$. Then there is a GELU network $\widehat\Psi$ of depth $Ct$, width $Cs$ and coefficient bound $(CM/\tau)^C$ such that, for every $x \in \cX$, $\widehat\Psi(x)$ is the output of a $\tau$-perturbation of $\Psi$ at $x$, where $C$ is an absolute constant.
\end{lemma}

\begin{proof}
We represent the affine coordinates exactly and approximate the positive parts and products by GELU layers, and then choose the accuracy of these approximations.

Since $\GELU(x)-\GELU(-x) = x$, GELU layers implement affine maps exactly, and, since $\GELU(x) = x\Phi(x)$, for $\lambda > 0$ we have
\begin{equation}\label{eq:gelu-relu}
\sup_{x \in \R}\left|\frac{\GELU(\lambda x)}{\lambda}-x_+\right| = \sup_{x \in \R}|x|\Phi(-\lambda|x|) \leq \frac{C}{\lambda}.
\end{equation}
We implement the approximate layer $\widehat\Psi_l$ on two alternating blocks of coordinates $(z,z')$ by
\[
(z,z') \mapsto (z,z')+\{0,\widehat\Psi_l(z)-z'\}.
\]
By Corollary~3.7 of \cite{yakovlev2025} with accuracy parameter $c\eta/M^3$, a GELU network of constant size with coefficients bounded by $(CM/\eta)^C$ computes a product $\widehat{xy}$ with
\begin{equation}\label{eq:gelu-product}
\sup_{x,y \in [-M,M]}|\widehat{xy}-xy| \leq \eta.
\end{equation}
With $\lambda \coloneqq 2C/\tau$ in \eqref{eq:gelu-relu}, valid on all of $\R$, and $\eta \coloneqq \tau$ in \eqref{eq:gelu-product}, valid as all product factors lie in $[-M,M]$, the resulting network implements a $\tau$-perturbation with depth $Ct$, width $Cs$ and coefficient bound $(CM/\tau)^C$, which proves the lemma.
\end{proof}

\begin{lemma}\label{lem:chebyshev}
Let $\kappa > 1$, $M \geq 1$ and $r \in \N$, and let $f$ be analytic on $E_\kappa \coloneqq \left\{z \in \mathbb C:|z-1|+|z+1| < \kappa+\kappa^{-1}\right\}$ with $\sup_{z \in E_\kappa}|f(z)| \leq M$. Then there is $\Psi \in \cM_{Cr,C}$ with coefficients bounded by $C_\kappa M$ such that, for all $x \in [-1,1]$, $\hat x \in \R$ and $\tau \geq 0$ with $r^2(\delta+\tau) \leq 1/9$, where $\delta \coloneqq |\clip_{[-1,1]}(\hat x)-x|$, every $\tau$-perturbation of $\Psi$ at $\hat x$ has intermediate values entering products bounded by $C_\kappa M$ and an output $\hat f$ with
\[
|\hat f-f(x)| \leq \frac{2M\kappa^{-r}}{\kappa-1}+C_\kappa M(\delta+\tau),
\]
where $C$ is an absolute constant and $C_\kappa$ depends only on $\kappa$.
\end{lemma}

\begin{proof}
We evaluate the truncated Chebyshev series of $f$ by its recurrence at a clipped input and bound the errors through Chebyshev polynomials of the second kind.

Let $T_j$ and $U_j$ denote the Chebyshev polynomials of the first and second kinds, with $U_{-1} \coloneqq 0$, and $p_r \coloneqq \sum_{j = 0}^rc_jT_j$ the truncation of the Chebyshev series $f = \sum_{j \geq 0}c_jT_j$ on $[-1,1]$. By Theorems~8.1 and~8.2 of \cite{trefethen2013}, we have
\[
|c_0| \leq M,\quad |c_j| \leq 2M\kappa^{-j}\ \ \text{for } j \geq 1\quad\text{and}\quad\sup_{x \in [-1,1]}|f(x)-p_r(x)| \leq \frac{2M\kappa^{-r}}{\kappa-1}.
\]
The map clips $\hat x$ to $\tilde x \coloneqq -1+(\hat x+1)_+-(\hat x-1)_+$, so that $|\tilde x-x| \leq \delta+2\tau$, runs $\hat T_0 \coloneqq 1$, $\hat T_1 \coloneqq \tilde x$ and $\hat T_{j+1} \coloneqq 2\widehat{\tilde x\hat T_j}-\hat T_{j-1}$, and returns $\hat f \coloneqq \sum_{j = 0}^rc_j\hat T_j$. The errors $e_j \coloneqq \hat T_j-T_j(x)$ satisfy $e_0 = 0$, $e_1 = \tilde x-x$ and
\[
e_{j+1} = 2xe_j-e_{j-1}+\sigma_j,\quad\text{where}\quad|\sigma_j| \leq 2|\tilde x-x|\,|\hat T_j|+2\tau.
\]
Since $U_l$ solves the same recurrence, we have
\[
e_j = e_1U_{j-1}(x)+\sum_{i = 1}^{j-1}\sigma_iU_{j-1-i}(x),
\]
and, since $|U_l| \leq l+1$ on $[-1,1]$, if $|\hat T_i| \leq 2$ for $i < j$, it follows that
\[
|e_j| \leq j(\delta+2\tau)+\frac{j^2}{2}(4\delta+10\tau) \leq 9j^2(\delta+\tau) \leq 1,
\]
so that, by induction, $|\hat T_j| \leq 2$ and $|e_j| \leq 9j^2(\delta+\tau)$ for $j \leq r$. Consequently,
\[
|\hat f-f(x)| \leq \sum_{j = 1}^r|c_j|\,|e_j|+|p_r(x)-f(x)| \leq 18M(\delta+\tau)\sum_{j \geq 1}j^2\kappa^{-j}+\frac{2M\kappa^{-r}}{\kappa-1},
\]
and the map, which retains $\tilde x$, two iterates and the partial sum, has depth $Cr$ and constant width, which proves the lemma with $C_\kappa$ depending only on $\kappa$.
\end{proof}

If $f$ is analytic on a complex neighbourhood of a compact interval $[a,b]$, then $y \mapsto f\{(a+b)/2+(b-a)y/2\}$ is analytic and bounded on $E_\kappa$ for some $\kappa > 1$. Therefore, applying Lemma \ref{lem:chebyshev} with degree $\lceil r/\log\kappa\rceil$ to the exponential on $[-C,C]$ and the reciprocal on $[c,C]$, we obtain approximations with error at most $C(e^{-r}+\delta+\tau)$ whenever $(Cr)^2(\delta+\tau) \leq 1/9$.

\begin{lemma}\label{lem:raw-basis-efficient}
For all $K \geq 1$ and $\zeta \in (0,1/2)$, there is a GELU network $\widehat b_{K,\zeta}:\R^{d_z} \to \R^K$ of depth $C\log(CK/\zeta)$, width $CK$ and coefficient bound $(CK/\zeta)^C$ such that
\[
\sup_{z \in \cZ}\|\widehat b_{K,\zeta}(z)-b_{(K)}(z)\|_\infty \leq \zeta,
\]
where $C$ does not depend on $K$ and $\zeta$, and $b_{(K)} = (b_1,\cdots,b_K)^\T$ and the sample space $\cZ = \cV\times\cZ_D \subset \R^{d_z}$ are as in Section \ref{app:prior}.
\end{lemma}

\begin{proof}
For $z = (V,D) \in \cZ$ and $k \in [K]$, we have $b_k(z) = \prod_{\ell = 1}^{d+1}f_\ell$, where $f_\ell \coloneqq \varphi_{r^{(k)}_\ell}\{\Phi_\ell(V_\ell)\}$ for $\ell \in [d]$, with $\Phi_\ell$ equal to $\Phi$ or to the identity on $[0,1]$, and $f_{d+1} \coloneqq \chi_{\eta^{(k)}}(D)$. We approximate these factors in turn and multiply them.

We begin with the normal distribution function. For $\xi \in (0,1/2)$ and $y \in \R$, let
\[
\bar y \coloneqq \{2\log(8/\xi)\}^{\frac{1}{2}}\quad\text{and}\quad x \coloneqq \clip_{[-1,1]}(y/\bar y),\quad\text{so that}\quad|\Phi(y)-\Phi(\bar yx)| \leq \Phi(-\bar y) \leq \frac{\xi}{16}.
\]
For $t \in E_2$, where $|\mathrm{Im}\,t| \leq 3/4$ and $|t| \leq 5/4$, writing $\Phi(\bar yt) = 1/2+(2\pi)^{-\frac{1}{2}}\bar yt\int_0^1e^{-\frac{(s\bar yt)^2}{2}}\,ds$, we obtain
\[
|\Phi(\bar yt)| \leq \frac{1}{2}+\frac{5\bar y}{4\sqrt{2\pi}}e^{\frac{9\bar y^2}{32}} \leq \frac{C}{\xi}.
\]
Applying Lemma \ref{lem:chebyshev} to $t \mapsto \Phi(\bar yt)$ with $\kappa = 2$, input $y/\bar y$, degree $\lceil C\log(C/\xi)\rceil$ and $\tau \coloneqq c\xi^2$, for which $\delta = 0$ and its condition holds, and Lemma \ref{lem:gelu-arithmetic} with $\cX = \R$, we obtain $\widehat\Phi_\xi$ of depth $C\log(C/\xi)$, constant width and coefficients polynomial in $\xi^{-1}$ with
\[
\|\widehat\Phi_\xi-\Phi\|_\infty \leq \frac{\xi}{8}+\frac{\xi}{16} \leq \frac{\xi}{4}.
\]

We next approximate the trigonometric functions. For $|x| \leq M$ with $M \geq 1$, let
\[
j \coloneqq \lceil\log_2 M\rceil,\quad s_i \coloneqq \sin(2^{i-j}x)\quad\text{and}\quad c_i \coloneqq \cos(2^{i-j}x)\quad\text{for } 0 \leq i \leq j,
\]
and let $R \geq 1$ satisfy $R^2\tau \leq 1/9$. As $|\sin t|,|\cos t| \leq \cosh(3/4) \leq 2$ on $E_2$, by Lemma \ref{lem:chebyshev} with input $2^{-j}x$ and degree $R$, followed by projections to $[-1,1]$, we compute $\hat s_0$ and $\hat c_0$, and then approximate the recursion
\[
\begin{pmatrix}s_{i+1}\\c_{i+1}\end{pmatrix} = \begin{pmatrix}2s_ic_i\\c_i^2-s_i^2\end{pmatrix}
\]
for $0 \leq i \leq j$,
with $\tau$-accurate products and coordinatewise projections to $[-1,1]$. Define $$e_i \coloneqq \max\{|\hat s_i-s_i|,|\hat c_i-c_i|\}.$$  Then, we have $e_0 \leq 4\cdot2^{-R}+C\tau$. Since $|\hat s_i|,|\hat c_i| \leq 1+2\tau$ and $e_i \leq 5/2$, it holds that
\[
e_{i+1} \leq (4+4\tau)e_i+4\tau \leq 4e_i+14\tau,
\]
so that, as $4^j \leq 4M^2$,
\begin{equation}\label{eq:trig-error}
e_j \leq 4^je_0+14\tau\sum_{i = 0}^{j-1}4^i \leq 4M^2(e_0+5\tau) \leq 4M^2(4\cdot2^{-R}+C\tau).
\end{equation}
For $R \coloneqq \lceil\log_2(128M^2/\xi)\rceil$ and $\tau \coloneqq c\xi/M^2$ with a sufficiently small $c$, we have $R^2\tau \leq 1/9$ and $e_j \leq \xi/4$, and the map computing the trigonometric functions has depth $O(R+j)$ and constant width.

Finally, for $M = 4\pi m_K$, we use distribution function error $c\zeta/M$ and trigonometric error $c\zeta$, so that, by the mean value theorem, each tensor factor $f_\ell$ satisfies
\[
|\widehat f_\ell-f_\ell| \leq Cc\zeta+\|\varphi_{r^{(k)}_\ell}'\|_\infty\,\frac{c\zeta}{M} \leq Cc\zeta,\quad\text{for }\ell \in [d],
\]
where the trigonometric arguments remain in $[-M,M]$ because the approximate transforms lie in $[-1,2]$. On the finite set $\cZ_D$, we approximate the discrete factor $f_{d+1}$ to error $c\zeta$ by Lemma \ref{lem:gelu-arithmetic}, using the exact tent representation
\[
\chi_\eta(D) = \sum_{a \in \cZ_D}\chi_\eta(a)\left(1-\frac{\|D-a\|_1}{\delta_D}\right)_+,\quad\text{where}\quad\delta_D \coloneqq \min_{a,a' \in \cZ_D,\ a \ne a'}\|a-a'\|_1,
\]
with $\delta_D \coloneqq 1$ if $|\cZ_D| = 1$. Multiplying the factors with accuracy $c\zeta$ for each product, by a telescoping product bound, we obtain, for all $z \in \cZ$ and $k \in [K]$,
\begin{align*}
|[\widehat b_{K,\zeta}(z)]_k-b_k(z)| &\leq \sum_{\ell = 1}^{d+1}|\widehat f_\ell-f_\ell| \prod_{h < \ell}|\widehat f_h|\prod_{h > \ell}|f_h|+c\zeta\sum_{\ell = 0}^{d}C^\ell\\
&\leq C\max_{1 \leq \ell \leq d+1}|\widehat f_\ell-f_\ell|+Cc\zeta\\ & \leq Cc\zeta \leq \zeta.
\end{align*}
Since $M \leq CK$ by \eqref{eq:frequency-bound}, evaluating the $K$ basis functions in parallel requires depth $C\log(CK/\zeta)$ and width $CK$, with coefficients polynomially bounded in $K/\zeta$, which completes the proof.
\end{proof}

\begin{lemma}\label{lem:gelu-projection}
For all $K \geq 1$ and $M,\lambda \geq 1$, there is a GELU network $\widehat P:\R^K \to \R^K$ of depth $C\log(2K)$, width $CK$ and coefficient bound $(CM\lambda K)^C$ such that
\[
\sup_{z \in [-M,M]^K}\|\widehat P(z)-\operatorname{proj}_{\Theta_K}(z)\| \leq \frac{C\sqrt K\log(2K)}{\lambda},
\]
where $C$ does not depend on $K$, $M$ and $\lambda$, and $\operatorname{proj}_{\Theta_K}(z) \coloneqq \argmin_{\theta \in \Theta_K}\|\theta-z\|$ is the Euclidean projection onto the set $\Theta_K = \left\{\theta \in \R^K:\|\theta\|_1 \leq B\right\}$ of Section \ref{app:prior}.
\end{lemma}

\begin{proof}
We compute the projection onto $\Theta_K$ by soft thresholding at a level that requires only sorting, prefix averages and maxima.

For $z \in [-M,M]^K$, let $u_{(1)} \geq \cdots \geq u_{(K)}$ denote $|z_1|,\cdots,|z_K|$ in decreasing order. The threshold representation of \cite{duchi2008} reads
\[
t = \max\left\{0,\max_{j \in [K]}\frac{\sum_{i = 1}^j u_{(i)}-B}{j}\right\}\quad \text{and} \quad P_i(z) = (z_i-t)_+-(-z_i-t)_+.
\]
Since $\sum_i(u_{(i)}-t)_+ = \max_{0 \leq j \leq K}\{\sum_{i = 1}^j u_{(i)}-jt\}$, we have $t = 0$ if $\|z\|_1 \leq B$ and $\|P(z)\|_1 = B$ otherwise, and these conditions together with $z-P(z) \in t\,\partial\|P(z)\|_1$ identify $P(z)$ as the Euclidean projection. We compute $P$ in the stages
\[
z \mapsto (u_{(1)},\cdots,u_{(K)}) \mapsto \left(\frac{\sum_{i = 1}^j u_{(i)}-B}{j}\right)_{j \in [K]} \mapsto t \mapsto P(z),
\]
using the depth-$C\log(2K)$ sorting network of \cite{ajtai1983} with zero padding, one affine layer of prefix averages and a binary tree of maxima. This map has width $CK$ and coefficients bounded by $(CM\lambda K)^C$, and satisfies
\[
\|\widehat P(z)-P(z)\| \leq \sqrt K\,\|\widehat P(z)-P(z)\|_\infty \overset{\text{(i)}}{\leq} C\sqrt K\max_j e_j \overset{\text{(ii)}}{\leq} \frac{C\sqrt K\log(2K)}{\lambda},
\]
where $e_j$ is the maximum-norm error after $j$ stages, (i) follows from the Lipschitz dependence of soft thresholding on $t$, and (ii) from $e_{j+1} \leq e_j+C/\lambda$, as each stage is nonexpansive in the maximum norm up to the error $C/\lambda$ in \eqref{eq:gelu-relu}, which proves the lemma.
\end{proof}

For $\theta \in \R^K$ with $\|\theta\|_1 \leq B+1$ and $a \in \cZ_D$, we define the $1$-periodic trigonometric polynomial
\[
g_{\theta,a}(x) \coloneqq \sum_{k = 1}^K\theta_k\chi_{\eta^{(k)}}(a)\prod_{\ell = 1}^d\varphi_{r^{(k)}_\ell}(x_\ell)\quad\text{for } x \in \R^d,
\]
with $(r^{(k)},\eta^{(k)})$ as in \eqref{eq:max-frequency}, so that $\theta^\T b_{(K)}(v,a) = g_{\theta,a}\{\Phi_1(v_1),\cdots,\Phi_d(v_d)\}$ for $v \in \cV$. We call a map elementary if it is a finite composition of affine maps, products $(y,y') \mapsto yy'$, reciprocals $y \mapsto 1/y$ of inputs with $|y| \geq c$ and univariate analytic functions, namely functions of one variable that are analytic on $\left\{z \in \mathbb C:\operatorname{dist}(z,I) < c\right\}$ for a compact interval $I$ containing their inputs, where $c > 0$ is the constant of Definition \ref{def:integral-queries}.

\begin{definition}[Integral representation of queries]\label{def:integral-queries}
A query family $k \in \cK$ admits an integral representation if there exist $s_k \in \N$, a box $\cB_k \subset \R^{s_k}$, an elementary readout $R_k:\cB_k\times\cU_k \to \R^{d_k}$ and constants $c,C > 0$ such that, for all $K \geq 1$, $\theta \in \R^K$ with $\|\theta\|_1 \leq B+1$ and $u \in \cU_k$, the following conditions hold, with numbers of operations and fields, auxiliary dimensions, weights and univariate analytic functions independent of $K$.
\begin{enumerate}[label=(I\arabic*),ref=(I\arabic*)]
\item\label{cond:IQ1} The fields of family $k$ are functions $f:\R^p \to \R$, $p \in \N$, generated from the functions $x \mapsto e^{g_{\theta,a}(x)}$ on $\R^d$, $a \in \cZ_D$, by finitely many constant extensions $(t,x) \mapsto f_1(x)$ in added coordinates $t \in \R^{p_1}$ and operations
\begin{equation}\label{eq:field-operations}
f_1+f_2,\quad f_1f_2,\quad f_1/f_2\quad\text{and}\quad\int_{[0,1]^{p_1}}w(t)f_1(t,\cdot)\,dt,
\end{equation}
where $\inf_{x \in \R^p}|f_2(x)| \geq c$ and $w \in L_1([0,1]^{p_1})$.
\item\label{cond:IQ2} A root $v_k(\theta,u) \in \cB_k$ of an equation map $F_k(\theta,\cdot,u):\cB_k \to \R^{s_k}$ represents the query through
\begin{equation}\label{eq:integral-query-representation}
\beta_{(k,u)}(\P_\theta) = R_k\{v_k(\theta,u),u\}\quad\text{and}\quad F_k\{\theta,v_k(\theta,u),u\} = 0,
\end{equation}
where $\P_\theta$ is the tilt in \eqref{eq:exp-tilt}, and $F_k(\theta,v,u)$ is an elementary function of $(v,u)$ and of the values $f(x)$ and the integrals $\int_{[x^-,x^+]}f$ over rectangles $[x^-,x^+] \coloneqq \prod_{i = 1}^p[x^-_i,x^+_i]$ of finitely many fields $f:\R^p \to \R$ at points $x,x^-,x^+ \in [0,1]^p$ that are elementary functions of $(v,u)$.
\item\label{cond:IQ3} The root satisfies $\operatorname{dist}\{v_k(\theta,u),\partial\cB_k\} \geq c$, and, for all $v \in \cB_k$, the Jacobian $\partial_vF_k$ of $F_k$ in $v$ at $(\theta,v,u)$ satisfies
\begin{equation}\label{eq:integral-query-curvature}
\frac{\partial_vF_k+(\partial_vF_k)^\T}{2} \succeq cI\quad\text{and}\quad\|\partial_vF_k\|_{\op} \leq C.
\end{equation}
\end{enumerate}
\end{definition}

\begin{lemma}\label{lem:analytic-fields}
Suppose that a query family satisfies condition \ref{cond:IQ1} of Definition \ref{def:integral-queries}. Then there exist constants $c,C > 0$ such that, for all $K \geq 1$, $\theta \in \R^K$ with $\|\theta\|_1 \leq B+1$, fields $f:\R^p \to \R$ of this family and $h \in \left\{f,\partial_{\theta_1}f,\cdots,\partial_{\theta_K}f\right\}$, the function $h$ extends analytically to $\left\{x+\mathrm iy:x,y \in \R^p,\,\|y\|_\infty < c/m_K\right\}$ and is bounded there by $C$, with $m_K$ as in \eqref{eq:max-frequency}. If moreover $p \leq d$, then, for all odd $M_g \in \N$ and $e \in \R^{M_g^p}$,
\[
\|h-I_{M_g}h\|_\infty \leq Cm_K^p e^{-\frac{cM_g}{m_K}} \quad\text{and}\quad \|I_{M_g}e\|_\infty \leq M_g^p\|e\|_\infty,
\]
where $I_{M_g}$ denotes trigonometric interpolation of $1$-periodic functions on $\R^p$, and of vectors $e$ of nodal values, at the nodes $\left\{0,1/M_g,\cdots,(M_g-1)/M_g\right\}^p$.
\end{lemma}

\begin{proof}
We prove analyticity and the uniform bounds by induction over the operations in \ref{cond:IQ1}, and then derive both interpolation bounds from the decay of the Fourier coefficients.

Since $g_{\theta,a}$ has frequencies at most $m_K$ and absolute Fourier coefficients summing to at most $C_\infty\|\theta\|_1$, with $C_\infty$ the basis bound of Section \ref{app:prior}, at $x+\mathrm iy$ with $x,y \in \R^d$ and $\|y\|_\infty \leq c/m_K$ we have
\[
|g_{\theta,a}|+|b_k|+|e^{g_{\theta,a}}|+|b_ke^{g_{\theta,a}}| \leq C,\quad\text{where}\quad b_k = \partial_{\theta_k}g_{\theta,a}.
\]
By Fubini's theorem, integration over $t$ against $w$ commutes with contour integration, so that it preserves analyticity in the retained coordinates by Morera's theorem, Theorem~10.17 of \cite{rudin1987}, and uniform bounds since
\[
\left\|\int w(t)f(t,\cdot)\,dt\right\|_\infty \leq \|w\|_1\|f\|_\infty.
\]
By Cauchy's estimate in Theorem~10.26 of \cite{rudin1987}, the first spatial derivatives of each denominator $\varrho$ are bounded by $Cm_K$ on a narrower strip, and, shrinking its width by a constant factor, we obtain
\[
|\varrho(x+\mathrm i y)-\varrho(x)| \leq Cm_K\|y\|_1 \leq \frac{1}{2}\inf_x|\varrho(x)|\quad\text{and hence}\quad |\varrho(x+\mathrm i y)| \geq \frac{1}{2}\inf_x|\varrho(x)| > 0.
\]
Consequently, reciprocals and their coefficient derivatives $\partial_{\theta_k}(\varrho^{-1}) = -\varrho^{-2}\partial_{\theta_k}\varrho$ remain bounded, and, by induction over the sequence of operations, every field satisfies the analytic bounds. For the Fourier coefficients $\widehat f_r$, $r \in \mathbb Z^p$, of $f$, we have
\[
\|f-I_{M_g}f\|_\infty \overset{\text{(i)}}{\leq} 2\sum_{\|r\|_\infty > \frac{M_g-1}{2}}|\widehat f_r| \overset{\text{(ii)}}{\leq} C\sum_{\|r\|_\infty > \frac{M_g-1}{2}}e^{-\frac{c\|r\|_1}{m_K}} \leq Cm_K^p e^{-\frac{cM_g}{m_K}},
\]
where (i) holds because each omitted coefficient aliases into one retained frequency, and (ii) by a contour shift as in Theorem~4.1 of \cite{wright2015}. For nodal errors $e$ with discrete Fourier coefficients $\widehat e_r$, we have
\[
|\widehat e_r| \leq M_g^{-p}\sum_{j = 1}^{M_g^p}|e_j| \leq \|e\|_\infty\quad\text{and hence}\quad\|I_{M_g}e\|_\infty \leq \sum_{\|r\|_\infty \leq \frac{M_g-1}{2}}|\widehat e_r| \leq M_g^p\|e\|_\infty.
\]
Since these estimates apply to each coefficient derivative, the lemma follows.
\end{proof}

\begin{lemma}\label{lem:query-computation}
Let $k \in \cK$ be a query family with an integral representation as in Definition \ref{def:integral-queries}. Then there exist constants $A,c,C > 0$ such that, for every $K \geq 1$, the following hold.
\begin{enumerate}[label=(\roman*),nosep]
\item For all $\theta \in \R^K$ with $\|\theta\|_1 \leq B+1$ and $u,u' \in \cU_k$, we have $\|\beta_{(k,u)}(\P_\theta)-\beta_{(k,u')}(\P_\theta)\| \leq CK^A\|u-u'\|$.
\item For every integer $r \geq A\log(2K)$, there is $\Psi \in \cM_{t,s}$ with $t \leq Cr^2$, $s \leq CKr^d$, input $(\theta,u) \in \R^K\times\cU_k$ and coefficients bounded by $C(Kr)^A$ such that, for all $\tau \in [0,c(Kr)^{-A}]$, $\theta \in \R^K$ with $\|\theta\|_1 \leq B+1$ and $u \in \cU_k$, every $\tau$-perturbation of $\Psi$ at $(\theta,u)$ has intermediate values of maximum norm at most $C(Kr)^A$ and an output $(\widehat\beta,\widehat D)$ with
\begin{equation}\label{eq:stable-query-error}
\|\widehat\beta-\beta_{(k,u)}(\P_\theta)\|+\|\widehat D-D_{(k,u)}(\theta)\|_F \leq C\{e^{-r}+(Kr)^A\tau\},
\end{equation}
where $D_q$, with $q = (k,u)$, is the Jacobian of $\theta \mapsto \beta_q(\P_\theta)$ in Assumption \ref{cond:general-queries}(ii).
\end{enumerate}
\end{lemma}

\begin{proof}
We will prove this lemma by discretising the fields, and then solve the resulting equation by contraction and differentiate the implicit solution.

\paragraph{Step 1: Field bound \eqref{eq:query-field-accuracy}.}
Let $H \geq 3$, and choose odd $M_g \asymp C_Hm_Kr$, so that the grid size satisfies $S = M_g^d \leq C_HKr^d$. Since differentiation adds polynomial frequency factors to the bounds of Lemma \ref{lem:analytic-fields}, every field $f$ satisfies
\begin{equation}\label{eq:query-field-accuracy}
\max_{|\alpha| \leq 2}\left\{\|\partial_x^\alpha(f-I_{M_g}f)\|_\infty+\max_{i \in [K]}\|\partial_x^\alpha\partial_{\theta_i}(f-I_{M_g}f)\|_\infty\right\} \leq C_H(Kr)^{A_0}e^{-Hr}.
\end{equation}
The discrete Fourier transform and the integration of the interpolant over $x \in [0,1]^p$ against $w$ are affine, with $|\int_{[0,1]^p}w(x)e^{2\pi\mathrm i j^\T x}\,dx| \leq \|w\|_1$. Storing real and imaginary parts separately, we reduce rectangle integrals to products of the factors
\[
\int_a^b e^{2\pi\mathrm i jx}\,dx = b-a\ \ \text{for } j = 0\quad\text{and}\quad \int_a^b e^{2\pi\mathrm i jx}\,dx = \frac{e^{2\pi\mathrm i jb}-e^{2\pi\mathrm i ja}}{2\pi\mathrm i j}\ \ \text{for } j \ne 0.
\]
By Lemma \ref{lem:chebyshev} with degree $\lceil C_Hr\rceil$, and by \eqref{eq:trig-error} with $M \leq C_Hm_Kr$ and $R = \lceil C_Hr\rceil$ at depth $C_Hr$, the univariate analytic functions with their first two derivatives and the Fourier phases are computed from approximations $\hat x$ of their inputs $x$ with errors at most
\[
C_H(e^{-Hr}+\tau)+C|\hat x-x|\quad\text{and}\quad C_H(Kr)^{A_0}(e^{-Hr}+\tau)+C_HKr|\hat x-x|,
\]
respectively, since these univariate functions are analytic on complex neighbourhoods of constant width around their input intervals.

\paragraph{Step 2: Bound \eqref{eq:query-oracle-accuracy} on $\widehat F_k$ and $\widehat J_k$.}
We store each field on its grid and evaluate the interpolants and their rectangle integrals as in Step~1. On the grid $\mathcal G \coloneqq \left\{0,1/M_g,\cdots,(M_g-1)/M_g\right\}^d$, the initial values depend linearly on $\theta$,
\[
g \coloneqq (g_{\theta,a}(z))_{(a,z) \in \cZ_D\times\mathcal G} = B_{\mathcal G}\theta,\quad\text{where}\quad (B_{\mathcal G})_{(a,z),:} \coloneqq \{\nabla_\theta g_{\theta,a}(z)\}^\T.
\]
Each field stage has at most $CS$ coordinates, and the affine matrices and their transposes have row sums polynomial in $Kr$, so that, by the scalar derivative bounds and \eqref{eq:query-field-accuracy}, the maximum errors $e_j$ of consecutive stages satisfy
\[
e_{j+1} \leq C_H(Kr)^{A_0}(e_j+e^{-Hr}+\tau).
\]
Differentiating each auxiliary coordinate forward and each equation component $F^{\rm grid}_{k,l}$ in reverse mode \citep{baur1983}, with transposed affine maps and the derivatives of the univariate functions, we obtain
\[
\nabla_\theta F^{\rm grid}_{k,l} = B_{\mathcal G}^\T \nabla_g F^{\rm grid}_{k,l},
\]
where each pass uses $O(S+K)$ coordinates. Approximating the derivatives of the univariate functions on enlarged ranges and applying the chain rule, we obtain the same error recursion for derivatives, and hence, with $J_k \coloneqq \partial_v F_k$,
\begin{equation}\label{eq:query-oracle-accuracy}
\|\widehat F_k-F_k\|+\|\widehat J_k-J_k\|_{\op}+\|\widehat F_{k,\theta}-\partial_\theta F_k\|_F \leq C_H(Kr)^{A_0}(e^{-Hr}+\tau) \eqqcolon \delta_r,
\end{equation}
where $\widehat F_k$, $\widehat J_k$ and $\widehat F_{k,\theta}$ are the computed values of $F_k$, $J_k$ and $\partial_\theta F_k$, and $A_0$ includes the constant number of stages and the Frobenius factor $\sqrt K$, independently of $H$. Each evaluation has depth $C_Hr$, and, by the differentiated field bounds, we have
\begin{equation}\label{eq:query-jacobian-modulus}
\|J_k(\theta,v,u)-J_k(\theta,w,u)\|_{\op}+\|\partial_\theta F_k(\theta,v,u)-\partial_\theta F_k(\theta,w,u)\|_F \leq CK^{A_0}\|v-w\|.
\end{equation}

\paragraph{Step 3: Root bound \eqref{eq:query-root-accuracy}.}
Let $v_* \coloneqq v_k(\theta,u)$ and $\cB^- \coloneqq \left\{v \in \cB_k:\operatorname{dist}(v,\partial\cB_k) \geq c/2\right\}$. Integrating \eqref{eq:integral-query-curvature} at $w+s(v-w)$ over $s \in [0,1]$ for $v,w \in \cB_k$, with $\gamma \coloneqq c/(2C^2)$, we obtain the contraction bound
\[
\|v-w-\gamma\{F_k(\theta,v,u)-F_k(\theta,w,u)\}\|^2 \leq (1-2\gamma c+\gamma^2C^2)\|v-w\|^2 \eqqcolon \rho^2\|v-w\|^2,
\]
where $\rho < 1$. Let $\widehat P^-$ approximate, with error $C\tau$, the projection onto $\cB^-$, which is nonexpansive and leaves $v_*$ unchanged. From a point $\widehat v_0 \in \cB^-$, we iterate
\[
\widehat v_{j+1} = \widehat P^-\{\widehat v_j-\gamma\widehat F_k(\theta,\widehat v_j,u)\}.
\]
For $C\tau < c/2$, the iterates stay in $\cB_k$, where $v \mapsto \operatorname{proj}_{\cB^-}\{v-\gamma F_k(\theta,v,u)\}$ contracts towards $v_*$ by $\rho$ and, by \eqref{eq:query-oracle-accuracy}, deviates from each update by at most $C\delta_r$, so that, by Lemma \ref{lem:perturbed-iterations}(i),
\begin{equation}\label{eq:query-root-accuracy}
\|\widehat v_J-v_*\| \leq C\rho^J+\frac{C\delta_r}{1-\rho}.
\end{equation}

\paragraph{Step 4: Implicit Jacobian \eqref{eq:query-implicit-jacobian}.}
We take $J = \lceil C_Hr\rceil$, so that $\rho^J \leq e^{-Hr}$. Since strong monotonicity ensures $\|J_k^{-1}\|_{\op} \leq c^{-1}$, by the implicit function theorem, Theorem~9.28 of \cite{rudin1976}, we have
\begin{equation}\label{eq:query-implicit-jacobian}
D_{(k,u)}(\theta) = -\partial_vR_k(v_*,u)\,J_k(\theta,v_*,u)^{-1}\partial_\theta F_k(\theta,v_*,u).
\end{equation}
We approximate the three factors at $\widehat v_J$, substitute them in \eqref{eq:query-implicit-jacobian} to define $\widehat D$ and let $\widehat\beta$ evaluate $R_k(\widehat v_J,u)$. For $\delta_r \leq c/2$, the exact and approximate Jacobians satisfy
\[
\|\widehat J_k^{-1}\|_{\op} \leq 2/c,\quad |\det\widehat J_k| \geq (c/2)^{s_k}\quad\text{and}\quad\widehat J_k^{-1}-J_k^{-1} = \widehat J_k^{-1}(J_k-\widehat J_k)J_k^{-1},
\]
and the determinant bound permits adjugate inversion to error $C(e^{-Hr}+\tau)$ at depth $C_Hr$ and constant additional width. It follows from this inverse comparison, \eqref{eq:query-oracle-accuracy}, \eqref{eq:query-jacobian-modulus} and \eqref{eq:query-root-accuracy} that
\[
\|\widehat\beta-\beta_{(k,u)}(\P_\theta)\|+\|\widehat D-D_{(k,u)}(\theta)\|_F \leq C_H(Kr)^{A_1}(e^{-Hr}+\tau).
\]
Since $K \leq e^{r/A}$, for $A \geq A_1+2$ we have $(Kr)^{A_1+2}e^{-Hr} \leq Ce^{-r} \leq C2^{-A}$. Hence, for a sufficiently large $A$ and $\tau \leq c(Kr)^{-A}$ with a sufficiently small $c$, the smallness conditions of Lemma \ref{lem:chebyshev} and \eqref{eq:trig-error} in Step~1 and on $\tau$ and $\delta_r$ in Steps~3 and~4 hold, and
\[
C_H(Kr)^{A_1}(e^{-Hr}+\tau) \lesssim e^{-r}+(Kr)^A\tau,
\]
which proves \eqref{eq:stable-query-error}. For the size, all coefficients and intermediate values remain polynomially bounded, and the $J$ updates of depth $O(r)$ and the final derivative evaluation reuse the grid coordinates, so that the map has
\[
\text{depth} \leq C(J+1)r \leq Cr^2\quad\text{and}\quad\text{width} \leq C(S+K) \leq CKr^d.
\]
Applying Lemma \ref{lem:perturbed-iterations}(ii) to $G = F_k(\theta,\cdot,u)$, which is strongly monotone on $\cB_k$ by \eqref{eq:integral-query-curvature}, $G' = F_k(\theta,\cdot,u')$ and their roots $v \coloneqq v_k(\theta,u)$ and $v' \coloneqq v_k(\theta,u')$, and using the Lipschitz dependence of the defining operations on the query inputs, we obtain
\[
c\|v-v'\| \leq \|F_k(\theta,v',u)-F_k(\theta,v',u')\| \leq CK^A\|u-u'\|,
\]
and, since the readout $R_k$ is Lipschitz, $\beta_{(k,u)}(\P_\theta)$ is $CK^A$-Lipschitz in $u$, which completes the proof.
\end{proof}

\subsubsection{Metric entropy and stochastic error}\label{app:pretraining-error}

%The lemmas below provide the curvature bound used in Step~1 of the proof of Theorem \ref{thm:TF-error}, the metric entropy bound of its Step~3 and the localised bound on the stochastic error of its Step~4. Lemma \ref{lem:central} shows that the score difference $g_f$ of \eqref{eq:score-difference}, centred at its conditional mean $h_f$, satisfies a Bernstein-type cumulant bound with variance proxy proportional to $h_f$. This variance bound localises the stochastic error, so that Lemma \ref{lem:general-comparison} controls the error at the rate of the metric entropy divided by $T$ through one exponential inequality over a loss cover, without peeling.
Throughout this subsection, expectations and probabilities refer to pretraining, and the context $X$, the conditional excess score $h_f$, the empirical mean $\hat\E_T$ and the moment error $e$ at $X$ are as in Section \ref{app:proof-tf-error}.

\begin{definition}[Entropy of the loss class]\label{def:loss-entropy}
For $\bar z \geq 0$, let
\[
G_{\bar z} \coloneqq \bigcup_{n \in \cN}\left\{(\cD_n,q) \in \cZ^n\times\cQ:\max_{i \in [n]}\|Z_i\|_\infty \leq \bar z\right\},
\]
and, for functions $\phi$ of $(X,\beta_0)$, let $\|\phi\|_{\bar z} \coloneqq \sup\left\{|\phi(X,\beta_0)|:X \in G_{\bar z},\ \|\beta_0\| \leq C_\beta\right\}$, where $C_\beta$ is the radius of the point readout in Definition \ref{def:transformer-output}, which also bounds the labels $\beta_q(\P)$. For $u > 0$, let
\[
\cH_{\mathrm{TF}}(\cN,\bar z,u) \coloneqq \log\min\left\{|\cF'|:\cF' \subseteq \cF_{\mathrm{TF}},\ \cF_{\mathrm{TF}} \subseteq \bigcup_{f' \in \cF'}\left\{f \in \cF_{\mathrm{TF}}:\|g_f-g_{f'}\|_{\bar z} \leq u\right\}\right\}.
\]
\end{definition}

\begin{lemma}[Curvature and Bernstein-type bound]\label{lem:central}
Suppose that Assumption \ref{cond:general-queries}(i) holds. Then there exist constants $c,C,\tau_0 > 0$, depending only on $d_0$, $v_-$, $v_+$ and $s_0$, such that, for every $f \in \cF_{\mathrm{TF}}$ and every context $X$, $e(f,f_\pi^\circ) \leq Ch_f$ and
\begin{equation}\label{eq:conditional-central}
\log\E\{e^{t(g_f-h_f)}\mid X\} \leq Ct^2h_f\quad\text{for } t \in [-\tau_0,\tau_0]\quad\text{and}\quad\E(e^{-\tau_0g_f}\mid X) \leq e^{-ch_f},
\end{equation}
where $g_f = g_f(X,\beta_q(\P))$, $h_f = h_f(X)$, $d_0 = \max_{k \in \cK}d_k$, $[v_-,v_+]$ is the eigenvalue range of the covariance readout in Definition \ref{def:transformer-output} and $s_0$ is the constant of Lemma \ref{lem:subgauss}.
\end{lemma}

\begin{proof}
We split the score difference into a centred quadratic term and a linear term in the posterior fluctuation, bound $h_f$ below by curvature and control both terms by Lemma \ref{lem:subgauss}.

For $X = (\cD_n,q)$ and $\beta_0 = \beta_q(\P)$, let $U \coloneqq \sqrt n(\beta_0-m_\pi)$, $D \coloneqq \sqrt n(f^\beta-m_\pi)$, $A \coloneqq (f^V)^{-1}$, $A_* \coloneqq (\clip S_\pi)^{-1}$ and $B \coloneqq A-A_*$, so that, by \eqref{eq:posterior-loss-regret}, with $F_C(A') = \tr(A'C)-\log\det A'$ as in \eqref{eq:clipped-precision},
\[
2h_f = D^\T AD+F_{S_\pi}(A)-F_{S_\pi}(A_*).
\]
Since $\E(U\mid X) = 0$ and $\Cov(U\mid X) = S_\pi$, expanding \eqref{eq:score}, we obtain
\begin{equation}\label{eq:conditional-score-decomposition}
g_f-h_f = \frac{1}{2}\{U^\T BU-\tr(BS_\pi)\}-D^\T AU.
\end{equation}
By the clip in Definition \ref{def:transformer-output}, $A$ lies in the precision interval $\mathcal I \coloneqq \left\{A':v_+^{-1}I \preceq A' \preceq v_-^{-1}I\right\}$, on which the Hessian of $F_{S_\pi}$ satisfies
\[
\nabla^2F_{S_\pi}(A')[B,B] = \|A'^{-\frac{1}{2}}BA'^{-\frac{1}{2}}\|_F^2 \geq v_-^2\|B\|_F^2.
\]
Since $A_*$ minimises $F_{S_\pi}$ on the convex set $\mathcal I \ni A$ by \eqref{eq:clipped-precision}, we have $\partial_tF_{S_\pi}(A_*+tB)|_{t = 0} \geq 0$, and, integrating $t \mapsto (1-t)\nabla^2F_{S_\pi}(A_*+tB)[B,B]$ over $t \in [0,1]$ in Taylor's formula, we obtain
\begin{equation}\label{eq:score-curvature}
h_f \geq \frac{\|D\|^2}{2v_+}+\frac{v_-^2}{4}\|B\|_F^2.
\end{equation}
Since $f^V-\clip S_\pi = -A^{-1}BA_*^{-1}$ and $\|A^{-1}\|_{\op},\|A_*^{-1}\|_{\op} \leq v_+$, by \eqref{eq:score-curvature} we have
\[
e(f,f_\pi^\circ) = \|D\|^2+\|f^V-\clip S_\pi\|_F^2 \leq \|D\|^2+v_+^4\|B\|_F^2 \leq \left(2v_++\frac{4v_+^4}{v_-^2}\right)h_f.
\]

By Lemma \ref{lem:subgauss} at size $n$, for $u \in \R^{d_k}$ with $\|u\| = 1$, $t \in \R$ and $m \geq 1$, we have
\begin{equation}\label{eq:fluctuation-moments}
\log\E(e^{tu^\T U}\mid X) \leq \frac{s_0^2t^2}{2}\quad\text{so that}\quad\max_{j \in [d_k]}\E(|U_j|^{2m}\mid X) \leq 2(2s_0^2)^m m!.
\end{equation}
For $\Upsilon \coloneqq U^\T BU-\tr(BS_\pi)$, by \eqref{eq:fluctuation-moments} and the Cauchy--Schwarz and Minkowski inequalities, we have
\[
\{\E(|\Upsilon|^m\mid X)\}^{\frac{1}{m}} \leq 2\sum_{i,j}|B_{ij}|\{\E(|U_iU_j|^m\mid X)\}^{\frac{1}{m}} \leq C(m!)^{\frac{1}{m}}\|B\|_F.
\]
Let $t_1 \coloneqq v_-/(4C\sqrt{d_0})$. Since $\E(\Upsilon\mid X) = 0$ and $\|B\|_F \leq 2\sqrt{d_0}/v_-$, expanding $e^{t\Upsilon} = \sum_{m \geq 0}(t\Upsilon)^m/m!$, we have, for $|t| \leq t_1$,
\begin{equation}\label{eq:quadratic-mgf}
\E(e^{t\Upsilon}\mid X) \leq 1+\sum_{m = 2}^\infty(|t|C\|B\|_F)^m \leq \exp(2C^2t^2\|B\|_F^2).
\end{equation}
Applying the conditional Cauchy--Schwarz inequality to \eqref{eq:conditional-score-decomposition}, we obtain, for $|t| \leq t_1$,
\begin{align*}
\log\E\{e^{t(g_f-h_f)}\mid X\} &\leq \frac{1}{2}\log\E(e^{t\Upsilon}\mid X)+\frac{1}{2}\log\E(e^{-2tD^\T AU}\mid X)\\
&\overset{\text{(i)}}{\leq} C t^2(\|B\|_F^2+\|D\|^2) \overset{\text{(ii)}}{\leq} C_1t^2h_f,
\end{align*}
where (i) follows from \eqref{eq:quadratic-mgf} and \eqref{eq:fluctuation-moments}, and (ii) from \eqref{eq:score-curvature}. Taking $0 < \tau_0 \leq \min\{t_1,(2C_1)^{-1}\}$, we conclude that $\log\E(e^{-\tau_0g_f}\mid X) \leq -\tau_0h_f+C_1\tau_0^2h_f \leq -\tau_0h_f/2$, which proves \eqref{eq:conditional-central} with $c = \tau_0/2$.
\end{proof}

For $\delta \in (0,1/2)$, let $\bar z \coloneqq \{2\log(16dC_0n_{\max}T/\delta)\}^{\frac{1}{2}}$, where $C_0$ is the upper density bound of $\cP$ in Section \ref{app:prior}. Since $\bar z > 1$, $d\P/d\P_0 \leq C_0$ for $\P \in \cP$ and $\P_0(|V_\ell| > \bar z) \leq 2e^{-\bar z^2/2}$ for the $d$ continuous coordinates $V_\ell$ of $Z$, by a union bound, we have
\begin{equation}\label{eq:pretraining-support}
T\Pr(X \notin G_{\bar z}) \leq 2dC_0T\sum_{n \in \cN}\pi_n(n)ne^{-\frac{\bar z^2}{2}} \leq \frac{\delta}{8},
\end{equation}
and we let $H_\delta \coloneqq \cH_{\mathrm{TF}}(\cN,\bar z,T^{-1})$ and $u_\delta \coloneqq \log(8/\delta)$.

\begin{lemma}[Localised stochastic error]\label{lem:general-comparison}
Suppose that Assumption \ref{cond:general-queries}(i) holds. Then there exists a constant $C$, depending only on $d_0$, $v_-$, $v_+$ and $s_0$, such that, for every $\delta \in (0,1/2)$, $\epsilon_0 \geq 0$ and deterministic $f^\circ \in \cF_{\mathrm{TF}}$ with $\E h_{f^\circ} \leq \epsilon_0$, on an event of pretraining probability at least $1-\delta$, every $\hat f$ satisfying \eqref{eq:erm} obeys
\[
\E\{1\wedge e(\hat f,f_\pi^\circ)\mid\cT\} \leq C\left(\epsilon_{\mathrm{opt}}+\frac{H_\delta+u_\delta}{T}+\frac{\epsilon_0}{\delta}\right),
\]
where $H_\delta$ and $u_\delta$ are as in \eqref{eq:pretraining-support}.
\end{lemma}

\begin{proof}
We bound the stochastic error in \eqref{eq:risk-decomposition} on a loss cover by an exponential inequality \citep{vanerven2015,grunwald2020}, bound the approximation error $\hat\E_T(g_{f^\circ})$, and then pass from the cover to $\hat f$ through \eqref{eq:erm}.

Let $c$ and $\tau_0$ be as in Lemma \ref{lem:central} and $\cJ(f) = \E(1\wedge h_f)$ as in Step~1 of the proof of Theorem \ref{thm:TF-error}, and let $c_2 \coloneqq 1-e^{-c}$. For deterministic $f \in \cF_{\mathrm{TF}}$, by \eqref{eq:conditional-central}, we have
\[
\E(e^{-\tau_0g_f}) \leq \E(e^{-ch_f}) \leq 1-c_2\cJ(f) \leq e^{-c_2\cJ(f)},
\]
where the second inequality uses $1-e^{-ch} \geq c_2(1\wedge h)$ for $h \geq 0$, and, by the independence of the triples $(\cD^{(t)}_{n_t},q_t,\beta_{0,t})$, $t \in [T]$,
\begin{equation}\label{eq:learn-exponential}
\E\exp\{c_2T\cJ(f)-\tau_0T\hat\E_T(g_f)\} = e^{c_2T\cJ(f)}\{\E(e^{-\tau_0g_f})\}^{T} \leq 1.
\end{equation}
By Definition \ref{def:loss-entropy}, there exist deterministic $f_1,\cdots,f_M \in \cF_{\mathrm{TF}}$ with $\log M \leq H_\delta$ and $\min_{j \in [M]}\|g_f-g_{f_j}\|_{\bar z} \leq T^{-1}$ for every $f \in \cF_{\mathrm{TF}}$. Applying Markov's inequality to \eqref{eq:learn-exponential} at each $f_j$ and a union bound, we obtain
\[
\Pr(\cE_1^c) \leq M\frac{\delta}{8M} = \frac{\delta}{8},\quad\text{where}\quad\cE_1 \coloneqq \left\{\max_{j \in [M]}\{c_2T\cJ(f_j)-\tau_0T\hat\E_T(g_{f_j})\} \leq \log(8M\delta^{-1})\right\}.
\]
Consequently, on $\cE_1$, we have
\begin{equation}\label{eq:net-comparison}
\cJ(f_j) \leq \frac{\tau_0}{c_2}\hat\E_T(g_{f_j})+\frac{\log(8M\delta^{-1})}{c_2T}\quad\text{for }j \in [M].
\end{equation}

Let $\tau \in (0,\tau_0]$ satisfy $C\tau \leq 1$, with $C$ as in \eqref{eq:conditional-central}. By \eqref{eq:conditional-central} and the independence of the $T$ triples in $\cT$, we have
\[
\E\{e^{\tau(g_{f^\circ}-2h_{f^\circ})}\mid X\} \leq e^{(-\tau+C\tau^2)h_{f^\circ}} \leq 1\quad\text{and}\quad\E\exp[\tau T\{\hat\E_T(g_{f^\circ})-2\hat\E_T(h_{f^\circ})\}] \leq 1.
\]
Since $\E h_{f^\circ} \leq \epsilon_0$, applying Markov's inequality to $\hat\E_T(h_{f^\circ})$ and to $\exp[\tau T\{\hat\E_T(g_{f^\circ})-2\hat\E_T(h_{f^\circ})\}]$, we obtain
\begin{equation}\label{eq:circ-score-tail}
\begin{aligned}
\Pr\left\{\hat\E_T(g_{f^\circ}) > \frac{16\epsilon_0}{\delta}+\frac{u_\delta}{\tau T}\right\} &\leq \Pr\left\{\hat\E_T(h_{f^\circ}) > \frac{8\epsilon_0}{\delta}\right\}+\Pr\left\{\hat\E_T(g_{f^\circ})-2\hat\E_T(h_{f^\circ}) > \frac{u_\delta}{\tau T}\right\}\\
&\leq \frac{\delta}{8}+e^{-u_\delta} = \frac{\delta}{4}.
\end{aligned}
\end{equation}

We define the event
\[
\cE \coloneqq \cE_1\cap\bigcap_{t \in [T]}\left\{(\cD^{(t)}_{n_t},q_t) \in G_{\bar z}\right\}\cap\left\{\hat\E_T(g_{f^\circ}) \leq \frac{16\epsilon_0}{\delta}+\frac{u_\delta}{\tau T}\right\}.
\]
By a union bound, \eqref{eq:pretraining-support} and \eqref{eq:circ-score-tail}, we have
\[
\Pr(\cE^c) \leq \frac{\delta}{8}+T\Pr(X \notin G_{\bar z})+\frac{\delta}{4} \leq \frac{\delta}{2}.
\]
On $\cE$, let $j \in [M]$ satisfy $\|g_{\hat f}-g_{f_j}\|_{\bar z} \leq T^{-1}$. Since $\|\beta_{0,t}\| \leq C_\beta$ and $(\cD^{(t)}_{n_t},q_t) \in G_{\bar z}$ for $t \in [T]$, by \eqref{eq:erm}, we have
\begin{equation}\label{eq:cover-erm}
\hat\E_T(g_{f_j}) \leq \hat\E_T(g_{\hat f})+T^{-1} \leq \hat\E_T(g_{f^\circ})+\epsilon_{\mathrm{opt}}+T^{-1}.
\end{equation}
Since $|h_{f_j}-h_{\hat f}| \leq \|g_{\hat f}-g_{f_j}\|_{\bar z} \leq T^{-1}$ on $G_{\bar z}$ and $x \mapsto 1\wedge x$ is $1$-Lipschitz with values in $[0,1]$, we have, on $\cE$,
\begin{align*}
\E\{1\wedge e(\hat f,f_\pi^\circ)\mid\cT\} &\overset{\text{(i)}}{\lesssim} \cJ(\hat f) \leq \cJ(f_j)+T^{-1}+\Pr(X \notin G_{\bar z})\\
&\overset{\text{(ii)}}{\leq} \frac{\tau_0}{c_2}\{\hat\E_T(g_{f^\circ})+\epsilon_{\mathrm{opt}}+T^{-1}\}+\frac{H_\delta+u_\delta}{c_2T}+\frac{1+\delta/8}{T}\\
&\overset{\text{(iii)}}{\lesssim} \epsilon_{\mathrm{opt}}+\frac{H_\delta+u_\delta}{T}+\frac{\epsilon_0}{\delta},
\end{align*}
where (i) follows from Lemma \ref{lem:central} and $1\wedge(Ch) \leq C(1\wedge h)$ for $C \geq 1$, (ii) from \eqref{eq:pretraining-support}, \eqref{eq:net-comparison}, \eqref{eq:cover-erm} and $\log(8M\delta^{-1}) \leq H_\delta+u_\delta$, and (iii) from the definition of $\cE$ and $T^{-1} \leq u_\delta/T$. Since $\Pr(\cE^c) \leq \delta/2$, the lemma follows.
\end{proof}

To control parameter perturbations of the loss class in Definition \ref{def:loss-entropy}, for the class $\cF_{\mathrm{TF}} = \cF_{\mathrm{TF}}(L,J,d_e,d_f,W)$ of Section \ref{app:architecture}, with $p$ free parameters and coefficient bound $W \geq 1$, and for $N \in \cN$ and $\bar z \geq 0$, we define
\[
L_0(N,\bar z) \coloneqq C(1+N)\{C(N+J+d_e+d_f+W)(1+\bar z)\}^{C(L+1)^2},
\]
where the constants $C$ depend only on $d_z$, $d_0$, the query encoding $\xi$ of Section \ref{app:architecture}, $C_\beta$ and $v_\pm$.

\begin{lemma}[Bound on the metric entropy]\label{lem:entropy}
For every $\bar z \geq 0$ and $u > 0$, the class $\cF_{\mathrm{TF}} = \cF_{\mathrm{TF}}(L,J,d_e,d_f,W)$ satisfies
\[
\cH_{\mathrm{TF}}(\cN,\bar z,u) \leq p\log\left\{1+\frac{4WL_0(n_{\max},\bar z)}{u}\right\}.
\]
\end{lemma}

\begin{proof}
We bound the propagation of a parameter perturbation through the blocks, deduce a Lipschitz bound on the loss in the parameters, and cover the parameter box $[-W,W]^p$ by a grid whose mesh is set by this Lipschitz bound.

At a context $(\cD_N,q) \in G_{\bar z}$ with $N \in \cN$, let $U \coloneqq C(N+J+d_e+d_f+W)(1+\bar z) \geq 2$. For parameters $\omega,\omega'$ in the set $\Omega \subseteq [-W,W]^p$ of Section \ref{app:architecture}, we compare the token arrays $H_l$ and $H_l'$ after block $l$ in the norm $\|H\|_{\max} \coloneqq \max_{i,j}|H_{ij}|$ through
\[
\zeta \coloneqq \|\omega-\omega'\|_\infty,\quad h_l \coloneqq \max\{1,\|H_l\|_{\max},\|H_l'\|_{\max}\}\quad \text{and} \quad e_l \coloneqq \|H_l-H_l'\|_{\max}.
\]
The affine embeddings satisfy $h_0 \leq U^C$ and $e_0 \leq U^C\zeta$, and, since the rows of $\softmax(\cdot)$ are probability vectors and $|\GELU(x)| = |x|\Phi(x) \leq |x|$, we have $h_{l+1} \leq U^Ch_l$, so that $h_l \leq U^{C(l+1)}$. For $a,v \in \R^m$, the derivative $\D\softmax(a)$ of $\softmax$ at $a$ satisfies
\[
\{\D\softmax(a)v\}_i = \softmax(a)_i\{v_i-\softmax(a)^\T v\},
\]
which implies that $\|\D\softmax(a)v\|_1 \leq 2\|v\|_\infty$, and, integrating $t \mapsto \D\softmax\{a'+t(a-a')\}(a-a')$ over $t \in [0,1]$, we have
\begin{equation}\label{eq:softmax-lipschitz}
\|\softmax(a)-\softmax(a')\|_1 \leq 2\|a-a'\|_\infty.
\end{equation}
For $j \in [J]$ and $i \in [N+1]$, the query--key products $a_r \coloneqq \{H_l\mathsf Q_j(H_l\mathsf K_j)^\T\}_{ir}/\sqrt{d_e}$, for $r$ with $\mathsf M_j(i,r) = 0$ in the masks of Section \ref{app:architecture}, and $a_r'$ under $(\omega',H_l')$ satisfy $\|a-a'\|_\infty \leq CU^Ch_l(e_l+h_l\zeta)$, and hence, by \eqref{eq:softmax-lipschitz} and $\sup_x|\GELU'(x)| \leq 2$, we have
\begin{equation}\label{eq:block-perturbation}
e_{l+1} \leq CU^C\{(1+h_l^2)e_l+(1+h_l^3)\zeta\} \leq U^{C(l+1)}(e_l+\zeta).
\end{equation}
Iterating \eqref{eq:block-perturbation} over $l$ and summing the exponents, we obtain
\begin{equation}\label{eq:token-perturbation}
e_l+\zeta \leq 2U^C\zeta\prod_{j = 0}^{l-1}2U^{C(j+1)} \leq U^{C(l+1)^2}\zeta.
\end{equation}

Let $(\beta,V) \coloneqq f_\omega(\cD_N,q)$ and $(\beta',V') \coloneqq f_{\omega'}(\cD_N,q)$. Since $V_t \coloneqq V'+t(V-V') \succeq v_-I$ for $t \in [0,1]$, integrating $t \mapsto \partial_t\log\det V_t = \tr\{V_t^{-1}(V-V')\}$ over $t \in [0,1]$ and using $V^{-1}-(V')^{-1} = V^{-1}(V'-V)(V')^{-1}$, we obtain
\[
|\log\det V-\log\det V'| \leq \sqrt{d_0}\,v_-^{-1}\|V-V'\|_F\quad\text{and}\quad\|V^{-1}-(V')^{-1}\|_F \leq v_-^{-2}\|V-V'\|_F.
\]
Since $\|\beta_0\|,\|\beta\|,\|\beta'\| \leq C_\beta$, it follows that
\begin{equation}\label{eq:loss-parameter-lipschitz}
|\ell_N(\beta_0;\beta,V)-\ell_N(\beta_0;\beta',V')| \leq C(1+N)\{\|\beta-\beta'\|+\|V-V'\|_F\} \overset{\text{(i)}}{\leq} C(1+N)U^{C(L+1)^2}\zeta,
\end{equation}
where (i) follows from \eqref{eq:token-perturbation} with $l = L$, the affine readout and the nonexpansive maps in Definition \ref{def:transformer-output}.

Finally, since $L_0(N,\bar z)$ increases with $N$, by \eqref{eq:loss-parameter-lipschitz}, we have
\[
\|g_{f_\omega}-g_{f_{\omega'}}\|_{\bar z} \leq L_0(n_{\max},\bar z)\|\omega-\omega'\|_\infty.
\]
We partition $[-W,W]^p$ into at most $\{1+4WL_0(n_{\max},\bar z)/u\}^p$ cells $R_s$ of maximum-norm diameter at most $u/L_0(n_{\max},\bar z)$ and choose $\omega_s \in \Omega\cap R_s$ whenever $\Omega\cap R_s \neq \emptyset$. Since every $\omega \in \Omega\cap R_s$ satisfies $\|g_{f_\omega}-g_{f_{\omega_s}}\|_{\bar z} \leq u$, by Definition \ref{def:loss-entropy}, we obtain
\[
\exp\{\cH_{\mathrm{TF}}(\cN,\bar z,u)\} \leq \left\{1+\frac{4WL_0(n_{\max},\bar z)}{u}\right\}^p,
\]
which is the asserted bound after taking logarithms.
\end{proof}

\subsection{Proof of Theorem \ref{thm:main}}\label{app:proof-main-theorem}

\begin{proof}
We first show that $\hat f$ is efficient, then approximate the distribution of the influence term $\sqrt N\P_N\psi^\star_{\P,q}$ on average under $\nu$, and finally deduce the normal limit and the coverage.

By \eqref{eq:explicit-learning-budget}, the class $\cF_{\mathrm{TF}}$ in the statement satisfies $\tilde r_N \lesssim r_N = o(1)$, with $\tilde r_N$ as in \eqref{eq:learning-budget}, and hence, by Theorem \ref{thm:target} with $K \to \infty$ and \eqref{eq:tracking-general}, the efficient benchmark $f_\star \coloneqq (\beta_q(\P)+\P_N\psi^\star_{\P,q},V_q^\star(\P))$ satisfies
\begin{equation}\label{eq:efficiency-split}
e(\hat f,f_\star) \leq 2e(\hat f,f_\pi)+2e(f_\pi,f_\star) = o_\nu(1),
\end{equation}
where $e$ is the moment error in \eqref{eq:moment-error}, and $e(\hat f,f_\star) = o_\nu(1)$ is equivalent to \eqref{eq:goal} and proves the second limit in \eqref{eq:efficient-normality}.

We compare the standardised influence term $Z_N \coloneqq N^{-\frac{1}{2}}\sum_{i = 1}^NV_q^\star(\P)^{-\frac{1}{2}}\psi^\star_{\P,q}(Z_i)$ with $G \sim \cN(0,I_{d_k})$ through
\[
\Delta_N(\P,q) \coloneqq \sup_{s \geq 0}|\Pr(\|Z_N\| \leq s)-\Pr(\|G\| \leq s)|
\]
and $\Delta_N^{\mathrm{BL}}(\P,q)$, the bounded Lipschitz distance $d_{\mathrm{BL}}$ between the distributions of $Z_N$ and $G$, where $\Delta_N(\P,q) = \Delta_N^{\mathrm{BL}}(\P,q) \coloneqq 0$ unless $\P \in \cP$ and $v_-I \preceq V_q^\star(\P) \preceq v_+I$, with $v_\pm$ the covariance readout bounds of Definition \ref{def:transformer-output}. We now show that
\begin{equation}\label{eq:normal-approximation}
\E_\nu\Delta_N+\E_\nu\Delta_N^{\mathrm{BL}} = o(1).
\end{equation}
Because the populations generated by the prior change with $N$, we first establish both limits along every sequence $(N_j,\P_j)_{j \geq 1}$ with $N_j \to \infty$, $\P_j \in \cP$ and $d\P_j/d\P_0 \to d\P/d\P_0$ in $L_\infty(\P_0)$ for some $\P \in \cP$, for each $q \in \cQ$.

By Assumption \ref{cond:general-queries}(iii), $\psi_j \coloneqq \psi^\star_{\P_j,q} \to \psi^\star_{\P,q}$ in $L_2(\P_0)$, so that $\|\psi_j\|^2$ is uniformly integrable under $\P_0$, and, for $\varepsilon > 0$, the variable $X_j \coloneqq V_q^\star(\P_j)^{-\frac{1}{2}}\psi_j(Z)$ with $Z \sim \P_j$ satisfies
\[
\E_{\P_j}\{\|X_j\|^2 1_{\{\|X_j\|>\varepsilon\sqrt{N_j}\}}\} \leq C\int\|\psi_j\|^2 1_{\{\|\psi_j\|>\varepsilon\sqrt{v_-N_j}\}}\,d\P_0 = o(1).
\]
Applying Lindeberg's theorem to the $N_j$ i.i.d.\ copies of $X_j$, which have mean zero and covariance $I$, and using the continuity of the distribution function of $\|G\|$, we obtain
\begin{equation}\label{eq:sequential-normal-limit}
\Delta_{N_j}(\P_j,q)+\Delta_{N_j}^{\mathrm{BL}}(\P_j,q) = o(1).
\end{equation}
Let $\vartheta$, $\cB_\infty$ and $\P_\vartheta$ be as defined before Lemma \ref{lem:prior-boundary}, and let $g_N$ be bounded functions with $\|g_N\|_\infty \leq h_N$, where $h_N \to 0$ is as in Assumption \ref{cond:simulator-prior}. Since the tilts $\P_{\theta,g}$ of Section \ref{app:simulator-prior}, with $d\P_{\theta,g}/d\P_\theta \propto e^g$, satisfy $d\P_{\vartheta_{1:K},g_N}/d\P_\vartheta \propto \exp(g_N-\sum_{k > K}\vartheta_kb_k)$, we have, on $\cB_\infty$,
\[
\left\|\frac{d\P_{\vartheta_{1:K},g_N}}{d\P_0}-\frac{d\P_\vartheta}{d\P_0}\right\|_\infty \leq C\Bigl(h_N+\sum_{k > K}|\vartheta_k|\Bigr) = o(1).
\]
Hence, by \eqref{eq:sequential-normal-limit}, $\Delta_N+\Delta_N^{\mathrm{BL}} \to 0$ at $(\P_{\vartheta_{1:K},g_N},q)$ on $\cB_\infty$, for every $q \in \cQ$.

Since $\Delta_N \leq 1$ and $\Delta_N^{\mathrm{BL}} \leq 2$, by \eqref{eq:series-coupling}, we have
\begin{equation}\label{eq:sieve-normal-approximation}
\E_{\pi_K\otimes\pi_q}\{\Delta_N(\P_{\theta,g_N},q)+\Delta_N^{\mathrm{BL}}(\P_{\theta,g_N},q)\} = o(1).
\end{equation}
Under the priors of Assumptions \ref{cond:simulator-prior} and \ref{cond:posterior-tails}, we have
\[
\begin{aligned}
\E_\nu(\Delta_N+\Delta_N^{\mathrm{BL}}) &\overset{\text{(i)}}{\leq} C\{\E_{\pi_P\otimes\pi_q}(\Delta_N+\Delta_N^{\mathrm{BL}})\}^{\frac{1}{2}}\\
&\overset{\text{(ii)}}{\leq} C\sup_g\{\E_{\pi_K\otimes\pi_q}\Delta_N(\P_{\theta,g},q)+\E_{\pi_K\otimes\pi_q}\Delta_N^{\mathrm{BL}}(\P_{\theta,g},q)\}^{\frac{1}{2}} \overset{\text{(iii)}}{=} o(1),
\end{aligned}
\]
where (i) follows from \eqref{eq:chi2-transfer}, (ii) from \eqref{eq:prior-reduction}, and (iii) from \eqref{eq:sieve-normal-approximation} along the sequences $g_N$ of Lemma \ref{lem:prior-reduction}, for which $\|g_N\|_\infty \leq h_N$ by Assumption \ref{cond:simulator-prior}.

Let $\mu_N$ denote the conditional distribution of $\sqrt N\{\hat\beta-\beta_q(\P)\}$ given $(\P,q,\cT)$. Since $\cD_N$ is independent of $\cT$ given $(\P,q)$ and $\sqrt N\P_N\psi^\star_{\P,q} = V_q^\star(\P)^{\frac{1}{2}}Z_N$, we have
\[
\begin{aligned}
\E\,d_{\mathrm{BL}}\bigl[\mu_N,\cN\{0,V_q^\star(\P)\}\bigr] &\leq \E\left[2\wedge\sqrt N\|\hat\beta-\beta_q(\P)-\P_N\psi^\star_{\P,q}\|\right]\\
&\quad+(1\vee v_+^{\frac{1}{2}})\,\E_\nu\Delta_N^{\mathrm{BL}} = o(1),
\end{aligned}
\]
where the equality follows from \eqref{eq:efficiency-split}, bounded convergence and \eqref{eq:normal-approximation}. This establishes the first limit in \eqref{eq:efficient-normality}.

Let $\chi_\alpha \coloneqq (\chi^2_{d_k,1-\alpha})^{\frac{1}{2}}$ and $B_N \coloneqq V_q^\star(\P)^{-\frac{1}{2}}\hat VV_q^\star(\P)^{-\frac{1}{2}}$. For every $(\P,q,\cT)$ and $0 < u \leq v_-/2$, on the event $\left\{2e(\hat f,f_\star) \leq u^2\right\}$, we have
\[
\|B_N^{-\frac{1}{2}}-I\|_{\op} \leq \frac{u}{v_-}\quad\text{and}\quad\|\sqrt N\,V_q^\star(\P)^{-\frac{1}{2}}\{\hat\beta-\beta_q(\P)\}-Z_N\| \leq \frac{u}{\sqrt{v_-}}.
\]
Since $\beta_q(\P) \in \hat\cC_N$ if and only if $\|B_N^{-\frac{1}{2}}\sqrt N\,V_q^\star(\P)^{-\frac{1}{2}}\{\hat\beta-\beta_q(\P)\}\| \leq \chi_\alpha$, by the triangle inequality, with $C_\alpha' \coloneqq 2\chi_\alpha/v_-+v_-^{-\frac{1}{2}}$, we have
\begin{align*}
&\,\P(\|Z_N\| \leq \chi_\alpha-C_\alpha'u)-\P\{2e(\hat f,f_\star) > u^2\}\\
\leq &\,\P\{\beta_q(\P) \in \hat\cC_N\} \leq \P(\|Z_N\| \leq \chi_\alpha+C_\alpha'u)+\P\{2e(\hat f,f_\star) > u^2\},
\end{align*}
where the probabilities are conditional on $\cT$. Since $\Pr(\|G\| \leq \chi_\alpha) = 1-\alpha$ and $\|G\|$ has a density bounded uniformly over $d_k \leq d_0$, the maximal query dimension of Section \ref{app:architecture}, replacing $\|Z_N\|$ by $\|G\|$ at cost $\Delta_N(\P,q)$, we obtain, for $0 < u \leq 1$,
\begin{equation}\label{eq:wald-perturbation}
|\P\{\beta_q(\P) \in \hat\cC_N\mid\cT\}-(1-\alpha)| \leq \Delta_N(\P,q)+C_\alpha u+\P\{2e(\hat f,f_\star) > u^2\mid\cT\},
\end{equation}
where $C_\alpha$ depends only on $d_0$, $v_-$, $v_+$ and $\alpha$, and the case $u > v_-/2$ holds once $C_\alpha \geq 2/v_-$.

By Markov's inequality and bounded convergence, since conditional probabilities lie in $[0,1]$, $e(\hat f,f_\star) = o_\nu(1)$ is equivalent to
\begin{equation}\label{eq:goal-conditional}
\E_\nu\left[\P\{e(\hat f,f_\star) > \varepsilon\mid\cT\}\right] = o_p(1) \quad \text{for every } \varepsilon > 0,
\end{equation}
where $o_p(1)$ refers to the distribution of $\cT$. Averaging \eqref{eq:wald-perturbation} over $\nu$, by \eqref{eq:normal-approximation} and \eqref{eq:goal-conditional}, we obtain, for $0 < u \leq 1$,
\begin{equation}\label{eq:coverage-average}
\E_\nu\left|\P\{\beta_q(\P) \in \hat\cC_N(\cD_N,q)\mid\cT\}-(1-\alpha)\right| \leq C_\alpha u+o_p(1).
\end{equation}
Letting $N \to \infty$ and then $u = o(1)$, we find that the left side of \eqref{eq:coverage-average} is $o_p(1)$, which is equivalent to \eqref{eq:goal-coverage} by the argument for \eqref{eq:goal-conditional} and completes the proof.
\end{proof}

\subsection{Efficiency of the target at a fixed population}\label{app:oracle}

The proof of Theorem \ref{thm:fixed-population} needs the target $f_\pi$ to be efficient at a fixed population, and this subsection establishes such efficiency for populations with smooth log densities. For $s > 0$, $0 \leq \Xi < \infty$ and $\eta \in (0,B)$, we define
\begin{equation}\label{eq:sobolev-class}
\cP_s(\Xi,\eta) \coloneqq \left\{\P:\frac{d\P}{d\P_0} \propto \exp\Bigl(\sum_{k \geq 1}\theta_kb_k\Bigr),\ \theta \in \R^{\N},\ \sum_{k \geq 1}k^{\frac{2s}{d}}\theta_k^2 \leq \Xi^2,\ \|\theta\|_1 \leq B-\eta\right\},
\end{equation}
which bounds the log density in a periodic Sobolev norm in the coordinates of Section \ref{app:prior}, as in the exponential series analysis of \cite{rivoirard2012}, and keeps the coefficients strictly inside the support $\Theta_K = \left\{\theta \in \R^K:\|\theta\|_1 \leq B\right\}$ of $\pi_K$. Proposition \ref{prop:spectral-truth} and Section \ref{app:single-population} take the sieve dimension in the window
\begin{equation}\label{eq:dimension-window}
K^{3\vee2a} = o(N)\quad\text{and}\quad\sqrt NK^{-\frac{s}{d}} = o(1),
\end{equation}
where the two conditions together force $s > d(3/2\vee a)$. For a fixed population $\P_\star$, we let $\nu \coloneqq \delta_{\P_\star}\otimes\pi_q$ in the rest of this section.

\begin{proposition}[Posterior moments at smooth populations]\label{prop:spectral-truth}
Suppose that Assumption \ref{cond:general-queries}(i)--(iii) holds, $s > 0$, $\Xi \geq 0$ and $\eta \in (0,B)$ are constants, and $K$ satisfies \eqref{eq:dimension-window}. Let $f_\pi = (m_\pi,S_\pi)$ denote the posterior moments under $\pi_K$ and $(g,V)$ the mode pair of Definition \ref{def:reference-estimator}. Then, for every $\P_\star \in \cP_s(\Xi,\eta)$ and $q \in \cQ$, with $\cD_N \sim \P_\star^{\otimes N}$,
\[
m_\pi = \beta_q(\P_\star)+\P_N\psi^\star_{\P_\star,q}+o_{\P_\star}(N^{-\frac{1}{2}})\quad\text{and}\quad S_\pi = V_q^\star(\P_\star)+o_{\P_\star}(1),
\]
and the same holds with $(g,V)$ in place of $(m_\pi,S_\pi)$. Consequently, $f_\pi$ and $(g,V)$ are efficient under $\nu = \delta_{\P_\star}\otimes\pi_q$.
\end{proposition}

\begin{proof}
Throughout this proof, expectations and stochastic orders refer to $\P_\star^{\otimes N}$. We fit a sieve population $\P_K$ to $\P_\star$ by matching moments, expand the posterior moments around $\P_K$, and then compare the projected gradient at $\P_K$ with the EIF at $\P_\star$.

For large $N$, $K^3 \leq N$ and $\|\Lambda_K^{-1}\|_{\op} = \sigma^{-2}K^{2a} \leq N$ by \eqref{eq:dimension-window}, so that the lemmas of Section \ref{app:proofs} hold. For $\|\phi\|_\infty\vee\|\phi'\|_\infty \leq C_\infty(B+1)$, with the basis bound $C_\infty$ of Section \ref{app:prior}, splitting the difference at $e^{\phi'}/\E_{\P_0}e^{\phi}$, by the mean value theorem, we have
\begin{equation}\label{eq:density-normalisation}
\begin{aligned}
\left\|\frac{e^\phi}{\E_{\P_0}e^\phi}-\frac{e^{\phi'}}{\E_{\P_0}e^{\phi'}}\right\|_{L_r(\P_0)} &\lesssim \|e^\phi-e^{\phi'}\|_{L_r(\P_0)}+|\E_{\P_0}(e^\phi-e^{\phi'})|\\
&\lesssim \|\phi-\phi'\|_{L_r(\P_0)}\quad\text{for } r \in \{2,\infty\}.
\end{aligned}
\end{equation}
Let $p_\star \coloneqq d\P_\star/d\P_0 \propto \exp(\sum_{k \geq 1}\theta_{\star k}b_k)$, with $(\theta_{\star k})_{k \geq 1}$ the coefficients of $\P_\star$ in \eqref{eq:sobolev-class}. In the notation of Section \ref{sec:target-efficiency}, $\theta_{\P_\star} = (\theta_{\star1},\cdots,\theta_{\star K})^\T \in \Theta_K$ and $g_{\P_\star,K} = \sum_{k > K}\theta_{\star k}b_k$, and, by orthonormality and the Cauchy--Schwarz inequality, we have
\begin{equation}\label{eq:sieve-remainder}
\|g_{\P_\star,K}\|_{L_2(\P_0)} \leq \Xi K^{-\frac{s}{d}}\quad\text{and}\quad\|g_{\P_\star,K}\|_\infty \lesssim K^{\frac{1}{2}-\frac{s}{d}}.
\end{equation}
In particular, by \eqref{eq:density-normalisation}, $\|d\P_{\theta_{\P_\star}}/d\P_0-p_\star\|_\infty \to 0$ as $K \to \infty$, so that $\P_\star \in \cP$.

Let
\[
\theta_K \coloneqq \argmin_{\theta \in \Theta_K}\{A_K(\theta)-\theta^\T\E_{\P_\star}b_{(K)}\}\quad\text{and}\quad\P_K \coloneqq \P_{\theta_K},
\]
where $A_K(\theta) = \log\E_{\P_0}e^{\theta^\T b_{(K)}}$ as in Section \ref{app:prior} and $\theta_K$ is unique by Lemma \ref{lem:geometry}. Since $\nabla_\theta\{A_K(\theta)-\theta^\T\E_{\P_\star}b_{(K)}\} = \E_{\P_0}\{(d\P_\theta/d\P_0-p_\star)b_{(K)}\}$, by strong convexity, the variational inequality at $\theta_K$, Bessel's inequality and \eqref{eq:density-normalisation}, we have
\begin{equation}\label{eq:fit-coefficients}
c\|\theta_K-\theta_{\P_\star}\| \leq \left\|\frac{d\P_{\theta_{\P_\star}}}{d\P_0}-p_\star\right\|_{L_2(\P_0)} \lesssim \|g_{\P_\star,K}\|_{L_2(\P_0)} \leq \Xi K^{-\frac{s}{d}}.
\end{equation}
Consequently, $\|\theta_K-\theta_{\P_\star}\|_1 \leq \sqrt K\|\theta_K-\theta_{\P_\star}\| \lesssim K^{\frac{1}{2}-\frac{s}{d}}$, so that $\|\theta_K\|_1 \leq B-\eta/2$ for large $K$, and, by the first-order condition at the interior point $\theta_K$ of $\Theta_K$, we have
\begin{equation}\label{eq:moment-matching}
\E_{\P_K}b_{(K)} = \nabla A_K(\theta_K) = \E_{\P_\star}b_{(K)}.
\end{equation}
Since $\log(d\P_K/d\P_0)-\log p_\star-(\theta_K-\theta_{\P_\star})^\T b_{(K)}+g_{\P_\star,K}$ is constant on $\cZ$, by \eqref{eq:density-normalisation}, orthonormality, \eqref{eq:sieve-remainder} and \eqref{eq:fit-coefficients}, we obtain, for large $K$,
\begin{equation}\label{eq:unpenalised-fit}
\left\|\frac{d\P_K}{d\P_0}-p_\star\right\|_{L_2(\P_0)} \lesssim K^{-\frac{s}{d}}\quad\text{and}\quad\left\|\frac{d\P_K}{d\P_0}-p_\star\right\|_\infty \lesssim K^{\frac{1}{2}-\frac{s}{d}}.
\end{equation}
Since $\lambda_k = \sigma^2k^{-2a}$, by \eqref{eq:sobolev-class} we have $\|\Lambda_K^{-1}\theta_{\P_\star}\|^2 = \sigma^{-4}\sum_{k \leq K}k^{4a}\theta_{\star k}^2 \leq \sigma^{-4}\Xi^2(1+K^{4a-\frac{2s}{d}})$, and hence, by \eqref{eq:fit-coefficients},
\begin{equation}\label{eq:fit-prior-score}
\begin{aligned}
\frac{\|\Lambda_K^{-1}\theta_K\|}{\sqrt N} & \leq \frac{\|\Lambda_K^{-1}\theta_{\P_\star}\|+\sigma^{-2}K^{2a}\|\theta_K-\theta_{\P_\star}\|}{\sqrt N} \\ & \lesssim \frac{1}{\sqrt N}+\frac{K^{2a}}{N}\sqrt NK^{-\frac{s}{d}} = o(1).
\end{aligned}
\end{equation}

Let $h \coloneqq \tilde\theta-\theta_K$ for the posterior mode $\tilde\theta$ of Section \ref{app:theory-technical}, and let $A$, $H$ and $U_N$ be as in the proof of Proposition \ref{prop:triangular} with $\theta_K$ in place of $\theta$, so that $U_N = \G_Nb_{(K)}$ by \eqref{eq:moment-matching}. Since $\Cov_{\P_\star}(b_{(K)}) \preceq CI$ by the density envelope of $\cP$, by Lemma \ref{lem:mode-localisation} at $\theta_K$ and \eqref{eq:fit-prior-score}, we have
\[
\E\|h\|^2 \lesssim \frac{\E\|U_N\|^2}{N}+\frac{\|\Lambda_K^{-1}\theta_K\|^2}{N^2} \lesssim \frac{K}{N}.
\]
Hence, as $K^3 = o(N)$, $\sqrt{NK}\|h\|^2+\sqrt K\|h\| = o_{\P_\star}(1)$, and, since $\|\tilde\theta\|_1 \leq B-\eta/2+\sqrt K\|h\|$ and $\sqrt K\rho_N = o(1)$ for the radius $\rho_N$ of Section \ref{app:theory-technical}, the event $\cE_N$ of regular contexts defined there satisfies
\begin{equation}\label{eq:smooth-regular}
\P_\star^{\otimes N}(\cE_N^c) \leq \P_\star^{\otimes N}\left\{\sqrt K(\|h\|+\rho_N) > \frac{\eta}{2}\right\} = o(1).
\end{equation}
On $\cE_N$, substituting \eqref{eq:mode-expansion} at $\theta_K$ into the first line of \eqref{eq:simulator-mean-expansion}, we obtain
\begin{equation}\label{eq:smooth-mean-error}
\begin{aligned}
&\,\|\sqrt N\{m_\pi-\beta_q(\P_K)\}-AH^{-1}U_N\|\\
\leq &\,\sqrt N\|m_\pi-g\|+\sqrt N\|g-\beta_q(\P_K)-Ah\|+\|AH^{-1}\|_{\op}\left\{\frac{\|\Lambda_K^{-1}\theta_K\|}{\sqrt N}+\sqrt N\|\mathrm{rem}(h)\|\right\}\\
\overset{\text{(i)}}{\lesssim} &\,\frac{K^{\frac{3}{2}}}{\sqrt N}+\sqrt{NK}\|h\|^2+\frac{\|\Lambda_K^{-1}\theta_K\|}{\sqrt N} \overset{\text{(ii)}}{=} o_{\P_\star}(1),
\end{aligned}
\end{equation}
where (i) follows from Lemma \ref{lem:laplace}, \eqref{eq:query-taylor}, \eqref{eq:score-identity} and Lemma \ref{lem:geometry}, and (ii) from $K^3 = o(N)$, $\sqrt{NK}\|h\|^2 = o_{\P_\star}(1)$ and \eqref{eq:fit-prior-score}. As in the proof of Proposition \ref{prop:triangular}, by \eqref{eq:score-identity}, the projected influence function $\psi^K_{\P_K,q}$ of Section \ref{app:posterior-expansion} satisfies $AI_K(\theta_K)^{-1}U_N = \G_N\psi^K_{\P_K,q}$, and
\[
\E\|A\{H^{-1}-I_K(\theta_K)^{-1}\}U_N\|^2 \lesssim \frac{K^{4a}}{N^2} = o(1),
\]
so that, by \eqref{eq:smooth-mean-error},
\begin{equation}\label{eq:smooth-mean-expansion}
\sqrt N\{m_\pi-\beta_q(\P_K)\} = \G_N\psi^K_{\P_K,q}+o_{\P_\star}(1).
\end{equation}
By Lemma \ref{lem:laplace} and the argument for \eqref{eq:simulator-covariance}, we also have
\begin{equation}\label{eq:smooth-covariance}
\|S_\pi-V\|_F+\|V-AI_K(\theta_K)^{-1}A^\T\|_F \lesssim \frac{K}{\sqrt N}+\sqrt K\|h\|+\frac{K^{2a}}{N} = o_{\P_\star}(1).
\end{equation}

Let $\Pi_K$ denote the componentwise $L_2(\P_0)$ projection onto $\spn(1,b_1,\cdots,b_K)$. Since $\Pi_K\psi^\star_{\P_K,q}-\E_{\P_K}\Pi_K\psi^\star_{\P_K,q}$ lies in the span of the coordinates of $s_{\P_K,K} = b_{(K)}-\E_{\P_K}b_{(K)}$, onto which $\psi^K_{\P_K,q}$ projects $\psi^\star_{\P_K,q}$ in $L_2(\P_K)$, and $\E_{\P_K}\psi^\star_{\P_K,q} = 0$, by the density envelope, we have
\begin{align*}
\|\psi^K_{\P_K,q}-\psi^\star_{\P_K,q}\|_{L_2(\P_K)} &\lesssim \|(I-\Pi_K)\psi^\star_{\P_K,q}\|_{L_2(\P_0)}\\
&\leq \|\psi^\star_{\P_K,q}-\psi^\star_{\P_\star,q}\|_{L_2(\P_0)}+\|(I-\Pi_K)\psi^\star_{\P_\star,q}\|_{L_2(\P_0)}.
\end{align*}
The first term is $o(1)$ by \eqref{eq:unpenalised-fit} and Assumption \ref{cond:general-queries}(iii), and the second by the completeness of $\{1,b_1,b_2,\cdots\}$ in $L_2(\P_0)$. Hence, by the density envelope and the triangle inequality, we obtain
\begin{equation}\label{eq:smooth-gradient-limit}
\|\psi^K_{\P_K,q}-\psi^\star_{\P_\star,q}\|_{L_2(\P_0)} = o(1).
\end{equation}

We now combine \eqref{eq:smooth-mean-expansion} and \eqref{eq:smooth-covariance} with \eqref{eq:smooth-gradient-limit}. By \eqref{eq:smooth-mean-expansion}, we have
\begin{equation}\label{eq:smooth-mean-limit}
\begin{aligned}
&\,\|\sqrt N\{m_\pi-\beta_q(\P_\star)\}-\G_N\psi^\star_{\P_\star,q}\|\\
\leq &\,\sqrt N\|\beta_q(\P_K)-\beta_q(\P_\star)\|+\|\G_N(\psi^K_{\P_K,q}-\psi^\star_{\P_\star,q})\|+o_{\P_\star}(1)\\
\overset{\text{(i)}}{\lesssim} &\,\sqrt NK^{-\frac{s}{d}}+O_{\P_\star}\bigl(\|\psi^K_{\P_K,q}-\psi^\star_{\P_\star,q}\|_{L_2(\P_0)}\bigr)+o_{\P_\star}(1) \overset{\text{(ii)}}{=} o_{\P_\star}(1),
\end{aligned}
\end{equation}
where (i) follows from \eqref{eq:query-lipschitz}, \eqref{eq:unpenalised-fit}, Chebyshev's inequality and the density envelope, and (ii) from \eqref{eq:dimension-window} and \eqref{eq:smooth-gradient-limit}.

Since $AI_K(\theta_K)^{-1}A^\T = \E_{\P_K}\{(\psi^K_{\P_K,q})^{\otimes2}\}$ by \eqref{eq:score-identity}, splitting the difference at $\E_{\P_\star}\{(\psi^K_{\P_K,q})^{\otimes2}\}$, we obtain
\[
\|AI_K(\theta_K)^{-1}A^\T-V_q^\star(\P_\star)\|_F \overset{\text{(i)}}{\lesssim} \left\|\frac{d\P_K}{d\P_0}-p_\star\right\|_\infty\|\psi^K_{\P_K,q}\|_{L_2(\P_0)}^2+\|\psi^K_{\P_K,q}-\psi^\star_{\P_\star,q}\|_{L_2(\P_0)} \overset{\text{(ii)}}{=} o(1),
\]
where (i) follows from the Cauchy--Schwarz inequality and Assumption \ref{cond:general-queries}(i), and (ii) from \eqref{eq:unpenalised-fit} and \eqref{eq:smooth-gradient-limit}. Hence, by \eqref{eq:smooth-covariance}, we have
\begin{equation}\label{eq:smooth-covariance-limit}
\|S_\pi-V_q^\star(\P_\star)\|_F+\|V-V_q^\star(\P_\star)\|_F = o_{\P_\star}(1).
\end{equation}
By \eqref{eq:smooth-mean-limit} and \eqref{eq:smooth-covariance-limit}, the remainder $\cR_N = \cR_N(\P_\star,q,\cD_N)$ of \eqref{eq:posterior-moment-remainder} satisfies $\cR_N = o_{\P_\star}(1)$, and, as $0 \leq \cR_N \leq 1$, by dominated convergence over $q \sim \pi_q$, we obtain $\E_\nu\E_\P\cR_N = o(1)$, which is the efficiency of $f_\pi$ by Markov's inequality.

For the mode pair, by Lemma \ref{lem:laplace}, $\sqrt N\|g-m_\pi\| \lesssim K^{\frac{3}{2}}/\sqrt N = o(1)$ on $\cE_N$, and hence, by \eqref{eq:smooth-regular}, \eqref{eq:smooth-mean-limit} and \eqref{eq:smooth-covariance-limit}, the pair $(g,V)$ satisfies \eqref{eq:goal} at $(\P_\star,q)$ and, by dominated convergence over $q$, is efficient under $\nu$, which completes the proof.
\end{proof}

\subsection{Proof of Theorem \ref{thm:fixed-population}}\label{app:single-population}

\begin{assumption}[Fixed population]\label{cond:fixed-population}
There exist constants $\kappa \in (0,(3\vee2a)^{-1})$, $s > d/(2\kappa)$, $\Xi \geq 0$ and $\eta \in (0,B)$ such that $K \asymp N^\kappa$ and $\P_\star \in \cP_s(\Xi,\eta)$ defined in \eqref{eq:sobolev-class}.
\end{assumption}

Assumption \ref{cond:fixed-population} requires the log density of $\P_\star$ to have periodic Sobolev smoothness above $d/(2\kappa)$ and $\|\theta_{\P_\star}\|_1 \leq B-\eta$ for every $K$, so that $\theta_{\P_\star}$ lies at $\ell_1$ distance at least $\eta$ from the boundary of $\Theta_K = \left\{\theta \in \R^K:\|\theta\|_1 \leq B\right\}$. Since $K \asymp N^\kappa$, $\kappa(3\vee2a) < 1$ and $\kappa s/d > 1/2$, we have $K^{3\vee2a} = o(N)$ and $\sqrt NK^{-\frac{s}{d}} = o(1)$, which is the window \eqref{eq:dimension-window} of Proposition \ref{prop:spectral-truth} and Lemma \ref{lem:small-ball}.

The proof bounds the overlap factor $\Gamma_N(\nu)$ in \eqref{eq:overlap-factor} through the prior mass of the set $B_N \coloneqq \left\{\P \in \cP:\chi^2(\P_\star\,\|\,\P) \leq N^{-1}\right\}$.
%, and Section \ref{app:fixed-resources} proves them.

\begin{lemma}\label{lem:small-ball}
Let $s > 0$, $\Xi \geq 0$ and $\eta \in (0,B)$ be constants, and suppose that $\pi_P(\cP_K) = 1$, Assumption \ref{cond:posterior-tails} holds and $K$ satisfies \eqref{eq:dimension-window}. Then, uniformly in $\P_\star \in \cP_s(\Xi,\eta)$,
\begin{equation}\label{eq:small-ball}
\log\pi_P(B_N) \geq -\frac{K}{2}\log N+a\log(K!)-\log\Gamma\Bigl(\frac{K}{2}+1\Bigr)-O(K).
\end{equation}
\end{lemma}

The second lemma constructs a network $f^\circ = f_\omega$ that approximates the posterior moments uniformly over contexts. Let $C^\circ \coloneqq (C_\omega+C_\omega^\T)/2$ denote the symmetrised raw covariance of $f^\circ$ in Definition \ref{def:transformer-output}, so that $f^{\circ,V} = \clip C^\circ$, and let $h_{f^\circ}$ denote the conditional excess score of $f^\circ$ in \eqref{eq:posterior-loss-regret}.

\begin{lemma}\label{lem:posterior-network}
Suppose that $\pi_P(\cP_K) = 1$ and Assumptions \ref{cond:general-queries}(i), \ref{cond:general-queries}(iv) and \ref{cond:posterior-tails} hold. Then there exists a constant $C > 0$, independent of $N$, $K$, $\cN$ and $\varepsilon$, such that, for every $\varepsilon \in (0,1)$, some deterministic $f^\circ \in \cF_{\mathrm{TF}}(L,1,d_e,d_f,W)$ with $p$ free parameters, $L \leq CK\log(Cn_{\max}K/\varepsilon)$ and $p+d_e+d_f+W \leq (Cn_{\max}K/\varepsilon)^{CK}$ satisfies, at every $(\cD_n,q) \in \cZ^n\times\cQ$ with $n \in \cN$,
\[
\sqrt n\|f^{\circ,\beta}-m_\pi\|+\|C^\circ-S_\pi\|_F \leq \varepsilon\quad\text{and}\quad0 \leq h_{f^\circ} \leq C\varepsilon^2.
\]
\end{lemma}

\begin{proof}[Proof of Theorem \ref{thm:fixed-population}]
We bound the overlap factor $\Gamma_N(\nu)$ in \eqref{eq:overlap-factor} by $e/\pi_P(B_N)$, which reduces the theorem to the budget \eqref{eq:single-budget} on the pretraining error $\tilde r_N$ in \eqref{eq:learning-budget}, and then verify this budget for the classes of Lemma \ref{lem:posterior-network}.

For $\P \in \cP$, let $p^{\otimes N} \coloneqq d\P^{\otimes N}/d\P_0^{\otimes N}$ and $p_\star^{\otimes N} \coloneqq d\P_\star^{\otimes N}/d\P_0^{\otimes N}$. We define
\[
\pi_B \coloneqq \pi_P(\cdot\mid B_N)\quad\text{and}\quad p_{\pi,N} \coloneqq \int p^{\otimes N}\,\pi_P(d\P),
\]
where $\pi_P(B_N) > 0$ under \eqref{eq:single-budget}. Since $\cL_{\nu,N}$ and $\cL_{\pi,N}$ in \eqref{eq:overlap-factor} have densities $p_\star^{\otimes N}$ and $p_{\pi,N}$ relative to $\P_0^{\otimes N}\otimes\pi_q$ and $p_{\pi,N} \geq \pi_P(B_N)\int p^{\otimes N}\,\pi_B(d\P)$, we have
\begin{equation}\label{eq:single-overlap}
\begin{aligned}
\Gamma_N(\nu) = \int\frac{(p_\star^{\otimes N})^2}{p_{\pi,N}}\,d\P_0^{\otimes N}
&\overset{\text{(i)}}{\leq} \frac{1}{\pi_P(B_N)}\int\!\!\int\frac{(p_\star^{\otimes N})^2}{p^{\otimes N}}\,d\P_0^{\otimes N}\,\pi_B(d\P)\\
&\overset{\text{(ii)}}{=} \frac{1}{\pi_P(B_N)}\int\{1+\chi^2(\P_\star\,\|\,\P)\}^N\,\pi_B(d\P)\\
&\overset{\text{(iii)}}{\leq} \frac{e}{\pi_P(B_N)},
\end{aligned}
\end{equation}
where (i) follows from Jensen's inequality for $x \mapsto 1/x$ and Tonelli's theorem, (ii) from independence and (iii) from the definition of $B_N$.

Let $\pi_P$ satisfy Assumptions \ref{cond:general-queries}--\ref{cond:posterior-tails} with $f_\pi$ efficient under $\nu$, and let the Transformer class have pretraining error
\begin{equation}\label{eq:single-budget}
\tilde r_N = o\{\pi_P(B_N)\}.
\end{equation}
By \eqref{eq:single-overlap}, \eqref{eq:single-budget} and \eqref{eq:learning}, we have, for the moment error $e(\cdot,\cdot)$ in \eqref{eq:moment-error},
\[
\Gamma_N(\nu)\tilde r_N \leq \frac{e\,\tilde r_N}{\pi_P(B_N)} = o(1)\quad\text{and}\quad e(\hat f,f_\pi^\circ) = o_\nu(1).
\]
As $\clip = \clip_{[v_-,v_+]}$ is nonexpansive and $\clip V_q^\star(\P_\star) = V_q^\star(\P_\star)$ under Assumption \ref{cond:covariance-bounds}(i), by the efficiency of $f_\pi$ under $\nu$, the benchmark $f_\star = (\beta_q(\P_\star)+\P_N\psi^\star_{\P_\star,q},V_q^\star(\P_\star))$ satisfies
\[
e(\hat f,f_\star)^{\frac{1}{2}} \leq e(\hat f,f_\pi^\circ)^{\frac{1}{2}}+e(f_\pi^\circ,f_\star)^{\frac{1}{2}} \leq e(\hat f,f_\pi^\circ)^{\frac{1}{2}}+e(f_\pi,f_\star)^{\frac{1}{2}} = o_\nu(1),
\]
which is \eqref{eq:goal} at $\P_\star$ and, as $\|\hat V-V_q^\star(\P_\star)\|_F \leq e(\hat f,f_\star)^{\frac{1}{2}}$, proves the second limit in \eqref{eq:single-efficiency}.

Since $\nu$ does not depend on $N$ and the summands $V_q^\star(\P_\star)^{-\frac{1}{2}}\psi^\star_{\P_\star,q}(Z_i)$ of $Z_N$ are i.i.d.\ with mean zero and identity covariance, by the central limit theorem, P\'olya's theorem and dominated convergence over $\pi_q$, we have
\begin{equation}\label{eq:single-normal}
\E_{\pi_q}\Delta_N(\P_\star,q)+\E_{\pi_q}\Delta_N^{\mathrm{BL}}(\P_\star,q) = o(1),
\end{equation}
where $Z_N$, $\Delta_N$ and $\Delta_N^{\mathrm{BL}}$ are as in the proof of Theorem \ref{thm:main}.
Let $\mu_N$ denote the conditional distribution of $\sqrt N\{\hat\beta-\beta_q(\P_\star)\}$ given $(q,\cT)$ and $d_{\mathrm{BL}}$ the bounded Lipschitz distance. As $\cD_N$ is independent of $\cT$, we have
\[
\begin{aligned}
\E\,d_{\mathrm{BL}}\bigl[\mu_N,\cN\{0,V_q^\star(\P_\star)\}\bigr] &\leq \E\left[2\wedge\sqrt N\|\hat\beta-\beta_q(\P_\star)-\P_N\psi^\star_{\P_\star,q}\|\right]\\
&\quad+(1\vee v_+^{\frac{1}{2}})\,\E_{\pi_q}\Delta_N^{\mathrm{BL}}(\P_\star,q) = o(1),
\end{aligned}
\]
where the first term vanishes by \eqref{eq:goal} and bounded convergence and the second by \eqref{eq:single-normal}, which proves the first limit in \eqref{eq:single-efficiency}. Averaging \eqref{eq:wald-perturbation} at $\P = \P_\star$ over $\pi_q$, by \eqref{eq:goal-conditional}, we obtain, for $0 < u \leq 1$,
\[
\E_{\pi_q}\left|\P_\star\{\beta_q(\P_\star) \in \hat\cC_N(\cD_N,q)\mid\cT\}-(1-\alpha)\right| \leq \E_{\pi_q}\Delta_N(\P_\star,q)+C_\alpha u+o_p(1),
\]
and, letting $N \to \infty$ and then $u = o(1)$, by \eqref{eq:single-normal} and the argument for \eqref{eq:goal-conditional}, we obtain \eqref{eq:single-coverage}.

Under the hypotheses of Theorem \ref{thm:fixed-population}, Assumptions \ref{cond:general-queries}--\ref{cond:posterior-tails} and \ref{cond:fixed-population} hold and $\pi_P$ is supported on $\left\{\P_\theta:\theta \in \Theta_K\right\} \subseteq \cP_K$, so that Proposition \ref{prop:spectral-truth} and Lemmas \ref{lem:small-ball} and \ref{lem:posterior-network} apply.

By Proposition \ref{prop:spectral-truth} and \eqref{eq:prior-stability} with $\epsilon_{\pi,N} = L_N^2/N = o(1)$, the posterior moments under $\pi_K$ and $f_\pi$ are efficient under $\nu$. By Lemma \ref{lem:small-ball} and Stirling's formula, we have
\[
\log\pi_P(B_N) \geq -\frac{K}{2}\log N+\Bigl(a-\frac{1}{2}\Bigr)K\log K-O(K),
\]
so that, as $a > 1$,
\begin{equation}\label{eq:small-ball-bounds}
N^{-\frac{K}{2}}K^{\frac{K}{4}} \leq \pi_P(B_N),
\end{equation}
and $N^{-\frac{K}{2}} = o\{\pi_P(B_N)\}$.

Let $\varepsilon \coloneqq N^{-1}(NK)^{-K}$. Since $n_{\max} \lesssim N$, we have $\log(Cn_{\max}K/\varepsilon) \leq 3K\log(CNK)$, so that the classes of Lemma \ref{lem:posterior-network} at accuracy $\varepsilon$, with $\Omega = [-W,W]^p$, satisfy
\[
L \leq C_1K^2\log(CNK)\quad\text{and}\quad p+W \leq (CNK)^{C_1K^2},
\]
where $C_1 \coloneqq \max(3C,1)$ for the constant $C$ of Lemma \ref{lem:posterior-network}. As these classes contain the network $f^\circ$ of Lemma \ref{lem:posterior-network}, by \eqref{eq:risk-benchmark}, we have
\[
\epsilon_{\mathrm{app}} \leq R(f^\circ)-R(f_\pi^\circ) = \E h_{f^\circ} \leq C\varepsilon^2 = o\{(NK)^{-K}\}.
\]
Since $t \mapsto t^{-1}\log(2pWn_{\max}t)$ decreases for $t \geq 2$, for $T \geq (CNK)^{C_*K^2}$ with $C_* \geq C_1+2$, we have
\[
\frac{pL^2\log(2pTWn_{\max})}{T} \leq \frac{C_1^2(3C_1+C_*)\{K^2\log(CNK)\}^3}{(CNK)^{(C_*-C_1)K^2}} = o\{(NK)^{-K}\},
\]
where the last step uses $(NK)^K \leq (CNK)^{K^2}$ and $y^3e^{-y} \to 0$ as $y = K^2\log(CNK) \to \infty$. Since $(NK)^{-K} \leq N^{-\frac{K}{2}}$ and $\epsilon_{\mathrm{opt}} = O(N^{-\frac{K}{2}})$, by \eqref{eq:learning-budget} and \eqref{eq:small-ball-bounds}, we obtain
\[
\tilde r_N = O(N^{-\frac{K}{2}}) = o\{\pi_P(B_N)\},
\]
which is \eqref{eq:single-budget}. As $\log(CNK) \asymp \log N$, these classes have $\exp\{O(K^2\log N)\}$ parameters and require $T \geq \exp(CK^2\log N)$, which proves the theorem by the reduction to \eqref{eq:single-budget}.
\end{proof}

\subsubsection{Proofs of Lemmas \ref{lem:small-ball} and \ref{lem:posterior-network}}\label{app:fixed-resources}

\begin{proof}[Proof of Lemma \ref{lem:small-ball}]
We compare $\chi^2(\P_\star\,\|\,\P_\theta)$ with $\|\theta-\theta_{\P_\star}\|$, find a ball inside the set $\left\{\theta \in \Theta_K:\P_\theta \in B_N\right\}$ and integrate the prior density over this ball. By Assumption \ref{cond:posterior-tails}, $c_w \leq d\pi_P/d\pi_K$, so that it suffices to prove \eqref{eq:small-ball} for $\pi_K$.

Let $\theta_{\P_\star}$ and $g_{\P_\star,K}$ be as in Section \ref{sec:target-efficiency}. For $\theta \in \Theta_K$, since $0 \leq A_K(\theta) \leq C_\infty\|\theta\|_1$ for $A_K$ and the basis bound $C_\infty$ of Section \ref{app:prior}, and likewise for $\P_\star$, the log densities of $\P_\star$ and $\P_\theta$ are bounded by $2C_\infty B$ and differ, up to an additive constant, by
\[
\phi_\theta \coloneqq g_{\P_\star,K}+(\theta_{\P_\star}-\theta)^\T b_{(K)},\quad\text{where}\quad\|\phi_\theta\|_{L_2(\P_0)}^2 = \|\theta-\theta_{\P_\star}\|^2+\|g_{\P_\star,K}\|_{L_2(\P_0)}^2
\]
by the orthonormality of $b_{(K)} = (b_1,\cdots,b_K)^\T$ and the orthogonality of $g_{\P_\star,K}$ to $b_1,\cdots,b_K$. By \eqref{eq:density-normalisation}, we have
\[
\Bigl\|\frac{d\P_\star}{d\P_0}-\frac{d\P_\theta}{d\P_0}\Bigr\|_{L_2(\P_0)} \leq C\|\phi_\theta\|_{L_2(\P_0)}.
\]
Since $d\P_\theta/d\P_0 \geq e^{-2C_\infty B}$, we obtain, for all $\theta \in \Theta_K$ and a constant $C_\chi > 0$ depending only on $B$ and $C_\infty$,
\begin{equation}\label{eq:chi2-coefficients}
\chi^2(\P_\star\,\|\,\P_\theta) \leq C_\chi\|\phi_\theta\|_{L_2(\P_0)}^2.
\end{equation}

Let $r_- \coloneqq (2C_\chi N)^{-\frac{1}{2}}$. By orthonormality, \eqref{eq:sobolev-class} and \eqref{eq:dimension-window}, for all sufficiently large $N$, uniformly in $\P_\star \in \cP_s(\Xi,\eta)$, we have
\[
\|g_{\P_\star,K}\|_{L_2(\P_0)} \leq \Xi K^{-\frac{s}{d}} \leq r_-\quad\text{and}\quad\sqrt Kr_- \leq \eta,
\]
and hence, by \eqref{eq:chi2-coefficients} and $\|\theta_{\P_\star}\|_1 \leq B-\eta$,
\[
\left\{\theta \in \R^K:\|\theta-\theta_{\P_\star}\| \leq r_-\right\} \subseteq \left\{\theta \in \Theta_K:\P_\theta \in B_N\right\}.
\]

It remains to integrate the prior density over this ball. As $\det\Lambda_K = \sigma^{2K}(K!)^{-2a}$ for $\Lambda_K = \diag(\sigma^2k^{-2a})_{k \in [K]}$ of Section \ref{app:prior}, the prior $\pi_K$ has the Lebesgue density
\[
\frac{(K!)^a}{Z_K(2\pi\sigma^2)^{\frac{K}{2}}}\exp\Bigl(-\frac{\theta^\T\Lambda_K^{-1}\theta}{2}\Bigr)\quad\text{on }\Theta_K,
\]
where $Z_K = \cN(0,\Lambda_K)(\Theta_K) \leq 1$ satisfies $\inf_KZ_K > 0$ by Lemma \ref{lem:prior-boundary}. By \eqref{eq:sobolev-class} and $s/d > a$, we have $\theta_{\P_\star}^\T\Lambda_K^{-1}\theta_{\P_\star} \leq \sigma^{-2}\Xi^2$, and hence, as $K^{2a}r_-^2 = o(1)$,
\[
\sup_{\|\theta-\theta_{\P_\star}\| \leq r_-}\theta^\T\Lambda_K^{-1}\theta \leq 2\sigma^{-2}(\Xi^2+K^{2a}r_-^2) \leq C.
\]
Multiplying the lower density bound by the ball volume $\pi^{\frac{K}{2}}r_-^K/\Gamma(K/2+1)$, we obtain
\[
\frac{K}{2}\log\frac{r_-^2}{2\sigma^2}-C \leq \log\pi_K(B_N)-a\log(K!)+\log\Gamma\Bigl(\frac{K}{2}+1\Bigr),
\]
which proves \eqref{eq:small-ball}, since $r_-^2 \asymp N^{-1}$.
\end{proof}

\begin{proof}[Proof of Lemma \ref{lem:posterior-network}]
We bound the Lipschitz constants of the posterior moments in the sufficient statistic and the query, interpolate the moments on a grid and evaluate the interpolants after one averaging head.

Let $n \in \cN$ and $k \in \cK$, with $\cU_k \subset \R^{\ell_k}$. For $t \in \R^K$, let $\E_t$ and $\Cov_t$ denote expectation and covariance under the distribution on $\Theta_K$ with Lebesgue density proportional to
\[
\theta \longmapsto w(\theta)\exp\Bigl\{n\theta^\T t-nA_K(\theta)-\frac{\theta^\T\Lambda_K^{-1}\theta}{2}\Bigr\},
\]
where $w \coloneqq w_\pi(\cdot\mid0)$ is the density in Assumption \ref{cond:posterior-tails}, and, for $q = (k,u)$, define
\[
m(t,q) \coloneqq \E_t\beta_q(\P_\theta)\quad\text{and}\quad S(t,q) \coloneqq n\Cov_t\{\beta_q(\P_\theta)\},
\]
so that $m_\pi(\cD_n,q) = m(\bbar,q)$ and $S_\pi(\cD_n,q) = S(\bbar,q)$ for $\bbar = \P_nb_{(K)} \in [-C_\infty,C_\infty]^K$. As $\Theta_K$ is compact, differentiating under the integral over $\theta \in \Theta_K$ by Theorem~A.5.3 of \cite{durrett2019}, where the centring derivatives cancel, we obtain, for $v \in \R^K$,
\[
\begin{aligned}
\partial_vm(t,q) &\coloneqq \frac{d}{d\lambda}m(t+\lambda v,q)\Big|_{\lambda = 0} = n\Cov_t\{\beta_q(\P_\theta),v^\T\theta\},\\
\partial_vS(t,q) &\coloneqq \frac{d}{d\lambda}S(t+\lambda v,q)\Big|_{\lambda = 0} = n^2\E_t\bigl[\{\beta_q(\P_\theta)-m(t,q)\}^{\otimes2}v^\T(\theta-\E_t\theta)\bigr].
\end{aligned}
\]
Since $\|\beta_q(\P_\theta)\| \leq C_\beta$ by the definition of $C_\beta$ in Section \ref{app:architecture}, $\|\theta\| \leq B$ on $\Theta_K$ and, by Lemma \ref{lem:query-computation}, $\|\beta_q(\P_\theta)-\beta_{q'}(\P_\theta)\| \leq CK^A\|u-u'\|$ for $q' \coloneqq (k,u')$, expanding the difference of the outer products, we obtain
\begin{gather*}
\sqrt n\|\partial_vm\|+\|\partial_vS\|_F \leq Cn^2\|v\|,\\
\sqrt n\|m(t,q)-m(t,q')\|+\|S(t,q)-S(t,q')\|_F \leq CnK^A\|u-u'\|.
\end{gather*}
For each coordinate $g$ of $(\sqrt n\,m,S)$, integrating $\lambda \mapsto \partial_{t'-t}g\{t+\lambda(t'-t),u\}$ over $\lambda \in [0,1]$, adding the second bound and taking $A \geq 1$, we find that, for $\bar L \coloneqq Cn_{\max}^2K^{A+1}$,
\begin{equation}\label{eq:posterior-moment-lipschitz}
|g(t,u)-g(t',u')| \leq \bar L\|(t,u)-(t',u')\|_\infty\quad\text{for}\quad t,t' \in \R^K\ \text{and}\ u,u' \in \cU_k.
\end{equation}

Let $\rho \coloneqq \varepsilon/(C_2\bar L)$ with $C_2 \geq 1$ chosen below, and let $\cX_\rho \subset \cX_k \coloneqq [-C_\infty,C_\infty]^K\times\cU_k$ contain one point of each nonempty $\cX_k\cap\rho(j+[0,1)^{K+\ell_k})$, $j \in \mathbb{Z}^{K+\ell_k}$, so that
\[
\max_{z \in \cX_k}\min_{z' \in \cX_\rho}\|z-z'\|_\infty \leq \rho\quad\text{and}\quad|\cX_\rho| \leq (C/\rho)^{K+\ell_k}.
\]
For each coordinate $g$, we define the McShane interpolant
\[
\widehat g(z) \coloneqq \min_{z' \in \cX_\rho}\{g(z')+\bar L\|z-z'\|_\infty\}\quad\text{for}\quad z \in \R^{K+\ell_k},
\]
which is $\bar L$-Lipschitz in the maximum norm and, by \eqref{eq:posterior-moment-lipschitz}, satisfies $g \leq \widehat g \leq g+2\bar L\rho$ on $\cX_k$. Consequently, for $z \in \cX_k$ and $\widehat z \in \R^{K+\ell_k}$,
\begin{equation}\label{eq:mcshane-error}
|\widehat g(\widehat z)-g(z)| \leq \bar L\|\widehat z-z\|_\infty+2\bar L\rho.
\end{equation}
As $|x| = x_++(-x)_+$, $\max(x,y) = y+(x-y)_+$ and $\min(x,y) = x-(x-y)_+$, balanced trees of affine maps and positive parts compute $\widehat g$. Since maxima and minima are $1$-Lipschitz in the maximum norm, perturbing these positive parts by $\tau$ changes the output by at most
\begin{equation}\label{eq:gate-perturbation}
(1+\bar L)\tau\log_2\{16(K+\ell_k)|\cX_\rho|\} \leq \frac{\varepsilon}{C_2}\quad\text{for}\quad\tau \leq \Bigl(\frac{Cn_{\max}K}{\varepsilon}\Bigr)^{-C},
\end{equation}
and, since all coefficients and intermediate values are bounded by $C(n_{\max}+\bar L)$, by Lemma \ref{lem:gelu-arithmetic} a GELU network of the same order of depth and width, with coefficients bounded by $(Cn_{\max}K/\varepsilon)^C$, implements such a perturbation $\widehat g_\tau$.

By Lemma \ref{lem:raw-basis-efficient} with accuracy $\rho$, the observation rows compute $\widehat b_{K,\rho}(Z_i)$, and one masked head with zero logits and all other value maps zero writes $\widehat t \coloneqq n^{-1}\sum_{i = 1}^n\widehat b_{K,\rho}(Z_i)$ into the query row. Since $\|\widehat t-\bbar\|_\infty \leq \rho$, by \eqref{eq:mcshane-error} and \eqref{eq:gate-perturbation}, we have
\[
|\widehat g_\tau(\widehat t,u)-g(\bbar,u)| \leq \frac{\varepsilon}{C_2}+3\bar L\rho = \frac{4\varepsilon}{C_2}.
\]
We evaluate these interpolants for all $n' \in \cN$ and $k \in \cK$ in parallel, multiply the mean coordinates of branch $n'$ by $(n')^{-\frac{1}{2}}$ and embed the outputs of family $k$ by the projections $P_k^\T$ and $M \mapsto P_k^\T MP_k$ of Section \ref{app:architecture}. Approximate products with the exact indicators
\[
1\{j = k\}\quad\text{in}\quad\xi(q,n)\quad\text{and}\quad1\{n = n'\} = (1-|n-n'|)_+,
\]
computed from the coordinate $n/n_{\max}$ of the query encoding $\xi(q,n)$ of Section \ref{app:architecture}, select the active branch with error at most $\varepsilon/(C_2\sqrt{n_{\max}})$ for $\tau \leq (Cn_{\max}K/\varepsilon)^{-C}$. Since $\clip_{C_\beta}(m_\pi) = m_\pi$ and $(S_\pi+S_\pi^\T)/2 = S_\pi$, for $C_2 \geq 10d_0$ with $d_0 = \max_{k \in \cK}d_k$, we obtain
\[
\sqrt n\|f^{\circ,\beta}-m_\pi\|+\|C^\circ-S_\pi\|_F \leq \frac{10d_0\varepsilon}{C_2} \leq \varepsilon\quad\text{and}\quad0 \leq h_{f^\circ} \leq C\varepsilon^2,
\]
where the second bound follows from \eqref{eq:raw-covariance-comparison} at size $n$. All these maps have depth at most $C\log\{(K+\ell_k)|\cX_\rho|/\rho\}$, width at most $C|\cN|(K+\ell_k)|\cX_\rho|$ and coefficients at most $(Cn_{\max}K/\varepsilon)^C$, and hence, by Lemma \ref{lem:gelu-arithmetic},
\[
L \leq CK\log\frac{Cn_{\max}K}{\varepsilon}\quad\text{and}\quad p+d_e+d_f+W \leq \Bigl(\frac{Cn_{\max}K}{\varepsilon}\Bigr)^{CK},
\]
since $|\cX_\rho| \leq (C/\rho)^{K+\ell_k}$, $\rho^{-1} \leq (Cn_{\max}K/\varepsilon)^{A+2}$ and $p \leq CLd_e(d_e+d_f)$. With the accuracy bounds above, $f^\circ$ thus has all the properties asserted in the lemma.
\end{proof}

\subsection{Lower bound at a fixed evaluation population}\label{app:fixed-lower-bound}

This subsection shows that the conclusion of Theorem \ref{thm:fixed-population} can fail for some approximate empirical risk minimiser when $T \leq \exp(cK\log N)$ for a small constant $c > 0$, even if all other conditions of the theorem hold. The reason is that all pretraining samples are then likely to be far from $\P_\star$, so that the empirical risk cannot distinguish an efficient Transformer from one biased at $\P_\star$. Let $\P_\star \coloneqq \P_{\theta_\star}$ for some $\theta_\star \in \R^{K_0}$ with $\|\theta_\star\|_1 < B$, identified with $(\theta_\star^\T,0,\cdots,0)^\T \in \Theta_K$ for $K \geq K_0$, such that $\|\beta_q(\P_\star)\| < C_\beta$ for $\pi_q$-almost every $q$, where $C_\beta$ is the radius of the point readout in Definition \ref{def:transformer-output}, and let $\nu \coloneqq \delta_{\P_\star}\otimes\pi_q$. Since $\P_\star \in \cP_s(\Xi,\eta)$ for every $s > 0$, with $\Xi \coloneqq (\sum_{k \leq K_0}k^{\frac{2s}{d}}\theta_{\star k}^2)^{\frac{1}{2}}$ and $\eta \coloneqq (B-\|\theta_\star\|_1)/2$, the population $\P_\star$ satisfies Assumption \ref{cond:fixed-population} whenever $K \asymp N^\kappa$ with $\kappa < (3\vee2a)^{-1}$.

We measure how close a sample $\cD_n = \{Z_i\}_{i = 1}^n$ with $n \in \cN$ is to $\P_\star$ by
\[
d_\star(\cD_n) \coloneqq \|\P_nb_{(K)}-s_\star\|_\infty,
\]
where $\P_nb_{(K)} \coloneqq n^{-1}\sum_{i = 1}^nb_{(K)}(Z_i)$ and $s_\star \coloneqq \E_{\P_\star}b_{(K)}$ average the sieve basis $b_{(K)}$ of Section \ref{app:prior}, which is bounded by $C_\infty$, over the sample and over $\P_\star$, respectively. With $\zeta \coloneqq C_\infty(\log N/N)^{\frac{1}{2}}$, we call $\cD_n$, and any context $(\cD_n,q)$, close to $\P_\star$ if $d_\star(\cD_n) \leq 2\zeta$ and far from $\P_\star$ if $d_\star(\cD_n) > 6\zeta$. When all pretraining samples are far from $\P_\star$, a map and its bump, defined below, have nearly the same empirical risk, although at most one of them can be efficient at $\P_\star$.

\begin{definition}[Bump]\label{def:bump}
A \textit{bump} of a map $f = (f^\beta,f^V)$ on contexts is a map $f' = (f'^\beta,f^V)$ that, up to an error $\tau \coloneqq \epsilon_{\mathrm{opt}}/N$ in norm, shifts the point estimate of $f$ by $N^{-\frac{1}{2}}e_1$ at contexts close to $\P_\star$, provided that $\|f^\beta\| \leq C_\beta-2N^{-\frac{1}{2}}$ there, and leaves it unchanged at contexts far from $\P_\star$, where $e_1$ is the first standard basis vector of $\R^{d_k}$ for $q = (k,u)$.
\end{definition}

\begin{proposition}[Lower bound on $T$ in Theorem \ref{thm:fixed-population}]\label{prop:fixed-lower-bound}
Suppose that all conditions of Theorem \ref{thm:fixed-population} except $T \geq \exp(CK^2\log N)$ hold with $\P_\star = \P_{\theta_\star}$, and that the Transformer class in \eqref{eq:erm} contains a bump of $\hat f$. Then there exists a constant $c > 0$ such that, if $T \leq \exp(cK\log N)$, some Transformer in this class that satisfies \eqref{eq:erm} with $\epsilon_{\mathrm{opt}} = O(N^{-\frac{K}{2}})$ violates \eqref{eq:single-efficiency}.
\end{proposition}

\begin{proof}
Let $\hat f'$ be the bump of $\hat f$ in the Transformer class $\cF_{\mathrm{TF}}$, and consider the event that all pretraining samples are far from $\P_\star$,
\[
\cE_\star \coloneqq \bigcap_{t \in [T]}\left\{d_\star(\cD^{(t)}_{n_t}) > 6\zeta\right\}.
\]
We bound $\Pr(\cE_\star^c)$ by splitting the simulated coefficients at distance $R_N$ from $\theta_\star$, show that $\hat f'$ nearly minimises $\hat R_T$ on $\cE_\star$, and compare $\hat f'^\beta$ with $\hat f^\beta$ at contexts close to $\P_\star$.

Let $c_I > 0$ satisfy $I_K(\theta) \succeq c_II$ for $\|\theta\|_1 \leq B+1$, as in Lemma \ref{lem:geometry}, and $R_N \coloneqq 24C_\infty c_I^{-1}(K\log N/N)^{\frac{1}{2}}$. By Assumptions \ref{cond:simulator-prior} and \ref{cond:posterior-tails} with $g_{\P,K} = 0$, we have $\cD^{(t)}_{n_t} \sim \P_\theta^{\otimes n_t}$ with $\theta \sim \pi_P$, $\pi_P(\Theta_K) = 1$ and $d\pi_P/d\pi_K \leq C_w$, so that, by the union bound over $t \in [T]$,
\begin{equation}\label{eq:lower-bound-split}
\Pr(\cE_\star^c) \leq T\Bigl[\pi_P\left\{\theta:\|\theta-\theta_\star\| < R_N\right\}+\sup_{n \in \cN,\,\theta \in \Theta_K:\|\theta-\theta_\star\| \geq R_N}\P_\theta^{\otimes n}\{d_\star(\cD_n) \leq 6\zeta\}\Bigr].
\end{equation}
For the second term, let $\Delta \coloneqq \theta-\theta_\star$. Since the log normaliser $A_K$ of Section \ref{app:prior} satisfies $\nabla A_K(\theta_\star) = s_\star$, by H\"older's inequality and Taylor's theorem, we have, whenever $d_\star(\cD_n) \leq 6\zeta$,
\[
\begin{aligned}
\log\frac{d\P_\theta^{\otimes n}}{d\P_\star^{\otimes n}}(\cD_n)
&= n\Delta^\T(\P_nb_{(K)}-s_\star)-n\{A_K(\theta)-A_K(\theta_\star)-\Delta^\T s_\star\}\\
&\leq \frac{nc_IR_N}{4}\|\Delta\|-\frac{nc_I}{2}\|\Delta\|^2,
\end{aligned}
\]
where the inequality also uses $\|\Delta\|_1 \leq \sqrt K\|\Delta\|$. Since $\P_\theta^{\otimes n}\{d_\star(\cD_n) \leq 6\zeta\}$ is at most the supremum of this likelihood ratio over this event and $n \geq n_{\min} \geq c_nN$ for a constant $c_n > 0$, we obtain, for $\|\Delta\| \geq R_N$,
\[
\P_\theta^{\otimes n}\{d_\star(\cD_n) \leq 6\zeta\} \leq \exp\Bigl(-\frac{nc_I}{4}\|\Delta\|^2\Bigr) \leq \exp\Bigl(-\frac{c_nc_IN}{4}R_N^2\Bigr) = N^{-C_2K},
\]
where $C_2 \coloneqq 144c_nC_\infty^2/c_I$. For the first term, as $d\pi_P/d\pi_K \leq C_w$, the Lebesgue density of $\theta_\P$ under $\pi_P$ is at most
\[
C_w\sup_{\theta \in \Theta_K}p_K(\theta) = \frac{C_w(K!)^a}{Z_K(2\pi\sigma^2)^{\frac{K}{2}}},
\]
where $p_K$ is the density of $\pi_K$ in the proof of Lemma \ref{lem:small-ball}, and, multiplying this bound by the volume $\pi^{\frac{K}{2}}R_N^K/\Gamma(K/2+1)$ of the ball, we obtain
\[
\begin{aligned}
\log\pi_P\left\{\theta:\|\theta-\theta_\star\| < R_N\right\}
&\leq \log\frac{C_w}{Z_K}+a\log(K!)-\log\Gamma\Bigl(\frac{K}{2}+1\Bigr)+\frac{K}{2}\log\frac{R_N^2}{2\sigma^2}\\
&\leq aK\log K-\frac{K}{2}\log N+CK\log\log N\\
&\leq -\frac{\epsilon}{4}K\log N,
\end{aligned}
\]
where we use $K! \leq K^K$, $\Gamma(K/2+1) \geq \{K/(2e)\}^{K/2}$, $\inf_KZ_K > 0$ and, for large $N$, $K^{2a} \leq N^{1-\epsilon}$ with $\epsilon \coloneqq (1-2a\kappa)/2 > 0$, by Assumption \ref{cond:fixed-population}. With $c \coloneqq \min(\epsilon/16,C_2/4)$ and $T \leq \exp(cK\log N) = N^{cK}$, the right side of \eqref{eq:lower-bound-split} is at most
\[
N^{cK}\bigl(N^{-\frac{\epsilon K}{4}}+N^{-C_2K}\bigr) \leq 2N^{-3cK} \leq N^{-cK}.
\]
On $\cE_\star$, by Definition \ref{def:bump}, $\hat f'^V = \hat f^V$ and $\|\hat f'^\beta-\hat f^\beta\| \leq \tau$ at $(\cD^{(t)}_{n_t},q_t)$ for all $t \in [T]$, so that, by \eqref{eq:loss-parameter-lipschitz} with $n_t$ in place of $N$ and by \eqref{eq:erm}, we obtain
\begin{equation}\label{eq:bump-erm}
\hat R_T(\hat f') \leq \hat R_T(\hat f)+C_\ell(1+n_{\max})\tau \leq \inf_{f \in \cF_{\mathrm{TF}}}\hat R_T(f)+C\epsilon_{\mathrm{opt}},
\end{equation}
where $C_\ell$ is the constant in \eqref{eq:loss-parameter-lipschitz} and $C \geq 1+C_\ell(1+n_{\max})/N$ is a constant, as $n_{\max} \lesssim N$.

Suppose next that $\hat f$ satisfies \eqref{eq:single-efficiency}. By Hoeffding's inequality and a union bound over the $K$ coordinates of $\P_Nb_{(K)}$, we have
\[
\P_\star^{\otimes N}\{d_\star(\cD_N) > 2\zeta\} \leq 2K\exp\Bigl(-\frac{N}{2C_\infty^2}\cdot\frac{4C_\infty^2\log N}{N}\Bigr) = \frac{2K}{N^2}.
\]
Since $\sqrt N\{\hat f^\beta-\beta_q(\P_\star)\} = O_\nu(1)$ by \eqref{eq:single-efficiency} and $\|\beta_q(\P_\star)\| < C_\beta$ for $\pi_q$-almost every $q$, by dominated convergence over $\pi_q$, we also have
\[
\Pr\nolimits_\nu\bigl(\|\hat f^\beta\| > C_\beta-2N^{-\frac{1}{2}}\bigr) \longrightarrow 0.
\]
On $\left\{d_\star(\cD_N) \leq 2\zeta,\ \|\hat f^\beta\| \leq C_\beta-2N^{-\frac{1}{2}}\right\}$, we have $\|\sqrt N(\hat f'^\beta-\hat f^\beta)-e_1\| \leq \sqrt N\tau \leq N^{-\frac{1}{2}}$ by Definition \ref{def:bump}, as $\epsilon_{\mathrm{opt}} \leq 1$ for large $N$, and hence
\[
\sqrt N\{\hat f'^\beta-\beta_q(\P_\star)\} = \sqrt N\{\hat f^\beta-\beta_q(\P_\star)\}+e_1+o_\nu(1).
\]
Conditional on $(q,\cT)$, the left side therefore converges in distribution to $\cN\{e_1,V_q^\star(\P_\star)\}$ in $\nu$-probability, so that $\hat f'$ violates \eqref{eq:single-efficiency}.

Finally, let $\hat f''$ equal $\hat f'$ and $\hat f$ on $\cE_\star$ and $\cE_\star^c$, respectively. By \eqref{eq:bump-erm} and \eqref{eq:erm}, both $\hat f \in \cF_{\mathrm{TF}}$ and $\hat f'' \in \cF_{\mathrm{TF}}$ satisfy \eqref{eq:erm} with $C\epsilon_{\mathrm{opt}} = O(N^{-\frac{K}{2}})$ in place of $\epsilon_{\mathrm{opt}}$. As $\Pr(\cE_\star^c) \leq N^{-cK} \to 0$, the map $\hat f''$ violates \eqref{eq:single-efficiency} whenever $\hat f$ satisfies it, which completes the proof.
\end{proof}

Transformer classes contain the bumps that the proposition requires. Let $\widehat x \coloneqq n^{-1}\sum_{i = 1}^n\widehat b_{K,\zeta}(Z_i)$ for the features $\widehat b_{K,\zeta}$ of Lemma \ref{lem:raw-basis-efficient}, so that $\|\widehat x-\P_nb_{(K)}\|_\infty \leq \zeta$. A path with its own pooling head can add $N^{-\frac{1}{2}}e_1\varphi_\star(\widehat x)$ to the point estimate, where
\[
\varphi_\star(x) \coloneqq \min\Bigl[1,\Bigl\{\frac{5}{2}-\frac{\|x-s_\star\|_\infty}{2\zeta}\Bigr\}_+\Bigr]\quad\text{for}\quad x \in \R^K
\]
satisfies $\varphi_\star(\widehat x) = 1$ at contexts close to $\P_\star$ and $\varphi_\star(\widehat x) = 0$ at contexts far from $\P_\star$.

Affine maps and positive parts with width $O(K)$ and depth $O(\log K)$ compute $\varphi_\star$, which the GELU networks of Lemma \ref{lem:gelu-arithmetic} approximate to accuracy $\sqrt N\tau$ with coefficients polynomial in $N/\tau$. As $\clip_{C_\beta}$ is nonexpansive and $\clip_{C_\beta}(\beta) = \beta$ whenever $\|\beta\| \leq C_\beta$, in particular for the shifted outputs at contexts with $\|f^\beta\| \leq C_\beta-2N^{-\frac{1}{2}}$, a Transformer class contains a bump of every member that leaves such a path unused.

\section{Proof of Proposition \ref{prop:query-verification}}\label{app:proofs-examples}

Throughout, the basis bound $C_\infty$, the $\ell_1$ radius $B$ of $\Theta_K$, the tilts $\P_\theta$ and the model $\cP$ are as in Section \ref{app:prior}, the densities $p \coloneqq d\P/d\P_0$ and $p' \coloneqq d\P'/d\P_0$ of $\P,\P' \in \cP$ satisfy $c_0 < p,p' < C_0$, and the quantities of the examples, such as $p_0$, $e_a$, $F_{a,\P}$, $\xi_{a,\P}$, $\lambda_l$ and $\Sigma_W$, are as in Section \ref{app:queries}.

\begin{proof}
We bound conditional moments, expand each query to first order in the density, from which parts~(i)--(iii) follow, and represent the queries as in Definition \ref{def:integral-queries}.

\paragraph{Step 1: Conditional moments and \eqref{eq:conditional-mean-lipschitz}.}
Let $\P,\P' \in \cP$, with $h' \coloneqq h(\P')$ for each quantity $h = h(\P)$ of the examples, and let $\delta \coloneqq \|p'-p\|_{L_2(\P_0)}$ and $\mathcal A \in \left\{\sigma(X),\sigma(D,X)\right\}$. Since $c_0 < p < C_0$, we have
\[
\frac{p}{\E_{\P_0}(p\mid\mathcal A)} \in \Bigl[\frac{c_0}{C_0},\frac{C_0}{c_0}\Bigr].
\]
Hence a real outcome $Y$, which is $\cN(0,1)$ and independent of the other coordinates under $\P_0$, has under $\P$ a conditional density in $[c\Phi',C\Phi']$ and bounded conditional moments given them, and
\begin{equation}\label{eq:qv-overlap}
c \leq e_a \leq 1-c,\quad p_G \geq c_0p_0\quad\text{and}\quad\inf_{y \in \cY}f_{a,\P}(y) \geq c,
\end{equation}
where $\cY \supseteq \left\{\xi_{a,\P}(u):\P \in \cP,\,a \in \{0,1\},\,u \in [\underline u,1-\underline u]\right\}$ is a compact interval. In the survival example, for $l \in [J-1]$,
\begin{equation}\label{eq:qv-survival-positivity}
\P(Y \geq l\mid X)\wedge\{1-\lambda_l(X)\} \geq \min_{a \in \cZ_D}\P\{(Y,\Delta) = a\mid X\} \geq c.
\end{equation}
Since $W$ is independent of $X$ under $\P_0$ and $\Var_{\P_0}(W) = I_{d_W}/12$, we have, for $v \in \R^{d_W}$,
\begin{equation}\label{eq:projection-gram-lower}
v^\T\Sigma_Wv = \inf_h\E_\P[\{v^\T W-h(X)\}^2] \geq c_0\inf_h\E_{\P_0}[\{v^\T W-h(X)\}^2] = \frac{c_0}{12}\|v\|^2,
\end{equation}
where the infima run over measurable $h$ with $h(X) \in L_2(\P_0)$, so that $\|\beta(\P)\| \lesssim 1$ by Cauchy--Schwarz. As $\beta_q(\P_0) = 0$ for treatment effects and projections, $\beta_u(\P_0) \leq J$ for mean survival times and $\|\psi^\star_{\P,q}(Z)\| \leq C(1+|Y|)$, we have
\begin{equation}\label{eq:qv-norms}
\|\beta_q(\P_0)\|+\|\psi^\star_{\P,q}\|_{L_2(\P)} \leq C.
\end{equation}

For measurable $f:\cZ \to \R$ with $\E_{\P_0}(f^2\mid\mathcal A) \leq C$, we have
\[
\E_{\P'}(f\mid\mathcal A)-\E_\P(f\mid\mathcal A) = \frac{\E_{\P_0}[\{f-\E_\P(f\mid\mathcal A)\}(p'-p)\mid\mathcal A]}{\E_{\P_0}(p'\mid\mathcal A)},
\]
and hence, by conditional Cauchy--Schwarz, uniformly over such $f$, including $1\{Y \leq y\}$ and $1\{(Y,\Delta) = a\}$,
\begin{equation}\label{eq:conditional-mean-lipschitz}
\|\E_{\P'}(f\mid\mathcal A)-\E_\P(f\mid\mathcal A)\|_{L_2(\P_0)} \leq C\delta.
\end{equation}

\paragraph{Step 2: Expansion \eqref{eq:query-frechet} and part (i).}
For a group effect, conditioning on $X$ and using $\E_{\P'}\{w_G(m_1'-m_0'-\beta_G')\} = 0$, we obtain
\begin{equation}\label{eq:qv-rem-group}
\beta_G'-\beta_G-\E_{\P'}\psi^\star_{\P,G} = \frac{(\beta_G'-\beta_G)(p_G-p_G')}{p_G}-\frac{1}{p_G}\E_{\P'}\Bigl\{w_G\sum_{a = 0}^1(2a-1)\frac{(e_a'-e_a)(m_a'-m_a)}{e_a}\Bigr\},
\end{equation}
and, by the same computation with $1\{Y \leq y\}$ in place of $Y$, a single arm $a \in \{0,1\}$ and $w_G \equiv 1$, we have
\begin{equation}\label{eq:qv-rem-cdf}
F_{a,\P'}(y)-F_{a,\P}(y)-\E_{\P'}\psi_{a,\P}(\cdot;y) = -\E_{\P'}\Bigl[\frac{\{e_a'(X)-e_a(X)\}\{F_a'(y\mid X)-F_a(y\mid X)\}}{e_a(X)}\Bigr].
\end{equation}
For the projection, whose contrasts follow by linearity, conditioning on $X$ to remove the cross terms in $\E_{\P'}\psi^\star_\P$, we obtain
\begin{equation}\label{eq:qv-rem-projection}
\beta'-\beta-\E_{\P'}\psi^\star_\P = \Sigma_W^{-1}(\Sigma_W-\Sigma_W')(\beta'-\beta)-\Sigma_W^{-1}\E_{\P'}[(m_W'-m_W)\{m_Y'-m_Y-(m_W'-m_W)^\T\beta\}].
\end{equation}
For a mean survival time, regarding $\mu_u$ as a polynomial in $(\lambda_l)_{l<u}$, let $\partial_l\mu_u \coloneqq \partial\mu_u/\partial\lambda_l$. By the EIF in Section \ref{app:queries}, we have
\[
\psi^\star_{\P,u}-\mu_u+\beta_u = \sum_{l<u}\frac{\partial_l\mu_u}{\P(Y \geq l\mid X)}\bigl[1\{Y = l,\Delta = 1\}-\lambda_l1\{Y \geq l\}\bigr],
\]
where $\E_{\P'}[1\{Y = l,\Delta = 1\}-\lambda_l1\{Y \geq l\}\mid X] = (\lambda_l'-\lambda_l)\P'(Y \geq l\mid X)$. Hence
\begin{align}
\beta_u'-\beta_u-\E_{\P'}\psi^\star_{\P,u} = {}&\E_{\P'}\Bigl\{\mu_u'-\mu_u-\sum_{l<u}\partial_l\mu_u(\lambda_l'-\lambda_l)\Bigr\}\notag\\
&-\E_{\P'}\Bigl\{\sum_{l<u}\partial_l\mu_u(\lambda_l'-\lambda_l)\frac{\P'(Y \geq l\mid X)-\P(Y \geq l\mid X)}{\P(Y \geq l\mid X)}\Bigr\},\label{eq:qv-rem-survival}
\end{align}
where the first expectation is a second order Taylor remainder of $\mu_u$, and $|\partial_l\mu_u| \leq J/c$ by \eqref{eq:qv-survival-positivity}.

Since $p_G$, $\E_\P\{w_G(m_1-m_0)\}$, $\Sigma_W$ and $\E_\P(\tilde W\tilde Y)$ integrate conditional moments against $p$, by \eqref{eq:qv-overlap}--\eqref{eq:conditional-mean-lipschitz}, we have, uniformly in $y \in \R$,
\[
|p_G'-p_G|+|\beta_G'-\beta_G|+\|\Sigma_W'-\Sigma_W\|+\|\beta'-\beta\|+\max_g\|g'-g\|_{L_2(\P_0)} \leq C\delta,
\]
where the maximum runs over $g \in \left\{e_a,m_a,F_a(y\mid\cdot),m_W,m_Y,\P(Y \geq l\mid\cdot),\lambda_l\right\}$ with $a \in \{0,1\}$ and $l \in [J]$. Hence, by \eqref{eq:qv-rem-group}--\eqref{eq:qv-rem-survival}, uniformly in $y$ and in $q$ other than quantile effects,
\begin{equation}\label{eq:qv-second-order}
|F_{a,\P'}(y)-F_{a,\P}(y)-\E_{\P'}\psi_{a,\P}(\cdot;y)| \leq C\delta^2\quad\text{and}\quad\|\beta_q(\P')-\beta_q(\P)-\E_{\P'}\psi^\star_{\P,q}\| \leq C\delta^2.
\end{equation}
For $a \in \{0,1\}$ and $u \in [\underline u,1-\underline u]$, let $\xi \coloneqq \xi_{a,\P}(u)$ and $\xi' \coloneqq \xi_{a,\P'}(u)$, both in $\cY$. Since $F_{a,\P'}(\xi') = u = F_{a,\P}(\xi)$ and $\E_\P\psi_{a,\P}(\cdot;y) = 0$, by the mean value theorem for $y \mapsto F_{a,\P}(y)$ and \eqref{eq:qv-second-order}, for some $\bar\xi \in [\xi\wedge\xi',\xi\vee\xi']$,
\[
f_{a,\P}(\bar\xi)(\xi'-\xi) = F_{a,\P}(\xi')-F_{a,\P'}(\xi') = -\int\psi_{a,\P}(\cdot;\xi')(p'-p)\,d\P_0+O(\delta^2).
\]
Hence $|\xi'-\xi| \leq C\delta$ by \eqref{eq:qv-overlap}, and, since $f_{a,\P}$ is continuous and $y \mapsto \psi_{a,\P}(\cdot;y)$ is continuous into $L_2(\P_0)$, we have
\[
\xi'-\xi = -\frac{1}{f_{a,\P}(\xi)}\int\psi_{a,\P}(\cdot;\xi)(p'-p)\,d\P_0+o(\delta).
\]
Since $\E_\P\psi^\star_{\P,q} = 0$, we have $\E_{\P'}\psi^\star_{\P,q} = \int\psi^\star_{\P,q}(p'-p)\,d\P_0$, and, subtracting the two arms for quantile effects, we obtain, for every $q \in \cQ$,
\begin{equation}\label{eq:query-frechet}
\beta_q(\P')-\beta_q(\P)-\int\psi^\star_{\P,q}(p'-p)\,d\P_0 = o(\delta)\quad\text{as}\quad\delta \to 0.
\end{equation}
Let $\P_\tau \in \cP$, with density $p_\tau \coloneqq d\P_\tau/d\P_0$, be differentiable in quadratic mean at $\P$ with score $s$, that is,
\[
\sqrt{p_\tau}-\sqrt p = \frac{\tau}{2}s\sqrt p+r_\tau\quad\text{with}\quad\|r_\tau\|_{L_2(\P_0)} = o(|\tau|),
\]
and hence, multiplying by $\sqrt{p_\tau}+\sqrt p$, we obtain
\[
p_\tau-p-\tau ps = r_\tau(\sqrt{p_\tau}+\sqrt p)+\frac{\tau}{2}s\sqrt p(\sqrt{p_\tau}-\sqrt p).
\]
Since $p_\tau,p \leq C_0$ and $\sqrt{p_\tau} \to \sqrt p$ in $L_2(\P_0)$, by dominated convergence along subsequences, $\|p_\tau-p-\tau ps\|_{L_2(\P_0)} = o(|\tau|)$. Hence, by \eqref{eq:query-frechet}, differentiating in $\tau$ at $\tau = 0$, we obtain
\[
\frac{d}{d\tau}\beta_q(\P_\tau)\Big|_{\tau = 0} = \int\psi^\star_{\P,q}ps\,d\P_0 = \E_\P(\psi^\star_{\P,q}s).
\]
As $\psi^\star_{\P,q}$ lies in $L_2^0(\P)$, the tangent space of $\cP$ at $\P$ in Section \ref{app:prior}, this gradient is the EIF of $\beta_q$ at $\P$, which, with \eqref{eq:qv-norms}, proves part~(i).

\paragraph{Step 3: Continuity of $\psi^\star_{\P,q}$ and part (iii).}
For group effects, mean survival times and projections, by \eqref{eq:qv-overlap}--\eqref{eq:conditional-mean-lipschitz} and the independence of $Y$ and the other coordinates under $\P_0$, each term $\phi_\P$ of $\psi^\star_{\P,q}$ satisfies $\|\phi_{\P'}-\phi_\P\|_{L_2(\P_0)} \leq C\delta$, so that
\begin{equation}\label{eq:gradient-lipschitz}
\|\psi^\star_{\P',q}-\psi^\star_{\P,q}\|_{L_2(\P_0)} \leq C\|p'-p\|_{L_2(\P_0)}.
\end{equation}
Since $\psi^\star_{\P,G}$ depends on $G$ only through $w_G$, $p_G$ and $\beta_G$, for rectangles $G$ and $G'$ we have
\[
\|\psi^\star_{\P,G'}-\psi^\star_{\P,G}\|_{L_2(\P_0)} \leq C\P_0(X \in G\mathbin{\triangle}G')^{\frac{1}{2}},
\]
which tends to zero as $\|u'-u\| \to 0$ for the endpoint vectors $u$ and $u'$ of $G$ and $G'$, while $\|M'\psi^\star_\P-M\psi^\star_\P\|_{L_2(\P_0)} \leq \|M'-M\|_{\op}\|\psi^\star_\P\|_{L_2(\P_0)}$ for contrasts and the horizons $u \in \{2,\cdots,J\}$ form a finite set.

For quantile effects, by Step~2 and the conditional density bounds of Step~1, we have
\[
|\xi_{a,\P'}(u')-\xi_{a,\P}(u)| \leq C(\delta+|u'-u|)\quad\text{and}\quad\sup_{y \in \R}|f_{a,\P'}(y)-f_{a,\P}(y)| \leq C\|p'-p\|_\infty.
\]
Since $f_{a,\P}$ is continuous and $\|1\{Y \leq y'\}-1\{Y \leq y\}\|_{L_2(\P_0)}^2 = |\Phi(y')-\Phi(y)|$, the $L_2(\P_0)$ norm of every term of $\psi^\star_{\P',(k,u')}-\psi^\star_{\P,(k,u)}$ tends to zero as $\|p'-p\|_\infty+|u'-u| \to 0$, which proves part~(iii).

\paragraph{Step 4: Lipschitz bound on $D_q$ and part (ii).}
For $\theta,\theta' \in \R^K$ with $\|\theta\|_1,\|\theta'\|_1 \leq B+1$ and $v,w \in \R^K$, let $p_\theta \coloneqq d\P_\theta/d\P_0$, $s_{\theta,v} \coloneqq v^\T\{b_{(K)}-\E_{\P_\theta}b_{(K)}\}$ and $\partial_v\varphi(\theta) \coloneqq (d/dt)\varphi(\theta+tv)|_{t = 0}$, so that
\[
\partial_vp_\theta = p_\theta s_{\theta,v}\quad\text{and}\quad\partial_w\partial_vp_\theta = p_\theta\{s_{\theta,v}s_{\theta,w}-v^\T I_K(\theta)w\},
\]
where $I_K(\theta) = \Cov_{\P_\theta}\{b_{(K)}\}$. Since $|s_{\theta,v}| \leq 2C_\infty\sqrt K\|v\|$ and $\|s_{\theta,v}\|_{L_2(\P_0)} \leq C\|v\|$, by Lemma \ref{lem:geometry} and, for \eqref{eq:qv-tilt-lipschitz}, integrating $t \mapsto \partial_{\theta'-\theta}\partial_vp_{\theta+t(\theta'-\theta)}$ over $t \in [0,1]$, we have
\begin{gather}
\|\partial_vp_\theta\|_{L_2(\P_0)} \leq C\|v\|\quad\text{and}\quad\|\partial_vp_\theta\|_\infty \leq C\sqrt K\|v\|,\label{eq:qv-tilt-bounds}\\
\|\partial_vp_{\theta'}-\partial_vp_\theta\|_{L_2(\P_0)} \leq C\sqrt K\|v\|\|\theta'-\theta\|.\label{eq:qv-tilt-lipschitz}
\end{gather}
Since $\|p_{\theta+\tau v}-p_\theta-\tau\partial_vp_\theta\|_\infty = O(\tau^2)$, by \eqref{eq:query-frechet} we have $D_q(\theta)v = \int\psi^\star_{\P_\theta,q}\partial_vp_\theta\,d\P_0$ for the Jacobian $D_q(\theta) = \partial\beta_q(\P_\theta)/\partial\theta^\T$, and hence
\begin{equation}\label{eq:jacobian-split}
\{D_q(\theta')-D_q(\theta)\}v = \int(\psi^\star_{\P_{\theta'},q}-\psi^\star_{\P_\theta,q})\partial_vp_{\theta'}\,d\P_0+\int\psi^\star_{\P_\theta,q}(\partial_vp_{\theta'}-\partial_vp_\theta)\,d\P_0.
\end{equation}
By \eqref{eq:qv-norms}, \eqref{eq:gradient-lipschitz}, \eqref{eq:qv-tilt-bounds}, \eqref{eq:qv-tilt-lipschitz} and $\|p_{\theta'}-p_\theta\|_{L_2(\P_0)} \leq C\|\theta'-\theta\|$, we bound both integrals in norm by $C\sqrt K\|v\|\|\theta'-\theta\|$ for group effects, mean survival times and projections.

For quantile effects at level $u$, let $\xi \coloneqq \xi_{a,\P_\theta}(u)$ and $\xi' \coloneqq \xi_{a,\P_{\theta'}}(u)$ for $a \in \{0,1\}$, so that $|\xi'-\xi| \leq C\|\theta'-\theta\|$ by Step~2. Since
\[
\|\log p_{\theta'}-\log p_\theta\|_\infty \leq C\sqrt K\|\theta'-\theta\|\quad\text{and}\quad\|\partial_y\theta^\T b_{(K)}\|_\infty \leq Cm_K \leq C\sqrt K,
\]
where the last bound follows from \eqref{eq:frequency-bound} for the largest coordinate frequency $m_K$ among $b_1,\cdots,b_K$ and $d = 1+d_X \geq 2$, we have
\[
|f_{a,\P_{\theta'}}(\xi')-f_{a,\P_\theta}(\xi)| \leq C\sqrt K\|\theta'-\theta\|+C(1+m_K)|\xi'-\xi| \leq C\sqrt K\|\theta'-\theta\|.
\]
Moreover, with $e_a'(x) \coloneqq \P_{\theta'}(D = a\mid X = x)$, we have
\[
\psi_{a,\P_{\theta'}}(\cdot;\xi')-\psi_{a,\P_\theta}(\cdot;\xi) = \frac{1\{D = a\}}{e_a'(X)}\{1\{Y \leq \xi'\}-1\{Y \leq \xi\}\}+R_a,
\]
where the remainder $R_a$ satisfies $\|R_a\|_{L_2(\P_0)} \leq C\|\theta'-\theta\|$ by \eqref{eq:qv-overlap} and \eqref{eq:conditional-mean-lipschitz}, and, by \eqref{eq:qv-tilt-bounds} and $\Phi' \leq 1$,
\[
\Bigl|\int\frac{1\{D = a\}}{e_a'(X)}\{1\{Y \leq \xi'\}-1\{Y \leq \xi\}\}\partial_vp_{\theta'}\,d\P_0\Bigr| \leq C\sqrt K\|v\||\xi'-\xi|.
\]
Since $\psi^\star_{\P,u} = \sum_{a = 0}^1(1-2a)\psi_{a,\P}\{\cdot;\xi_{a,\P}(u)\}/f_{a,\P}\{\xi_{a,\P}(u)\}$, we obtain, for every $q \in \cQ$,
\[
\|D_q(\theta')-D_q(\theta)\|_{\op} = \sup_{v \in \R^K:\|v\| = 1}\|\{D_q(\theta')-D_q(\theta)\}v\| \leq C\sqrt K\|\theta'-\theta\|,
\]
which is part~(ii).

\paragraph{Step 5: Integral representations and part (iv).}
Let $x_0 \coloneqq \Phi(y)$, let $x \in [0,1]^{d_X}$ denote the value of $X$, let $p_a(x_0,x) \coloneqq e^{g_{\theta,a}(x_0,x)}$, with $g_{\theta,a}$ as in Section \ref{app:comparator-primitives}, be the generating fields of condition \ref{cond:IQ1}, and define
\[
\varrho_a(x) \coloneqq \int_0^1p_a(x_0,x)\,dx_0\quad\text{and}\quad\mu_a(x) \coloneqq \int_0^1\Phi^{-1}(x_0)p_a(x_0,x)\,dx_0,
\]
where $\Phi^{-1} \in L_1([0,1])$. With $\varrho \coloneqq (\varrho_0+\varrho_1)/2$, we have $d\P_{\theta,X}/dx \propto \varrho$ for the distribution $\P_{\theta,X}$ of $X$ under $\P_\theta$, and $m_a = \mu_a/\varrho_a$, so that, cancelling the factor $e^{-A_K(\theta)}$, we have
\[
\beta_G(\P_\theta) = \frac{\int_G\varrho(\mu_1/\varrho_1-\mu_0/\varrho_0)\,dx}{\int_G\varrho\,dx}.
\]
Since $\varrho_a \geq c$ and $\int_G\varrho\,dx \geq cp_0$, the equation map and the readout of condition \ref{cond:IQ2},
\[
F_k(\theta,v,u) \coloneqq v-\beta_G(\P_\theta)\quad\text{and}\quad R_k(v,u) \coloneqq v,
\]
represent group effects on a box, with rectangle endpoints as integration limits and polynomial selectors for finite collections.

For mean survival times, $V = X$ and $a \in \cZ_D$ ranges over the values of $(Y,\Delta)$, so that, with the fields $p_a(x) \coloneqq e^{g_{\theta,a}(x)}$,
\[
\frac{d\P_{\theta,X}}{dx} \propto \sum_ap_a\quad\text{and}\quad\P_\theta\{(Y,\Delta) = a\mid X = x\} = \frac{p_a(x)}{\sum_{a'}p_{a'}(x)}.
\]
Hence, for $l \in [J]$,
\[
1-\lambda_l = \frac{p_{(l,0)}+\sum_{j > l}\{p_{(j,0)}+p_{(j,1)}\}}{\sum_{j \geq l}\{p_{(j,0)}+p_{(j,1)}\}},\quad\text{where } p_{(J,0)} \coloneqq 0,
\]
is a ratio of fields whose denominator is at least $c$, so that $S(j\mid\cdot)$ is a field and
\[
\beta_u(\P_\theta) = \sum_{j = 0}^{u-1}\frac{\int S(j\mid x)\sum_ap_a(x)\,dx}{\int\sum_ap_a(x)\,dx}
\]
is represented like a group effect, with polynomial selectors for the horizon $u$.

For the projection, $\cZ_D$ is a singleton and fields also depend on $w$. Integrating $p_a\,dx_0\,dw$ against $(1,w,w^{\otimes2})$ and $(1,w)\Phi^{-1}(x_0)$, dividing by the covariate density and integrating products in $x$, we express $\Sigma_W$ and $\E_{\P_\theta}(\tilde W\tilde Y)$ by rectangle integrals of fields in
\[
F_k(\theta,v,u) \coloneqq \Sigma_Wv-\E_{\P_\theta}(\tilde W\tilde Y)\quad\text{and}\quad R_k(v,u) \coloneqq Mv,
\]
where $M$ is the encoded contrast or the identity, and condition \ref{cond:IQ3} holds on a sufficiently large box by \eqref{eq:projection-gram-lower} and the bounded second moments. For quantile effects, the field
\[
h_a(x_0) \coloneqq \frac{\int\varrho(x)p_a(x_0,x)/\varrho_a(x)\,dx}{\int\varrho(x)\,dx} \in [c,C]
\]
satisfies $F_{a,\P_\theta}(y) = \int_0^{\Phi(y)}h_a(x_0)\,dx_0$. On a box $[-H,H]^2 \supseteq \left\{v \in \R^2:\operatorname{dist}(v,\cY^2) \leq c\right\}$, we represent the quantile effect by
\[
F_k(\theta,v,u) \coloneqq \begin{pmatrix}F_{0,\P_\theta}(v_0)-u\\F_{1,\P_\theta}(v_1)-u\end{pmatrix}\quad\text{and}\quad R_k(v,u) \coloneqq v_1-v_0,
\]
with $\Phi$ as a univariate analytic function. Condition \ref{cond:IQ3} holds on the box since $\partial_vF_k = \diag\{h_a(\Phi(v_a))\Phi'(v_a)\}_{a = 0,1}$ has diagonal entries in $[c,C]$. All bounds are uniform in $K \geq 1$, $\theta \in \R^K$ with $\|\theta\|_1 \leq B+1$ and $u \in \cU_k$, which proves part~(iv).

Finite stacks satisfy parts~(i)--(iii) coordinatewise and part~(iv) with product boxes and block-diagonal $\partial_vF_k$, which preserve \eqref{eq:integral-query-curvature}, with $\sum_{a \in \{0,1\}}p_a/2$ in place of $p_a$ in the projection of a stack with treatment effects. This completes the proof.
\end{proof}

\section{Experimental details}\label{app:experiments}

\subsection{Transformer architecture: details}

The implementation uses a 20-layer pre-norm Transformer of width 384, with six attention heads and a learned dataset-level readout; it has approximately 20 million parameters.  Observation rows have no positional encoding, making the readout invariant to their ordering. Scalar experiments use separate point and variance heads, with the variance mapped through a softplus function plus $10^{-3}$ to ensure positivity. 

\subsection{Setup of priors}\label{app:exp-prior}
\paragraph{ATE.}\phantomsection\label{app:ate-details} A base table is first drawn with equal probability from one of three sources, namely OpenML tables, a graph-based synthetic table generator and a random multilayer perceptron generator, which are described below.
\begin{itemize}[leftmargin=*]
    \item \emph{OpenML tables}.  A dataset is drawn uniformly from the 337 OpenML datasets curated by \cite{balazadeh2025}, which exclude every table that shares covariates or outcomes with the real-data benchmarks, including IHDP. Categorical columns are integer-coded and missing values are imputed by column medians.
    \item \emph{Graph-based generator}.  Tables are drawn from the structural causal model prior of TabICLv2 \citep{qu2026}, which builds a random directed acyclic graph with at most 32 nodes and assigns each node a function from several families, including Gaussian processes, neural networks, tree ensembles and linear maps. 
    \item \emph{Random multilayer perceptron generator}.  In the spirit of the TabPFN prior \citep{hollmann2023}, a few independent standard Gaussian variables are passed through a random multilayer perceptron with one to three layers, whose activation is drawn from the hyperbolic tangent, sine, ReLU and identity functions.  The columns of the table are sampled from the inputs and all intermediate activations, and up to a third of them are discretised into two to ten quantile bins to mimic categorical features.
\end{itemize}

After $d_X \in \{2,\dotsc,100\}$ covariate columns are selected, two further columns give raw potential-outcome means $\widetilde\mu_0$ and $\widetilde\mu_1$.  With $\lambda \coloneqq \E(\widetilde\mu_1-\widetilde\mu_0)$, heterogeneity $\gamma \sim \mathrm{Unif}(0,1)$ and offset $\kappa \sim \mathrm{Unif}(-5,5)$, the conditional means are
\[
 \mu_0 = \widetilde\mu_0,
 \quad
 \mu_1 = \widetilde\mu_0+\gamma(\widetilde\mu_1-\widetilde\mu_0)
 +(1-\gamma)\lambda+\kappa,
\]
so the ATE label is $\tau = \lambda+\kappa$.  Independent arm-specific Gaussian errors are multiplied by covariate-dependent scales to introduce heteroskedasticity.  Treatment is sampled with probability
\[
 \xi\,\operatorname{expit}\{g(X)\}+(1-\xi)\1\{g(X) > 0\},
 \quad \xi \sim \mathrm{Unif}(0,1),
\]
where $g$ is drawn as a constant, a linear map or a single-hidden-layer random network.  Probabilities are clipped to $[10^{-3},1-10^{-3}]$, and draws with numerically empty arms are regenerated.

\paragraph{QTE.}\phantomsection\label{app:qte-details} The data-generating process follows that of the ATE. The main difference is that the QTE has no closed form, so the training label is computed by Monte Carlo: each data-generating process generates $M$ additional units from the same mechanism and records both potential outcomes $Y_j(0)$ and $Y_j(1)$, $j=1,\dotsc,M$. The target at quantile level $u$ is $\tau(u) \coloneqq \hat Q_1(u)-\hat Q_0(u)$, where $\hat Q_d$ is the empirical quantile function of $\{Y_j(d)\}_{j=1}^M$. In our experiment, $M$ is set as 8192.

\paragraph{MST.}\phantomsection\label{app:mst-details} The survival generator draws a base table and selects a subset of its columns as covariates, as for the ATE.  One further column is selected at random and shifted to give a nonnegative failure time $T$.  Given a target censoring level $\rho \sim \mathrm{Unif}(0.05,0.9)$, the censoring times are sampled conditionally on the covariates.  Specifically, with $T_{\max}$ the largest simulated failure time in the dataset, the censoring times are drawn as
\[
 C_i \sim \mathrm{Unif}\left[0,\frac{T_{\max}}
 {\rho\,\operatorname{expit}\{g(X_i)\}}\right],
\]
so that unit $i$ is censored with probability $\rho\operatorname{expit}\{g(X_i)\}T_i/T_{\max}$, and hence $\rho$ is an upper bound on the censoring rate.  The MST label is the sample mean $\frac{1}{n}\sum_{i = 1}^nT_i$ of the latent failure times.

\paragraph{PLR.}\phantomsection\label{app:plm-details} For the partially linear projection, we draw $\beta$ uniformly from $[-3,3]$.  Given nuisance covariates $X$, the target regressor is generated as
\[
 W = m_\psi(X)+V,
\]
where $m_\psi$ is a single-hidden-layer random network and $V$ is a zero-mean random variable with positive variance.  A further base-table column $\widetilde\mu$ and conditionally mean-zero noise $u$ form
\[
 Y = W\beta+\widetilde\mu+u.
\]

\subsection{Optimisation}
Training uses AdamW with moment coefficients $(0.9,0.95)$, learning rate $10^{-4}$, 1,000 warm-up steps followed by cosine decay, weight decay $0.01$, gradient clipping at $1$, and bfloat16 autocasting.  Each device processes 32 independently drawn data-generating processes per update. One context length is sampled uniformly from $\{800,1000,1200,1400,1600\}$ for each update. A single validation set is generated before training to select the checkpoint. The total number of updates is $5 \times 10^5$.

\subsection{Baselines: implementation}\label{app:baseline-implementation}
Every baseline receives exactly the same dataset $\cD_N$ as our estimator. Unless stated otherwise, nuisance functions are fitted on the full evaluation sample, and the interval of a baseline is the two-sided Wald interval $\hat\beta\pm z_{0.975}\widehat{\mathrm{SE}}$, where $z_{0.975}$ is the $0.975$ quantile of the standard normal distribution. Several baselines have no implemented standard error; for these, we report the plug-in standard errors described below.

\paragraph{ATE.}  Write $e(x)\coloneqq\P(D=1\mid X=x)$ for the propensity score, and recall that $m_d(x)=\E_\P(Y\mid D=d,X=x)$.  For the classical methods, random forests are fitted with scikit-learn, using the settings of the benchmark code released with CausalPFN.  Fitted propensity scores are clipped to $[10^{-3},1-10^{-3}]$.
\begin{itemize}[leftmargin=*]
 \item \emph{CausalPFN}.  We use the official package and pretrained checkpoint\footnote{\url{https://github.com/vdblm/CausalPFN}; checkpoint \url{https://huggingface.co/vdblm/causalpfn}.}. The point estimate is $N^{-1}\sum_{i=1}^N\{\hat m_1(X_i)-\hat m_0(X_i)\}$, where $\hat m_d(X_i)$ is the prediction of CausalPFN, and the interval is formed by the 2.5\% and 97.5\% quantiles of 1,000 ATE draws from its predictive distribution.
 \item \emph{Do-PFN}.  We use the publicly released code and checkpoint\footnote{\url{https://github.com/jr2021/Do-PFN}.}. The ATE is the average of the predicted conditional average treatment effects over the rows of $\cD_N$. Therefore, Do-PFN does not output an interval and is absent from the coverage panel of Figure \ref{fig:ate-results}.
 \item \emph{S-, T- and X-learners}.  We use EconML 0.17.0\footnote{\url{https://github.com/py-why/EconML}.} with random forests for all outcome models and, for the X-learner, a random-forest propensity model. Since EconML does not provide a standard error for these estimators, we plug the learner's own outcome regressions $\hat m_0,\hat m_1$ and effect function $\hat\tau$, together with a separately fitted random-forest propensity score $\hat e$, into the EIF, and take the empirical standard deviation of the resulting EIF values divided by $\sqrt N$ as the standard error. 
 \item \emph{IPW}.  We implement the normalised (H\'ajek) estimator, whose standard error is given by the sandwich estimator.
 \item \emph{AIPW}.  The nuisances are fitted by random forests with two-fold cross-fitting, where the folds are stratified by $D$.
 \item \emph{Difference in means}.  The variance is estimated by $\hat\sigma_1^2/N_1+\hat\sigma_0^2/N_0$, where $N_d$ and $\hat\sigma_d^2$ are the size and the sample variance of arm $d$.
\end{itemize}

\paragraph{MST.} Except for the complete-case mean, every baseline integrates an estimated survival function over $[0,\max_{1\leq i\leq N}Y_i]$.
\begin{itemize}[leftmargin=*]
 \item \emph{Complete-case mean}.  The average of $Y_i$ over the uncensored observations. 
 \item \emph{Kaplan--Meier estimator}.  The estimate is the area under the Kaplan--Meier curve.  Its variance is a Greenwood-type estimate, adjusted for the sample-dependent label by weighting each summand by one minus the Kaplan--Meier estimate of the censoring survival function just before the corresponding event time.
 \item \emph{IPCW}.  The survival function is estimated by weighting the at-risk indicators by the inverse of the Kaplan--Meier estimate of the censoring survival function. 
 \item \emph{Cox model}.  We use lifelines 0.30.3\footnote{\url{https://github.com/CamDavidsonPilon/lifelines}.} with the Breslow baseline hazard.  The estimate averages the integrated predicted survival curves over the sample, and its variance is computed from the influence function of this plug-in estimator, with an analogous adjustment for the sample-dependent label. 
 \item \emph{Random survival forest}.  We use scikit-survival 0.28.0\footnote{\url{https://github.com/sebp/scikit-survival}.}, and the estimate is computed in the same way as for the Cox model. 
\end{itemize}

\paragraph{PLR.}  The three baselines are our own implementations.  Each regresses the residual of $Y$ on the residual of $W$, after removing estimated functions of $X$ from both, and uses the heteroskedasticity-robust sandwich variance in the $\chi^2_1$ Wald interval.
\begin{itemize}[leftmargin=*]
 \item \emph{OLS}.  The residuals come from least-squares regressions on $(X,1)$, so the estimate is the coefficient of $W$ in the least-squares regression of $Y$ on $(W,X,1)$.
 \item \emph{Robinson residualisation}.  The conditional means $m_W$ and $m_Y$ are fitted on the full sample by least squares on $(X,1)$, followed by a random forest fitted to the least-squares residuals.
 \item \emph{Double machine learning}.  The same learners as for Robinson residualisation, combined with two-fold cross-fitting.
\end{itemize}

\paragraph{QTE.}  The four classical estimators are our own implementations.  The propensity score and the conditional distribution functions are fitted by logistic regression (scikit-learn), and the propensity score is clipped to $[0.01,0.99]$.
\begin{itemize}[leftmargin=*]
 \item \emph{Unadjusted}.  The difference between the arm-wise empirical quantiles.
 \item \emph{IPW}.  The difference between the arm-wise quantiles weighted by the inverse propensity score.
 \item \emph{Distribution regression}.  For each arm, the conditional distribution function is fitted by logistic regressions of $\1\{Y\leq y\}$ on $X$ over a grid of 64 thresholds $y$; it is then averaged over the sample and inverted to give the marginal quantile.
 \item \emph{Doubly robust}.  The marginal distribution function from distribution regression is augmented by an inverse-propensity-weighted residual term before inversion.
 \item \emph{CausalPFN}.  With the settings used for the ATE, CausalPFN predicts $\hat m_d(X_i)$.  The distribution of $Y(d)$ is approximated by the empirical distribution of $\hat m_d(X_j)$ plus the centred residuals of arm $d$, and the estimate is the difference of the resulting quantiles.
\end{itemize}
All standard errors are computed from estimated influence functions, with the quantile density estimated by a difference quotient of the method's own quantile function.  Distribution regression and the doubly robust estimator use the efficient influence function evaluated at the fitted nuisances, while CausalPFN uses an influence function that treats $\hat m_0$ and $\hat m_1$ as known.

\subsection{Runtime analysis: details}

For Figure \ref{fig:runtime}, each method is applied to 100 held-out datasets at each context length, one dataset per call, after one unrecorded warm-up call for each method and length.  The recorded time covers what is run on a new dataset: preprocessing, host-to-device transfer, any per-dataset model fitting, prediction and, where available, the computation of the standard error. Our estimator, CausalPFN and Do-PFN run on the GPU, whereas the classical estimators run on the CPU with BLAS, OpenMP and joblib parallelism restricted to a single thread.

\subsection{Computational infrastructure}\label{app:computation}
All computations were run on NVIDIA A100-SXM4 GPUs (Ampere architecture, 80\,GB memory) with 18 cores of an Intel Xeon Platinum 8360Y CPU (2.40\,GHz) and 256\,GB of memory.  The software environment consists of Python 3.11.15 and PyTorch 2.13.0\footnote{\url{https://pytorch.org}.}, which is used to implement the neural-network-based methods.

\subsection{Numerical results for mean survival time}
\label{sec:exp-mst}

The mean survival time is the expected time to an event, such as death, relapse or system failure; its estimation must account for incomplete follow-up caused by right censoring.  Let $T$ and $C$ denote the latent failure and censoring times, and let $X$ contain baseline covariates.  Their joint law induces an observed-data distribution $\P$ for $Z=(Y,\Delta,X)$, where
 $Y \coloneqq \min(T,C)$ and $\Delta \coloneqq \1\{T\leq C\}$.
At the population level, the mean failure time $\beta_{\mathrm{MST}}(\P)=\E(T)$ is identified from $\P$ under conditionally independent censoring and suitable positivity and support conditions.  The evaluation data are $\cD_N=\{(Y_i,\Delta_i,X_i)\}_{i=1}^N$. 

We compare our estimator with the complete-case mean $\sum_{i=1}^N\Delta_iY_i/\sum_{i=1}^N\Delta_i$, the Kaplan--Meier estimator \citep{kaplan1958}, an inverse-probability-of-censoring-weighted estimator \citep{satten2001}, and plug-in estimators based on the Cox model \citep{cox1972} and a random survival forest \citep{ishwaran2008}.  In Figure \ref{fig:mst-results}, our estimator has the lowest RMSE throughout, and its decrease broadly follows the $N^{-1/2}$ reference trend.  Its Wald coverage remains close to the nominal 95\% level.  The rate and coverage patterns are qualitatively consistent with Theorem \ref{thm:main}.

\begin{figure}[H]
\centering
\includegraphics[width=\linewidth]{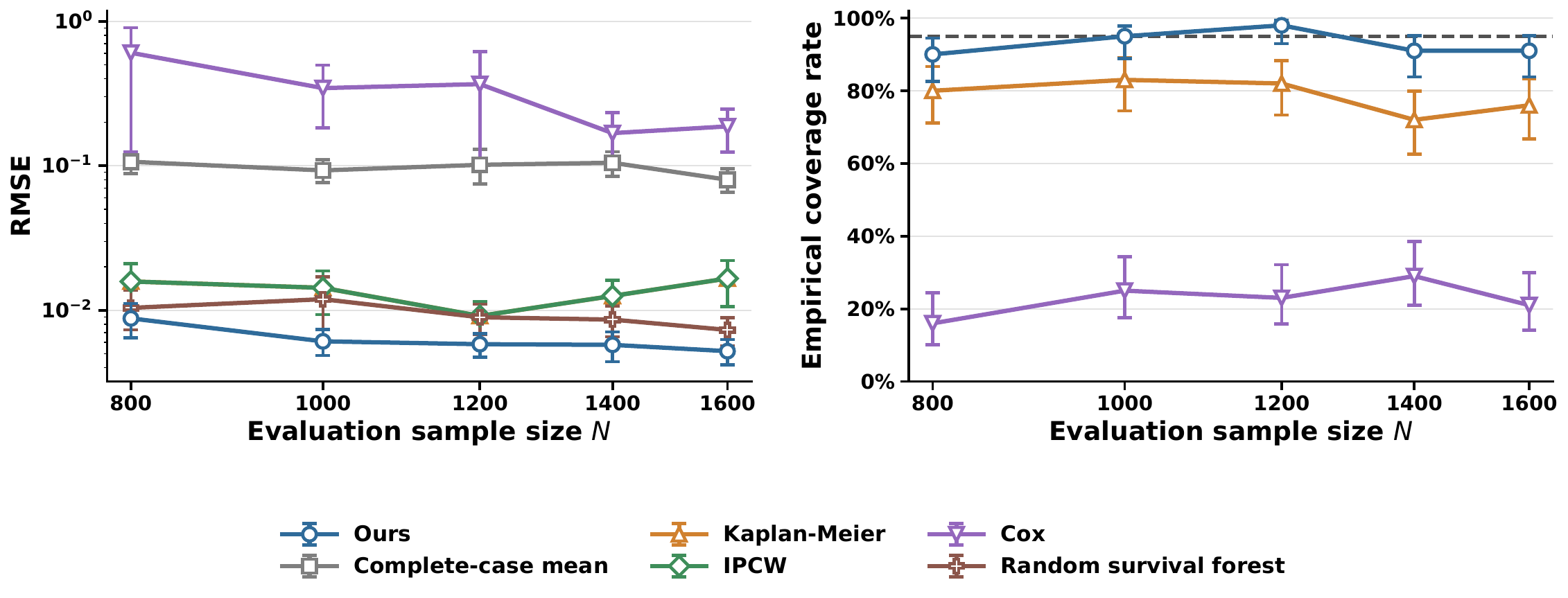}
\caption{Performance on mean-survival-time estimation.}
\label{fig:mst-results}
\end{figure}

\subsection{Additional numerical results for ATE and QTE}

\begin{figure}[H]
\includegraphics[width=\linewidth]{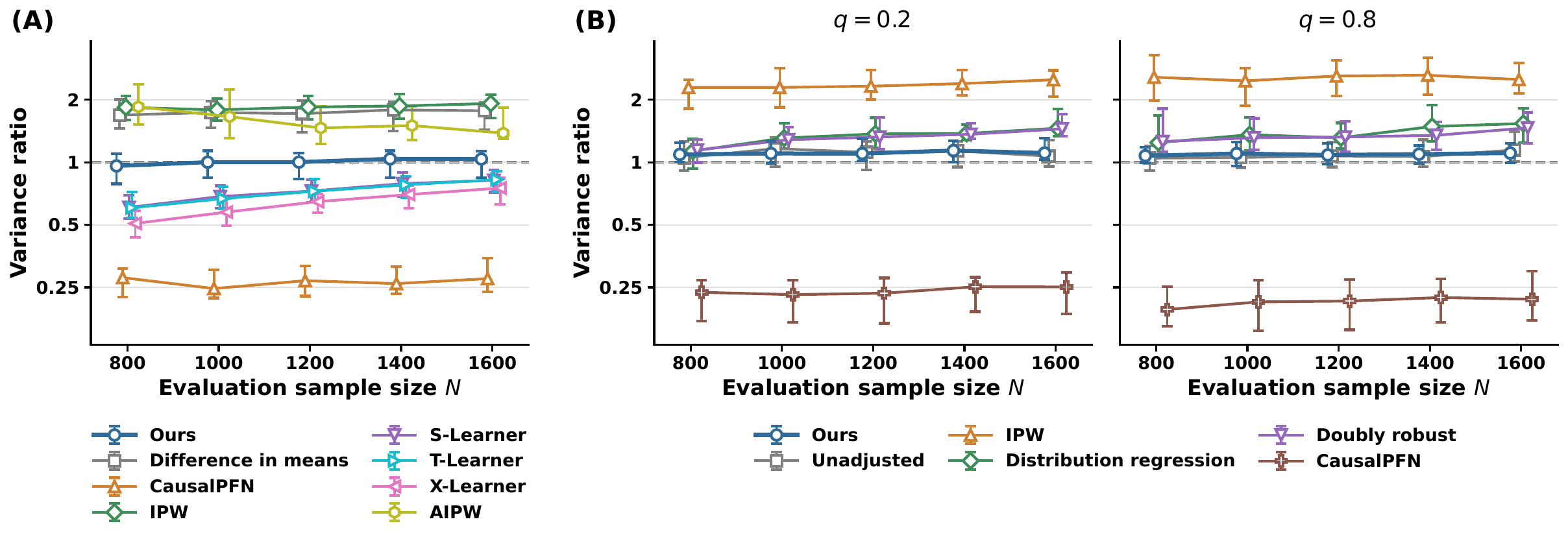}
\caption{Median ratio of each method's estimated variance, $\hat V$, to the efficiency bound $V_q^\star(\P)$. Error bars are pointwise 95\% bootstrap intervals that resample complete simulation replicates. Panels (A) and (B) are for the ATE and QTE, respectively. The dashed line marks a ratio of 1.}\label{fig:variance-ratio}
\end{figure}

In this section, we additionally compare each method's variance estimate $\widehat V$ with the efficiency bound $V_q^\star(\P)$. Since the simulator specifies the propensity score and the conditional law of each potential outcome, the bound can be easily computed from the asymptotic variances given by \citet{hahn1998} for the ATE and by \citet{firpo2007} for the QTE.\footnote{For the MST and PLR, the bound involves the conditional law of the failure time and the conditional variance of a base-table column given the covariates, respectively. The simulator does not specify them in closed form, so the bound $V_q^\star(\P)$ have to be estimated. We therefore do not consider this comparison for these two tasks.} Figure~\ref{fig:variance-ratio} shows that the variance ratio of our method is very close to 1, which provides additional evidence for the consistency $\widehat V \to V_q^\star(\P)$ established in Theorem~\ref{thm:main}. Note that some methods even have ratios below 1, which explains why their confidence intervals fail to attain the nominal coverage level.

\clearpage
\renewcommand{\refname}{References}
\putbib

\end{bibunit}
\end{document}